\PassOptionsToPackage{hyphens}{url}
\documentclass{egpubl}
\usepackage{eg2027}

\newif\ifanon
\anonfalse
\ConferenceSubmission
\makeatletter
\copyrightTextTitPag{}\copyrightTextRunPag{}
\def\ps@titlepage{\let\@mkboth\@gobbletwo\def\@oddhead{}\def\@oddfoot{}\let\@evenhead\@oddhead\let\@evenfoot\@oddfoot}
\makeatother

\usepackage[utf8]{inputenc}
\usepackage[T1]{fontenc}
\usepackage{dfadobe}

\usepackage{cite}
\BibtexOrBiblatex
\electronicVersion
\PrintedOrElectronic

\usepackage{graphicx}
\usepackage{egweblnk}

\usepackage{amsmath,amssymb}
\usepackage{booktabs}
\usepackage{multirow}
\usepackage{cleveref}
\usepackage{subcaption}
\usepackage{xcolor}
\usepackage{enumitem}
\usepackage{colortbl}
\usepackage[normalem]{ulem}
\usepackage{float}
\usepackage{tikz}
\usetikzlibrary{arrows.meta}
\usepackage{stfloats}
\usepackage{array}
\usepackage{listings}
\usepackage{algorithm}
\usepackage{algpseudocode}
\lstdefinestyle{jsonbox}{%
  basicstyle=\scriptsize\ttfamily,
  backgroundcolor=\color{gray!10},
  frame=single,rulecolor=\color{gray!40},
  breaklines=true,breakatwhitespace=false,
  columns=flexible,keepspaces=true,
  showstringspaces=false,
  aboveskip=6pt,belowskip=6pt,
  escapeinside={(*@}{@*)},
}

\newcommand{\todoslot}[1]{}
\newif\ifarxiv
\arxivfalse
\ifarxiv
  \makeatletter
  \renewcommand\ps@titlepage{\let\@mkboth\@gobbletwo
    \def\@oddhead{}\def\@evenhead{}\def\@oddfoot{}\def\@evenfoot{}%
    \let\sectionmark\EmptySectionmark \let\subsectionmark\EmptySubsectionmark}
  \def\@oddhead{}\def\@evenhead{}\def\@oddfoot{}\def\@evenfoot{}%
  \copyrightTextTitPag{}\copyrightTextRunPag{}%
  \makeatother
\fi

\usepackage{xr}%
\newcommand{\supp}[1]{supplementary~\Cref{S-#1}}%
\newlength{\qw}
\newcommand{\qI}[1]{\includegraphics[width=\qw]{#1}}
\newcommand{\qIn}[2]{\begin{tikzpicture}[baseline,inner sep=0]\node[anchor=south west,inner sep=0](img){\includegraphics[width=\qw]{#1}};\node[anchor=south east,inner sep=1.5pt,font=\tiny\itshape,text=black]at(img.south east){N\,=\,#2};\end{tikzpicture}}
\newcommand{\qIz}[1]{\includegraphics[width=\qw,trim={60pt 60pt 60pt 60pt},clip]{#1}}
\newcommand{\qIzn}[2]{\begin{tikzpicture}[baseline,inner sep=0]\node[anchor=south west,inner sep=0](img){\includegraphics[width=\qw,trim={60pt 60pt 60pt 60pt},clip]{#1}};\node[anchor=south east,inner sep=1.5pt,font=\tiny\itshape,text=black]at(img.south east){N\,=\,#2};\end{tikzpicture}}
\newcommand{\qIair}[1]{\raisebox{0.12\qw}{\includegraphics[width=\qw,trim={30pt 80pt 15pt 80pt},clip]{#1}}}
\newcommand{\qIairn}[2]{\raisebox{0.12\qw}{\begin{tikzpicture}[baseline,inner sep=0]\node[anchor=south west,inner sep=0](img){\includegraphics[width=\qw,trim={30pt 80pt 15pt 80pt},clip]{#1}};\node[anchor=south east,inner sep=1.5pt,font=\tiny\itshape,text=black]at(img.south east){N\,=\,#2};\end{tikzpicture}}}
\newcommand{\qIs}[2]{\begin{tikzpicture}[baseline,inner sep=0]\path[use as bounding box](0,0)rectangle(\qw,\qw);\node[anchor=center,inner sep=0]at(0.5\qw,0.5\qw){\includegraphics[width=#1\qw]{#2}};\end{tikzpicture}}
\newcommand{\qIsn}[3]{\begin{tikzpicture}[baseline,inner sep=0]\path[use as bounding box](0,0)rectangle(\qw,\qw);\node[anchor=center,inner sep=0](img)at(0.5\qw,0.5\qw){\includegraphics[width=#1\qw]{#2}};\node[anchor=south east,inner sep=1.5pt,font=\tiny\itshape,text=black]at(\qw,0){N\,=\,#3};\end{tikzpicture}}
\newcommand{\qIfailed}{\makebox[\qw][c]{\raisebox{0.5\qw}{\small\textit{failed}}}}
\newcommand{\qlab}[1]{\raisebox{0.43\qw}{\rotatebox[origin=c]{90}{\small #1}}}

\ifanon \usepackage{lineno} \linenumbers \fi
\title[Training-free Primitive Shape Abstraction]%
      {Prompting Image Generators for\\Training-free Primitive Shape Abstraction}

\ifanon
\author[Submission 1146]{Submission 1146}
\else
\author[G.~Kobsik \& T.~Elsner \& L.~Kobbelt]{Gregor Kobsik \qquad Tim Elsner \qquad Leif Kobbelt}
\fi

\begin{document}

\maketitle

\begin{abstract}
  Compact primitive abstractions represent 3D shapes with a few geometric primitives while preserving recognizable components.
  Learned methods depend on their training classes, and optimization-based methods split shapes geometrically rather than into parts.
  We instead reuse the visual part knowledge of pretrained models without task-specific training or fine-tuning.
  A vision-language model names parts in multi-view renders, and an unmodified image generator paints color-coded part masks.
  Reprojection and spatial clustering recover 3D instances, and a classical optimizer fits one tapered and bent superquadric per part.
  With five to eight primitives per object, the abstractions match the Chamfer distance of the strongest learned baseline on HumanPrim, improve on it by 10\,\% on Toys4K, and have the lowest overlap among compact methods, while chair legs, backrest bars and wheels remain separate primitives.
  Our accuracy also transfers better than theirs to objects outside the learned methods' ShapeNet training classes.
  Replacing the generated masks with part labels from the 3D segmentation methods \mbox{P3-SAM} or PartField lowers IoU by 7 to 17 points under the same fitter.
  Further studies relate the remaining volumetric error to part granularity and to parts that the rendered views observe from one side only.


\end{abstract}

\begin{figure*}[t]
  \centering
  \setlength{\fboxsep}{0pt}%
  \newdimen\sw    \sw=0.155\linewidth
  \newdimen\imght \imght=2.9cm
  \newdimen\titlht \titlht=2.4em
  \newdimen\capht  \capht=3.8em
  \newdimen\capwd  \capwd=\dimexpr\sw+0.7cm\relax
  \newdimen\titwd  \titwd=\dimexpr\sw+10.0cm\relax
  \newdimen\robotht \robotht=1.2cm
  \tikzset{badge/.style={circle, fill=black!75, draw=white, line width=0.5pt,
    text=white, font=\scriptsize\bfseries, inner sep=2pt}}%
  \providecommand{\pstage}[5]{}%
  \renewcommand{\pstage}[5]{%
    \begin{minipage}[t]{\sw}\centering
      \makebox[0pt][c]{\parbox[b][\titlht][b]{\titwd}{\centering\small\bfseries #1}}\\[3pt]%
      \parbox[c][\imght][s]{\sw}{\centering%
        \vfill%
        \begin{tikzpicture}[baseline=(img.base)]%
          \node[inner sep=0] (img)
            {\includegraphics[width=\sw,height=\imght,keepaspectratio]{#2}};%
          \node[badge, anchor=north west, xshift=4pt, yshift=-4pt]
            at (img.north west) {#5};%
        \end{tikzpicture}%
      }\\[5pt]%
      \makebox[0pt][c]{\parbox[t][\capht][t]{\capwd}{\centering\scriptsize #4}}%
    \end{minipage}%
  }%
  \providecommand{\pipearrow}{}%
  \renewcommand{\pipearrow}{%
    \begin{minipage}[t]{0.6cm}\centering
      \parbox[b][\titlht][b]{0.6cm}{}\\[3pt]%
      \parbox[c][\imght][c]{0.6cm}{\centering%
        \tikz[baseline=-0.5ex]%
          \draw[-{Stealth[length=5pt,width=4pt]},line width=2pt,color=gray!60]%
            (0,0)--(0.45cm,0);%
      }\\[3pt]%
      \parbox[t][\capht][t]{0.6cm}{}%
    \end{minipage}%
  }%
  \providecommand{\vlmbridge}{}%
  \renewcommand{\vlmbridge}{%
    \begin{minipage}[t]{2cm}\centering
      \parbox[b][\titlht][b]{2cm}{}\\[3pt]%
      \parbox[c][\imght][c]{2cm}{\centering%
        \begin{tikzpicture}[baseline=-0.5ex]%
          \draw[-{Stealth[length=5pt,width=4pt]},line width=2pt,color=gray!60]%
            (0.25cm,0)--(1.75cm,0);%
          \node[above=0pt, overlay] at (1cm,0) {%
            \IfFileExists{images/workflow/vlm_robot.jpeg}{%
              \includegraphics[height=\robotht]{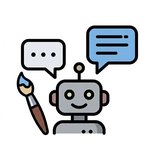}%
            }{%
              \colorbox{gray!20}{\scriptsize\bfseries VLM+Gen}%
            }%
          };%
          \node[below=5pt, overlay] at (1cm,0) {\scriptsize\bfseries VLM\,+\,Gen};%
        \end{tikzpicture}%
      }\\[3pt]%
      \parbox[t][\capht][t]{2cm}{}%
    \end{minipage}%
  }%
  \providecommand{\imgstack}[5]{}%
  \renewcommand{\imgstack}[5]{%
    \begin{tikzpicture}[baseline,
        img/.style={inner sep=0, draw=gray!55, line width=0.6pt}]%
      \node[img] at (15pt,15pt)
        {\includegraphics[width=\dimexpr\sw-16pt\relax]{#1}};%
      \node[img] at (10pt,10pt)
        {\includegraphics[width=\dimexpr\sw-16pt\relax]{#2}};%
      \node[img] at (5pt,5pt)
        {\includegraphics[width=\dimexpr\sw-16pt\relax]{#3}};%
      \node[img] (front) at (0pt,0pt)
        {\includegraphics[width=\dimexpr\sw-16pt\relax]{#4}};%
      \node[badge, anchor=north west, xshift=4pt, yshift=0pt]
        at (current bounding box.north west) {#5};%
    \end{tikzpicture}%
  }%
  \begin{minipage}[t]{\sw}\centering
    \makebox[0pt][c]{\parbox[b][\titlht][b]{\titwd}{\centering\small\bfseries Multi-View Render}}\\[3pt]%
    \parbox[c][\imght][c]{\sw}{\centering%
      \imgstack{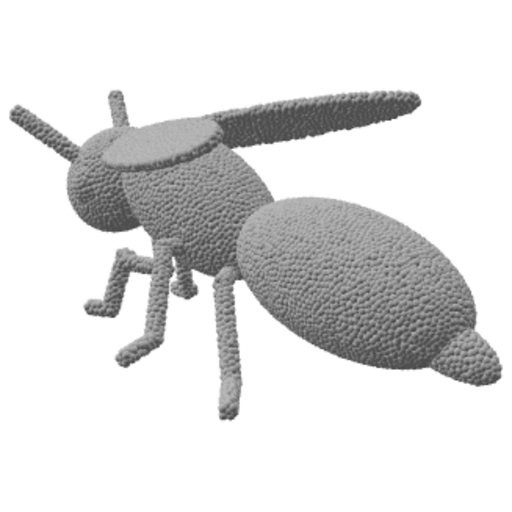}%
               {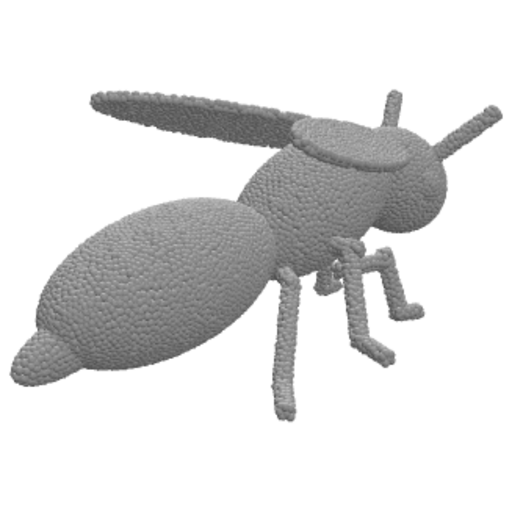}%
               {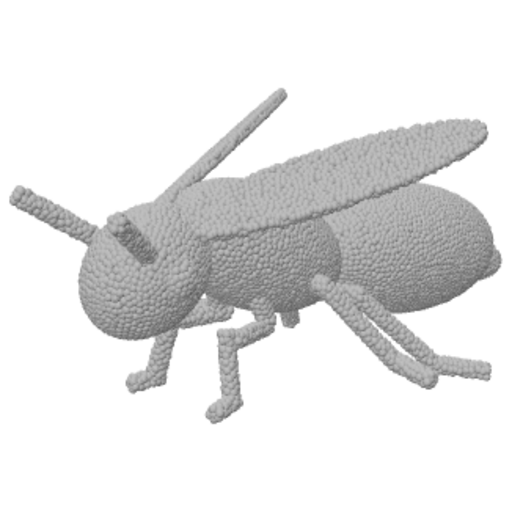}%
               {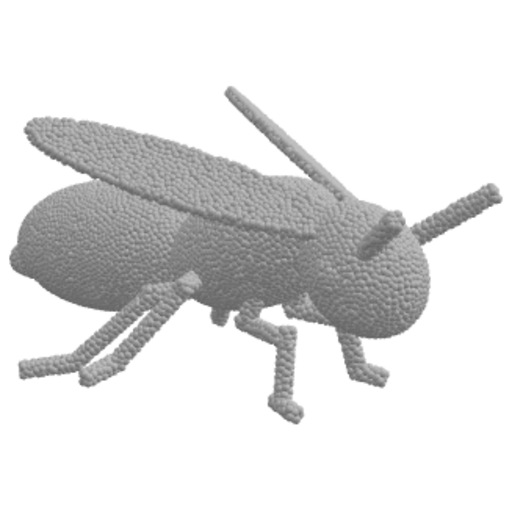}{1}%
    }\\[5pt]%
    \makebox[0pt][c]{\parbox[t][\capht][t]{\capwd}{\centering\scriptsize Render four perspective views of a single 3D object.}}%
  \end{minipage}%
  \vlmbridge
  \begin{minipage}[t]{\sw}\centering
    \makebox[0pt][c]{\parbox[b][\titlht][b]{\titwd}{\centering\small\bfseries Generative Segmentation}}\\[3pt]%
    \parbox[c][\imght][c]{\sw}{\centering%
      \imgstack{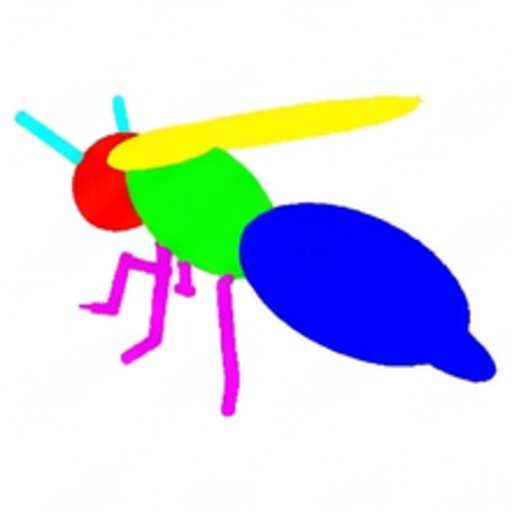}%
               {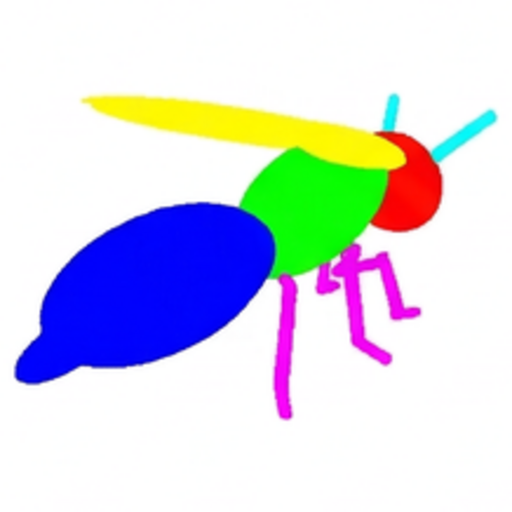}%
               {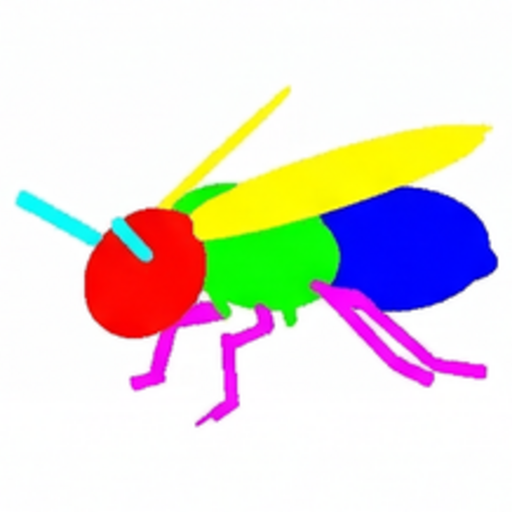}%
               {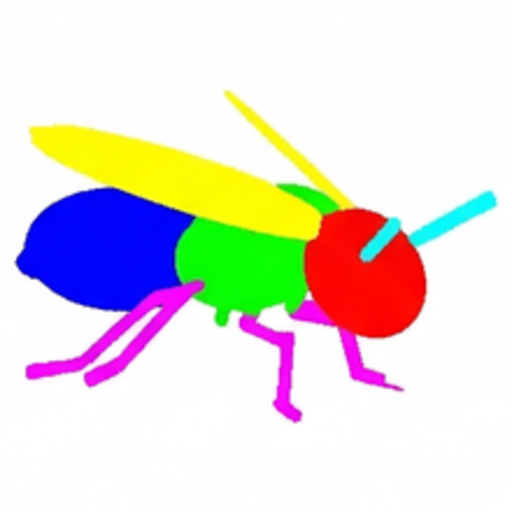}{2}%
    }\\[5pt]%
    \makebox[0pt][c]{\parbox[t][\capht][t]{\capwd}{\centering\scriptsize Prompt a generative image model to paint a color-coded part mask.}}%
  \end{minipage}%
  \pipearrow
  \pstage{3D Reprojection}%
         {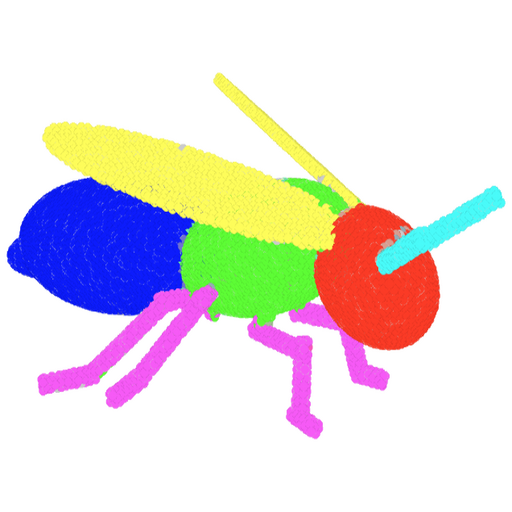}{2.9cm}%
         {Project 2D labels onto 3D geometry by voting across views.}{3}%
  \pipearrow
  \pstage{Clustering}%
         {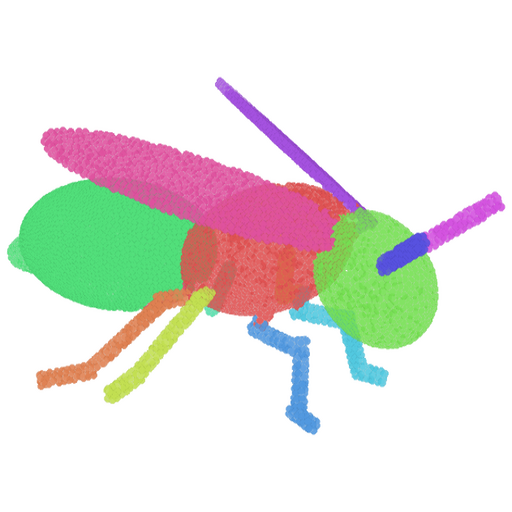}{2.9cm}%
         {Extract part point clouds by color-restricted spatial clustering.}{4}%
  \pipearrow
  \pstage{Abstraction}%
         {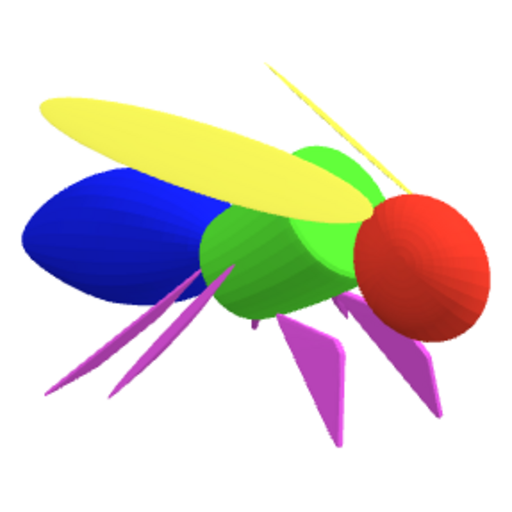}{2.9cm}%
         {Fit one superquadric primitive per part by parameter optimization.}{5}%
  \vspace{-20pt}
  \caption{%
    \textbf{Pipeline overview.}
    Given a 3D object, our pipeline extracts semantic part primitives in five steps:
    \textbf{(1)~Multi-View Render:} We render four perspective views and prompt a
    vision-language model (VLM) to identify the object's semantic parts.
    \textbf{(2)~Generative Segmentation:} A generative image model paints a
    color-coded part mask from that analysis.
    \textbf{(3)~3D Reprojection:} Pixel labels are projected back onto the 3D geometry
    by voting across views.
    \textbf{(4)~Clustering:} Color-restricted spatial clustering removes label noise
    and yields clean part point clouds.
    \textbf{(5)~Abstraction:} Parallel multi-start Chamfer-distance optimization fits
    one superquadric primitive per cluster.%
  }
  \label{fig:pipeline}
\end{figure*}
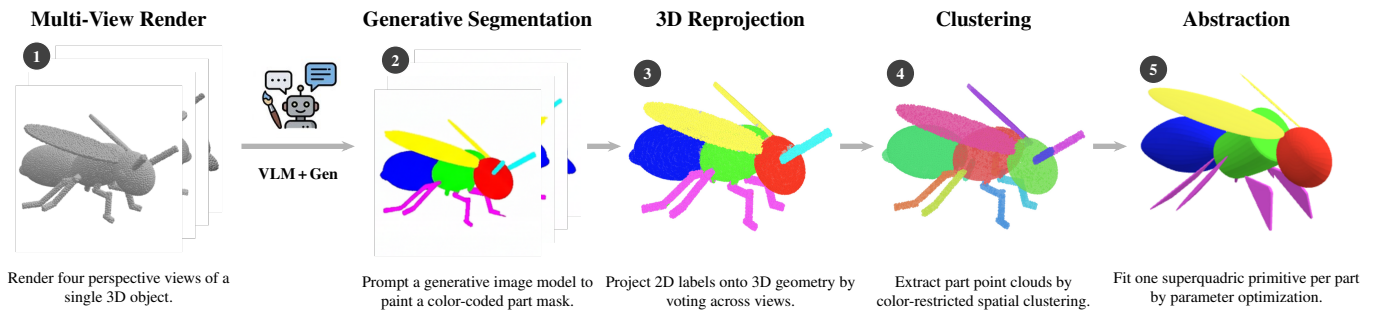

\section{Introduction}
\label{sec:introduction}

General-purpose pretrained models make knowledge reusable across tasks through prompting.
In natural language processing, one model can translate and answer questions without task-specific parameter updates~\cite{brown2020language}.
Image generators have likewise seen how objects decompose into parts across large image collections, and instruction-following models can paint such a decomposition as a color-coded mask.
We investigate this principle for compact 3D shape abstraction: can an existing image generator supply the visual part knowledge without specialized 3D training?

Shape abstraction represents complex geometry with simple primitives for robotic grasping, collision detection and shape editing.
Part-oriented abstractions should also preserve recognizable components, which low surface error alone does not ensure.
Methods trained on 3D shapes predict primitive assemblies directly~\cite{tulsiani2017learning, paschalidou2019superquadrics, fedele2025superdec, ye2025primitiveanything, tavernini2026superflex}, but transfer depends on their training distribution.
Optimization-based methods~\cite{liu2022ems, liu2023marching, wang2025lightsq, ganeshan2025superfrusta} avoid task-specific training but partition geometrically: a chair leg may merge with the seat or split across primitives (\Cref{fig:qual_humanprim}).
We instead reuse pretrained visual knowledge to connect semantic part grouping with classical geometric fitting.

Our pipeline separates part prediction from fitting (\Cref{fig:pipeline}).
A vision-language model examines multi-view renders and proposes a part vocabulary with one color per type, and an image generator paints the corresponding masks.
Reprojection transfers labels to the geometry, spatial connectivity separates instances, and optimization fits one superquadric per component.
All chair legs, for example, share a color and are separated in 3D, without learned lifting or consistent instance identities across views.
Neither model is fine-tuned, and no 3D component is trained.

Rendering, applying a 2D model and backprojecting predictions is established in 3D segmentation~\cite{liu2023partslip, zhong2024meshsegmenter}, as is superquadric recovery by optimization~\cite{solina1990recovery, paschalidou2019superquadrics}.
Our contribution couples them through \emph{generated part masks}, which determine the points fitted by each primitive.
This makes pretrained part knowledge usable for abstraction and lets us compare generated masks with learned 3D segmenters and human part annotations under the same fitter.

We evaluate on HumanPrim and Toys4K, with ShapeNet results in the supplementary material.
With 5--8 primitives, our method matches SuperFlex's Chamfer distance on HumanPrim, improves on it by 10\,\% on Toys4K, and achieves the lowest overlap among compact methods, while keeping components separate that learned baselines merge.
SuperFlex retains a volumetric IoU advantage, which controlled studies link to one-sided part thickness and granularity rather than the fitter.
Our accuracy also transfers better to objects outside the learned methods' training classes.
With a shared fitter, the generated masks give higher IoU than the 3D segmenters P3-SAM and PartField, while PartSAM, trained on 3D part data, reaches lower CD on both benchmarks and higher IoU on HumanPrim with 1.4--1.7$\times$ more primitives (\Cref{sec:seg_source}).

\section{Related Work}
\label{sec:related_work}

\subsection{Lifting 2D Segmentation to 3D}
\label{sec:lifting}

Foundation segmentation models such as SAM~\cite{kirillov2023sam} enabled class-agnostic 2D segmentation, and a broad family of methods lifts 2D foundation-model predictions to 3D by reprojection or feature aggregation.
PartSLIP~\cite{liu2023partslip}, MeshSegmenter~\cite{zhong2024meshsegmenter}, and Sun et al.'s study of GPT-4 Vision for zero-shot point clouds~\cite{qi2024gpt4vzeroshot} all render or project the 3D input to 2D, apply a vision foundation model, and backproject the resulting labels or features.
Our pipeline follows the same render--apply--backproject skeleton but differs in what the 2D model is and what it returns.
We prompt a \emph{generative} model to paint a part mask instead of reading \emph{features} out of a discriminative backbone, and we fit a primitive to each part instead of stopping at backprojected labels.
SAMPart3D~\cite{yang2024sampart3d} and COPS~\cite{garosi2025cops} remove the need for part-label supervision.
You et al.~\cite{you2024img2cad} combine component identification by a vision-language model with a trained parameter estimator to reverse-engineer parametric CAD programs from a single image.
Their output space (CAD operations instead of volumetric primitives, cf.\ the BenchCAD benchmark~\cite{zhang2026benchcad}) makes a direct comparison to primitive abstraction ill-defined.
Native 3D segmenters such as PartField~\cite{qian2025partfield}, P3-SAM~\cite{li2025p3sam}, and Point-SAM~\cite{zhou2025pointsam} operate directly on point clouds or voxels and skip the render-and-reproject round trip.
Point-SAM in particular offers a promptable interface that is a category-agnostic alternative to our generative decomposition stage.
Most recently, SegviGen~\cite{li2026segvigen} steers a native 3D generative model to produce part segmentations by injecting 2D color guidance into its denoising process.
Across all these methods, 3D part understanding is obtained either by reprojecting discriminative model predictions or by training native 3D networks.
We instead prompt an unmodified generative image model to paint part masks and transfer them to 3D by geometric reprojection, without task-specific training or a learned lifting module.
\paragraph*{Concurrent work.}
Gabeur et al.~\cite{gabeur2026imagegen} showed concurrently that image generators trained at scale are strong generalist vision learners.
After lightweight instruction tuning, their generator reaches state-of-the-art segmentation and metric-depth results by reframing perception as image generation.
Their finding and ours point in the same direction but differ in mechanism and target.
They fine-tune the generator for pixel-level 2D perception, whereas we use a production model as is, through prompting only, and lift its part-level output into a 3D primitive decomposition.

\subsection{Learning-Based Shape Abstraction}
\label{sec:learning}

Learning-based abstraction started from supervised cuboid prediction~\cite{tulsiani2017learning, zou2017prnn} and moved to self-supervised decompositions into cuboids~\cite{sun2019hierarchicalcuboid, yang2021jointcuboid, kobsik2025f2c}, superquadrics and deformable parts~\cite{paschalidou2019superquadrics, paschalidou2021neuralparts, fedele2025superdec, tavernini2026superflex} and sweep surfaces~\cite{zhao2024sweepnet}.
A second line treats abstraction as sequence generation.
PASTA~\cite{li2024pasta} and Primitive Anything~\cite{ye2025primitiveanything} generate primitive assemblies auto-regressively, the latter trained on human-crafted abstractions, and Tian et al.~\cite{tian2025llmprimitives} fine-tune an LLM to emit primitive parameters.
These methods are strong within their training distribution but require category-specific training data and typically assume a canonical orientation.
Our training-free pipeline needs neither training data nor alignment preprocessing.

\subsection{Superquadric Recovery and Shape Primitives}
\label{sec:primitives}

Superquadrics~\cite{barr1981superquadrics, pentland1986parts} interpolate between cuboids, ellipsoids, cylinders and octahedra with two shape parameters.
Solina and Bajcsy~\cite{solina1990recovery} added global tapering and bending and a recovery method from range images, which Leonardis et al.~\cite{leonardis1997superquadrics} extended to multi-part recovery and Jakli\v{c} et al.~\cite{jaklic2000superquadrics} surveyed.
Optimization-based multi-primitive fitting continues this line.
EMS~\cite{liu2022ems} casts it as expectation maximization, Marching Primitives~\cite{liu2023marching} fits from signed distance functions, Light-SQ~\cite{wang2025lightsq} partitions a watertight mesh by SDF carving before fitting, and SuperFrusta~\cite{ganeshan2025superfrusta} fits a unified 8-parameter primitive by residual refinement.
Multi-view variants fit textured or silhouette-matched superquadrics by differentiable rendering~\cite{monnier2023dbw, alaniz2023iterative} or combine them with Gaussian splatting~\cite{gao2025partgs}.
All of these partition an object geometrically and lack the cross-instance semantic knowledge needed to identify parts consistently across a class.
Zhao et al.~\cite{zhao2025uspc} report a ``seg + fit'' baseline with the same two-stage structure as ours: it clusters Diff3F features~\cite{dutt2024diff3f} with $k$-means and fits superquadrics by principal component analysis.
Our pipeline replaces both components, with generated part masks instead of feature clustering and multi-start optimization instead of a closed-form fit.

\section{Methodology}
\label{sec:methodology}

\noindent Our pipeline consists of three stages: (1) multi-view rendering and generative part segmentation, (2) reprojection and color-restricted clustering, and (3) superquadric primitive fitting via parallel multi-start optimization.

\subsection{Multi-View Capture and Generative Segmentation}
\label{sec:segmentation}

We follow the established render, apply and backproject scheme for lifting 2D predictions to 3D and render the object from four ring views, $90^\circ$ apart at a slight elevation, and tile them into one composite image.
The six-view layout of \Cref{sec:results} adds a top and a bottom view (all parameters in \supp{tab:hyperparams}).
The colors, the number and granularity of parts, and the semantic assignment (all chair legs map to one part type) must agree across views.
We therefore separate semantic analysis from mask generation into two model calls.

The first call is the \emph{analysis}.
A vision-language model (Gemini 3 Flash) receives the multi-view image and a structured prompt asking it to identify the object, decompose it into semantic parts (e.g., legs, seat, back), and assign a distinct color to each.
Its output is a structured JSON mapping part names to colors that anchors the second call.
We assign one color per semantic part \emph{type} rather than per instance (left and right legs share a color) for two reasons.
The image models we tested cannot reliably distinguish left from right across opposing views, and per-view instance identities tend to be inconsistent across viewpoints, so instance-colored masks cannot be reprojected into a coherent 3D labeling.
The instance-color experiment is reported in \Cref{S-sec:instance_seg}.

The second call is the \emph{mask generation}.
A generative image model (Nano Banana~2~\cite{nanobana2}) receives the same multi-view image together with the part-to-color mapping from the first call and paints a color-coded part mask.
Because the part-to-color mapping is fixed upfront, the model only has to follow it, which makes cross-view color consistency achievable.
Separating analysis from generation also makes the decomposition explicit.
The part list is a human-readable JSON that can be inspected and edited before the more expensive generation proceeds, whereas learned methods encode the part structure implicitly in their weights.

\subsection{Reprojection and Color-Restricted Clustering}
\label{sec:clustering}

The mask is reprojected onto the geometry by voting across views.
Each visible surface point collects the mask pixels around its projection in every view in which it is seen and takes the most frequent label.
Points seen in no view stay unlabeled and are discarded with the background.
The part names of the analysis step are used only in the prompt, and the fitter sees colors.
Before voting, every mask pixel is snapped to a global palette obtained by mean-shift in CIELAB over all views (bandwidth $\Delta E = 40$), so that a part painted in slightly different shades across views receives one palette color.

The generated masks are imperfect, and colors bleed at part boundaries, so the reprojected labels contain noise.
Voting across views already suppresses a poor mask in any single view.
To extract clean per-part point clouds, we additionally apply color-restricted spatial clustering.
Points are grouped by palette color, background points are discarded, and within each color group we extract spatially connected components with DBSCAN~\cite{ester1996dbscan} using a minimum neighborhood size of one, i.e.\ single-linkage components over the $r$-neighborhood graph ($r = 0.05$, objects normalized to $[-1,1]^3$, KD-tree queries).
Clusters below a minimum size ($1\,\%$ of the object's points) are discarded as outliers, since small disconnected fragments typically come from mask noise or color bleed.

\subsection{Superquadric Primitive Fitting}
\label{sec:fitting}

Given the part point clouds from the previous stage, this stage is a plain parameter optimization that fits one superquadric per point cloud, independently and without further subdivision.
Superquadric fitting is, however, a non-convex problem.
The loss landscape contains many local minima, and a single initialization frequently converges to a non-optimal solution.
We address this with a parallel multi-start strategy.

Each point cloud segment is first normalized to a unit cube, then fitted with a superquadric parameterized by up to 15 values: size $(a_1, a_2, a_3)$, position $(p_x, p_y, p_z)$, rotation $(\alpha, \beta, \gamma)$ as Euler angles, shape $(\epsilon_1, \epsilon_2)$, and optional tapering $(k_x, k_y)$ and bending $(k_b, \alpha_b)$ deformations~\cite{solina1990recovery}.
We initialize from an approximate minimum-volume bounding box (ApproxMVBB,~\cite{barequet2001mvbb}) and generate nine initial poses by combining three primitive types (a cuboid with $\epsilon_1 = \epsilon_2 = 0.1$, an ellipsoid with $\epsilon_1, \epsilon_2 = 1.0$, and a cylinder with $\epsilon_1 = 0.1, \epsilon_2 = 1.0$) with three axis-aligned orientations obtained by rotating the bounding box axes.
The orientation variation is necessary because deformations act along the principal axis.
Each pose is started rigid, tapered, bent, bent in the opposite direction, and tapered and bent, giving 45 candidates per segment.

All candidates are optimized in parallel using L-BFGS~\cite{liu1989lbfgs} (80 iterations, 2048 input points and 2500 surface samples per segment), minimizing, as in~\cite{paschalidou2019superquadrics}, the bidirectional Chamfer distance between the point cloud $\mathcal{P}$ and points sampled from the superquadric surface $\mathcal{S}$:
\begin{equation}
  \mathcal{L} = \frac{1}{|\mathcal{P}|} \sum_{\mathbf{p} \in \mathcal{P}} \min_{\mathbf{s} \in \mathcal{S}} \|\mathbf{p} - \mathbf{s}\|^2 + \lambda \sum_{\mathbf{s} \in \mathcal{S}} w_\mathbf{s} \min_{\mathbf{p} \in \mathcal{P}} \|\mathbf{s} - \mathbf{p}\|^2
  \label{eq:chamfer}
\end{equation}
The forward term pulls the primitive surface towards every input point, so no part of the segment is left unfitted.
The backward term, weighted by $\lambda = 0.1$, pulls every sampled primitive surface point towards the data and thus penalizes primitive surface area where there are no points, i.e.\ over-extension and inflated volume.
Surface samples are drawn from the triangulated parametric grid of the superquadric with probability proportional to triangle area and uniform barycentric coordinates, with equal weights $w_\mathbf{s} = 1/|\mathcal{S}|$, so the backward term is a Monte-Carlo estimate of the area-weighted mean over the primitive surface instead of a mean over the non-uniform parametric grid.
The samples are redrawn with a fixed seed in every iteration and remain differentiable through the barycentric interpolation.
Both terms are surface distances, so a segment observed from one side only (a mattress top, a table top) admits a zero-thickness optimum with perfect Chamfer distance but no volume.
We therefore floor every axis $a_i$ at $2\,\%$ of the segment's largest bounding-box half-extent.
The floor mainly binds in these degenerate cases and is neutral for most segments (\supp{sec:fitter_floor}).

Parameters are optimized in an unconstrained latent space and mapped to valid ranges, e.g.\ through $\tanh$ for the bend ($k_b \in (-1, 1)$) and taper ($k_x, k_y \in (-2, 2)$) parameters, where $0$ is the undeformed primitive.
After convergence, we rank the candidates by the Chamfer objective, keep the ten lowest, and among these select the one with the largest axis product $a_1 a_2 a_3$, a proxy for its volume, which is then de-normalized to the original coordinate frame.
The Chamfer objective alone does not fix the thickness of a part observed from one side, so candidates within a fraction of a percent in cost can differ several-fold in volume.
Preferring the larger one resolves the ambiguity at no extra cost (\supp{sec:selection} compares selection rules).

\begin{figure*}[p]
\centering
\setlength{\qw}{0.096\textwidth}
\setlength{\tabcolsep}{0.5pt}
\renewcommand{\arraystretch}{0.3}
\begin{tabular}{c@{\hspace{2pt}} cccccccccc}
\qlab{EMS} &
\qIn{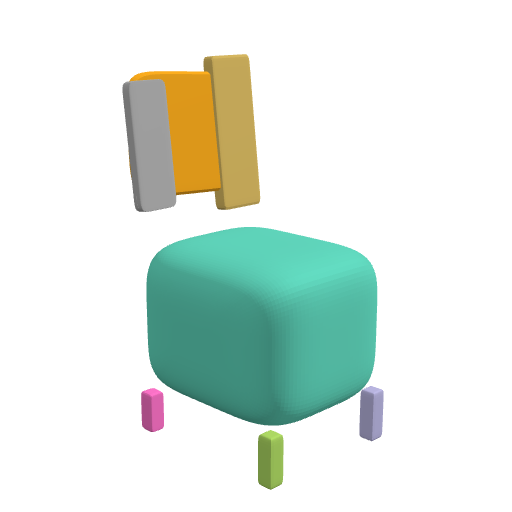}{8} &
\qIn{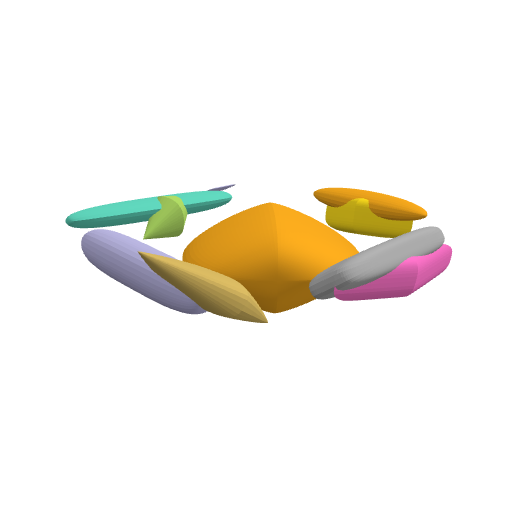}{8} &
\qIn{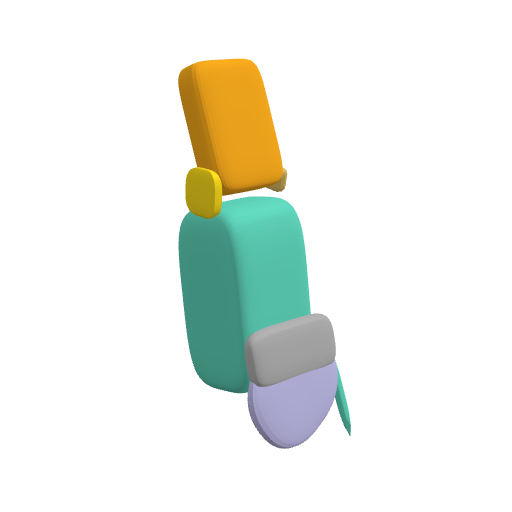}{8} &
\qIzn{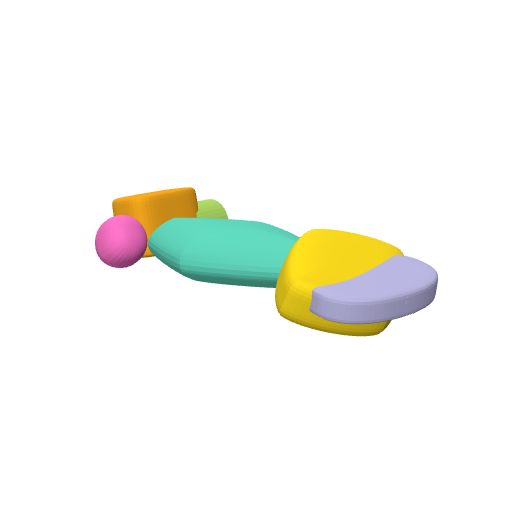}{6} &
\qIzn{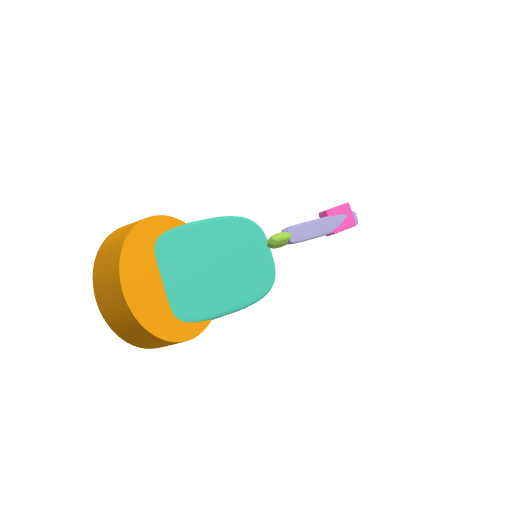}{5} &
\qIzn{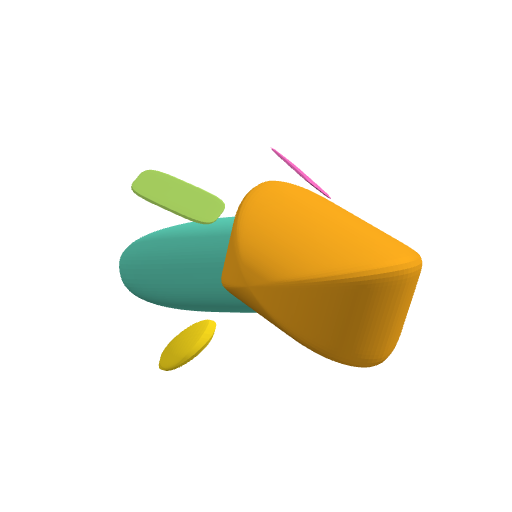}{6} &
\qIn{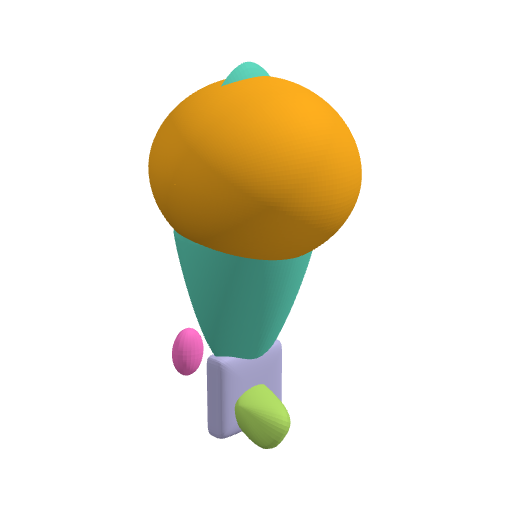}{5} &
\qIn{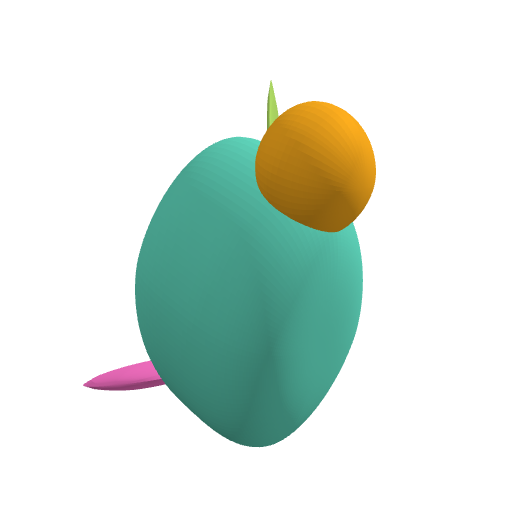}{6} &
\qIn{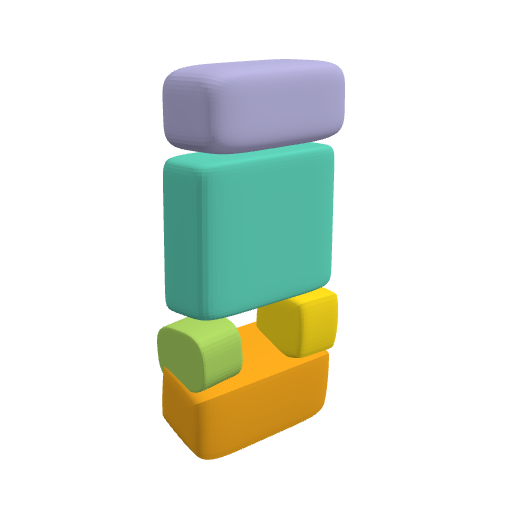}{6} &
\qIn{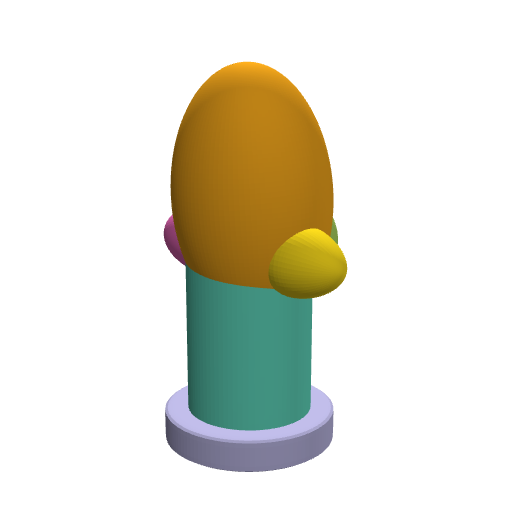}{6} \\
\qlab{Light-SQ} &
\qIn{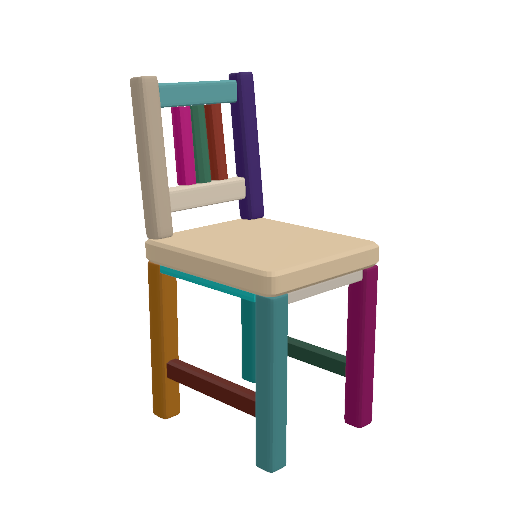}{18} &
\qIn{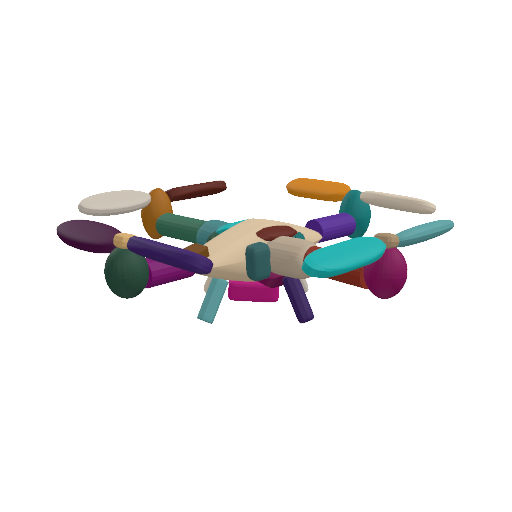}{37} &
\qIn{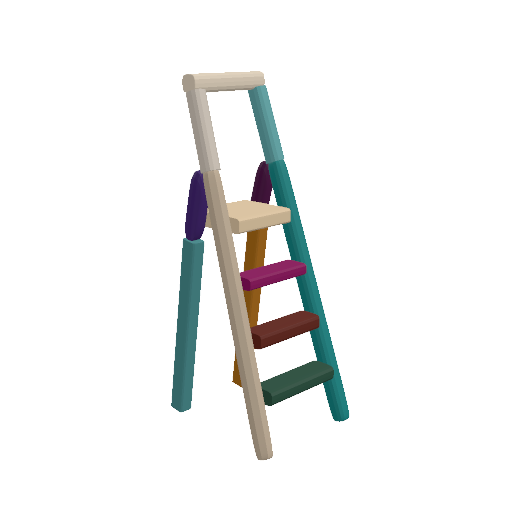}{13} &
\qIzn{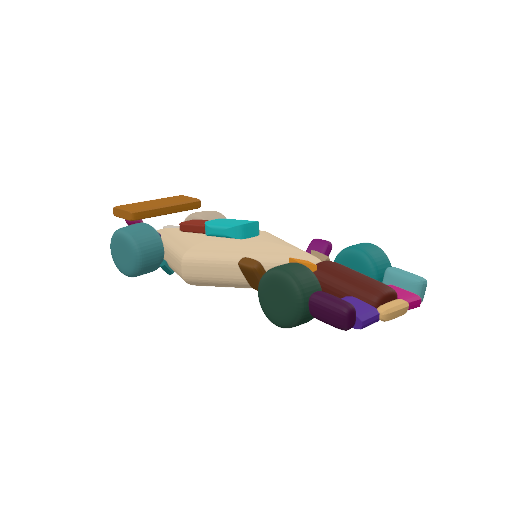}{25} &
\qIzn{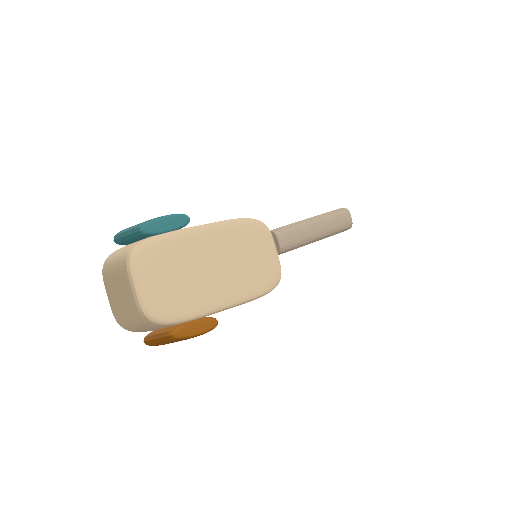}{4} &
\qIzn{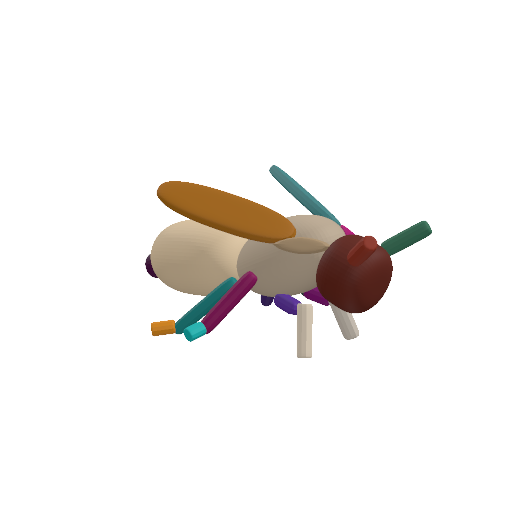}{22} &
\qIn{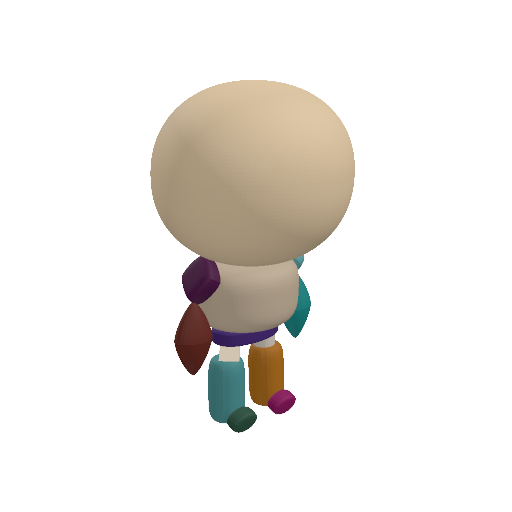}{13} &
\qIn{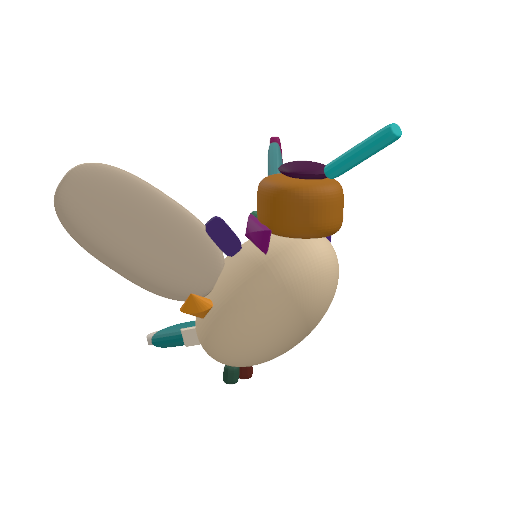}{20} &
\qIn{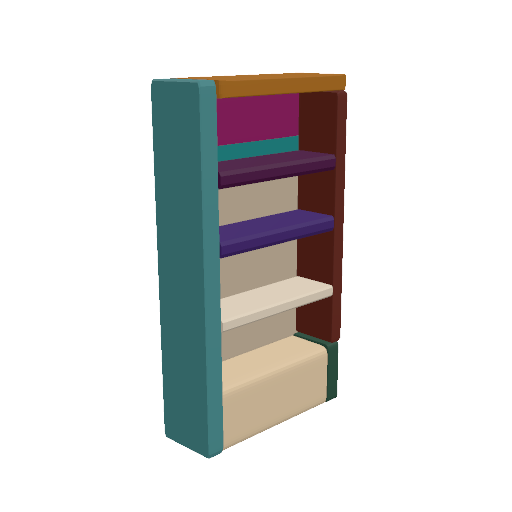}{11} &
\qIn{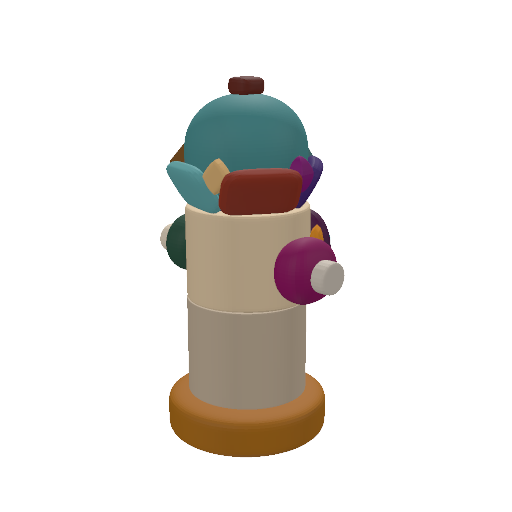}{30} \\
\midrule
\qlab{PrimAny} &
\qIn{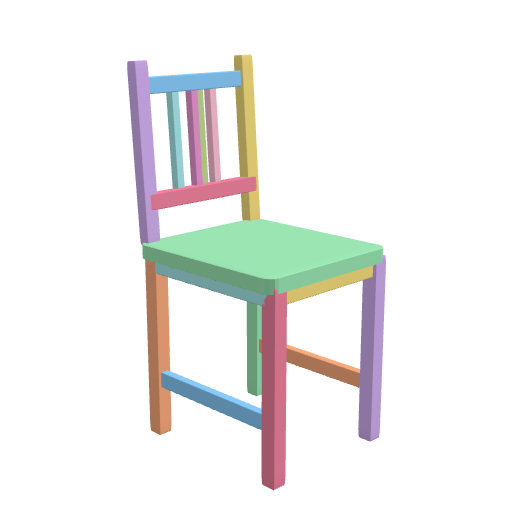}{19} &
\qIn{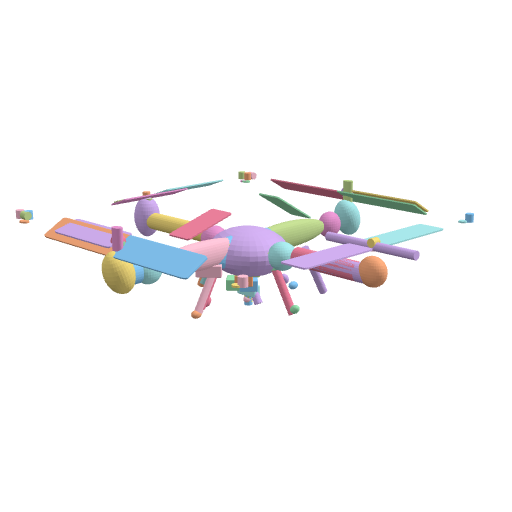}{107} &
\qIn{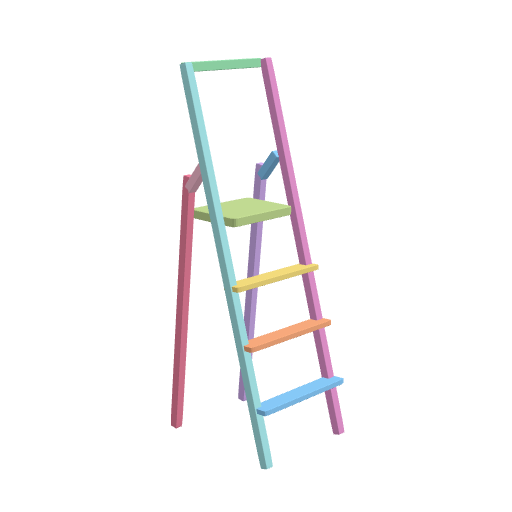}{11} &
\qIzn{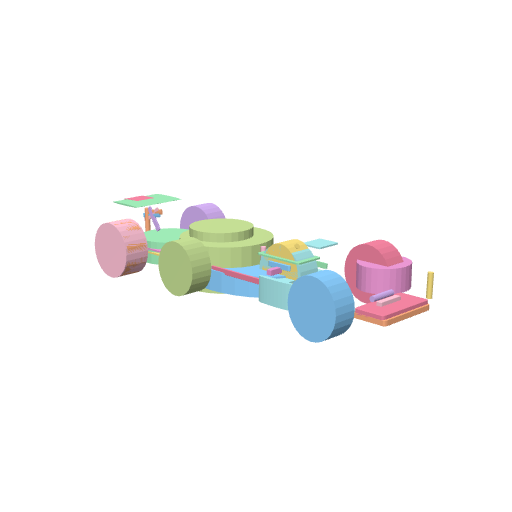}{40} &
\qIzn{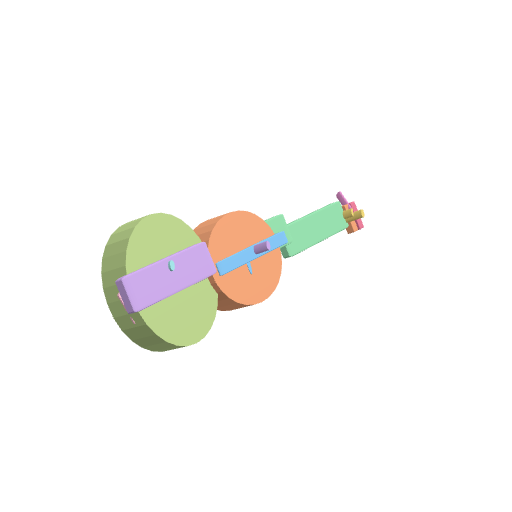}{16} &
\qIzn{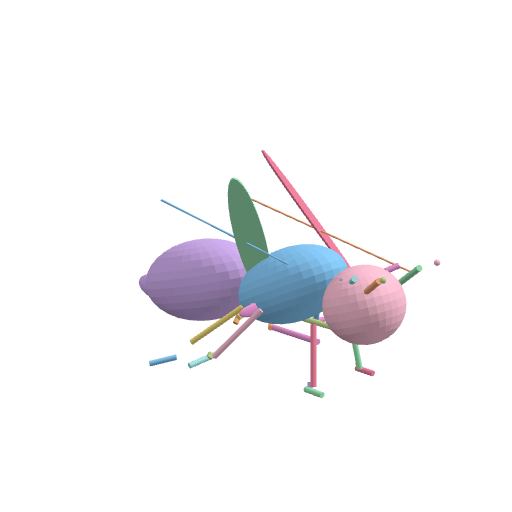}{40} &
\qIn{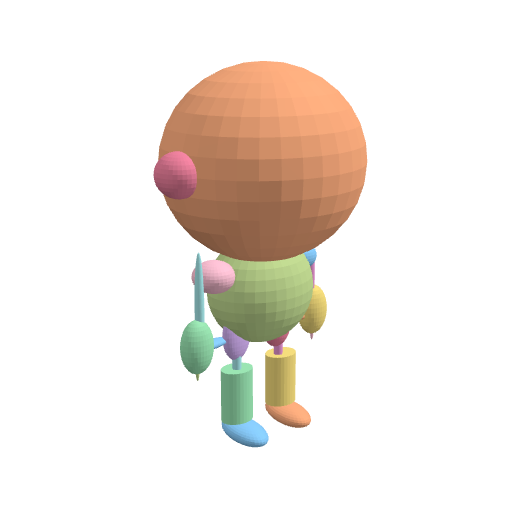}{22} &
\qIn{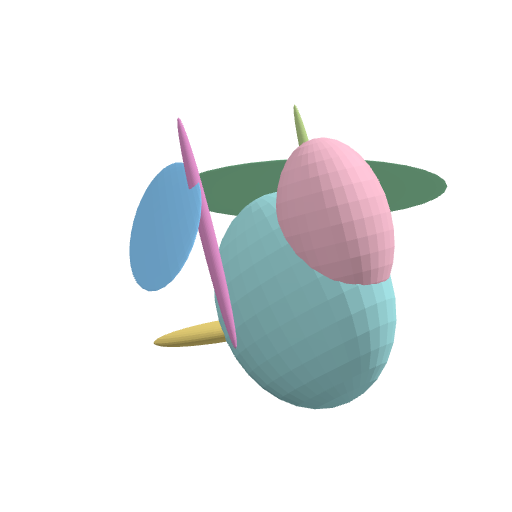}{11} &
\qIn{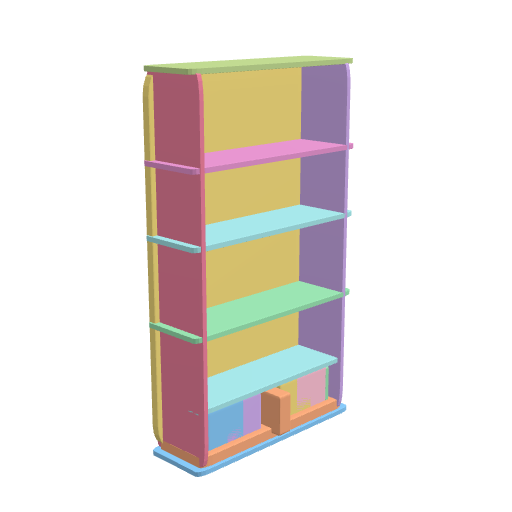}{17} &
\qIn{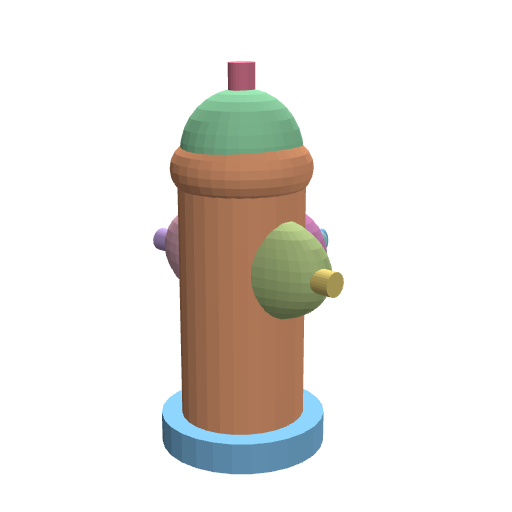}{12} \\
\qlab{F2C} &
\qIn{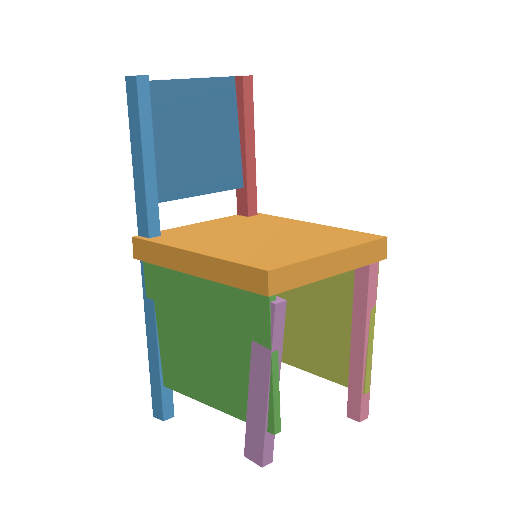}{8} &
\qIn{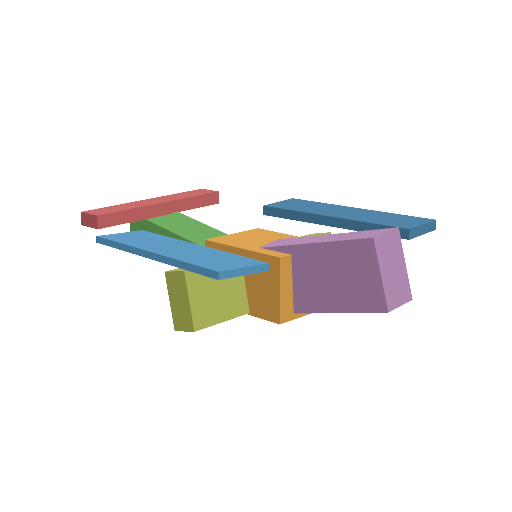}{7} &
\qIn{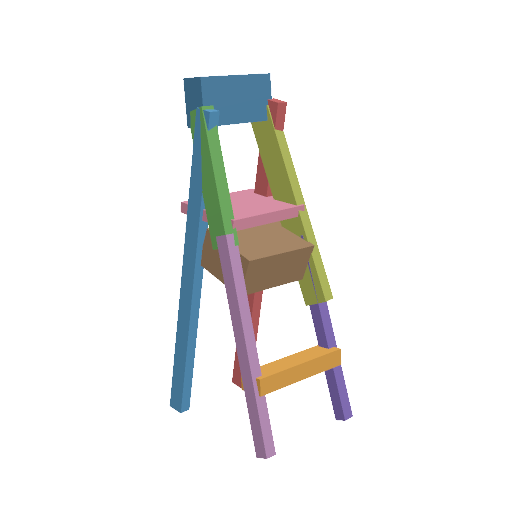}{10} &
\qIzn{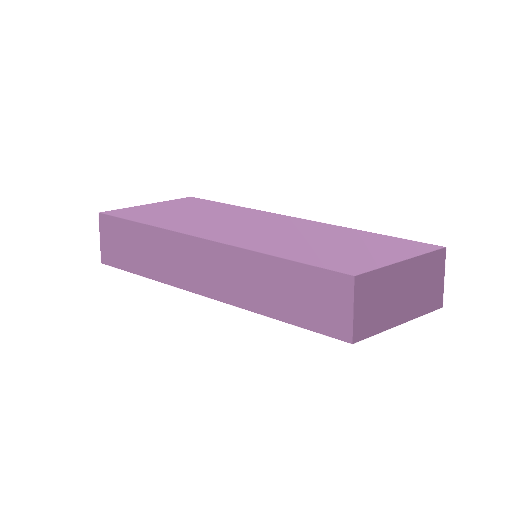}{1} &
\qIzn{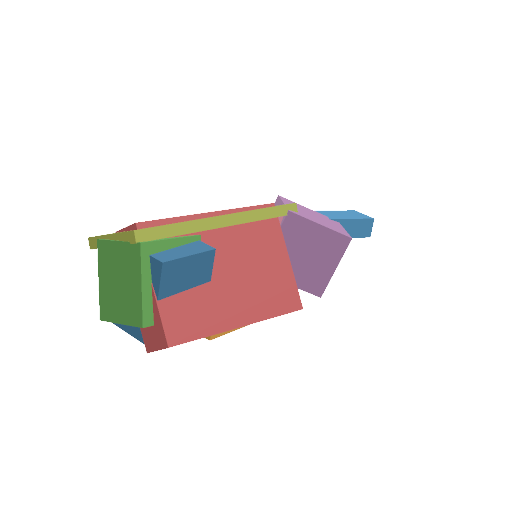}{8} &
\qIzn{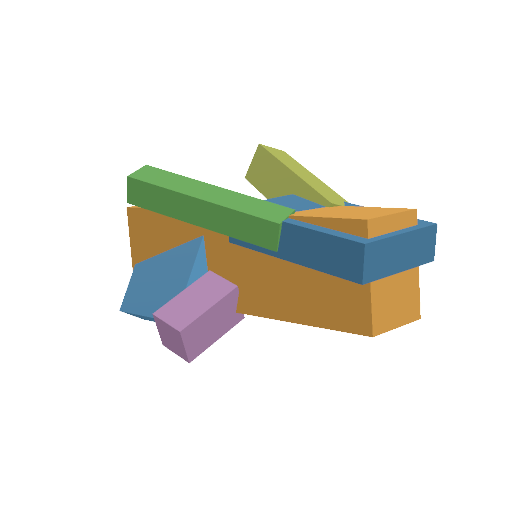}{8} &
\qIn{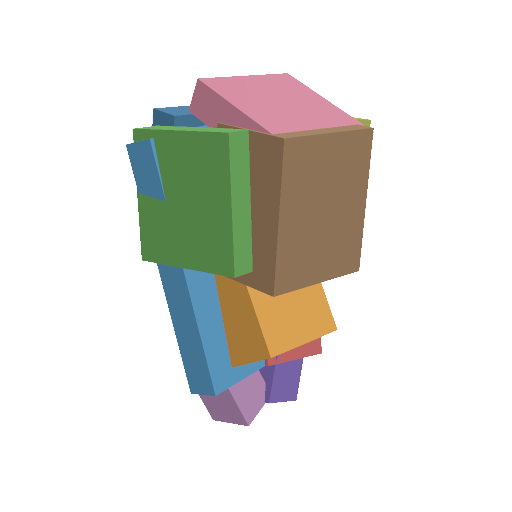}{10} &
\qIn{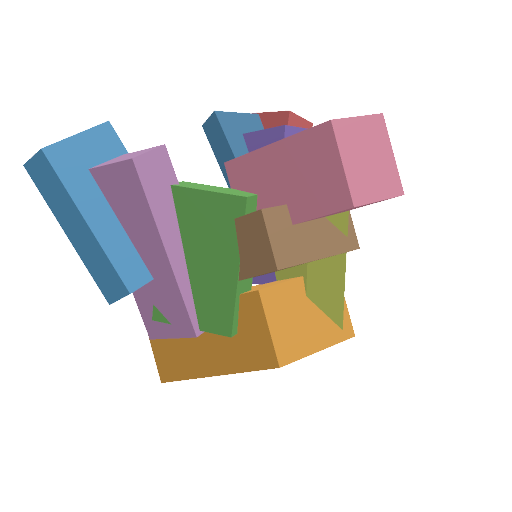}{10} &
\qIn{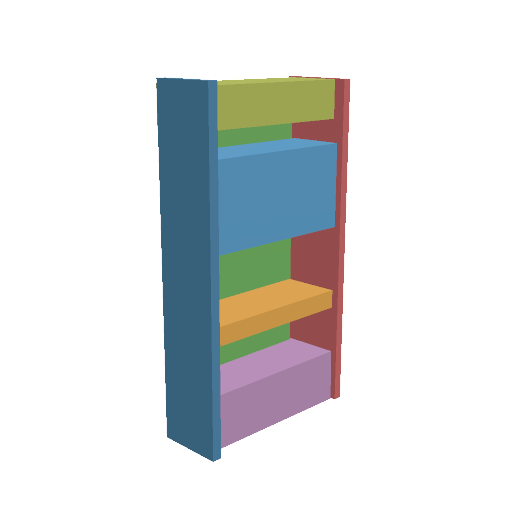}{7} &
\qIn{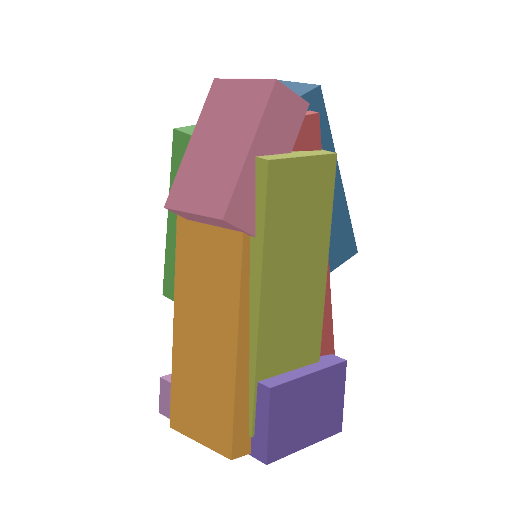}{9} \\
\qlab{SuperDec} &
\qIn{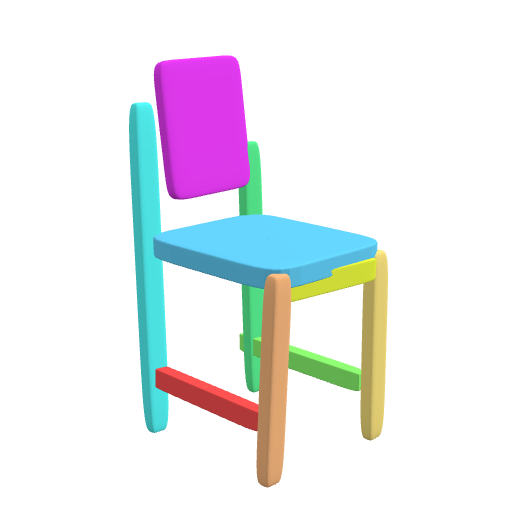}{10} &
\qIn{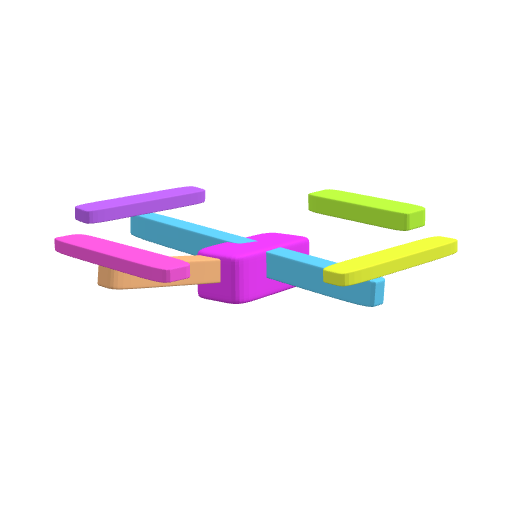}{7} &
\qIn{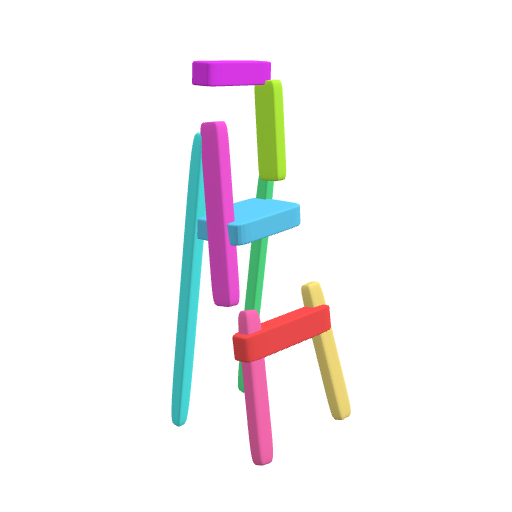}{9} &
\qIzn{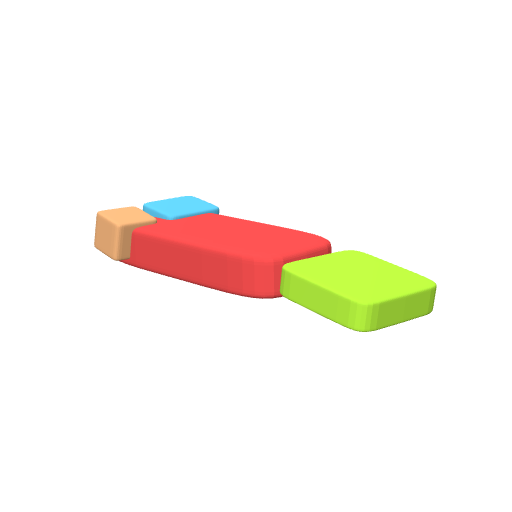}{4} &
\qIzn{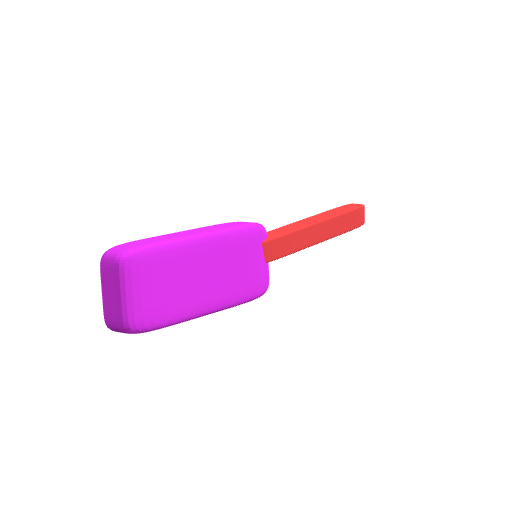}{2} &
\qIzn{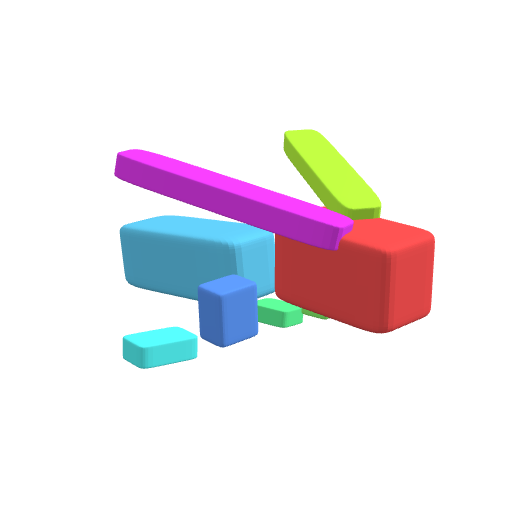}{8} &
\qIn{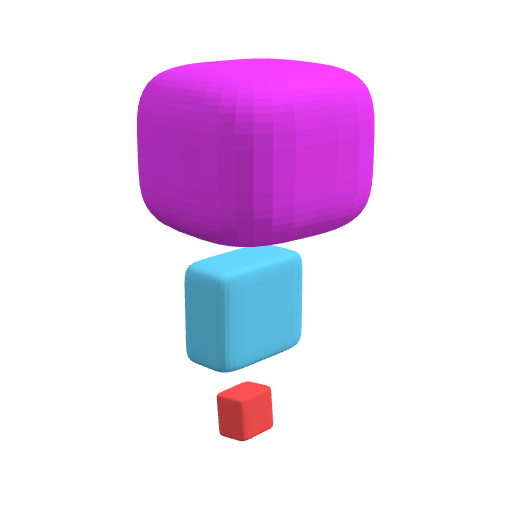}{3} &
\qIn{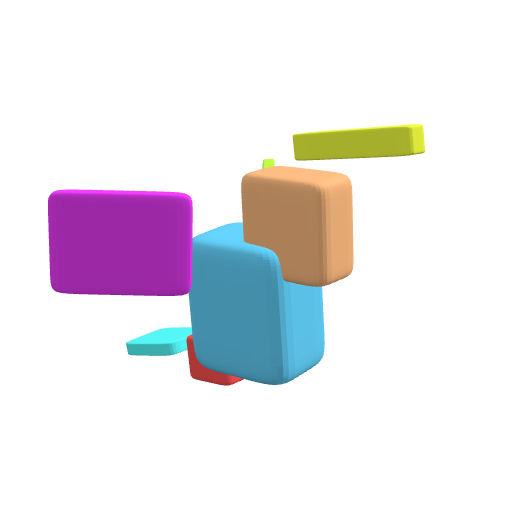}{7} &
\qIn{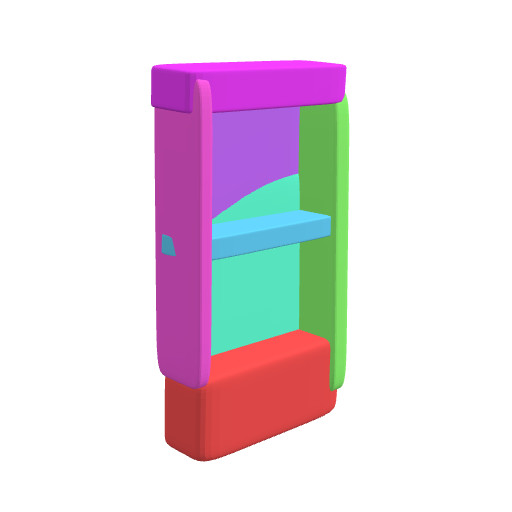}{7} &
\qIn{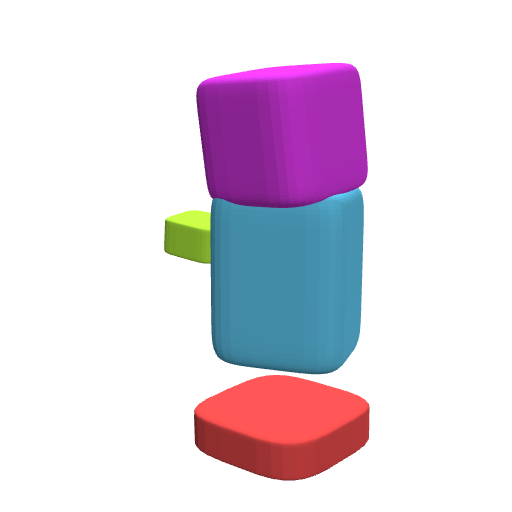}{4} \\
\qlab{SuperFlex} &
\qIn{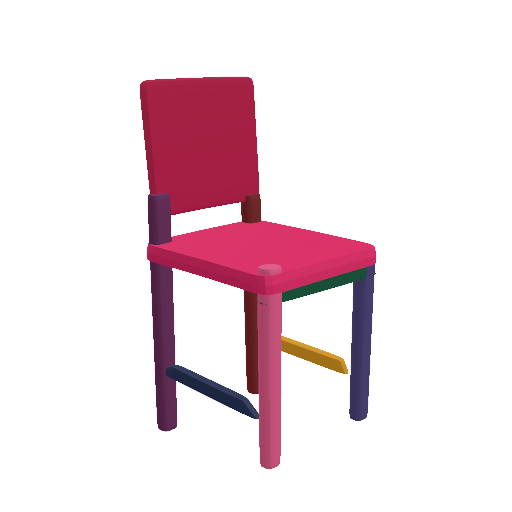}{10} &
\qIn{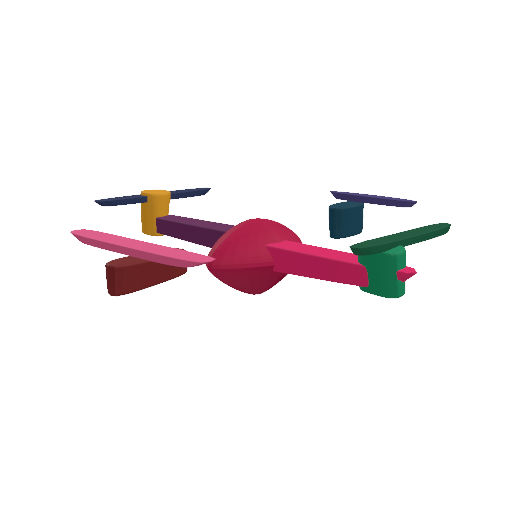}{11} &
\qIn{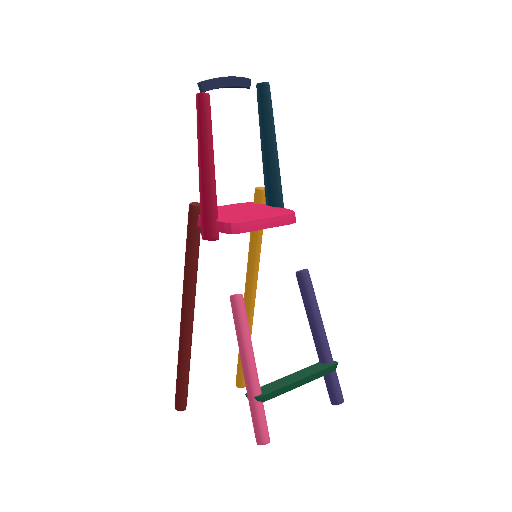}{10} &
\qIzn{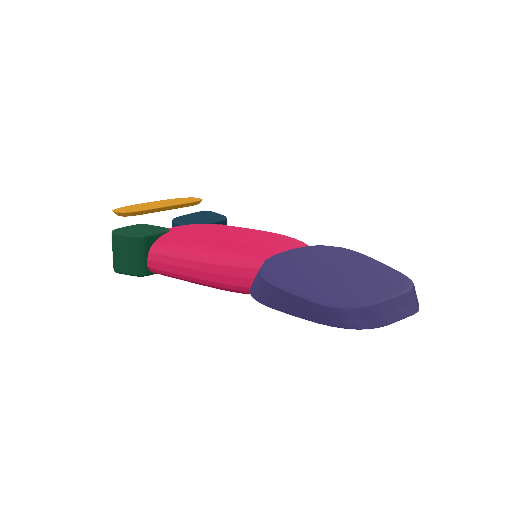}{5} &
\qIzn{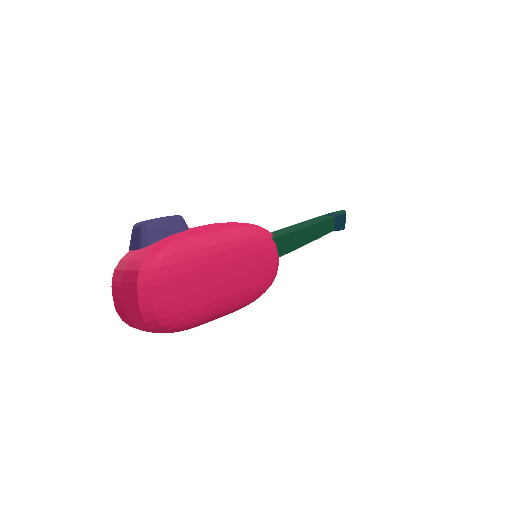}{4} &
\qIzn{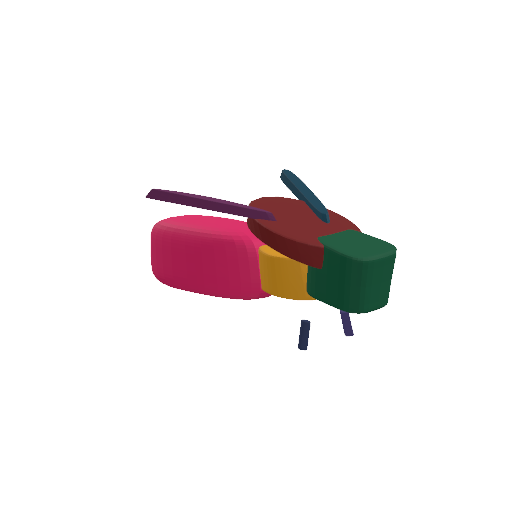}{8} &
\qIn{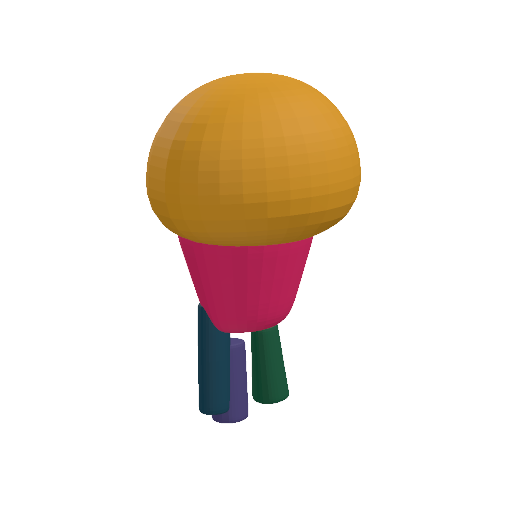}{5} &
\qIn{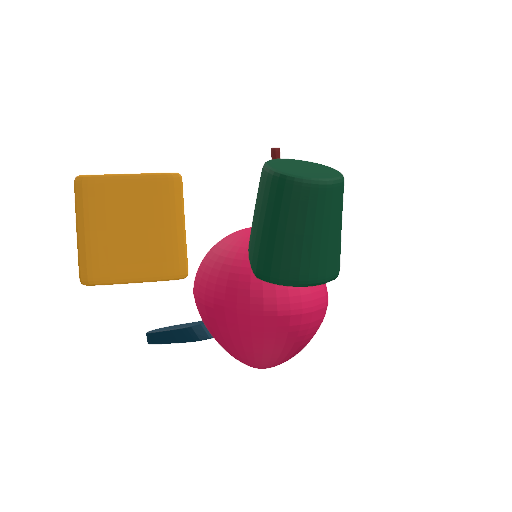}{6} &
\qIn{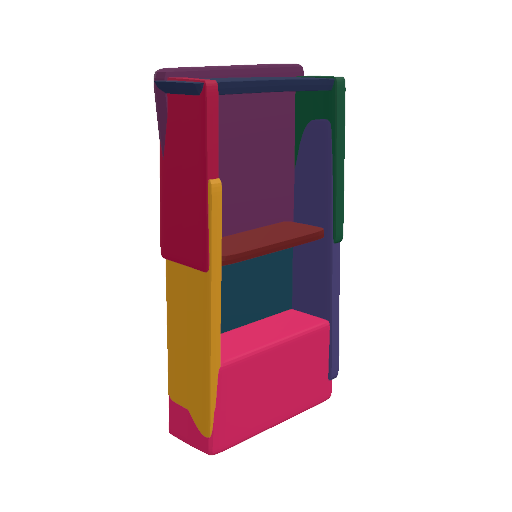}{9} &
\qIn{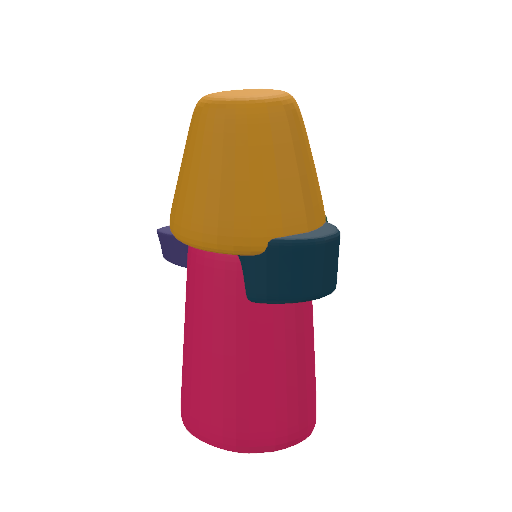}{5} \\
\midrule
\qlab{P3-SAM$^\ast$} &
\qIn{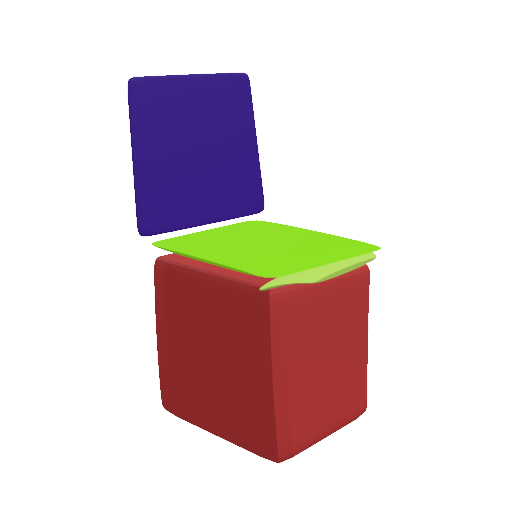}{5} &
\qIn{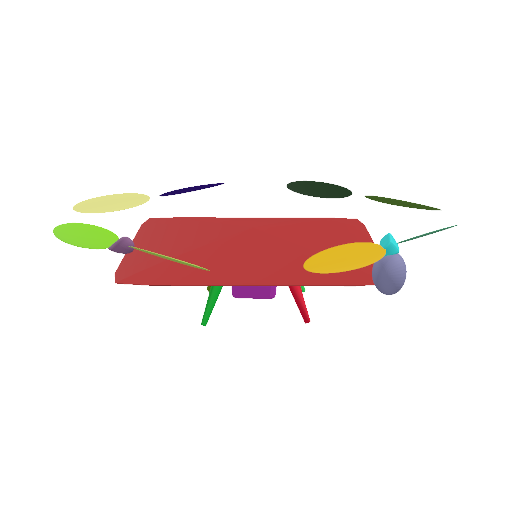}{17} &
\qIn{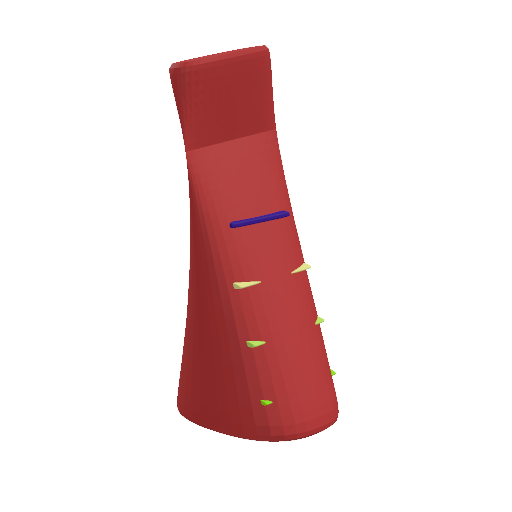}{5} &
\qIzn{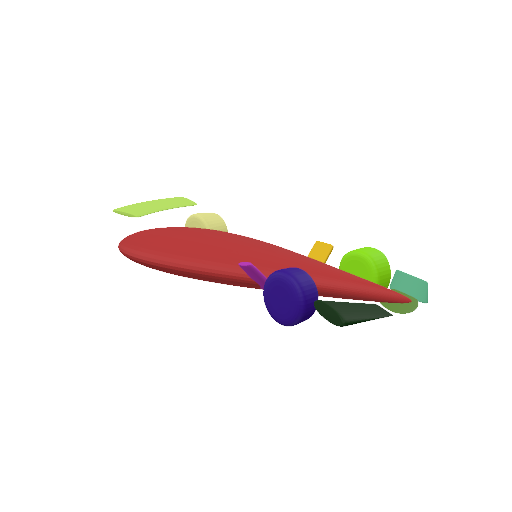}{10} &
\qIzn{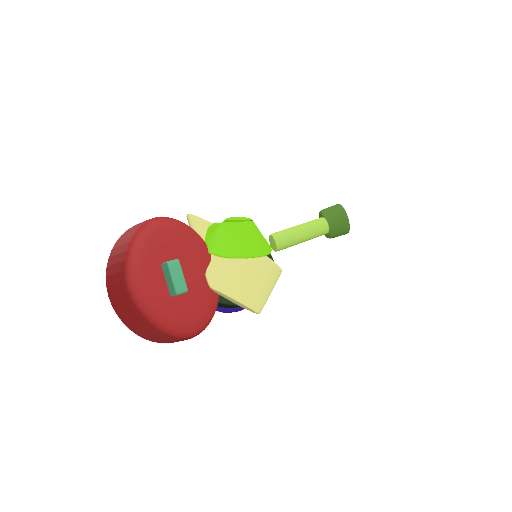}{8} &
\qIzn{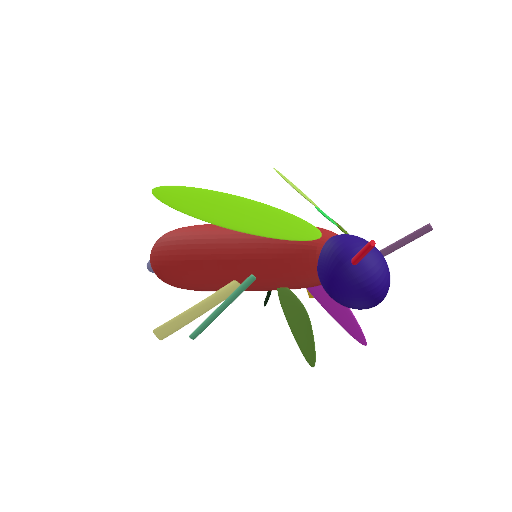}{15} &
\qIn{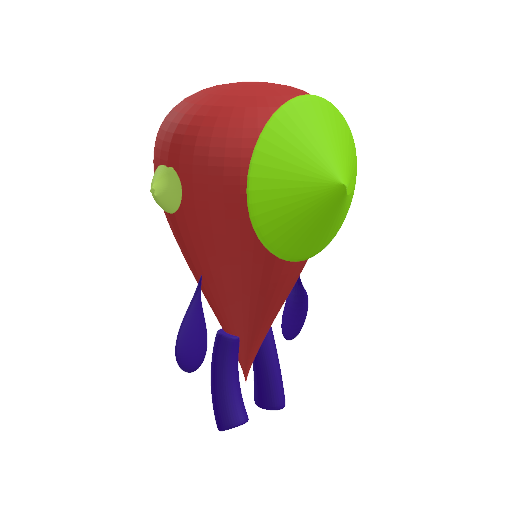}{7} &
\qIn{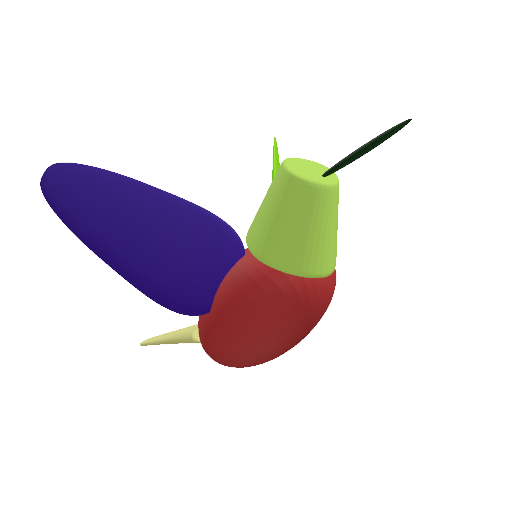}{6} &
\qIn{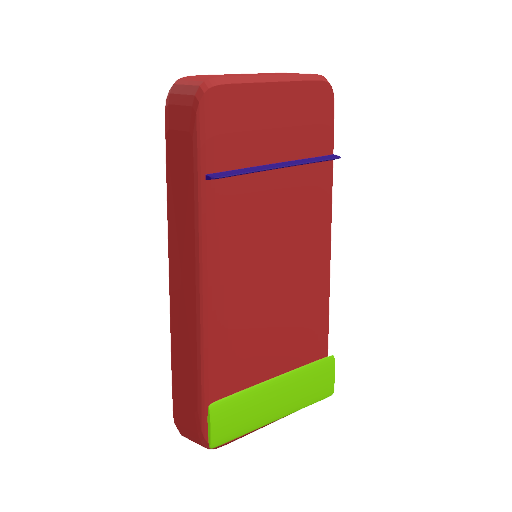}{3} &
\qIn{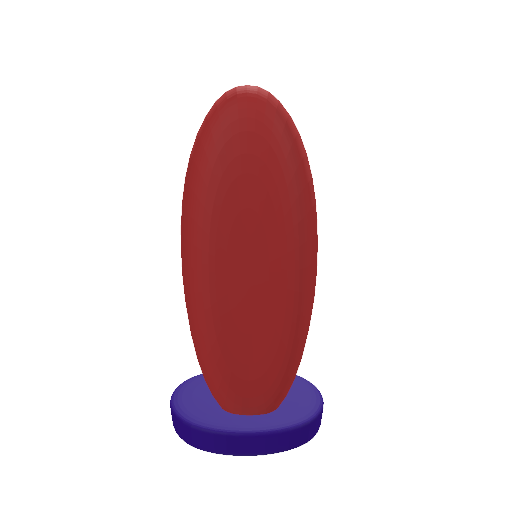}{2} \\
\qlab{PartField$^\ast$} &
\qIn{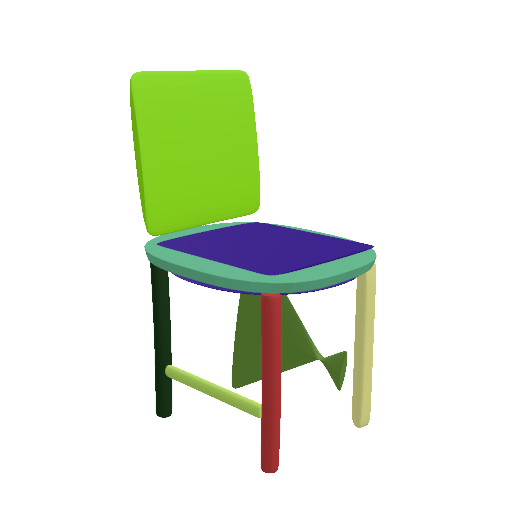}{9} &
\qIn{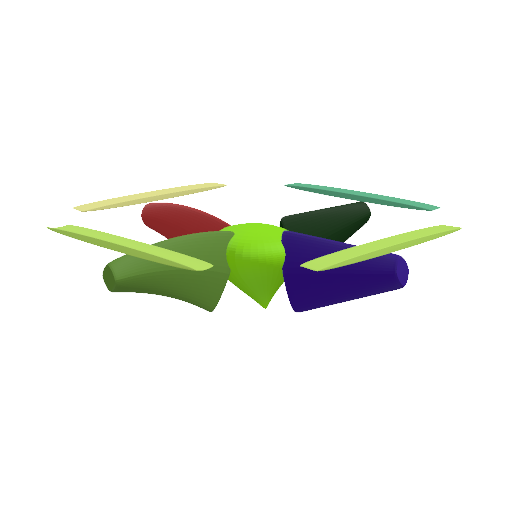}{9} &
\qIn{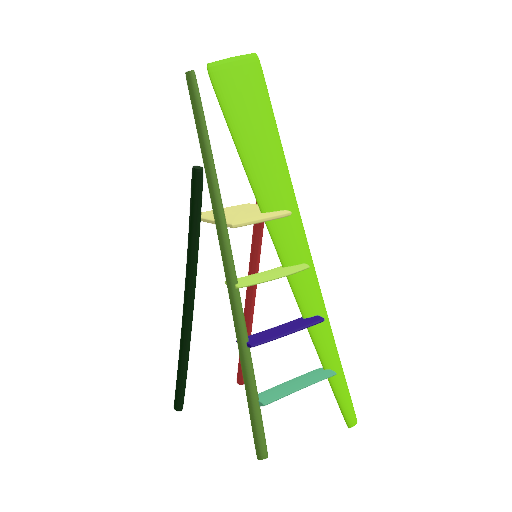}{8} &
\qIzn{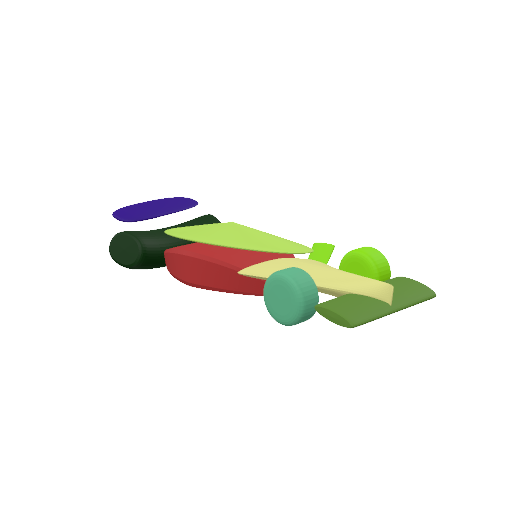}{9} &
\qIzn{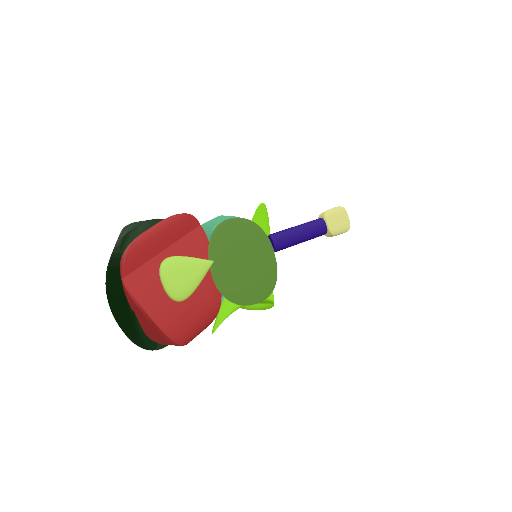}{8} &
\qIzn{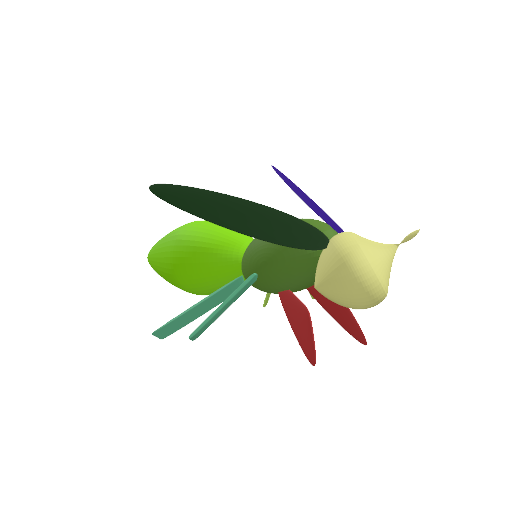}{11} &
\qIn{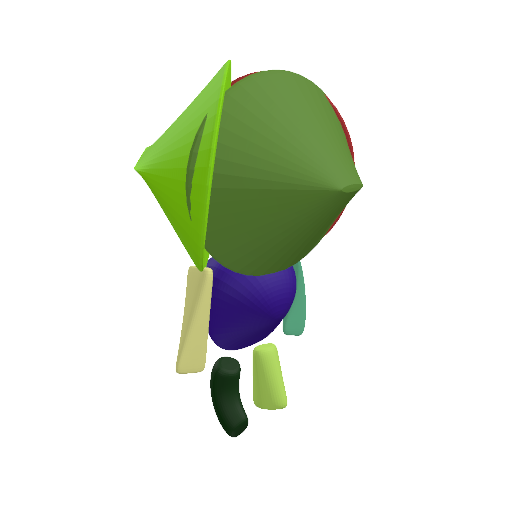}{8} &
\qIn{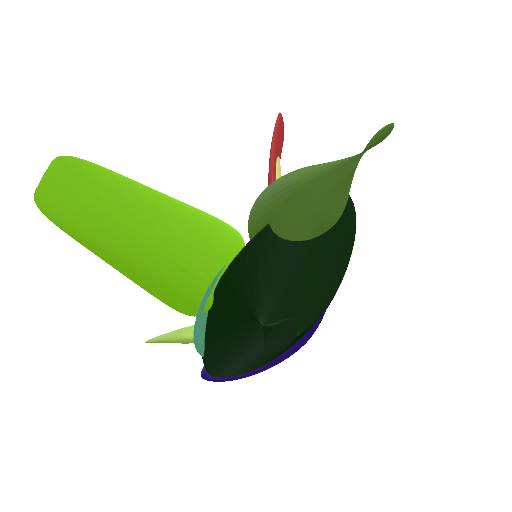}{8} &
\qIn{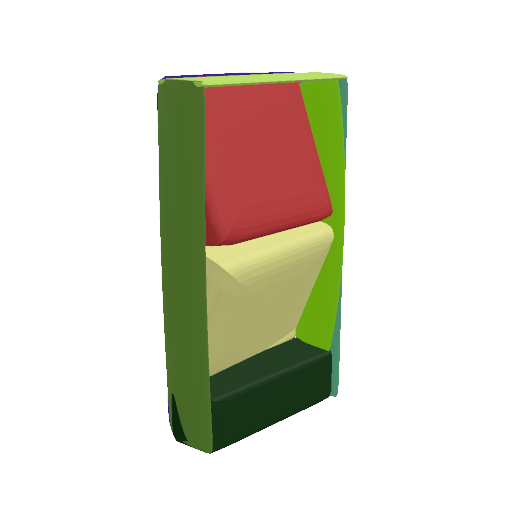}{8} &
\qIn{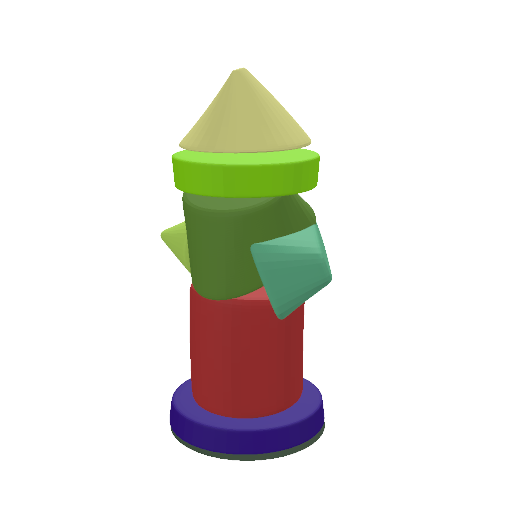}{8} \\
\qlab{PartSAM$^\ast$} &
\qIn{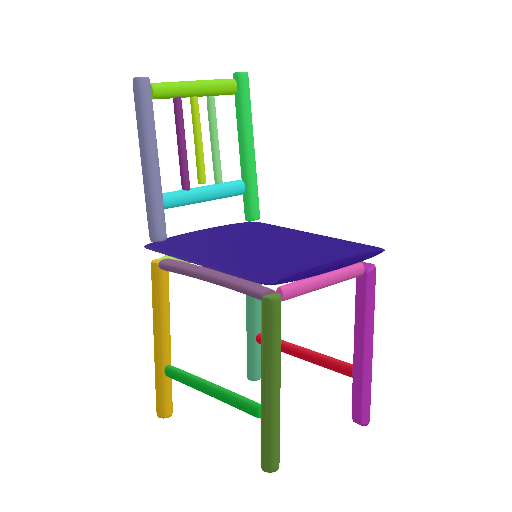}{19} &
\qIn{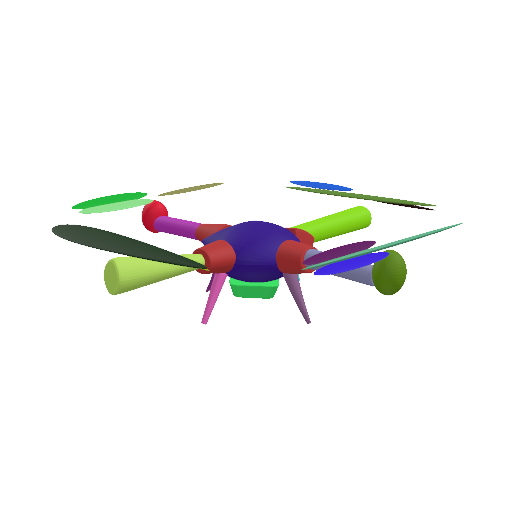}{27} &
\qIn{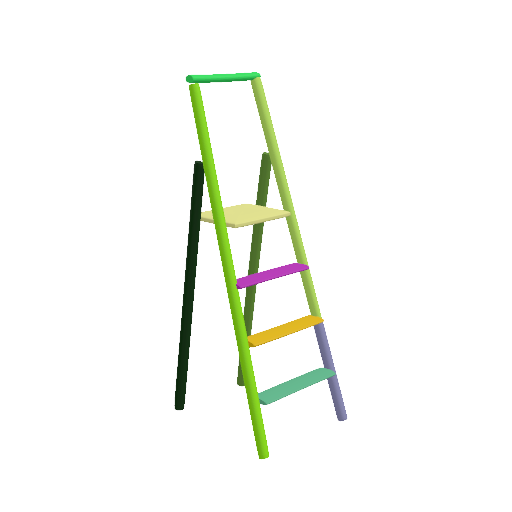}{10} &
\qIzn{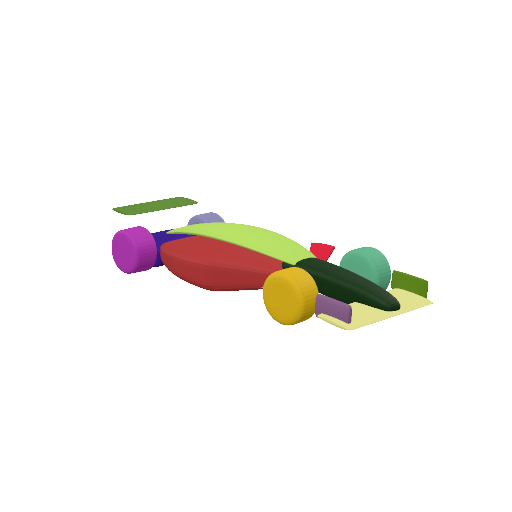}{14} &
\qIzn{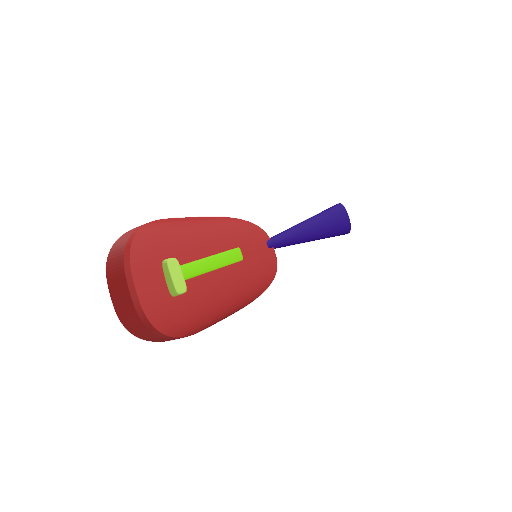}{4} &
\qIzn{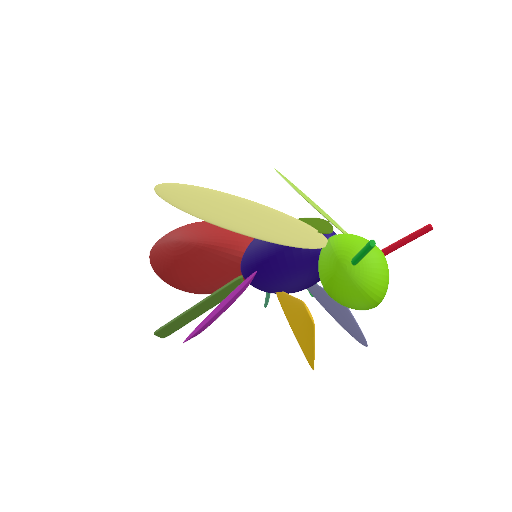}{14} &
\qIn{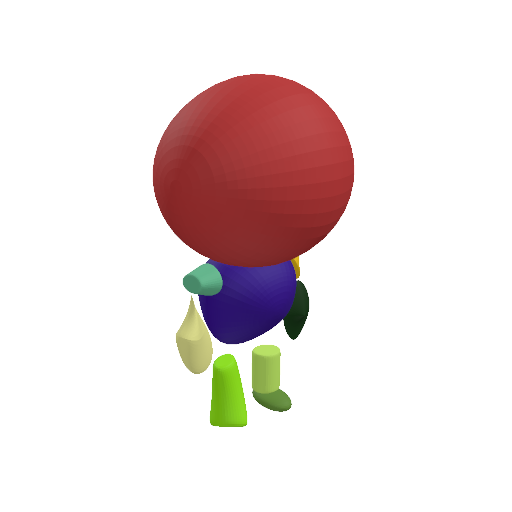}{9} &
\qIn{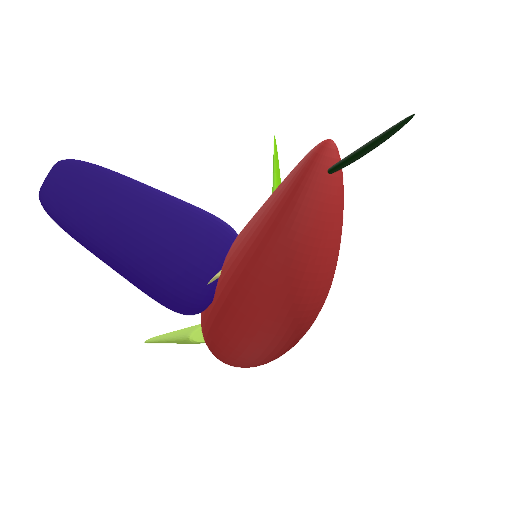}{6} &
\qIn{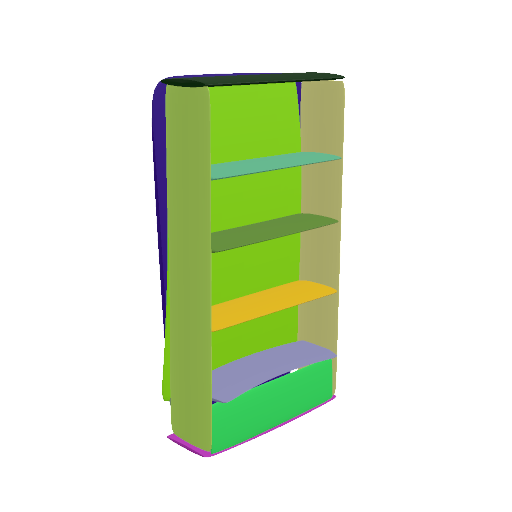}{11} &
\qIn{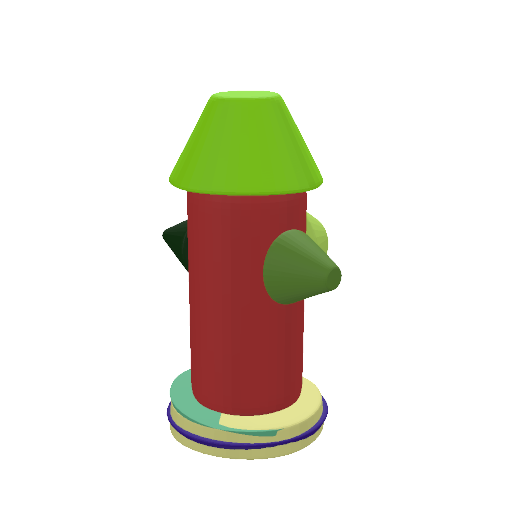}{8} \\
\midrule
\qlab{Ours} &
\qIn{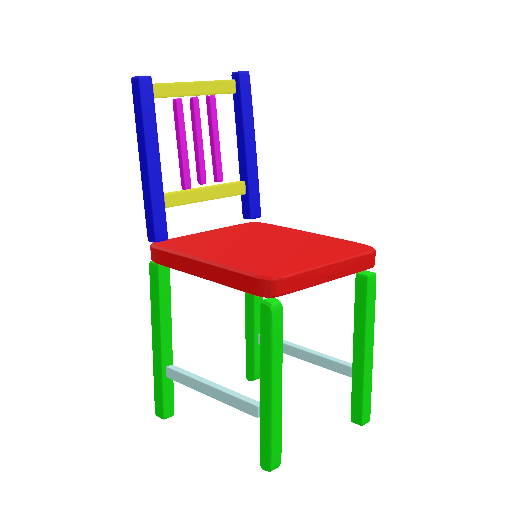}{14} &
\qIn{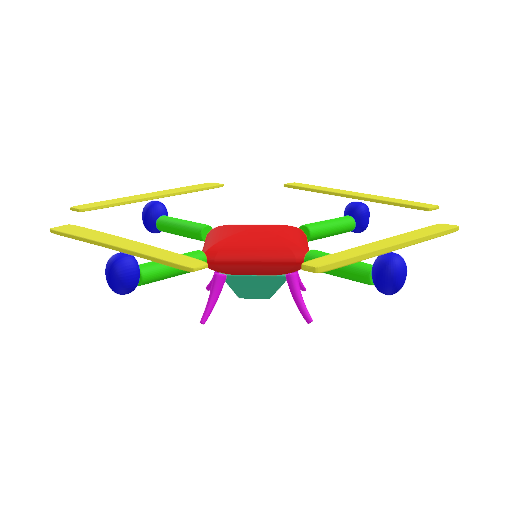}{18} &
\qIn{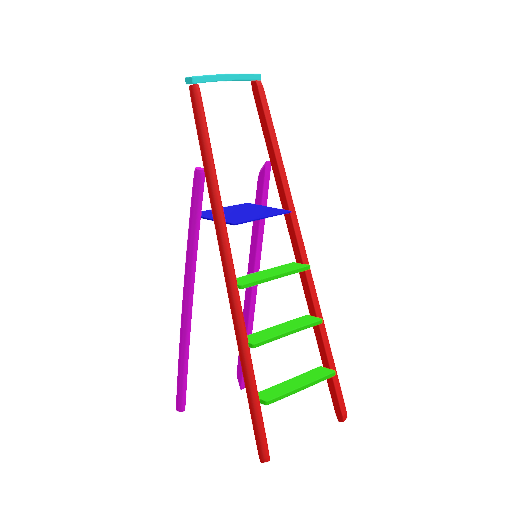}{9} &
\qIzn{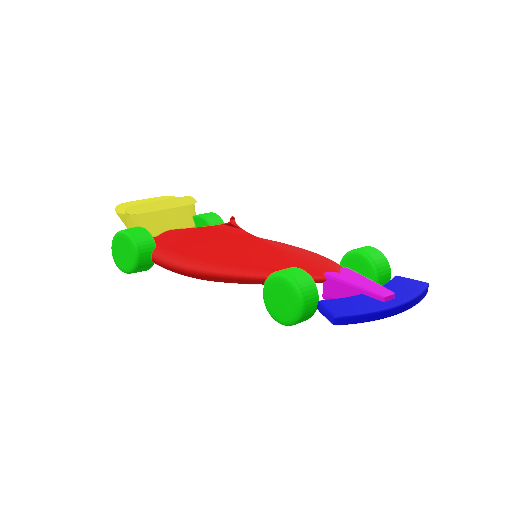}{8} &
\qIzn{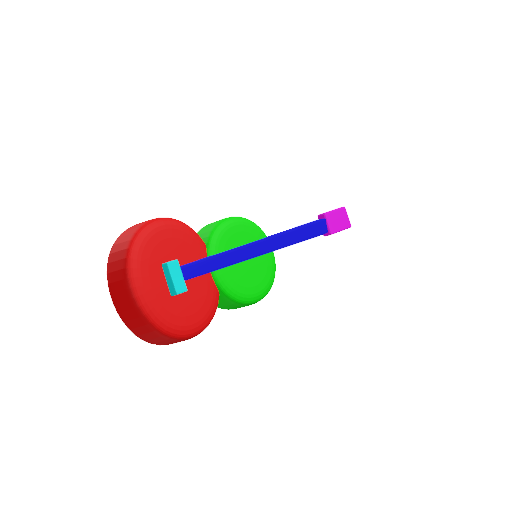}{5} &
\qIzn{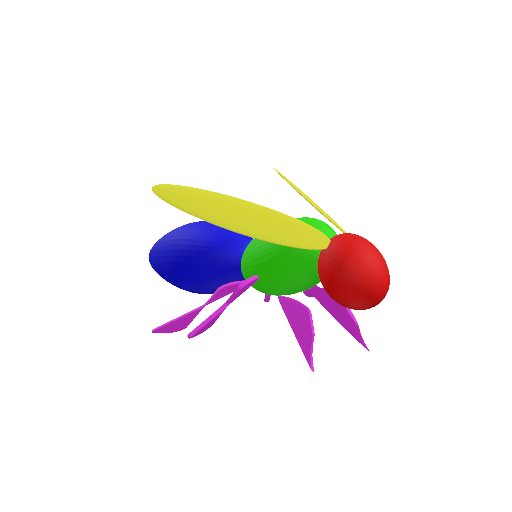}{11} &
\qIn{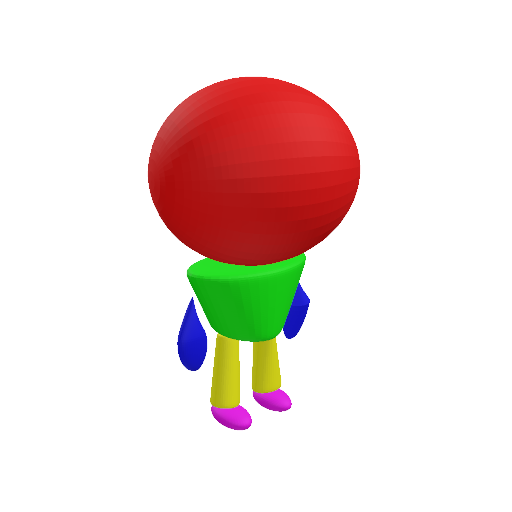}{8} &
\qIn{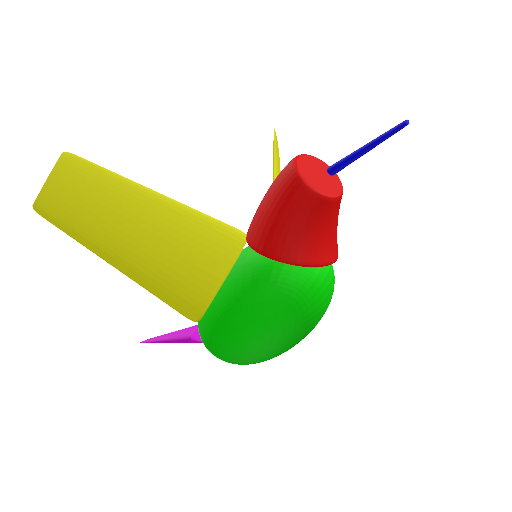}{6} &
\qIn{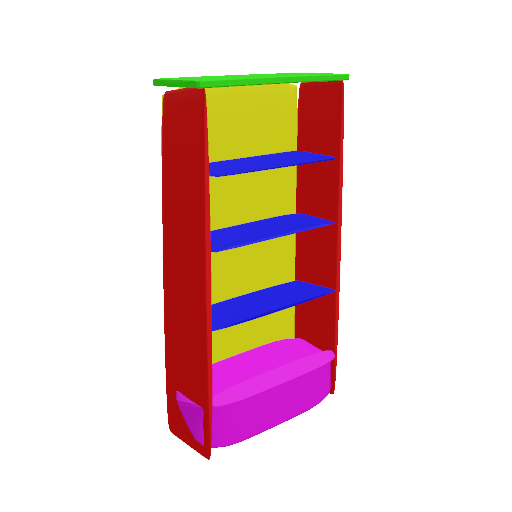}{8} &
\qIn{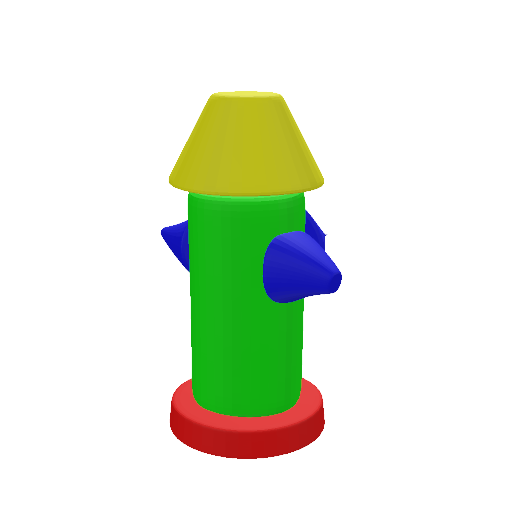}{6} \\
\qlab{GT} &
\qI{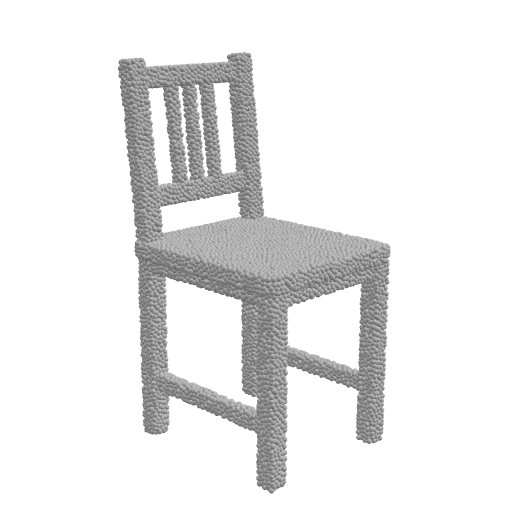} &
\qI{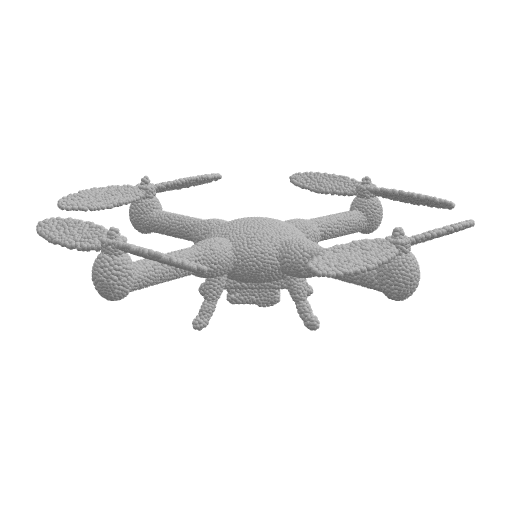} &
\qI{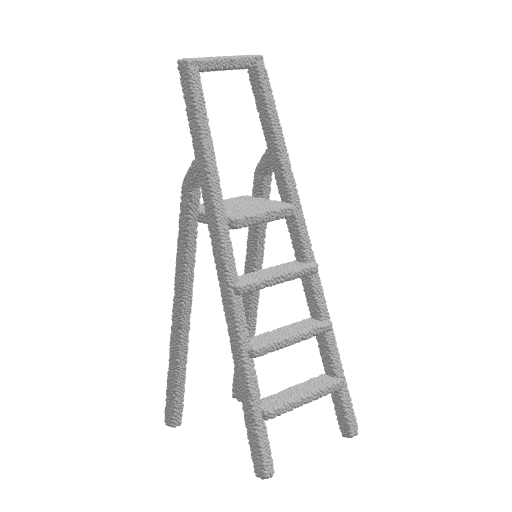} &
\qIz{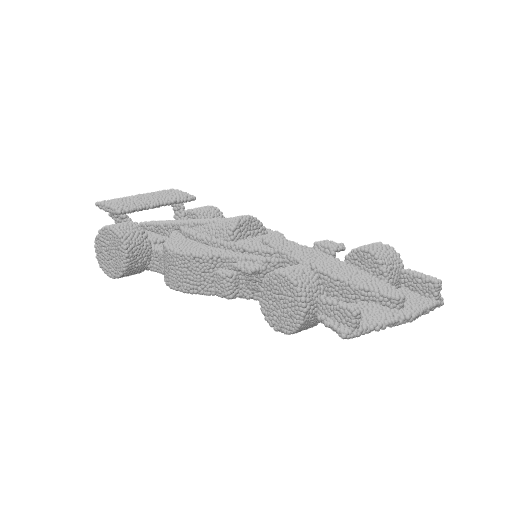} &
\qIz{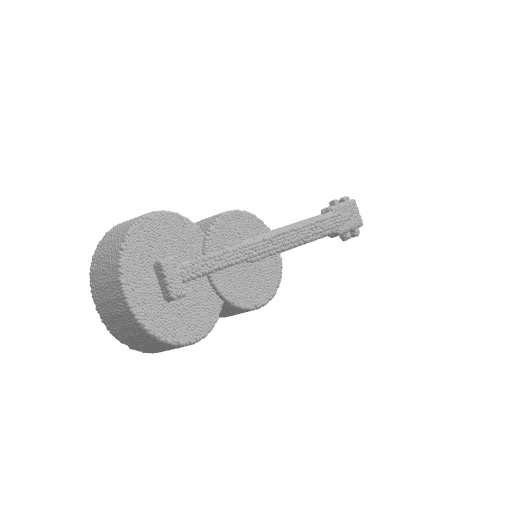} &
\qIz{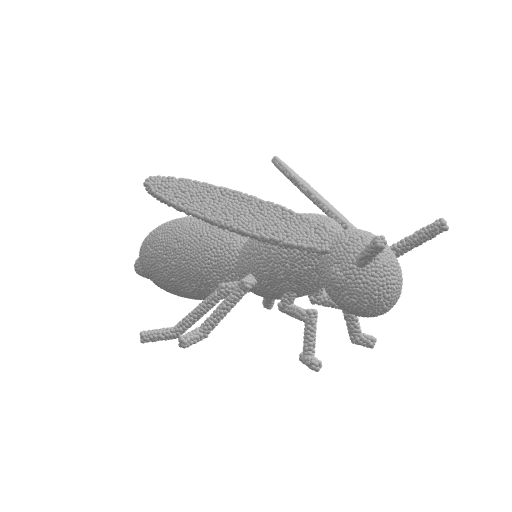} &
\qI{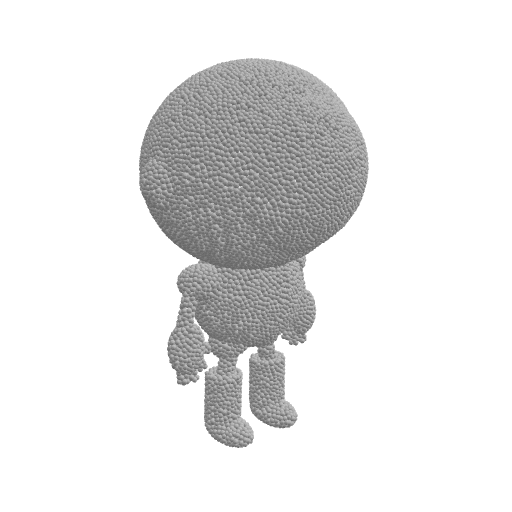} &
\qI{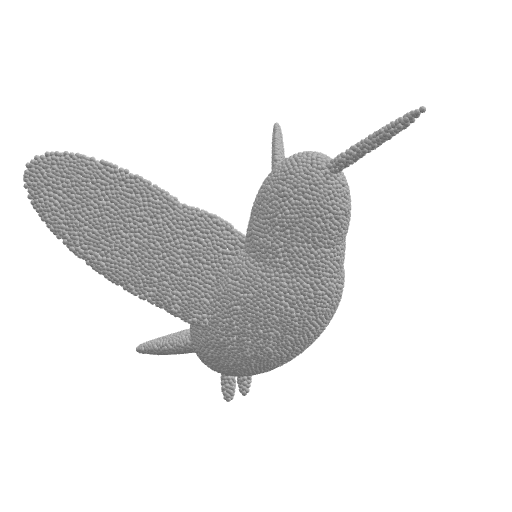} &
\qI{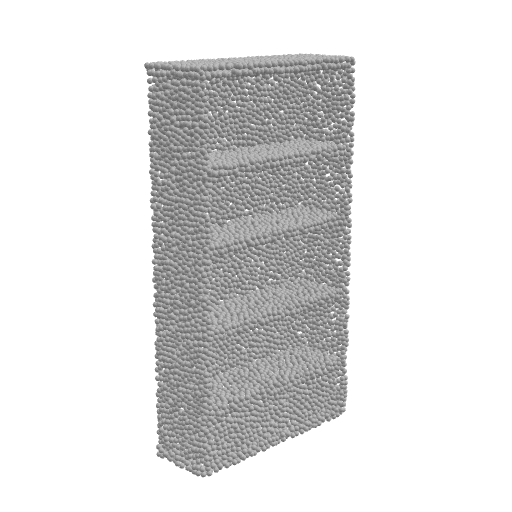} &
\qI{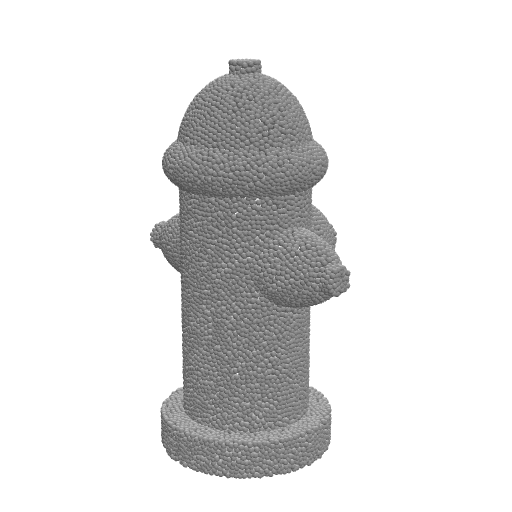} \\
\end{tabular}
\caption{\textbf{Qualitative comparison on HumanPrim.} One row per method and one column per object, with primitive counts in the lower right corners. Ours uses four views. Its 5--18 primitives retain components such as chair backrest bars, ladder rungs and car wheels. Primitive Anything and Light-SQ often use more primitives on these examples, while EMS merges several structures into bulkier elements. The compact learned baselines retain different levels of part detail. $^\ast$Part labels of that segmenter fitted with our clustering and fitter (\Cref{sec:seg_source}).}
\label{fig:qual_humanprim}
\end{figure*}

\begin{figure*}[p]
\centering
\setlength{\qw}{0.096\textwidth}
\setlength{\tabcolsep}{0.5pt}
\renewcommand{\arraystretch}{0.3}
\begin{tabular}{c@{\hspace{2pt}} cccccccccc}
\qlab{EMS} &
\qIn{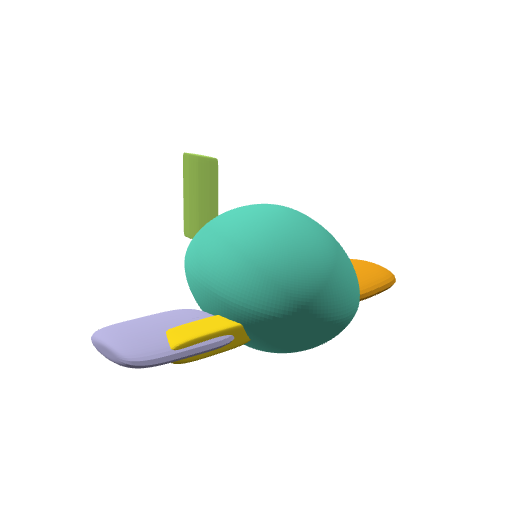}{6} &
\qIn{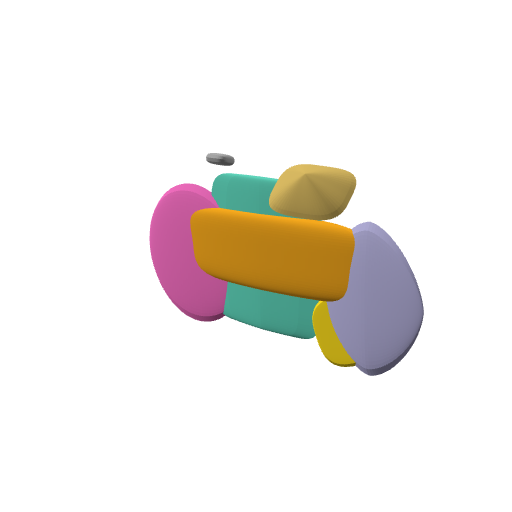}{9} &
\qIn{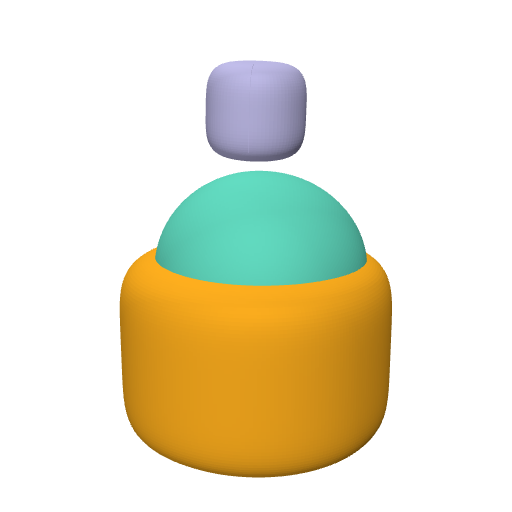}{3} &
\qIn{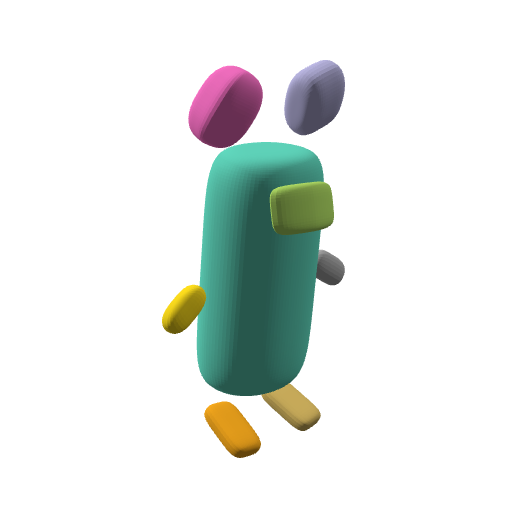}{9} &
\qIn{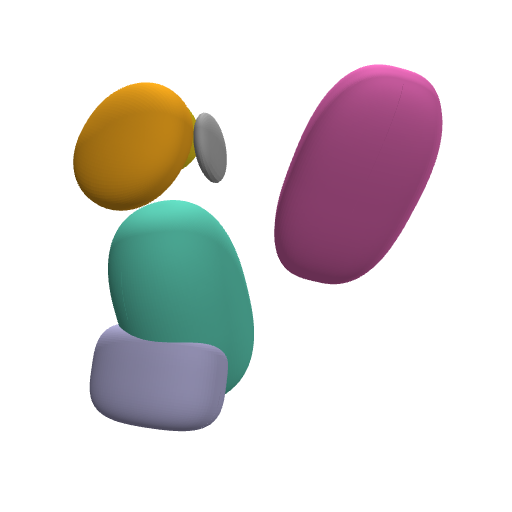}{8} &
\qIn{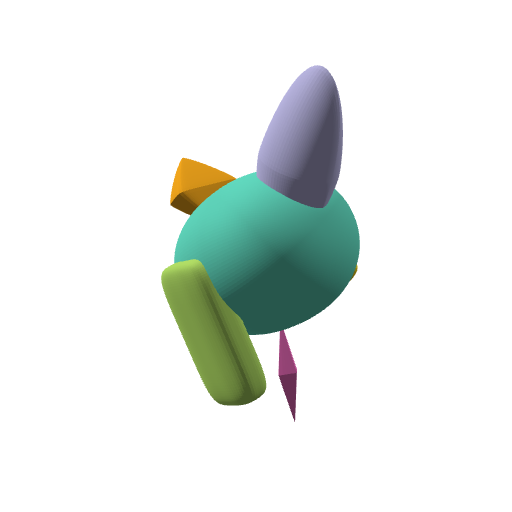}{6} &
\qIn{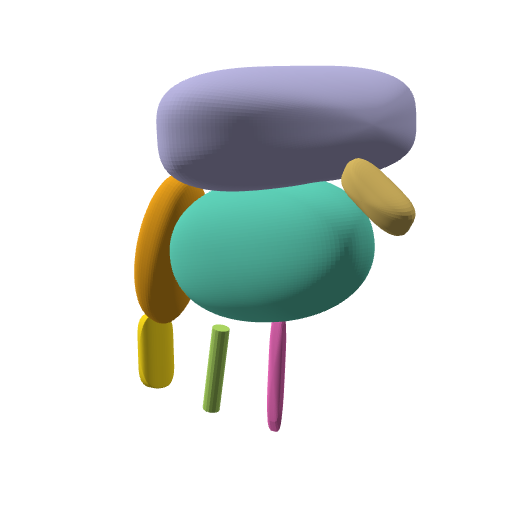}{7} &
\qIzn{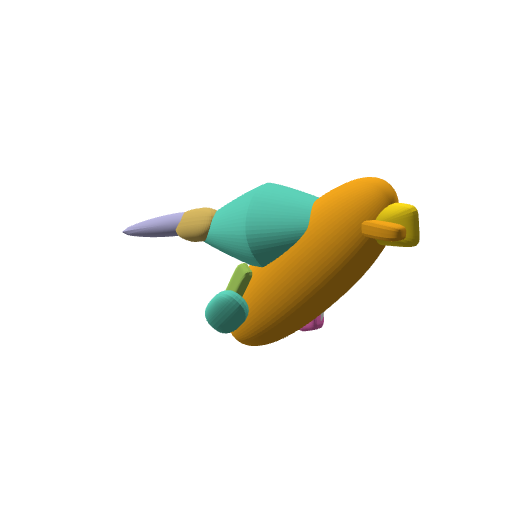}{10} &
\qIn{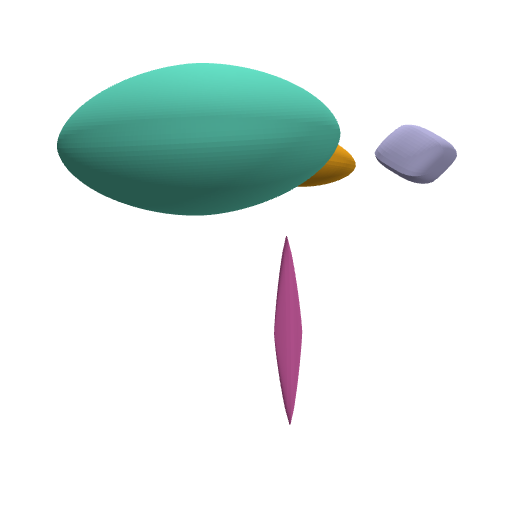}{5} &
\qIn{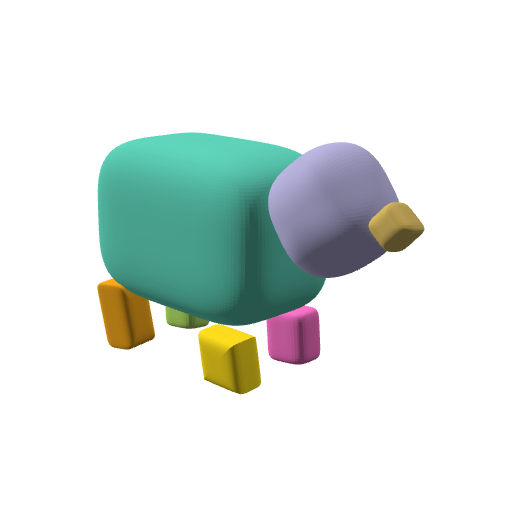}{7} \\
\qlab{Light-SQ} &
\qIn{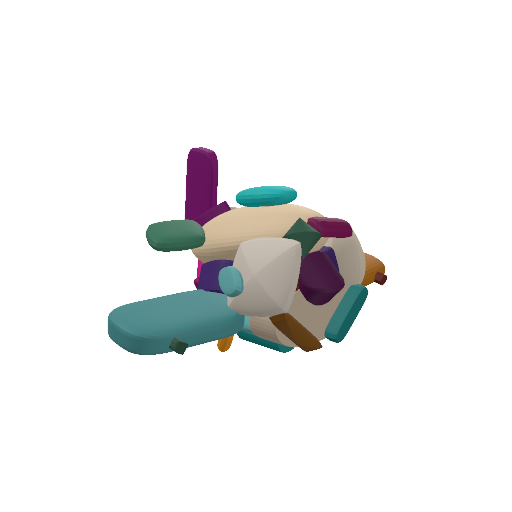}{36} &
\qIn{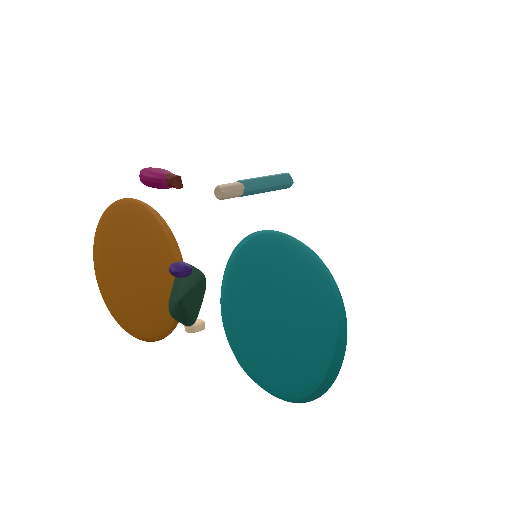}{9} &
\qIn{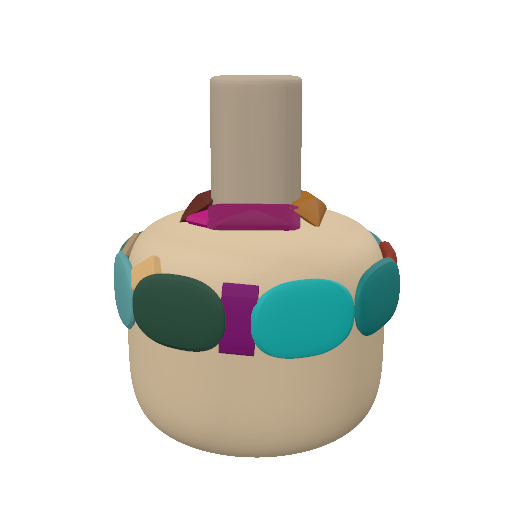}{22} &
\qIn{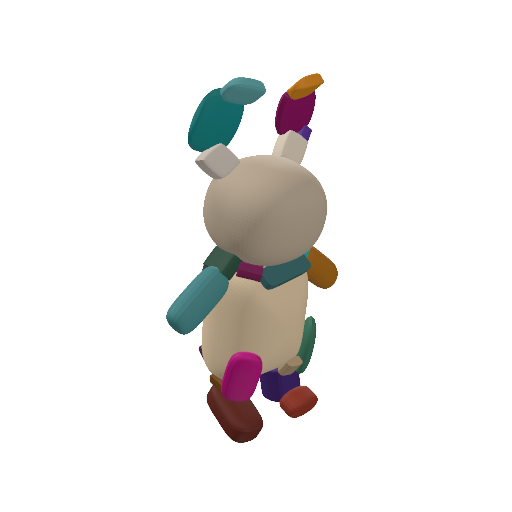}{27} &
\qIn{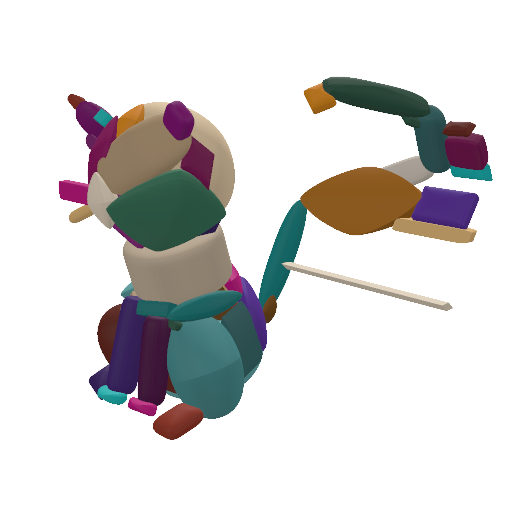}{69} &
\qIn{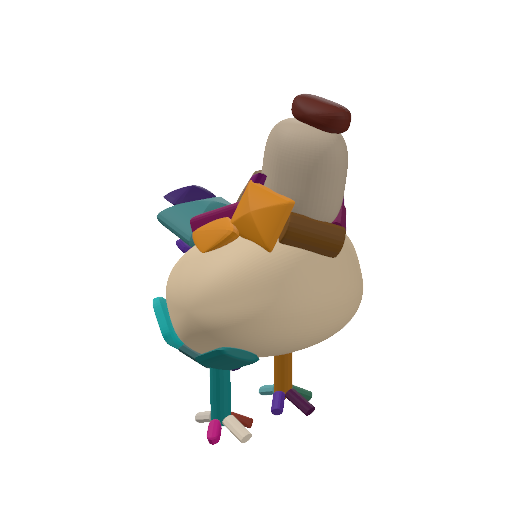}{41} &
\qIn{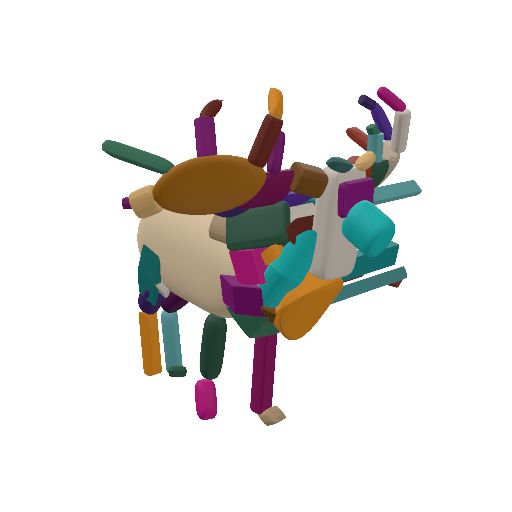}{74} &
\qIzn{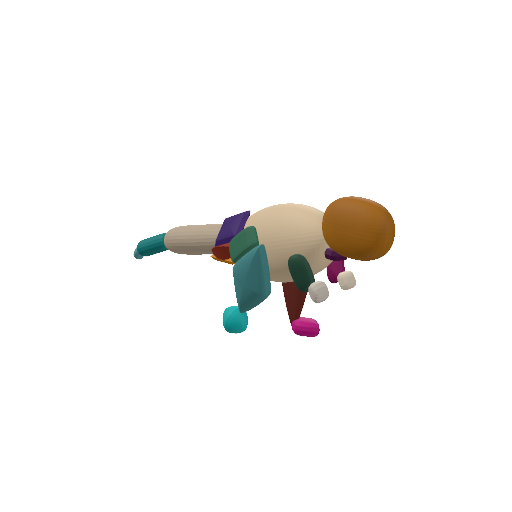}{20} &
\qIn{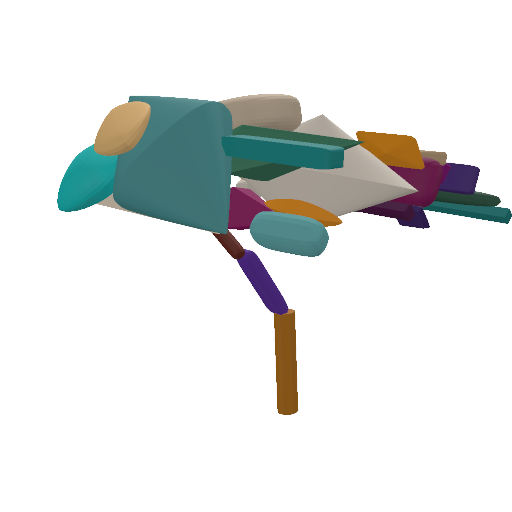}{34} &
\qIn{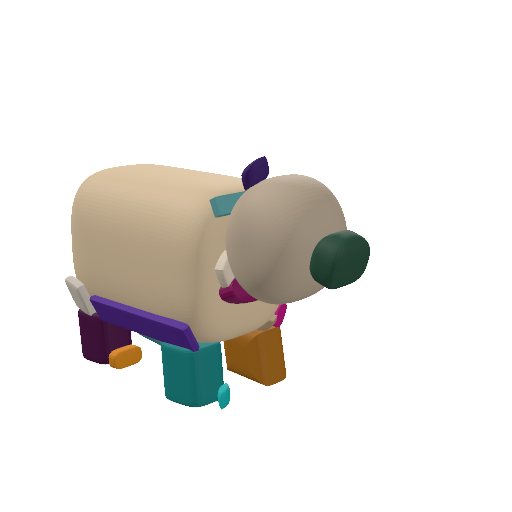}{22} \\
\midrule
\qlab{PrimAny} &
\qIn{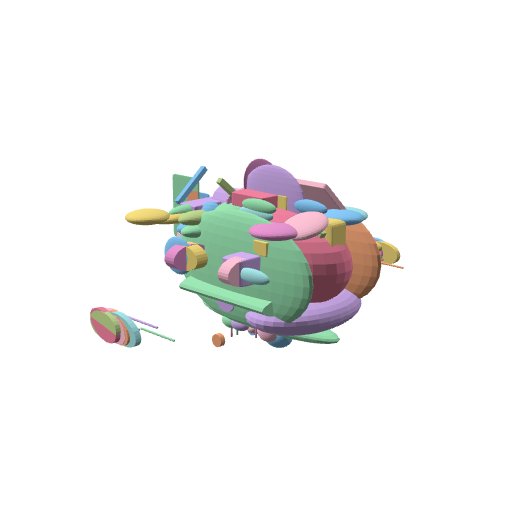}{300} &
\qIn{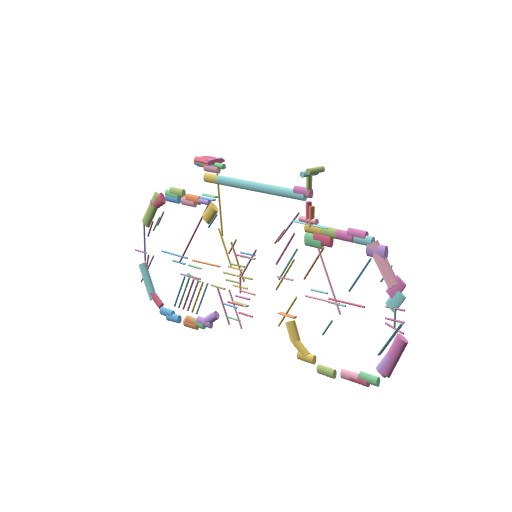}{408} &
\qIn{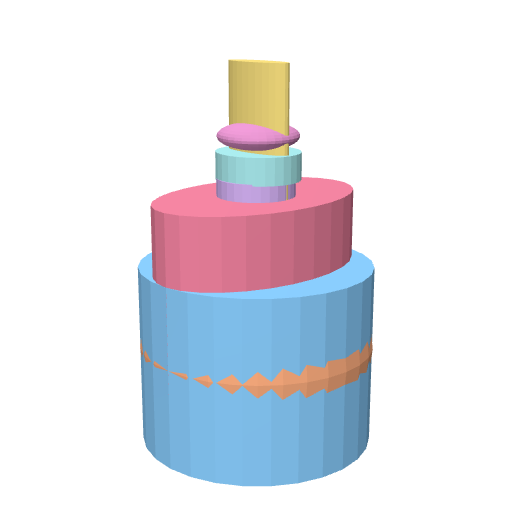}{18} &
\qIn{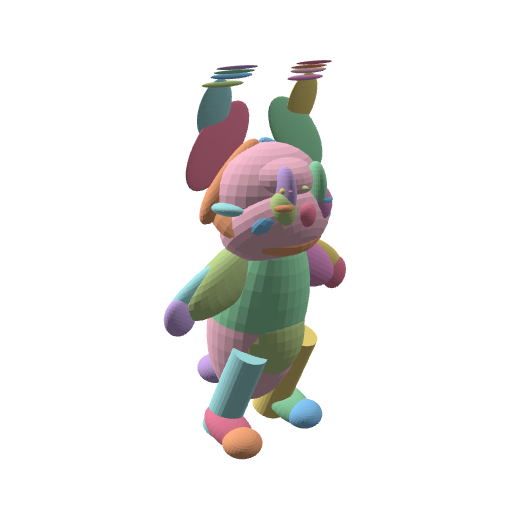}{54} &
\qIn{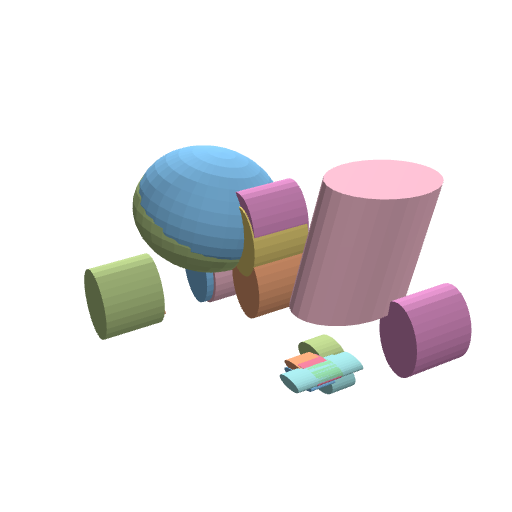}{428} &
\qIn{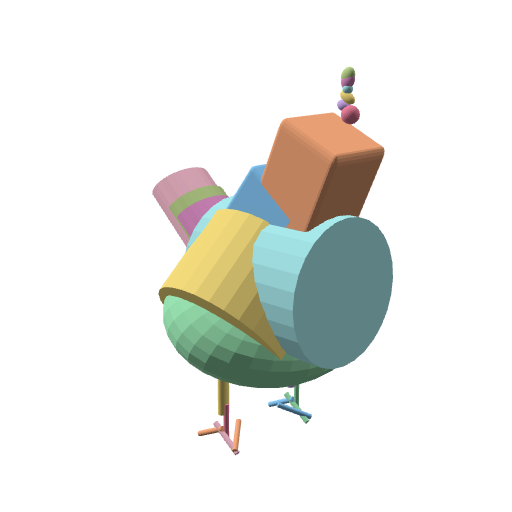}{62} &
\qIn{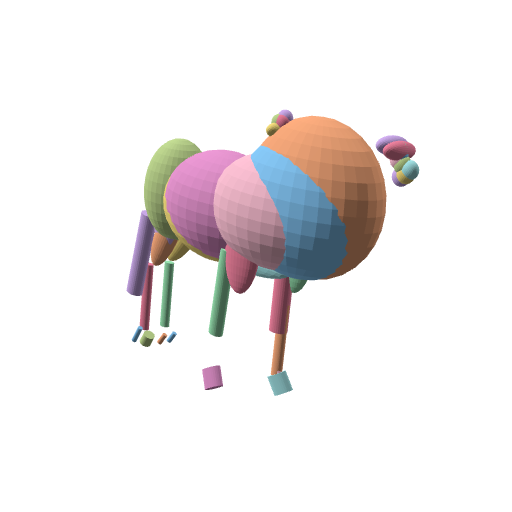}{170} &
\qIzn{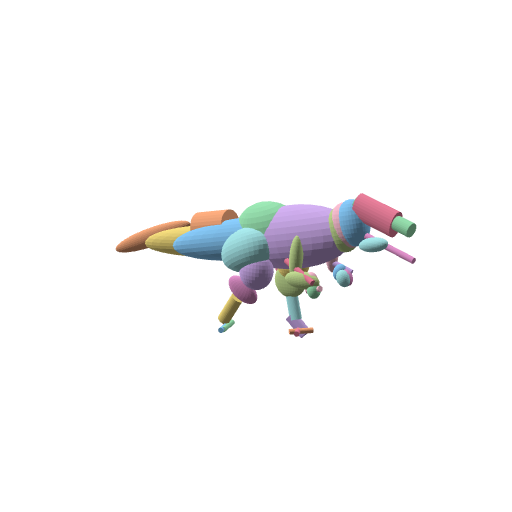}{68} &
\qIn{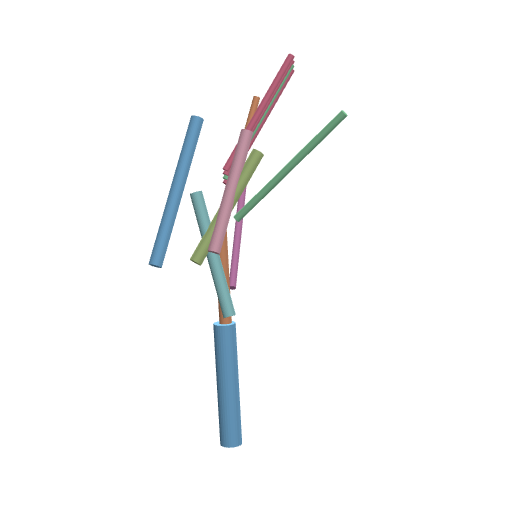}{432} &
\qIn{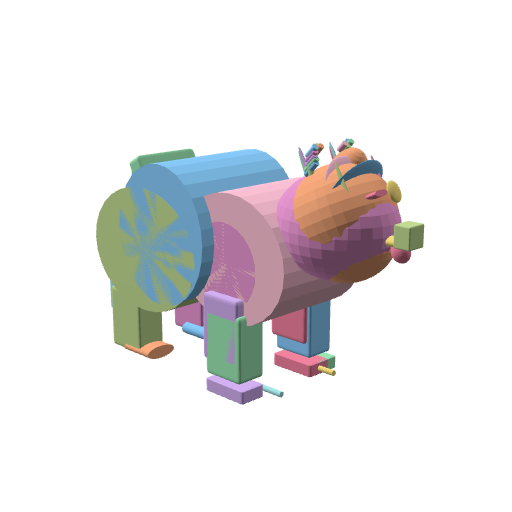}{195} \\
\qlab{F2C} &
\qIn{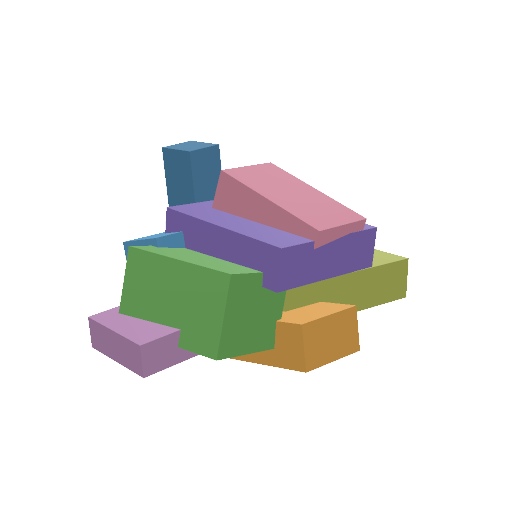}{9} &
\qIn{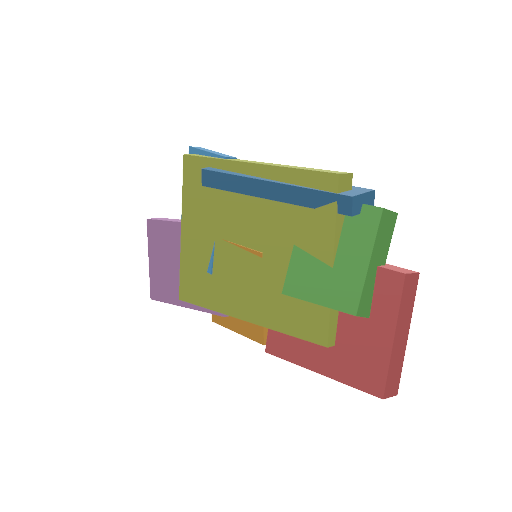}{7} &
\qIn{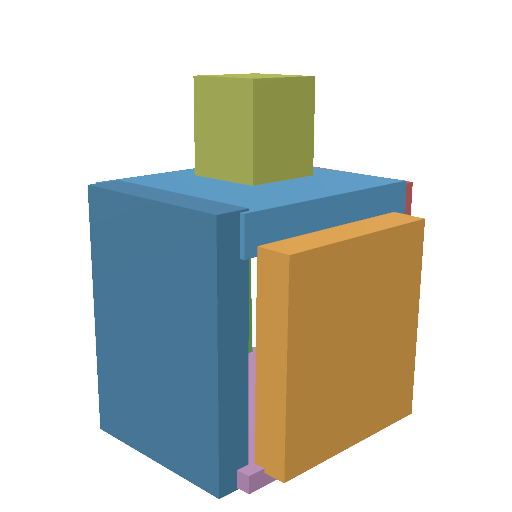}{7} &
\qIn{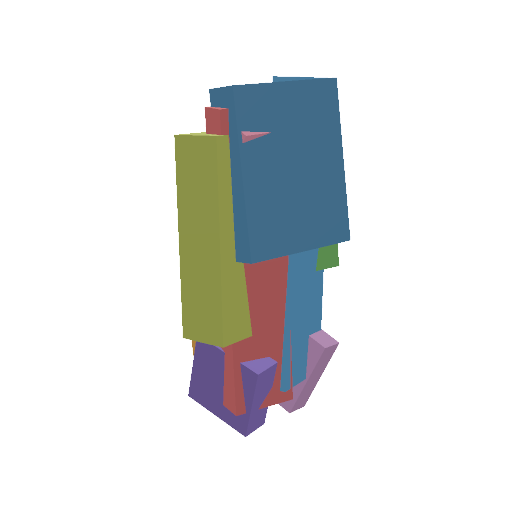}{10} &
\qIn{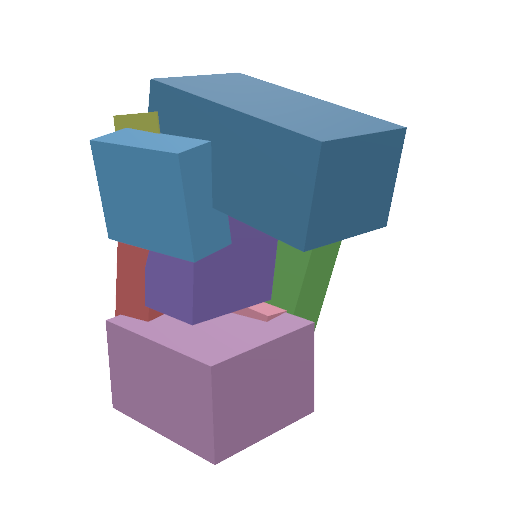}{9} &
\qIn{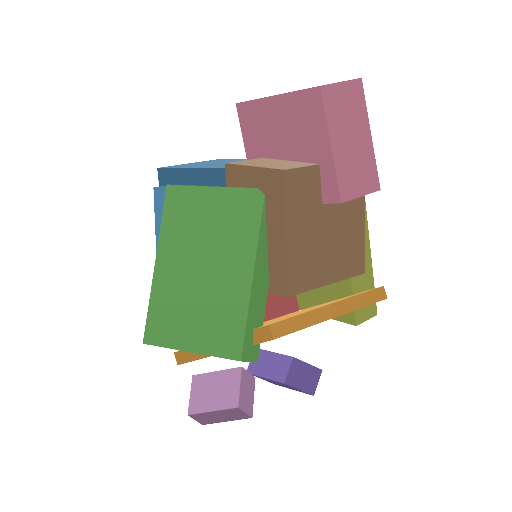}{10} &
\qIn{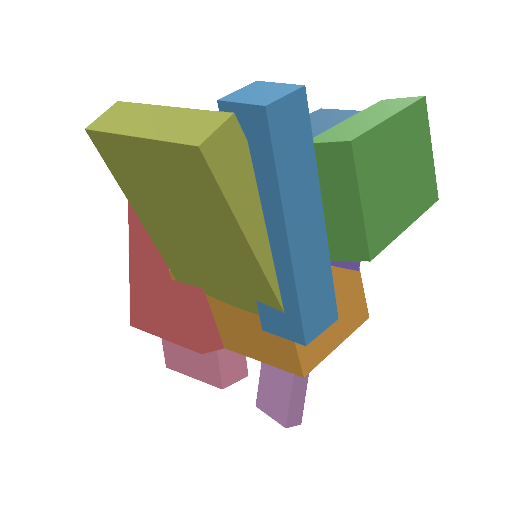}{9} &
\qIzn{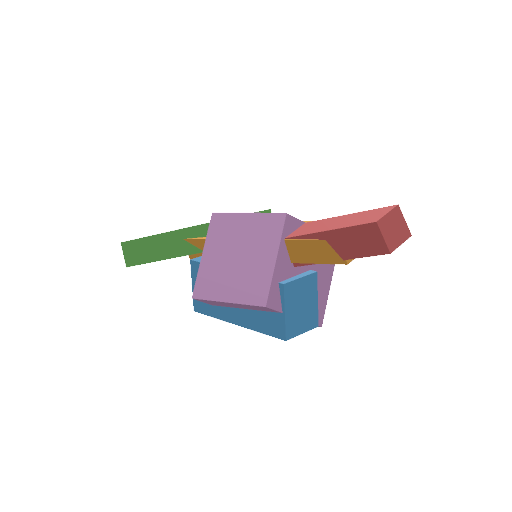}{5} &
\qIn{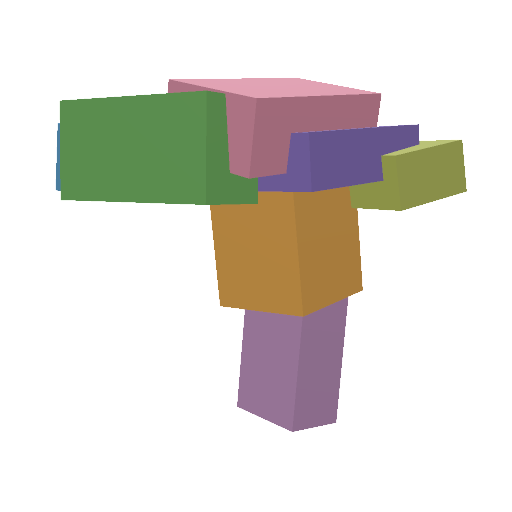}{9} &
\qIn{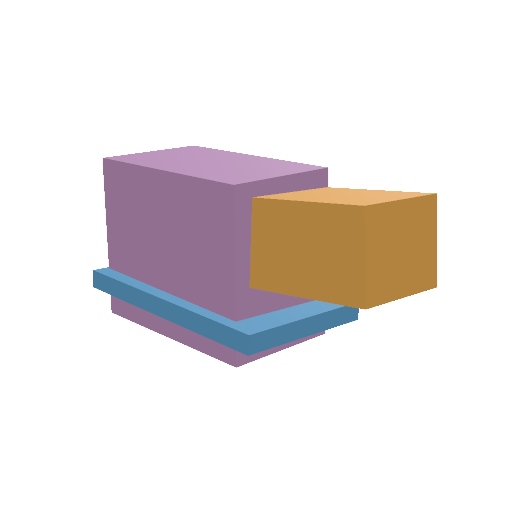}{4} \\
\qlab{SuperDec} &
\qIn{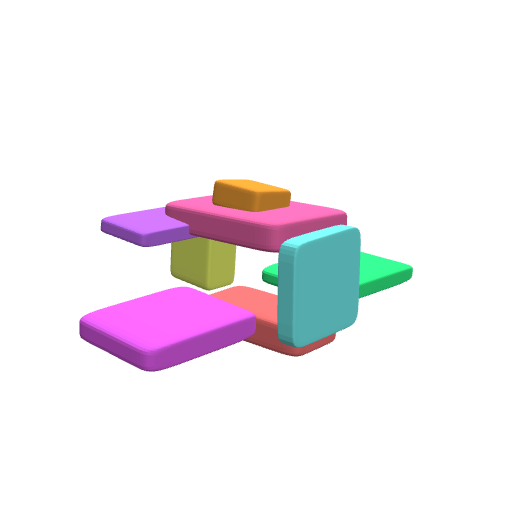}{8} &
\qIn{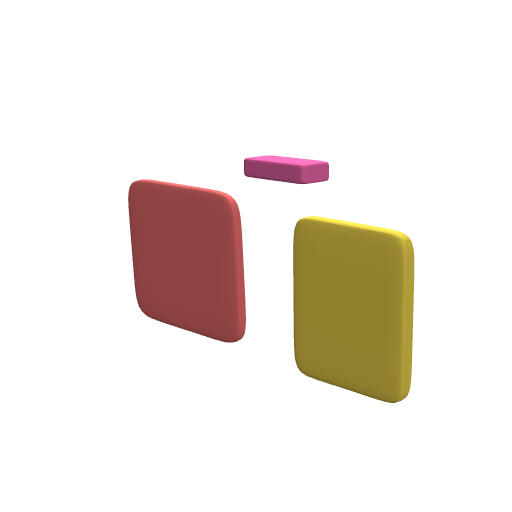}{3} &
\qIn{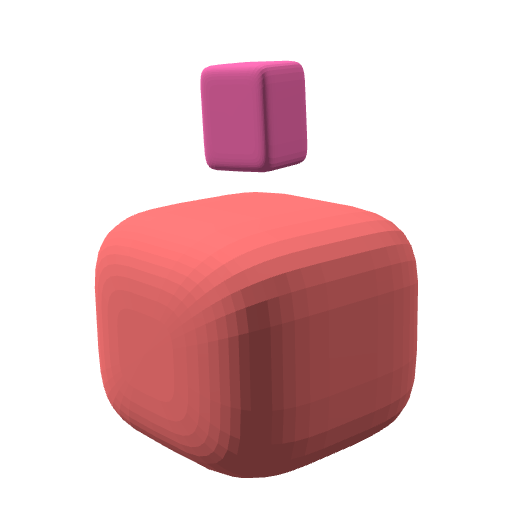}{2} &
\qIn{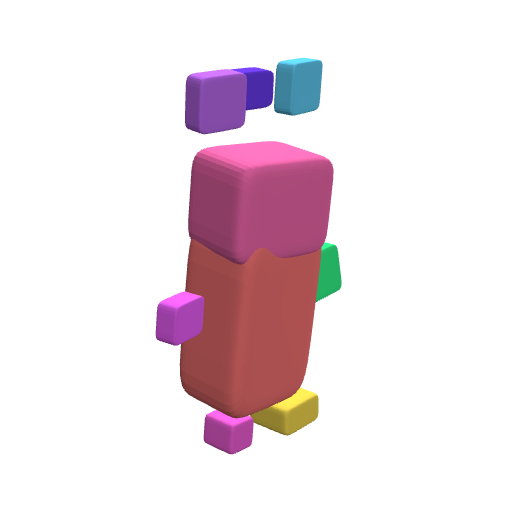}{9} &
\qIn{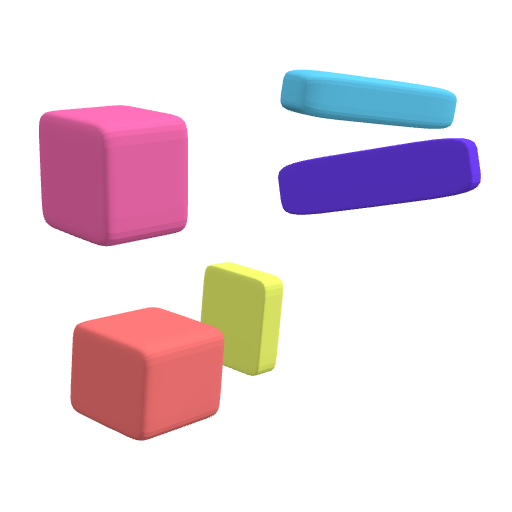}{5} &
\qIn{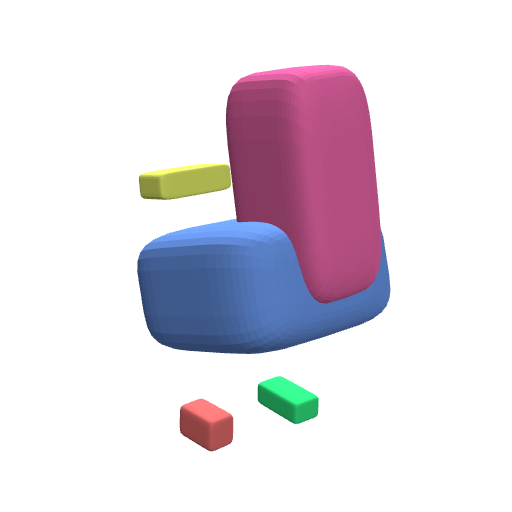}{5} &
\qIn{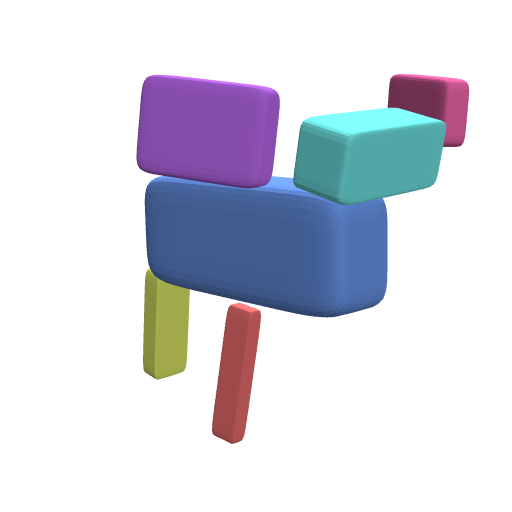}{6} &
\qIzn{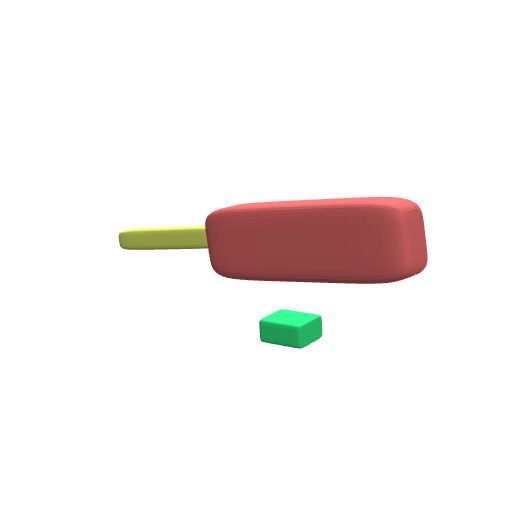}{3} &
\qIn{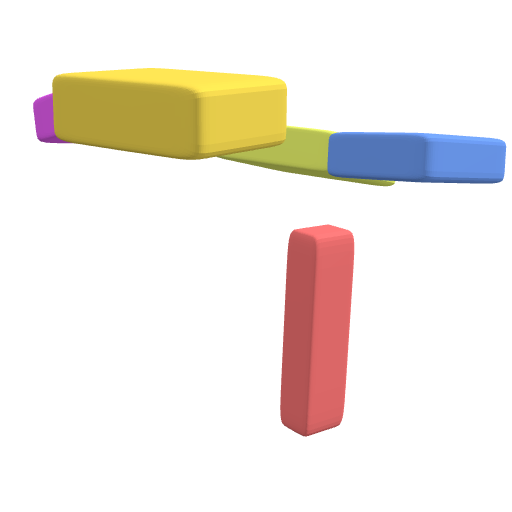}{5} &
\qIn{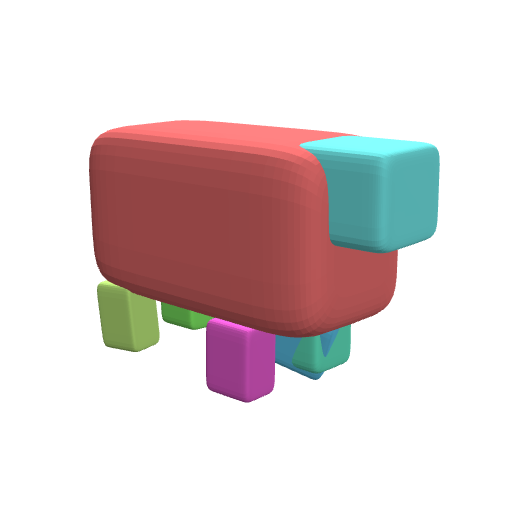}{8} \\
\qlab{SuperFlex} &
\qIn{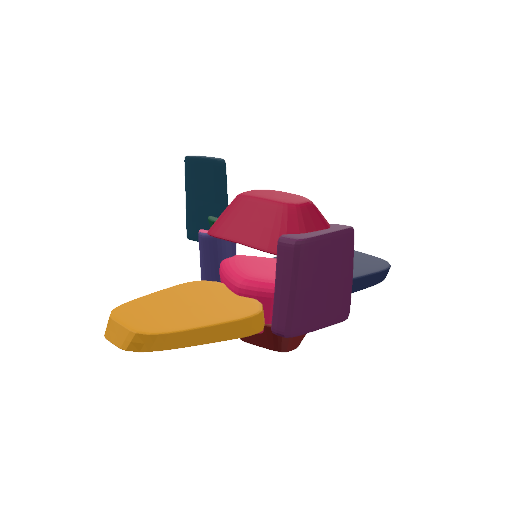}{10} &
\qIn{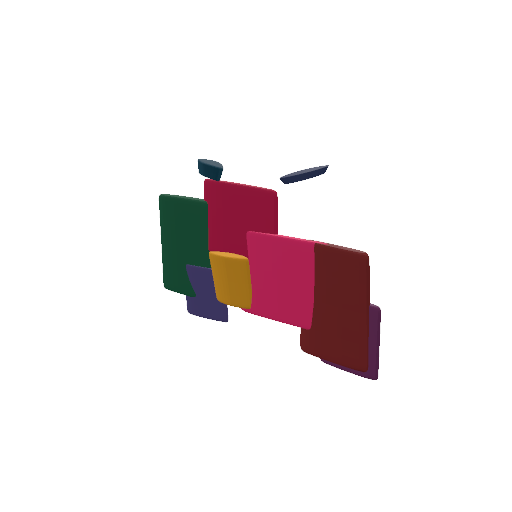}{9} &
\qIn{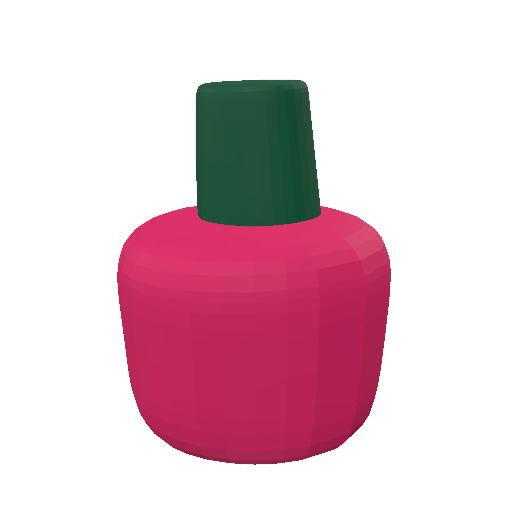}{2} &
\qIn{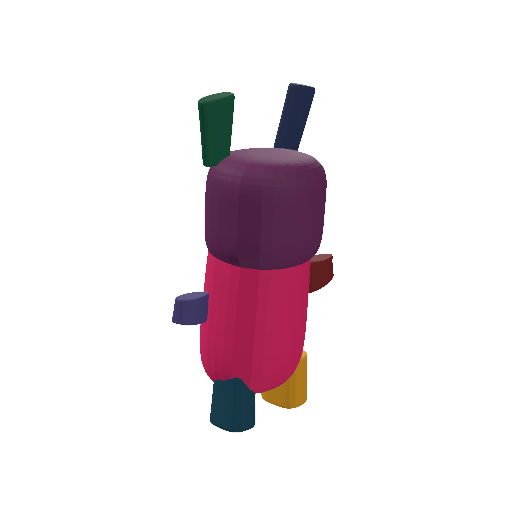}{8} &
\qIn{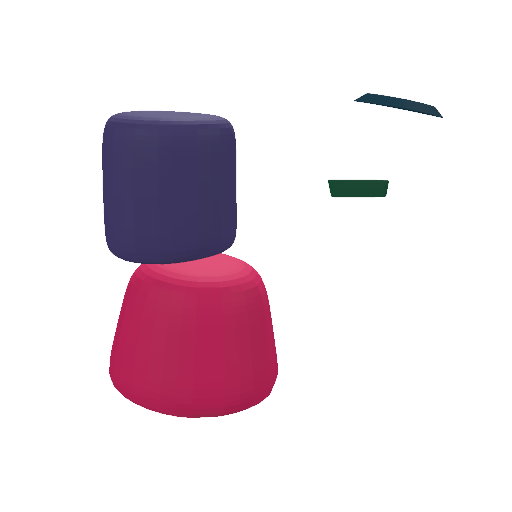}{4} &
\qIn{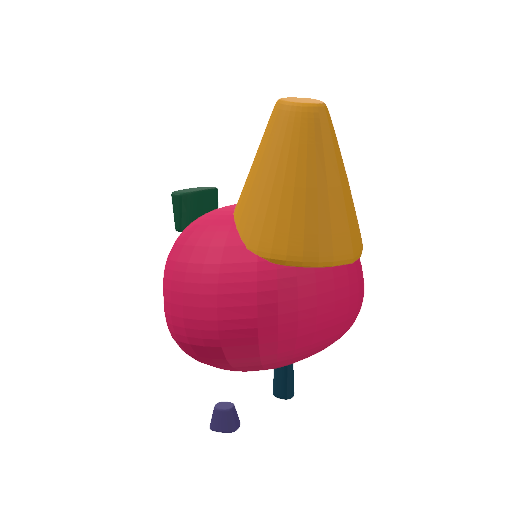}{5} &
\qIn{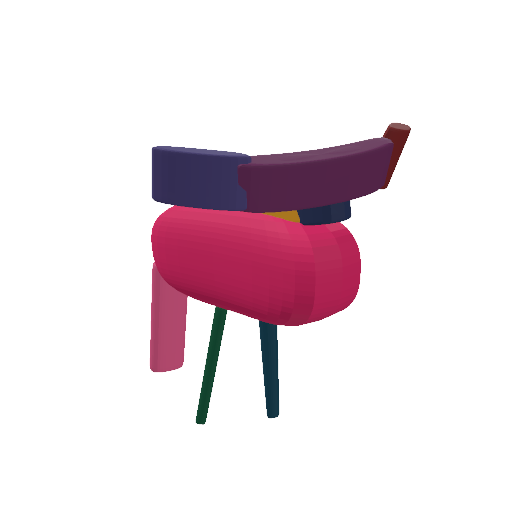}{10} &
\qIzn{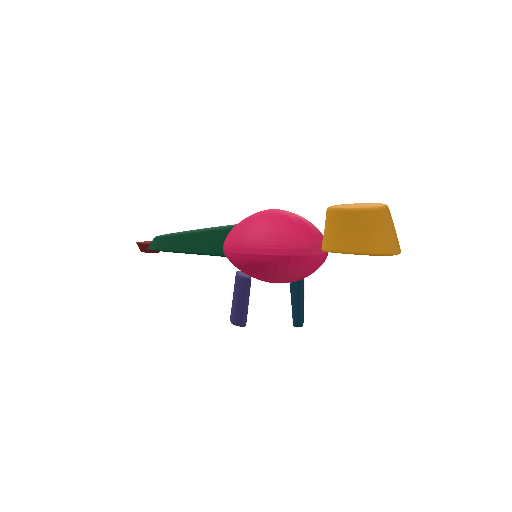}{6} &
\qIn{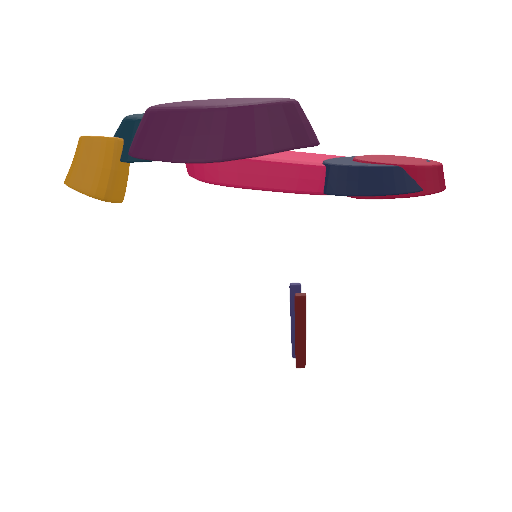}{9} &
\qIn{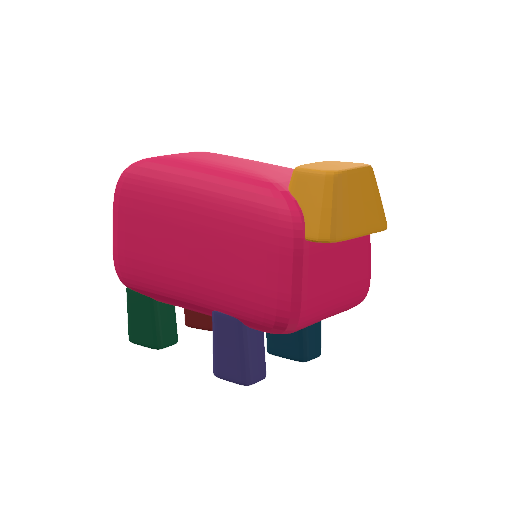}{6} \\
\midrule
\qlab{P3-SAM$^\ast$} &
\qIn{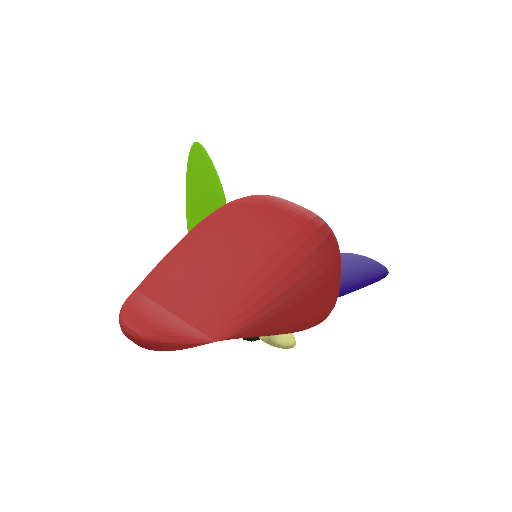}{6} &
\qIn{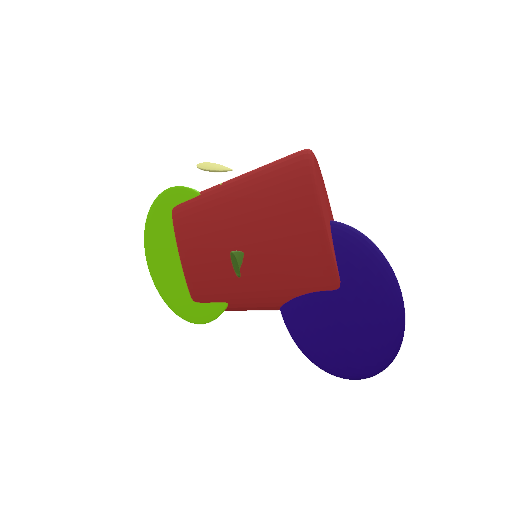}{6} &
\qIn{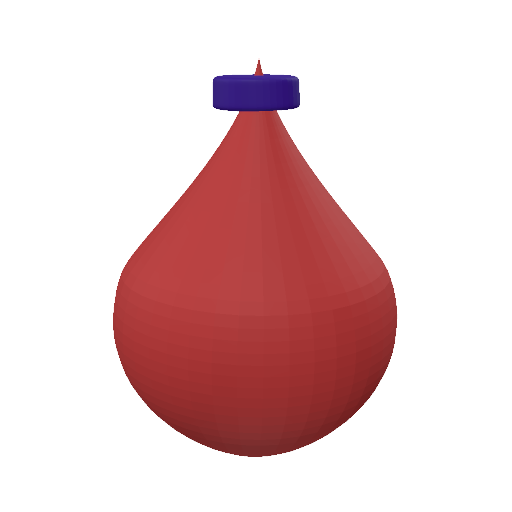}{2} &
\qIn{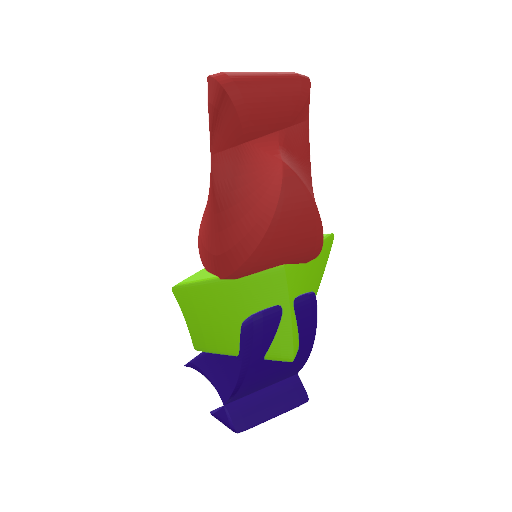}{3} &
\qIn{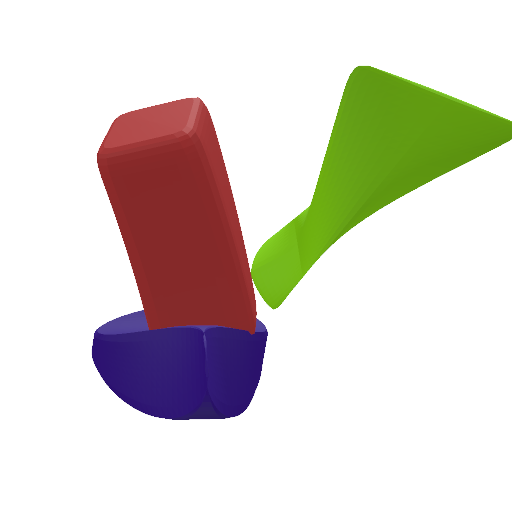}{3} &
\qIn{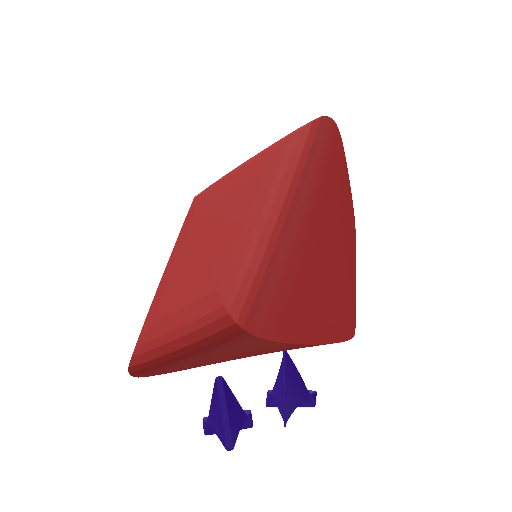}{3} &
\qIn{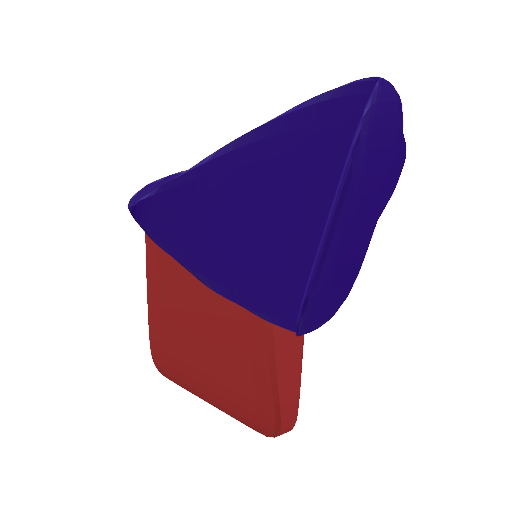}{2} &
\qIzn{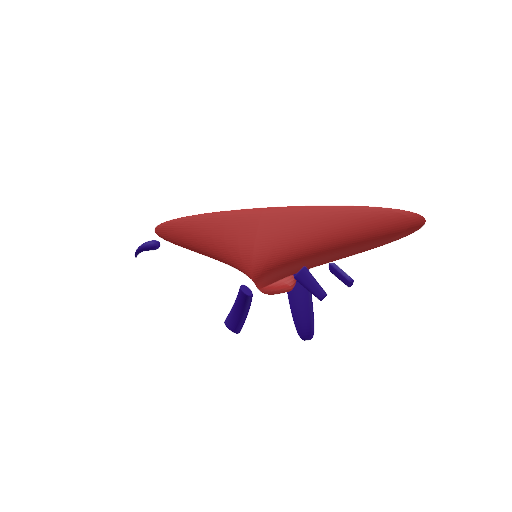}{6} &
\qIn{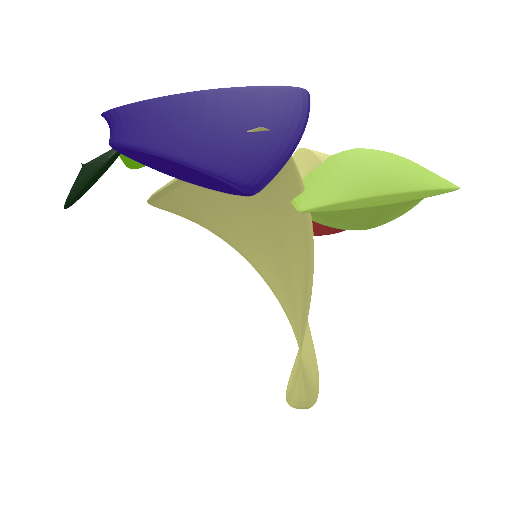}{7} &
\qIn{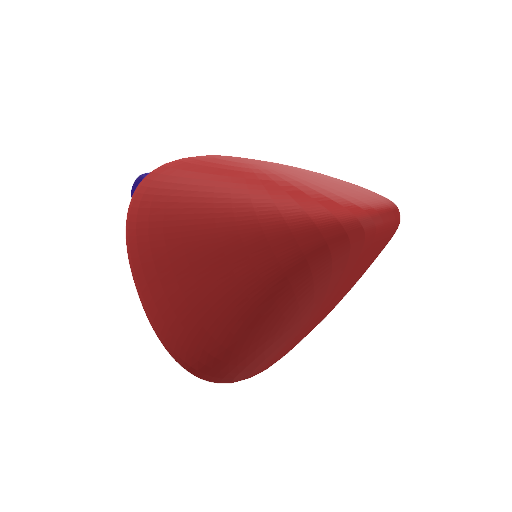}{2} \\
\qlab{PartField$^\ast$} &
\qIn{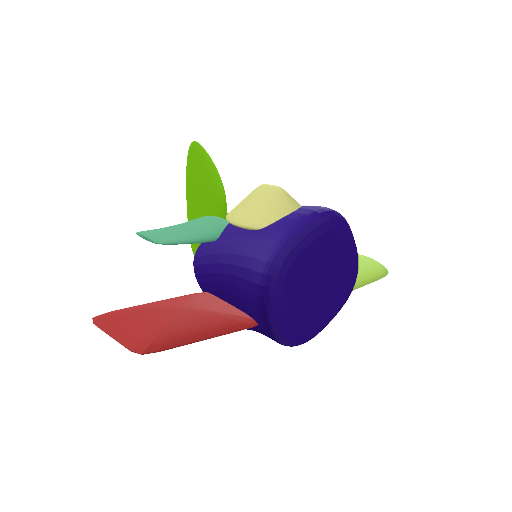}{8} &
\qIn{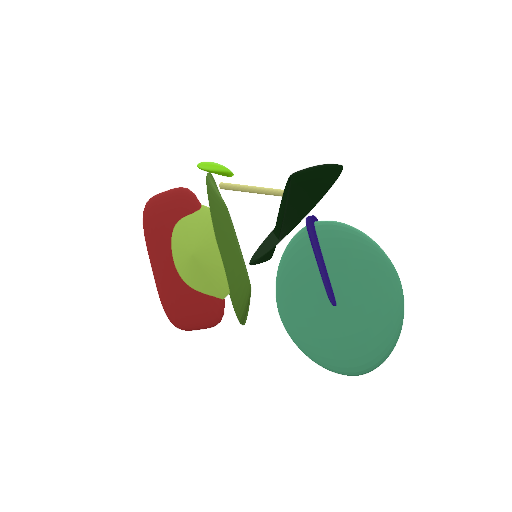}{8} &
\qIn{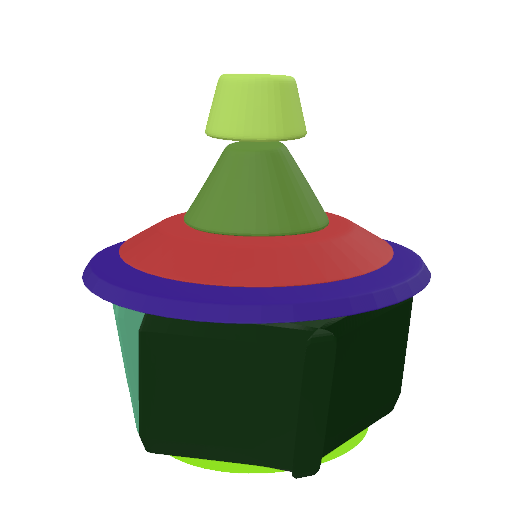}{8} &
\qIn{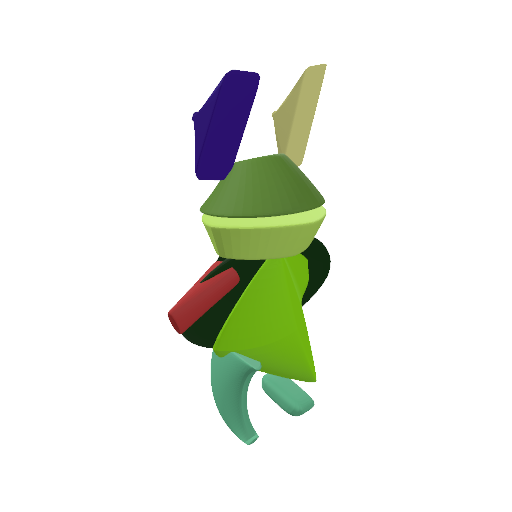}{9} &
\qIn{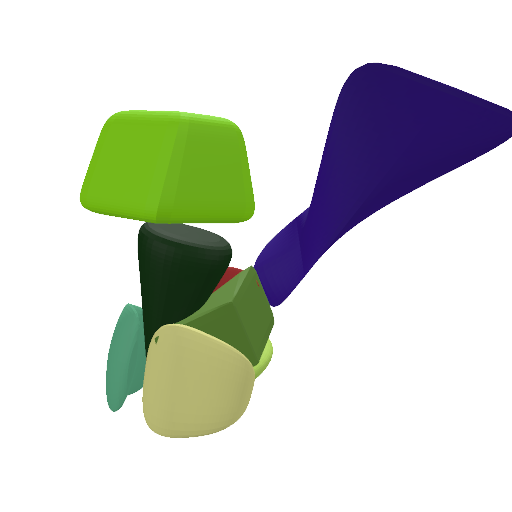}{8} &
\qIn{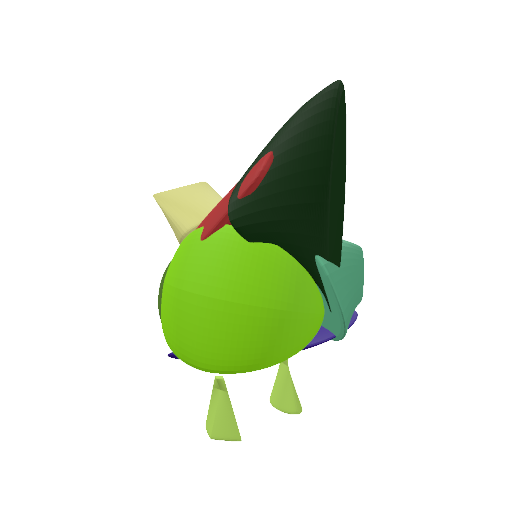}{9} &
\qIn{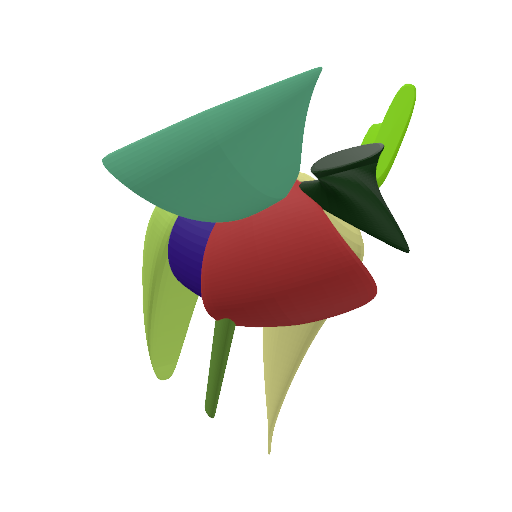}{8} &
\qIzn{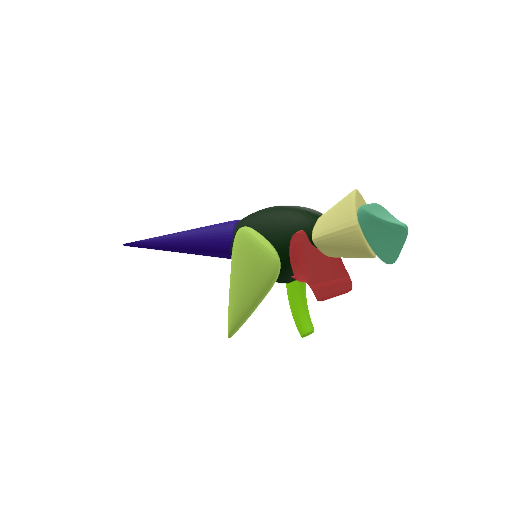}{8} &
\qIn{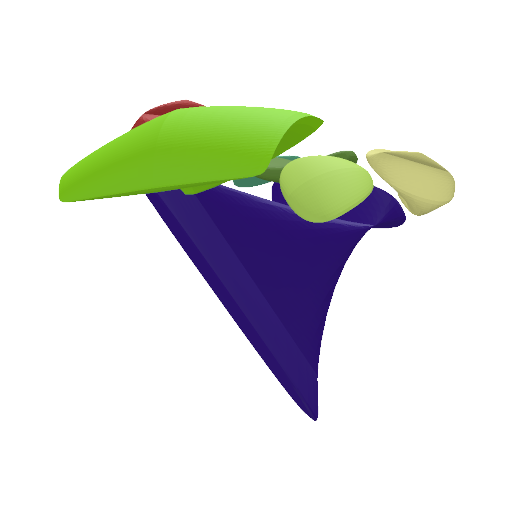}{8} &
\qIn{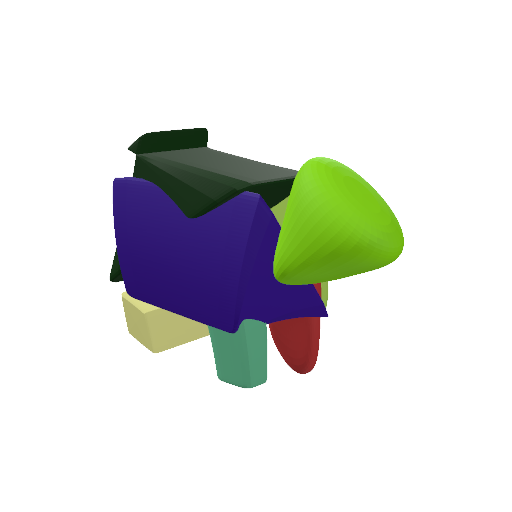}{8} \\
\qlab{PartSAM$^\ast$} &
\qIn{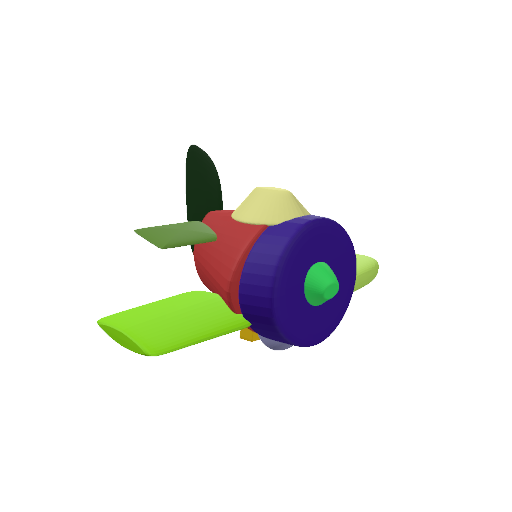}{12} &
\qIn{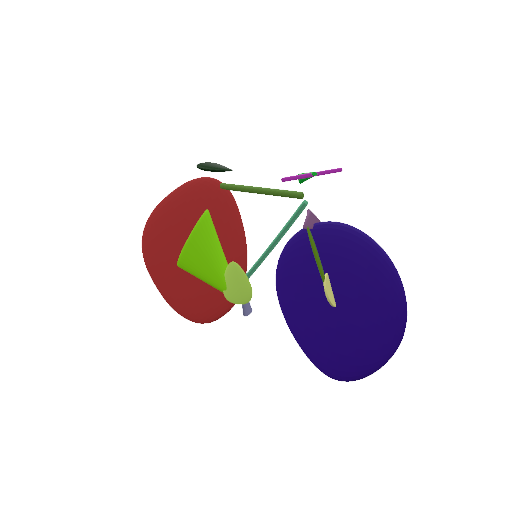}{15} &
\qIn{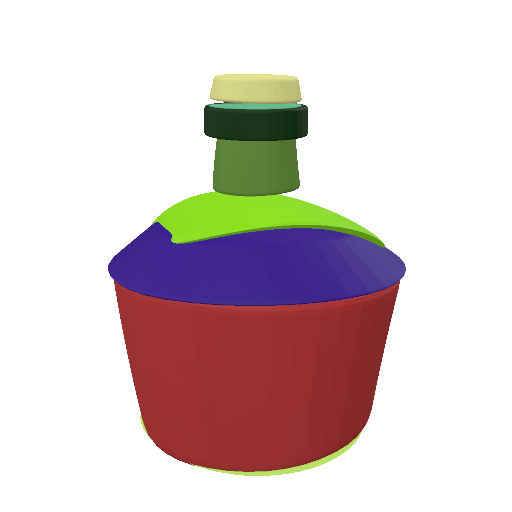}{8} &
\qIn{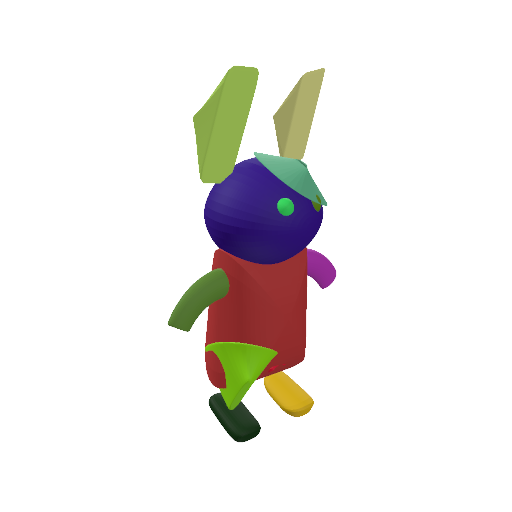}{14} &
\qIn{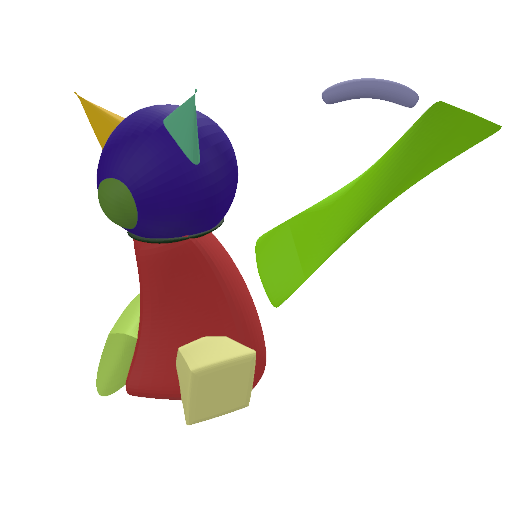}{10} &
\qIn{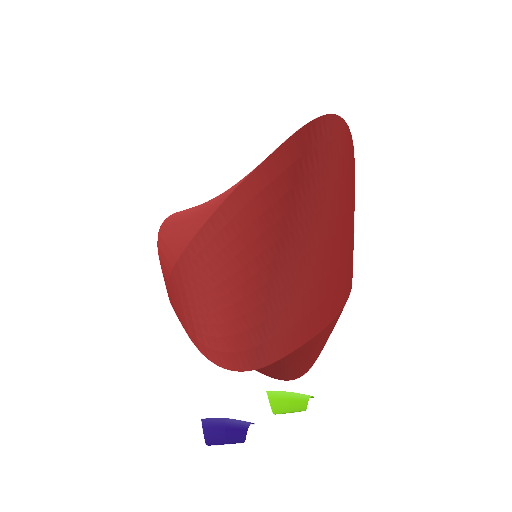}{3} &
\qIn{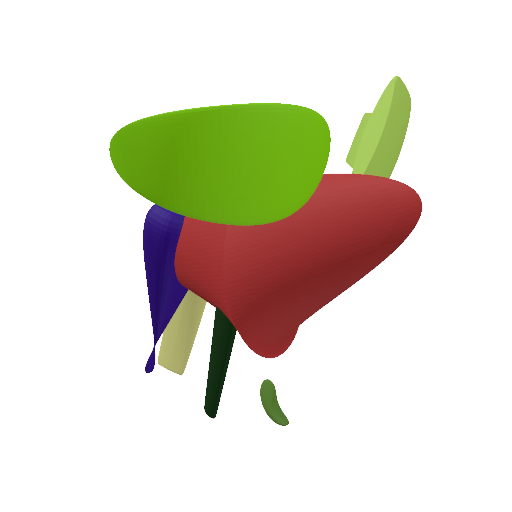}{7} &
\qIzn{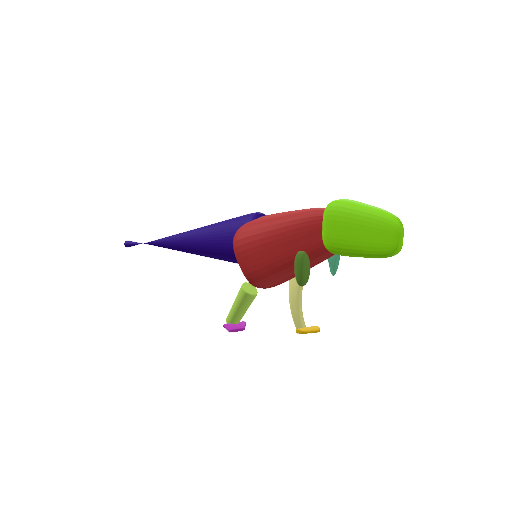}{9} &
\qIn{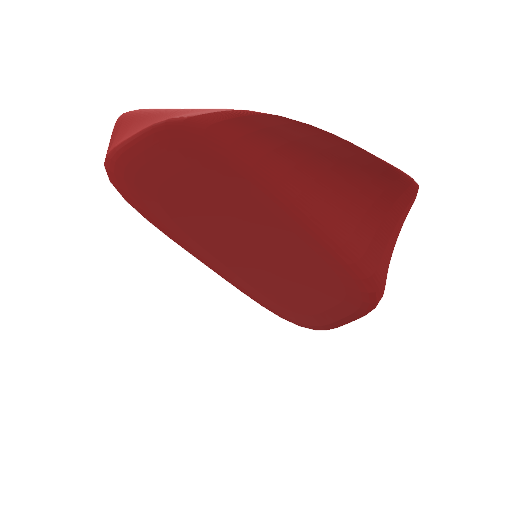}{1} &
\qIn{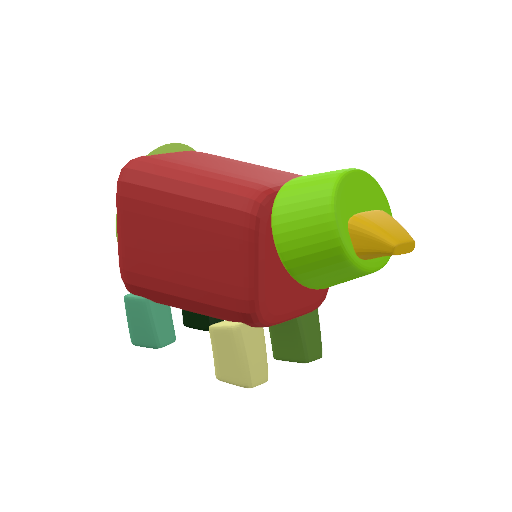}{9} \\
\midrule
\qlab{Ours} &
\qIn{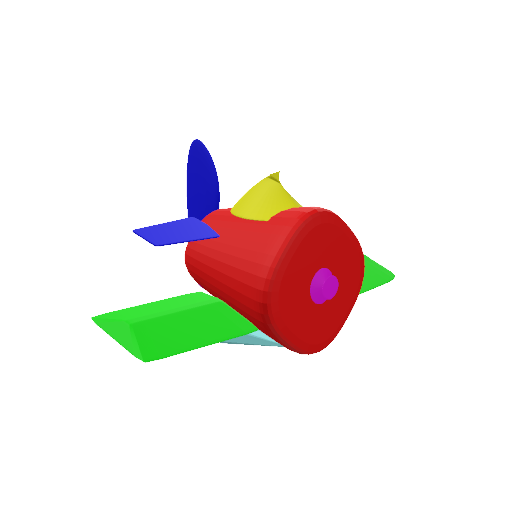}{10} &
\qIn{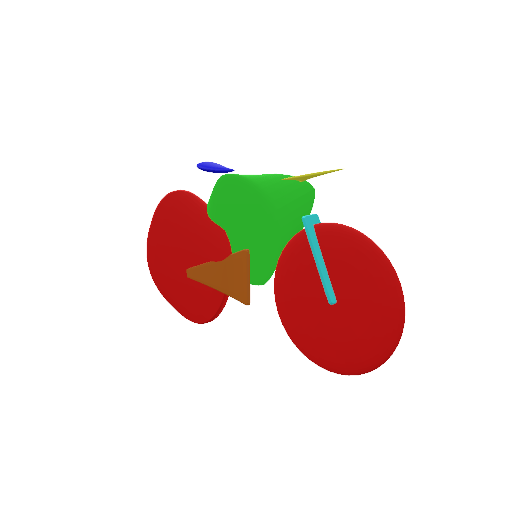}{7} &
\qIn{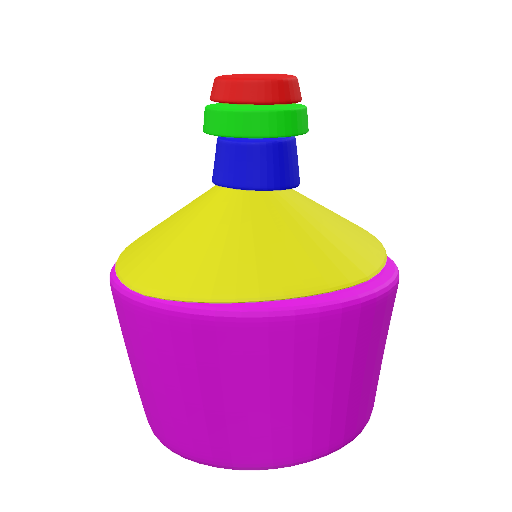}{5} &
\qIn{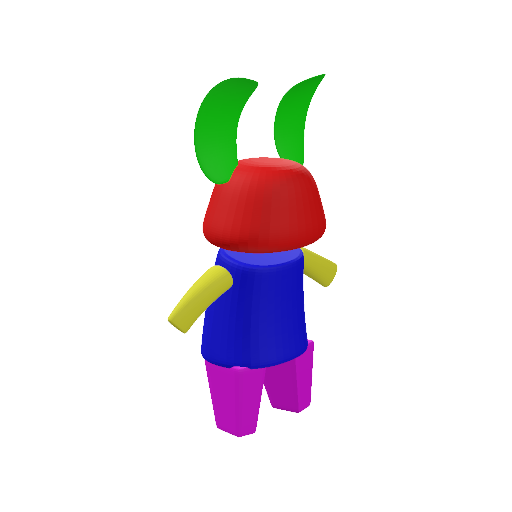}{9} &
\qIn{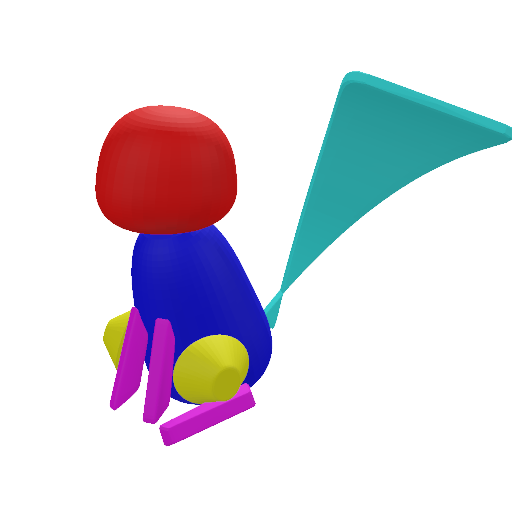}{8} &
\qIn{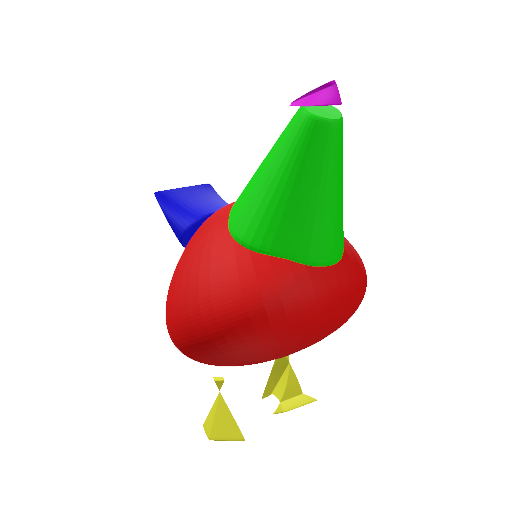}{6} &
\qIn{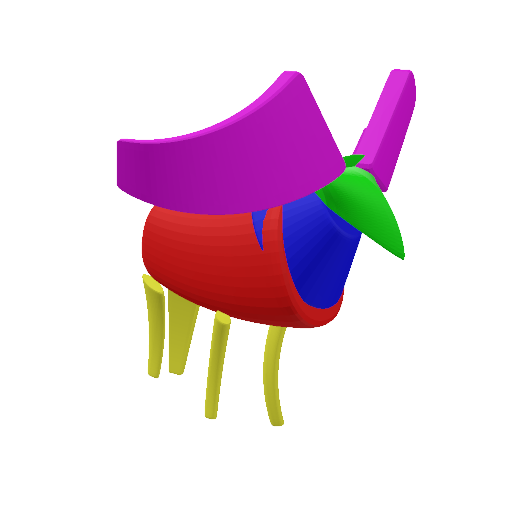}{9} &
\qIzn{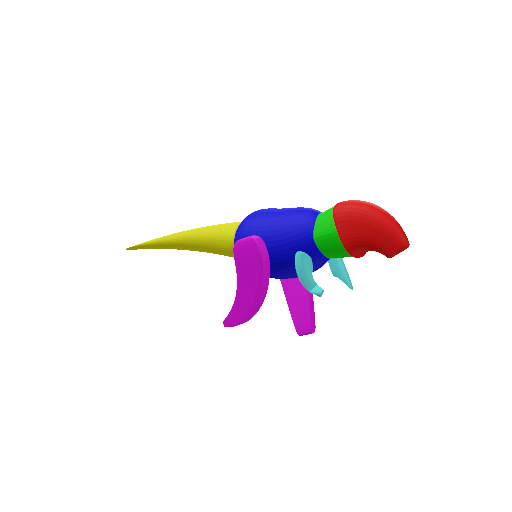}{8} &
\qIn{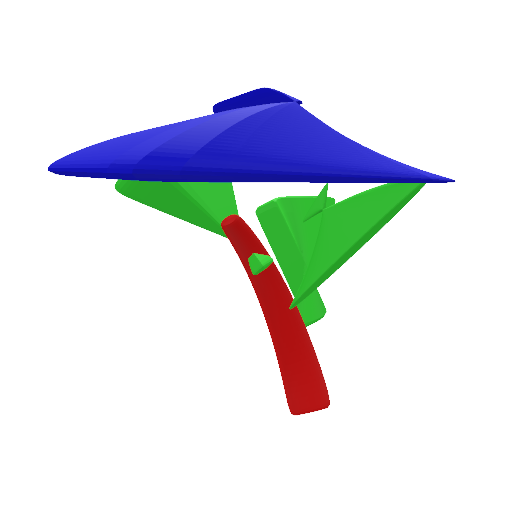}{5} &
\qIn{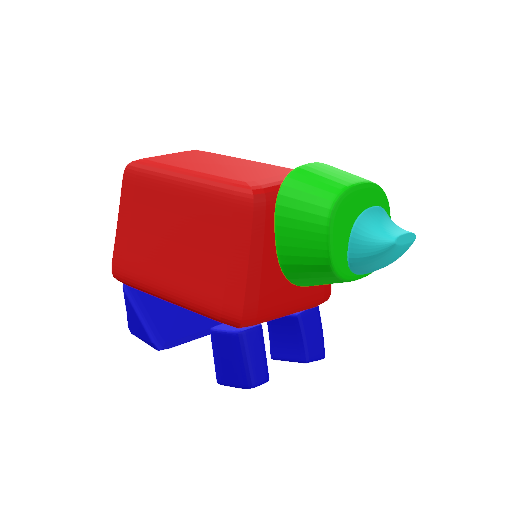}{7} \\
\qlab{GT} &
\qI{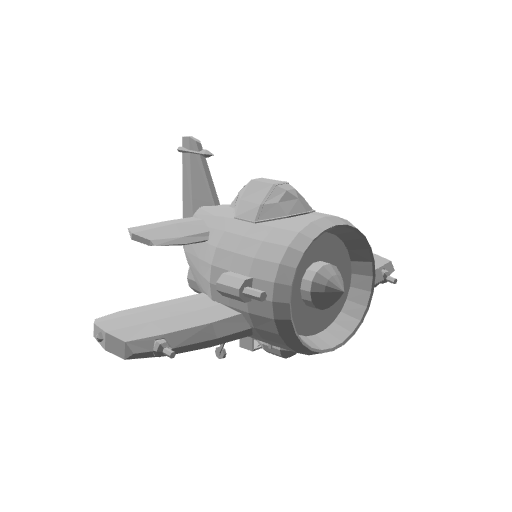} &
\qI{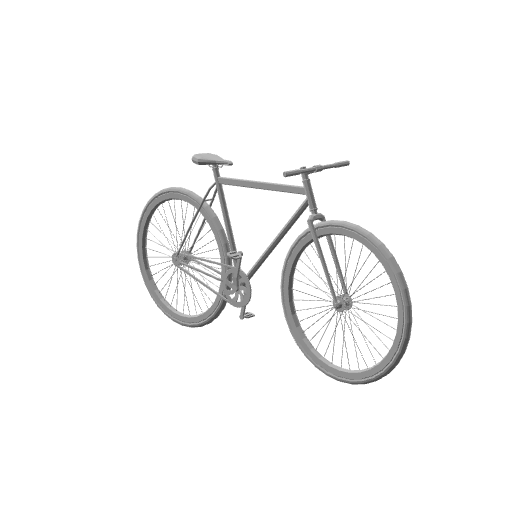} &
\qI{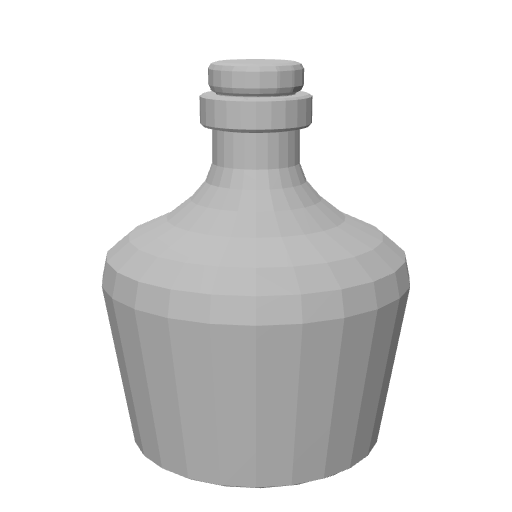} &
\qI{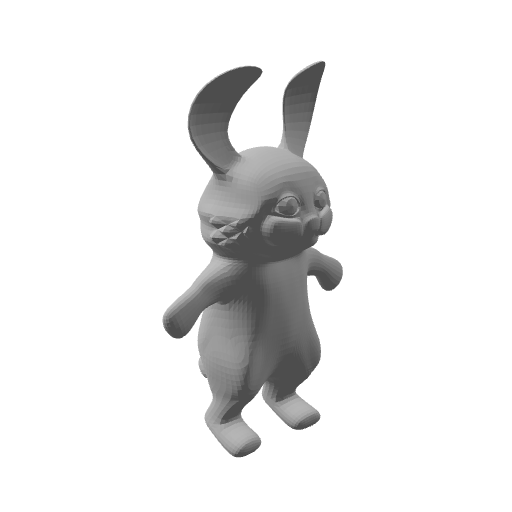} &
\qI{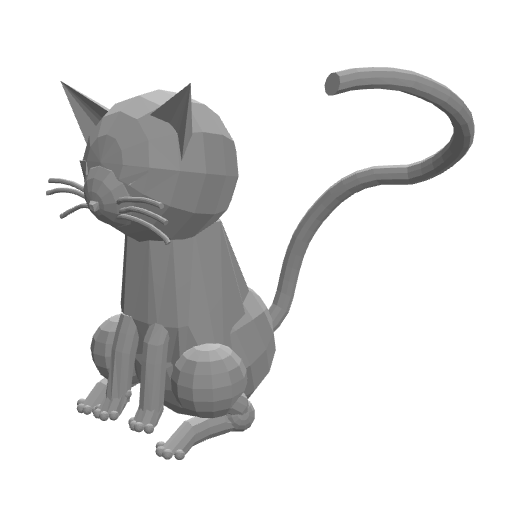} &
\qI{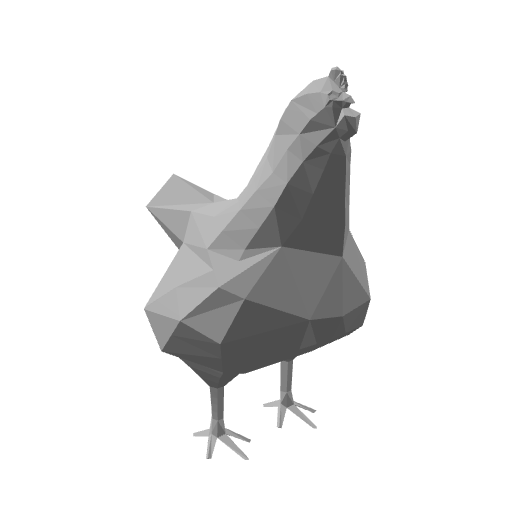} &
\qI{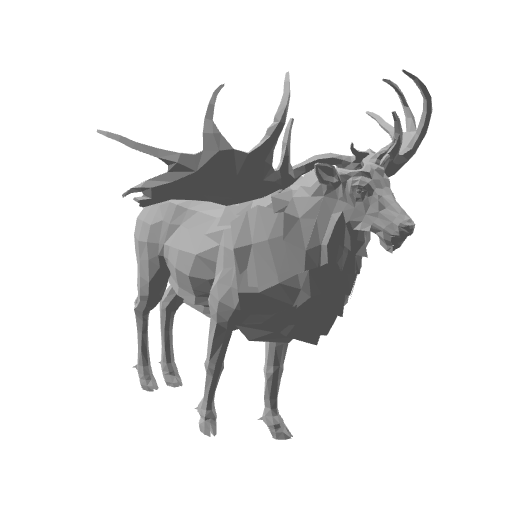} &
\qIz{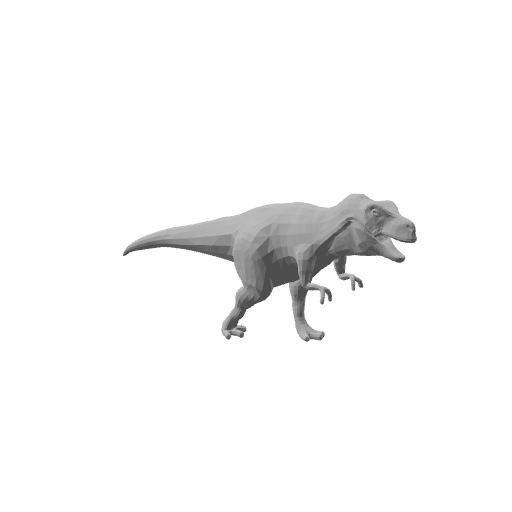} &
\qI{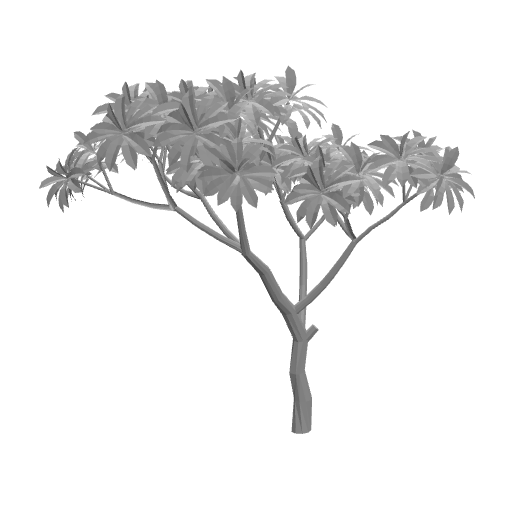} &
\qI{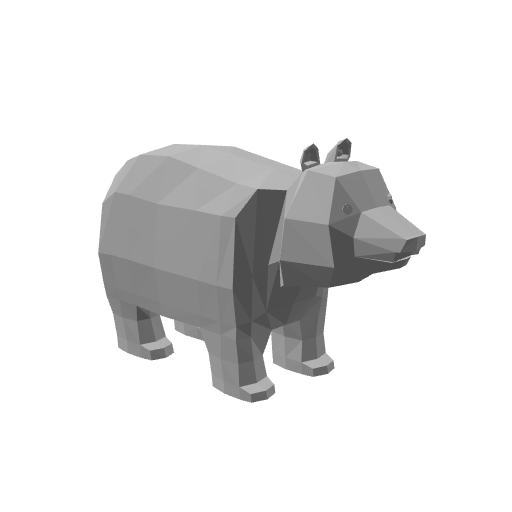} \\
\end{tabular}
\caption{\textbf{Qualitative comparison on Toys4K.} Ours uses four views. Our abstractions retain recognizable components such as ears, limbs and tails, although the cat's tail and the tree crown are geometrically coarse. SuperFlex produces compact abstractions but omits several appendages of the cat. Primitive Anything uses hundreds of primitives on several examples, while Light-SQ often subdivides the animals into many smaller primitives. $^\ast$Part labels of that segmenter fitted with our clustering and fitter (\Cref{sec:seg_source}).}
\label{fig:qual_toys4k}
\end{figure*}

\begin{table}[!th]
\centering
\small
\setlength{\tabcolsep}{9pt}
\begin{tabular}{@{}p{2.3cm} cccc@{}}
\toprule
Method & CD $\downarrow$ & IoU $\uparrow$ & OR $\downarrow$ & \#P \\
\midrule
\multicolumn{5}{c}{\emph{HumanPrim} (314 objects)} \\
EMS & 0.126 & 41.1\% & 1.191 & 7.3 \\
\textcolor{black!55}{Light-SQ} & \textcolor{black!55}{0.077} & \textcolor{black!55}{67.5\%} & \textcolor{black!55}{1.006} & \textcolor{black!55}{\underline{18.5}} \\
\midrule
\textcolor{black!55}{PrimAny} & \textcolor{black!55}{0.086} & \textcolor{black!55}{70.1\%} & \textcolor{black!55}{1.320} & \textcolor{black!55}{\underline{29.5}} \\
F2C & 0.127 & 38.5\% & 1.207 & \textbf{6.4} \\
SuperDec & 0.090 & 58.3\% & 1.052 & 6.6 \\
SuperFlex & 0.082 & \textbf{66.8\%} & 1.092 & 6.7 \\
\midrule
\rowcolor{gray!15}
Ours (4 views) & 0.082 & 63.2\% & \textbf{1.023} & 7.8 \\
\rowcolor{gray!15}
Ours (6 views) & \textbf{0.081} & 65.3\% & 1.038 & 7.7 \\
\midrule
\multicolumn{5}{c}{\emph{Toys4K} (315 objects)} \\
EMS & 0.113 & 59.0\% & 1.236 & 6.4 \\
\textcolor{black!55}{Light-SQ} & \textcolor{black!55}{0.125} & \textcolor{black!55}{66.1\%} & \textcolor{black!55}{1.009} & \textcolor{black!55}{\underline{24.7}} \\
\midrule
\textcolor{black!55}{PrimAny} & \textcolor{black!55}{0.145} & \textcolor{black!55}{45.9\%} & \textcolor{black!55}{2.080} & \textcolor{black!55}{\underline{74.9}} \\
F2C & 0.143 & 42.9\% & 1.249 & 6.7 \\
SuperDec & 0.104 & 52.2\% & 1.058 & 4.9 \\
SuperFlex & 0.096 & \textbf{67.6\%} & 1.095 & 4.9 \\
\midrule
\rowcolor{gray!15}
Ours (4 views) & 0.089 & 61.4\% & \textbf{1.032} & \textbf{4.6} \\
\rowcolor{gray!15}
Ours (6 views) & \textbf{0.086} & 63.7\% & 1.053 & 4.7 \\
\bottomrule
\end{tabular}
\caption{\textbf{Quantitative results.} Means over all objects of HumanPrim and Toys4K. Bold marks the best value per column and benchmark among the compact methods. Primitive Anything and Light-SQ (gray) use 2--16$\times$ more primitives (underlined). SuperFlex, trained on ShapeNet, leads in IoU on both benchmarks. Our method reaches the same mean CD as SuperFlex on HumanPrim (0.081 vs.\ 0.082), lower CD on Toys4K, and the lowest overlap rate of the compact methods on both.}
\label{tab:results}
\end{table}

\section{Evaluation}
\label{sec:evaluation}

We first describe datasets, baselines and metrics (\Cref{sec:setup}), then compare against the baselines on HumanPrim and Toys4K, including the split by training classes (\Cref{sec:results}).
We then analyze how view layout and surface coverage (\Cref{sec:view_layout}) affect the result, compare segmentation sources with a shared fitter (\Cref{sec:seg_source}), study annotated part decompositions (\Cref{sec:gt_seg}) and collect failure cases (\Cref{sec:failure_cases}).

\subsection{Experimental Setup}
\label{sec:setup}

\paragraph*{Datasets.}
We evaluate on two datasets.
HumanPrim~\cite{ye2025primitiveanything} (314 objects as 10k-point clouds) contains vehicles, aircraft, chairs, sofas and tables that largely fall into ShapeNet classes (62\,\% of the objects belong to one of the 13 training classes of SuperDec and SuperFlex by our category mapping) but carry more parts per object.
In Toys4K~\cite{stojanov21cvpr} (315 objects from 105 categories with ground-truth meshes) only 10\,\% of the objects belong to one of these classes, and the rest are animals, food, plants and tools.
Light-SQ requires a mesh and receives the Poisson reconstruction (depth 7) that is also the IoU ground truth for HumanPrim.
Results on a ShapeNet~\cite{chang2015shapenet} subset of airplanes, chairs and tables, classes the learned methods were trained on, are given in the supplementary material.

\paragraph*{Methods in comparison.}
We compare against six methods, two optimization-based and four learned:
EMS~\cite{liu2022ems} (optimization-based EM superquadric fitting, category-agnostic),
Light-SQ~\cite{wang2025lightsq} (optimization-based superquadric fitting on a signed distance field with convex-decomposition guidance, category-agnostic),
Primitive Anything (PrimAny)~\cite{ye2025primitiveanything} (auto-regressive transformer, trained on human-crafted abstractions),
Fine-to-Coarse (F2C)~\cite{kobsik2025f2c} (progressive cuboid refinement, trained on ShapeNet),
SuperDec~\cite{fedele2025superdec} (feed-forward superquadric decomposition, trained on ShapeNet), and
SuperFlex~\cite{tavernini2026superflex} (its successor with tapering and bending deformations, trained on ShapeNet).

\paragraph*{Metrics.}
We evaluate using four metrics.
\textbf{Chamfer distance (CD)} measures surface fidelity as the sum of the two directional mean nearest-neighbor distances (unsquared, $\ell_2$) between the original point cloud (or mesh) and points sampled from the fitted primitives, both sampled to 2048 points and normalized to the ground-truth shape extent $[-1, 1]^3$.
\textbf{Volumetric Intersection over Union (IoU)} measures the overlap between the volume enclosed by the fitted primitives and the ground-truth mesh ($128^3$ grid).
For HumanPrim point clouds, we obtain a watertight reference mesh by Poisson surface reconstruction. Toys4K and ShapeNet meshes are converted to watertight meshes before occupancy evaluation.
\textbf{Overlap Rate (OR)}, introduced by Wang et al.~\cite{wang2025lightsq}, is the summed volume of all primitives divided by the volume of their union.
OR $= 1.0$ means that no two primitives share any interior volume, and higher values indicate redundant overlap.
\textbf{Primitive count (\#P)} reports the mean number of primitives per abstraction.
Fewer primitives mean a more compact decomposition, but interpretability also depends on which object components the primitives preserve.
None of these metrics directly measures semantic part correctness, so our evidence for recognizable part structure is qualitative.
IoU rewards volume filling and rises for geometrically simple objects independently of the decomposition.
We compute all metrics ourselves with one evaluation protocol for every method, so the numbers can differ slightly from those reported in the original publications.

\subsection{Results}
\label{sec:results}

\Cref{tab:results} reports quantitative results on HumanPrim and Toys4K for four ring views and for six views, which add a top and a bottom view, using the same fitter and settings.
\Cref{fig:qual_humanprim,fig:qual_toys4k} show qualitative samples of the four-view variant.

\paragraph*{Surface fidelity at compact complexity.}
Our method achieves surface accuracy comparable to SuperFlex on HumanPrim and lower Chamfer distance on Toys4K, using an average of 4.6--7.8 primitives per object.
On HumanPrim, the six-view variant reaches CD 0.081 against 0.082 for SuperFlex and 0.090 for SuperDec, with 7.7 primitives against 6.7 and 6.6, respectively.
Light-SQ achieves the lowest CD overall (0.077) but uses 18.5 primitives on average, and Primitive Anything reaches 0.086 with 29.5.
On Toys4K, our six-view variant achieves the lowest CD among the methods of \Cref{tab:results} (0.086), approximately 10\,\% below SuperFlex (0.096), with slightly fewer primitives (4.7 vs.\ 4.9).
The four-view variant also has lower CD than every baseline (0.089) and the lowest mean primitive count (4.6).
Light-SQ and Primitive Anything use 24.7 and 74.9 primitives without reaching this accuracy.

\paragraph*{Preservation of part structure.}
The qualitative comparisons illustrate how the primitive budget is allocated to object components.
In \Cref{fig:qual_humanprim}, our chair abstraction retains individual backrest bars and legs, while the ladder preserves its open structure of rails and rungs, which EMS largely merges into bulkier primitives.
The chair uses 14 primitives against 10 for SuperFlex: the additional primitives retain finer structural components.
The sports car similarly retains distinct wheels and body components.
In \Cref{fig:qual_toys4k}, the rabbit's ears and limbs and the dinosaur's tail and legs remain recognizable as separate parts.
SuperFlex also produces a plausible rabbit, but omits much of the cat's appendage structure, which our abstraction retains.
Part preservation does not ensure accurate part geometry: our cat's tail becomes an overly broad surface, and the tree crown is strongly simplified.
Both variants have a lower overlap rate than every compact baseline on both benchmarks (1.023 and 1.038 on HumanPrim, 1.032 and 1.053 on Toys4K), without an explicit overlap penalty.

\paragraph*{Surface accuracy and volumetric agreement.}
SuperFlex achieves higher IoU on both benchmarks: 66.8\,\% against 65.3\,\% for our six-view variant on HumanPrim, and 67.6\,\% against 63.7\,\% on Toys4K.
Primitive Anything reaches the highest HumanPrim IoU (70.1\,\%) with 29.5 primitives, while Light-SQ reaches 67.5\,\% with 18.5.
The remaining gap in volume is consistent with the surface-based fitting objective: a primitive can approximate an observed surface while underestimating its thickness.
The following studies separate the contributions of segmentation, surface coverage and fitting to this gap.

\paragraph*{Transfer beyond the learned methods' training classes.}
The larger CD advantage on Toys4K coincides with its smaller share of objects belonging to the 13 ShapeNet training classes of SuperDec and SuperFlex: 10\,\% compared with 62\,\% on HumanPrim.
The within-benchmark split in \supp{tab:ood_split} provides further evidence.
On HumanPrim, SuperFlex's CD rises from 0.078 inside these classes to 0.089 outside them, and SuperDec's from 0.086 to 0.098.
Our four-view variant changes from 0.081 to 0.084 and the six-view variant from 0.079 to 0.085, so both achieve lower CD than these baselines outside the training classes.
On Toys4K, our six-view CD is 0.087 inside and 0.086 outside these classes, while SuperFlex retains its IoU lead on both benchmarks and subsets.
On ShapeNet itself, the training classes of the learned methods, SuperFlex leads the abstraction methods with CD 0.081 and IoU 63.3\,\% against our 0.095 and 48.6\,\% (\supp{tab:results_shapenet}).

\subsection{View Layout and Coverage}
\label{sec:view_layout}

\emph{How much does exposing additional surface improve the abstraction?}
The four ring views leave about 18\,\% of the surface unobserved on the 50 HumanPrim objects of this study.
Unobserved points receive no labels and do not enter the fitter, which can underestimate the thickness of parts seen from one side.
We compare layouts with 1--16 views using the same renderer, prompt template and fitter, and measure view coverage (VC) as the fraction of surface points visible in at least one view (\Cref{tab:view_coverage}).

Adding top and bottom views raises coverage from 81.6\,\% to 97.5\,\% and IoU from 68.4\,\% to 69.1\,\%, with unchanged CD (0.073) and approximately unchanged primitive count (7.5 to 7.6).
On the full benchmarks the six-view layout raises IoU by 2.1 and 2.3 points at similar primitive count and slightly higher overlap (\Cref{tab:results}), most on broad-surfaced categories: HumanPrim chairs, beds and sofas gain 7--10 points, whereas ladders and swings lose 18--20 points.
The nine- and sixteen-view layouts reach 99.1\,\% and 99.8\,\% coverage, but their IoU differs from six views by at most 0.3 percentage points and CD by at most 0.001, so returns diminish once most of the surface is visible.
Changing views also changes the generated masks, so the gains combine better coverage with masks generated from more views.

A single composite mask and four sequential masks yield similar results on this subset (CD 0.073 vs.\ 0.074, IoU 68.4\,\% vs.\ 68.2\,\%), but the composite needs one image-generation call instead of four.

\begin{table}[t]
\centering\small
\setlength{\tabcolsep}{5pt}
\begin{tabular}{@{}l c c c c c@{}}
\toprule
Layout & $V$ & VC $\uparrow$ & CD $\downarrow$ & IoU $\uparrow$ & \#P \\
\midrule
single view          & 1  & 46.5\,\% & 0.134 & 36.0\,\% & 5.7 \\
ring, 2 of 4 views   & 2  & 65.3\,\% & 0.085 & 57.9\,\% & 7.0 \\
ring, 3 of 4 views   & 3  & 78.3\,\% & 0.076 & 66.3\,\% & 7.3 \\
\rowcolor{gray!15}
ring                 & 4 & 81.6\,\% & \textbf{0.073} & 68.4\,\% & 7.5 \\
\rowcolor{gray!15}
ring + top/bottom    & 6  & 97.5\,\% & \textbf{0.073} & \textbf{69.1\,\%} & 7.6 \\
two rings + top      & 9  & 99.1\,\% & 0.074 & 68.8\,\% & 7.3 \\
four rings           & 16 & 99.8\,\% & 0.074 & 68.8\,\% & 7.2 \\
\bottomrule
\end{tabular}
\caption{%
  \textbf{View layout and surface coverage.}
  Means over 50 HumanPrim objects with identical renderer, prompt and fitter. VC = view coverage, the fraction of the surface visible in at least one view. Gray rows are the layouts of \Cref{tab:results}.
}
\label{tab:view_coverage}
\end{table}

\subsection{Segmentation Source with a Shared Fitter}
\label{sec:seg_source}

\emph{How much does the choice of part segmentation affect the resulting primitive abstraction?}
We replace our generated labels with P3-SAM~\cite{li2025p3sam}, PartField~\cite{qian2025partfield} or PartSAM~\cite{zhu2026partsam} predictions and retain the same color-restricted clustering and superquadric fitter for every row (\Cref{tab:seg_source}).
For P3-SAM, we follow the released inference default of 100k points, sampled from the Poisson reconstruction, and transfer its labels by nearest neighbors to the common 10k fitting points. PartField and PartSAM receive the original 10k point cloud.

\begin{table}[tbp]
\centering
\small
\setlength{\tabcolsep}{9pt}
\begin{tabular}{@{}p{2.3cm} cccc@{}}
\toprule
Segmentation & CD $\downarrow$ & IoU $\uparrow$ & OR $\downarrow$ & \#P \\
\midrule
\multicolumn{5}{c}{\textit{HumanPrim} ($n=314$)} \\
P3-SAM & 0.093 & 57.4\% & 1.031 & \textbf{6.1} \\
PartField ($k=8$) & 0.078 & 48.0\% & 1.046 & 9.0 \\
PartSAM & \textbf{0.072} & \textbf{70.6\%} & 1.080 & 11.0 \\
\rowcolor{gray!15}
Ours (4 views) & 0.082 & 63.2\% & \textbf{1.023} & 7.8 \\
\rowcolor{gray!15}
Ours (6 views) & 0.081 & 65.3\% & 1.038 & 7.7 \\
\midrule
\multicolumn{5}{c}{\textit{Toys4K} ($n=315$)} \\
P3-SAM & 0.099 & 57.0\% & 1.084 & \textbf{3.4} \\
PartField ($k=8$) & 0.087 & 48.9\% & 1.080 & 8.6 \\
PartSAM & \textbf{0.082} & 62.1\% & 1.111 & 8.0 \\
\rowcolor{gray!15}
Ours (4 views) & 0.089 & 61.4\% & \textbf{1.032} & 4.6 \\
\rowcolor{gray!15}
Ours (6 views) & 0.086 & \textbf{63.7\%} & 1.053 & 4.7 \\
\bottomrule
\end{tabular}
\caption{\textbf{Segmentation source.} Other segmenters' part decompositions fitted with our clustering and fitter. PartSAM, trained on 3D part data, reaches lower CD on both benchmarks and higher IoU on HumanPrim with 1.4--1.7$\times$ more primitives and the highest overlap. Bold marks the best value per column.}
\label{tab:seg_source}
\end{table}

With the fitter fixed, our six-view masks yield higher IoU than P3-SAM and PartField on HumanPrim and Toys4K, by 7.9 and 6.6 points over P3-SAM and by 17.3 and 14.8 points over PartField, and lower CD than P3-SAM on both.
PartSAM reaches higher IoU on HumanPrim (70.6\,\% vs.\ 65.3\,\%) and lower CD on both, but it does so with 8.0--11.0 primitives against our 4.7--7.7 and with the highest overlap rate of all rows. On Toys4K our IoU is higher (63.7\,\% vs.\ 62.1\,\%).
Our four-view masks give the lowest overlap rate on both benchmarks.
\Cref{fig:qual_humanprim,fig:qual_toys4k} show the corresponding abstractions: P3-SAM merges thin structures into their neighbors, PartField cuts objects into pieces of similar size regardless of structure, and PartSAM subdivides parts further, whereas the generated masks keep components such as chair legs and rungs as single primitives.

\subsection{Ground-Truth Segmentation Study}
\label{sec:gt_seg}

\begin{table}[t]
\centering
\setlength{\tabcolsep}{4pt}
\begin{tabular}{l ccc ccc}
\toprule
 & \multicolumn{3}{c}{Chair} & \multicolumn{3}{c}{Table} \\
\cmidrule(lr){2-4} \cmidrule(lr){5-7}
 & CD $\downarrow$ & IoU $\uparrow$ & \#P & CD $\downarrow$ & IoU $\uparrow$ & \#P \\
\midrule
Ours (4 views) & 0.109 & 41.2\% & 6.3  & 0.089 & 41.1\% & 5.3 \\
Ours (6 views) & 0.107 & 46.6\% & 6.3  & 0.095 & 41.2\% & 5.4 \\
\midrule
GT L1 & 0.095 & 50.7\% & 5.4  & 0.122 & 33.3\% & 2.7 \\
GT L2 & 0.084 & 55.3\% & 8.2  & \textbf{0.076} & \textbf{45.0\%} & 7.6 \\
GT L3 & \textbf{0.082} & \textbf{56.7\%} & 8.7 & \textbf{0.076} & 44.8\% & 7.9 \\
\bottomrule
\end{tabular}
\caption{\textbf{Ground-truth segmentation study.} Instance-split PartNet parts (levels L1--L3) replace the generative segmentation. Clustering, fitter and evaluation are identical to our pipeline, and all rows are scored against the ShapeNet meshes of the 94 chairs and 95 tables with PartNet annotation. Cleaner parts raise IoU at every level on chairs. On tables, L1 merges top and legs into 2.7 primitives and falls below our result, whereas L2--L3 with 8 primitives exceed it.}
\label{tab:gt_segmentation}
\end{table}

\emph{How does the choice of part decomposition affect abstraction quality with a fixed fitter?}
We replace the generative segmentation with human-annotated PartNet labels~\cite{mo2019partnet} at the three granularity levels L1--L3 on 94 chairs and 95 tables with corresponding ShapeNet meshes~\cite{chang2015shapenet}.
PartNet labels are semantic rather than instance-level, so we split each part into spatially connected components with the flood-fill radius and minimum-size filter of our pipeline, keep the fitter fixed, and score all rows against the same meshes (\Cref{tab:gt_segmentation}).
The study compares the resulting part point clouds, so it measures the decomposition together with the surface each decomposition covers.

On chairs, every PartNet level improves both metrics: L1 reaches CD 0.095 and IoU 50.7\,\% with fewer primitives (5.4) than our four-view result (0.109 / 41.2\,\%, 6.3), and L2--L3 raise IoU to 55--57\,\% with 8--9 primitives.
On tables, the coarse L1 merges top and legs into 2.7 primitives and falls below our pipeline (CD 0.122, IoU 33.3\,\%), whereas L2--L3 reach 0.076 / 45\,\% with 7.6--7.9 primitives.
The decomposition thus matters as much as the fitter, but its granularity must suit the primitive representation, and the annotated chairs show 4--10 IoU points of headroom over our six-view result for better masks at a suitable granularity.

\subsection{Failure Cases and Limitations}
\label{sec:failure_cases}

\begin{figure}[t]
  \centering
  \setlength{\fboxsep}{2pt}\setlength{\fboxrule}{0.5pt}%
  \providecommand{\fcimg}[1]{}
  \renewcommand{\fcimg}[1]{%
    \begin{minipage}[c]{0.28\linewidth}
      \centering\includegraphics[width=\linewidth,keepaspectratio]{#1}%
    \end{minipage}}
  \providecommand{\fcimgbox}[1]{}
  \renewcommand{\fcimgbox}[1]{%
    \begin{minipage}[c]{0.28\linewidth}
      \centering\fcolorbox{gray!60}{white}{%
        \includegraphics[width=\dimexpr\linewidth-5pt\relax,keepaspectratio]{#1}}%
    \end{minipage}}
  \providecommand{\fcarrow}{}
  \renewcommand{\fcarrow}{\hfill{\large$\rightarrow$}\hfill}
  \makebox[0.28\linewidth]{\small\textbf{Multi-View}}
  \makebox[0.06\linewidth]{}
  \makebox[0.28\linewidth]{\small\textbf{Segmentation}}
  \makebox[0.06\linewidth]{}
  \makebox[0.28\linewidth]{\small\textbf{Abstraction}}\\[2pt]
  \fcimgbox{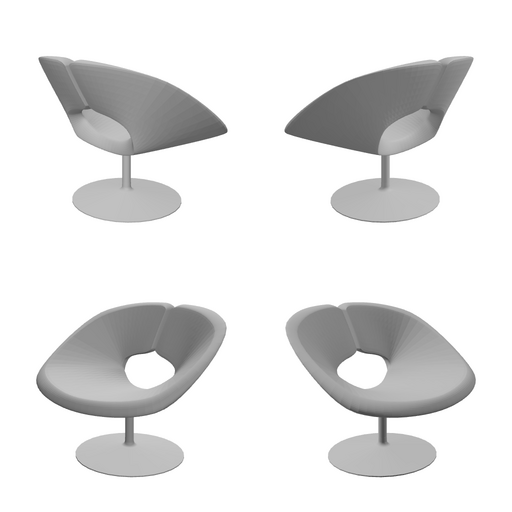}\fcarrow
  \fcimgbox{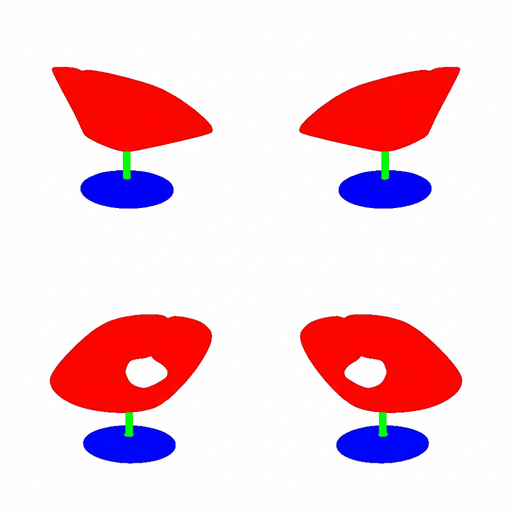}\fcarrow
  \fcimg{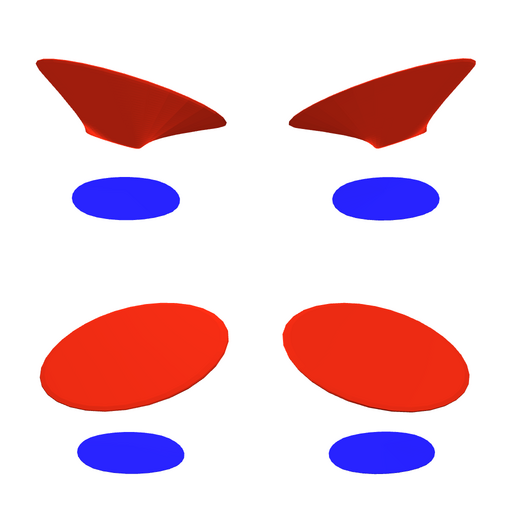}\\[6pt]
  \fcimgbox{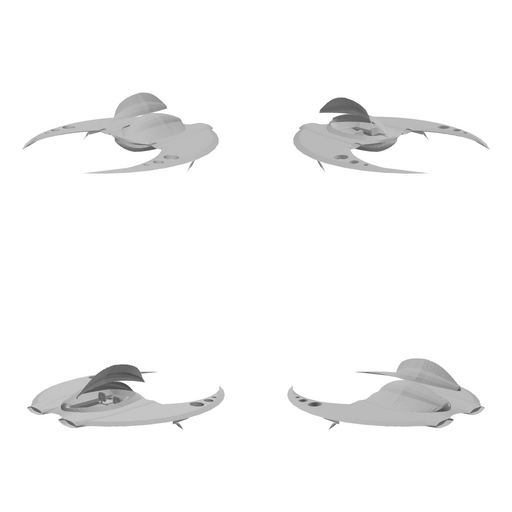}\fcarrow
  \fcimgbox{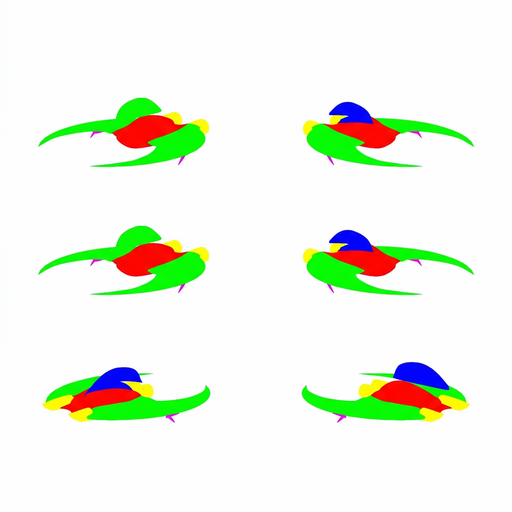}\fcarrow
  \fcimg{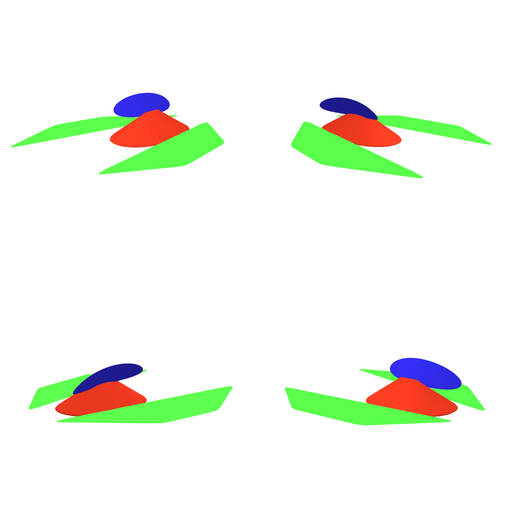}%
  \caption{%
    \textbf{Failure cases.}
    In the top row, color-restricted clustering wrongly removes the chair leg (green), which is absent from the final abstraction.
    In the bottom row, the generative image model hallucinates an additional phantom view. Its votes fall outside the object silhouette and are discarded, so the abstraction is unaffected.
  }
  \label{fig:failure_cases}
\end{figure}

\begin{figure}[htbp]
  \centering
  \newcommand{\segw}{0.185\linewidth}%

  \makebox[\linewidth]{%
    \makebox[\segw]{\small Run 1}%
    \hfill
    \makebox[\segw]{\small Run 2}%
    \hfill
    \makebox[\segw]{\small Run 3}%
    \hfill
    \makebox[\segw]{\small Run 4}%
    \hfill
    \makebox[\segw]{\small Run 5}%
  }

  \vspace{2pt}

  \includegraphics[width=\segw]{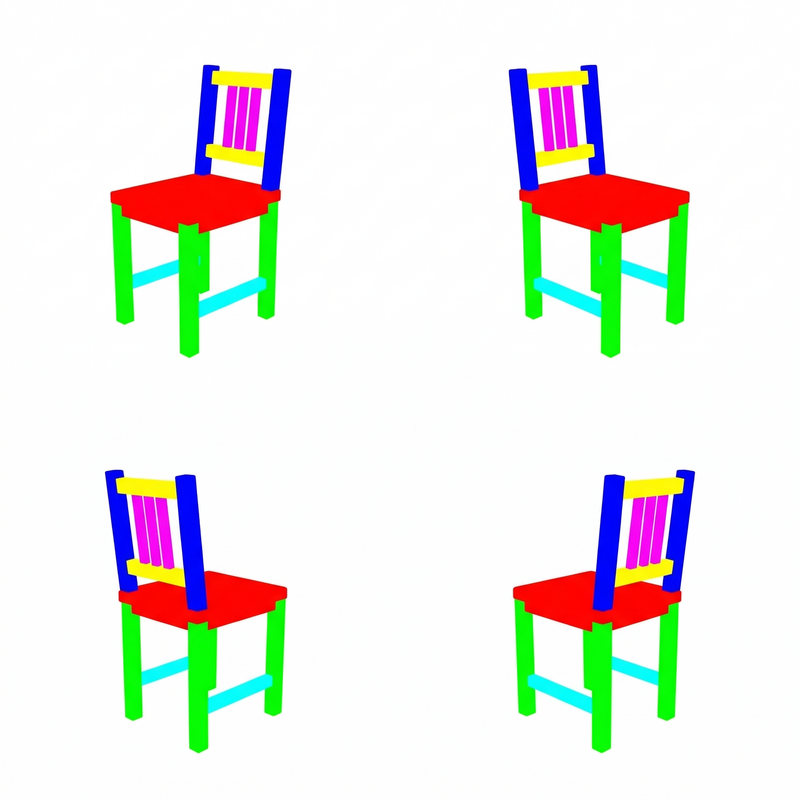}%
  \hfill
  \includegraphics[width=\segw]{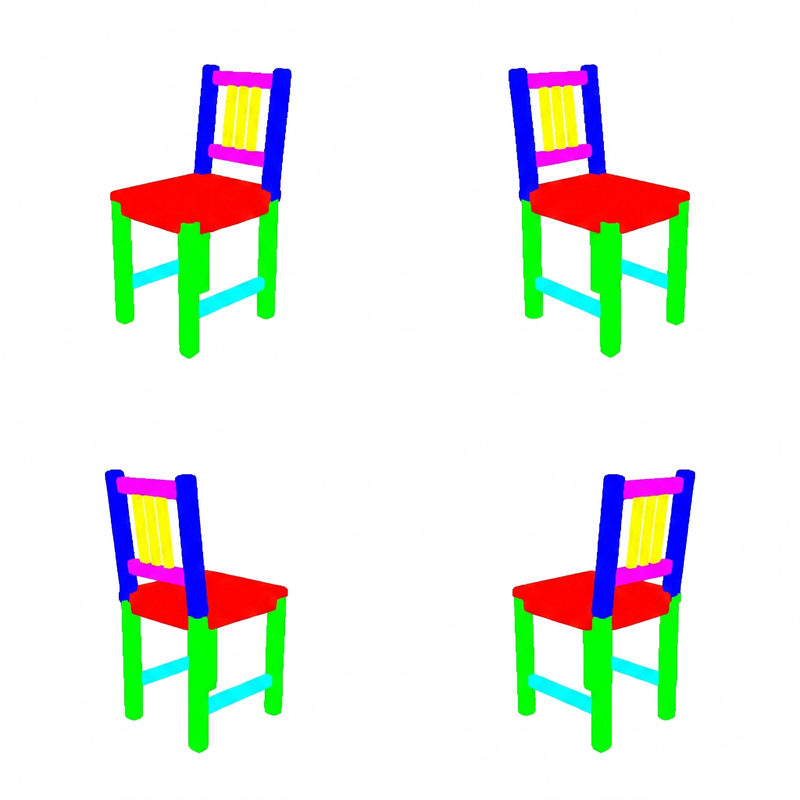}%
  \hfill
  \includegraphics[width=\segw]{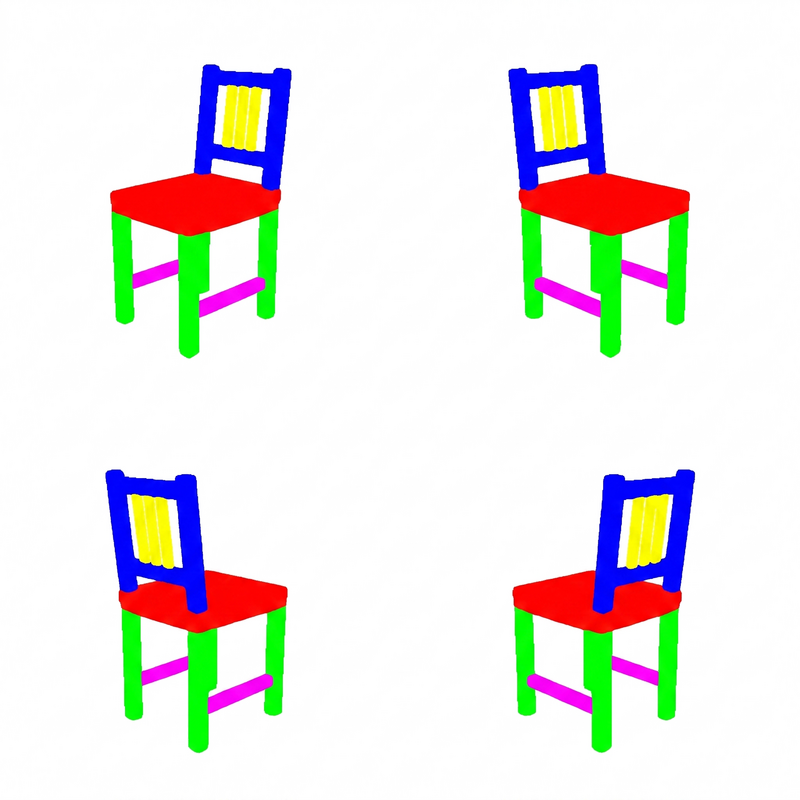}%
  \hfill
  \includegraphics[width=\segw]{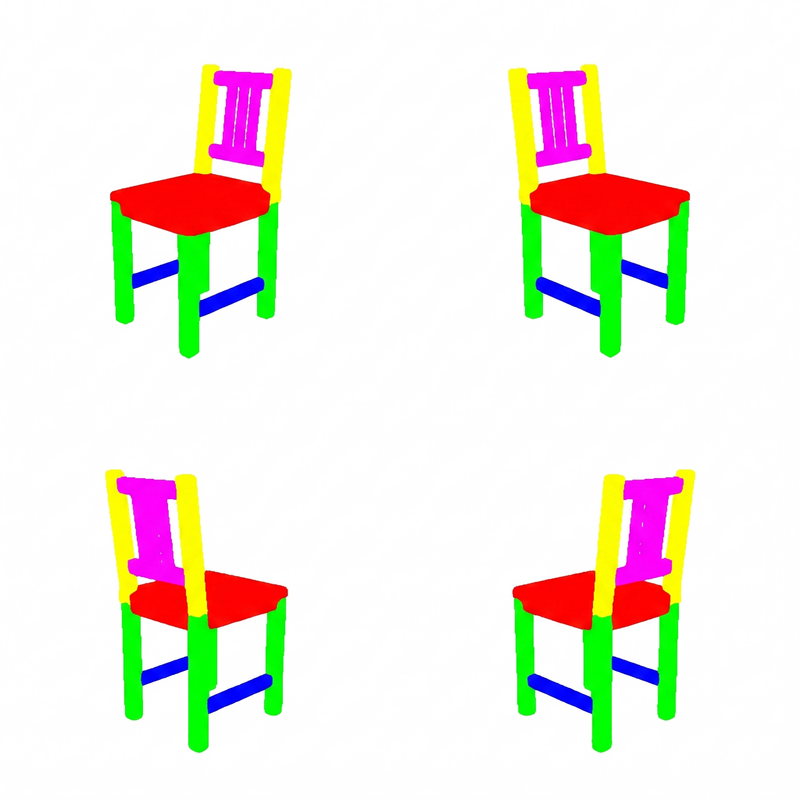}%
  \hfill
  \includegraphics[width=\segw]{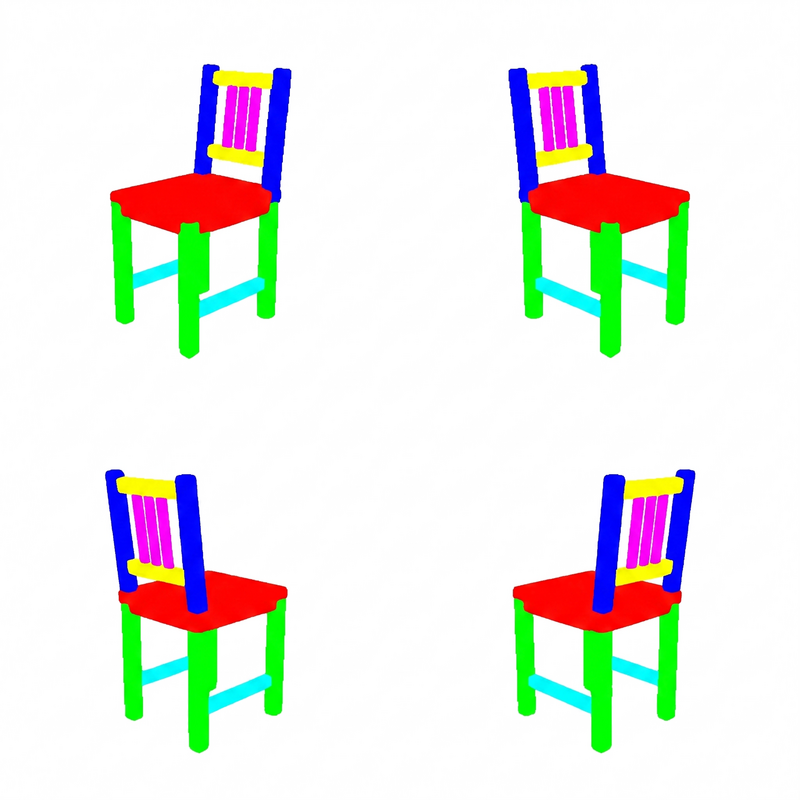}%

  \vspace{2pt}

  \includegraphics[width=\segw]{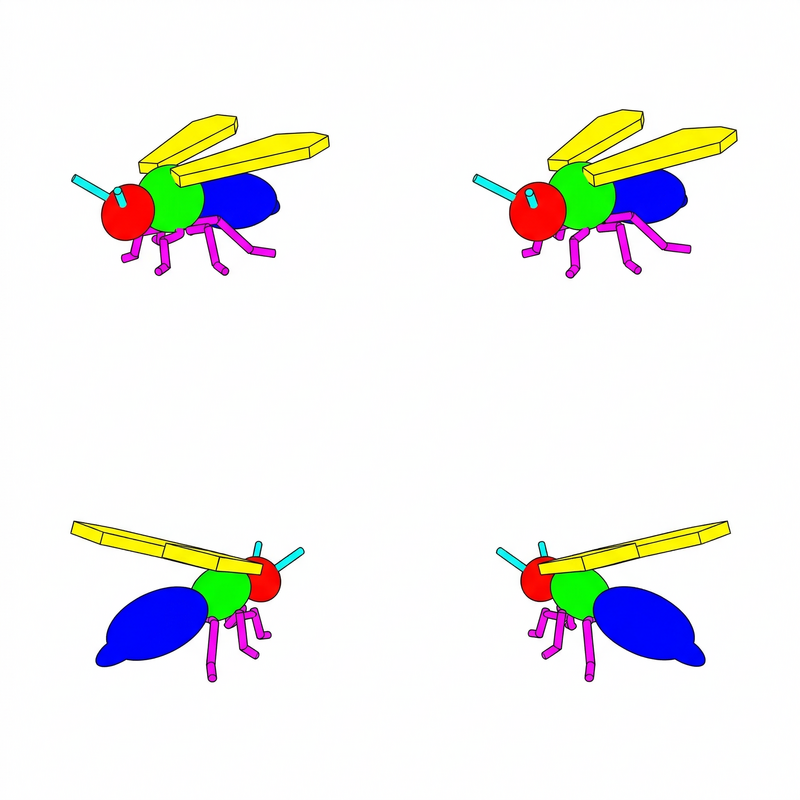}%
  \hfill
  \includegraphics[width=\segw]{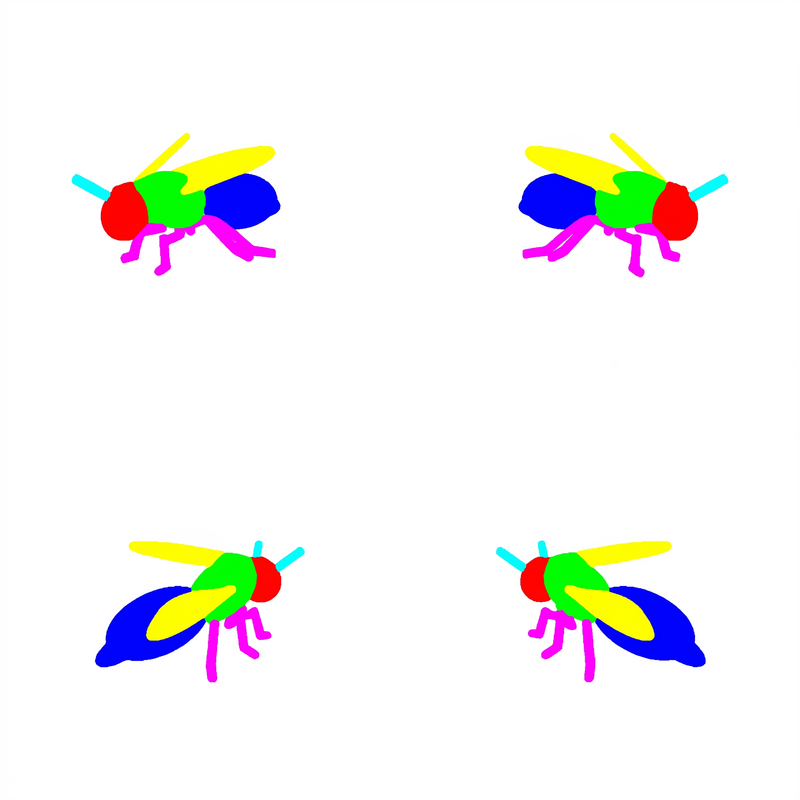}%
  \hfill
  \includegraphics[width=\segw]{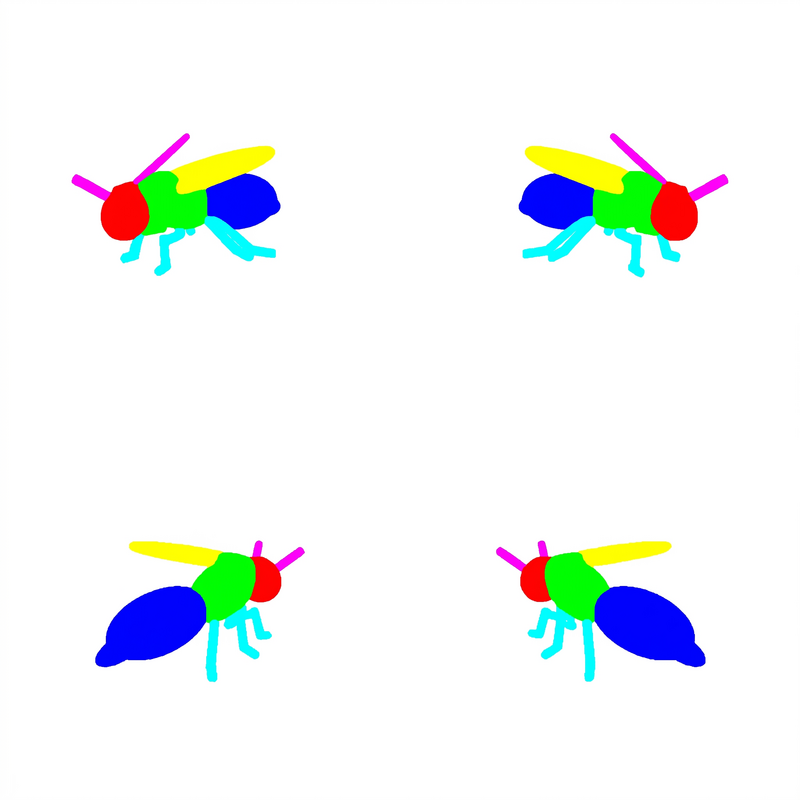}%
  \hfill
  \includegraphics[width=\segw]{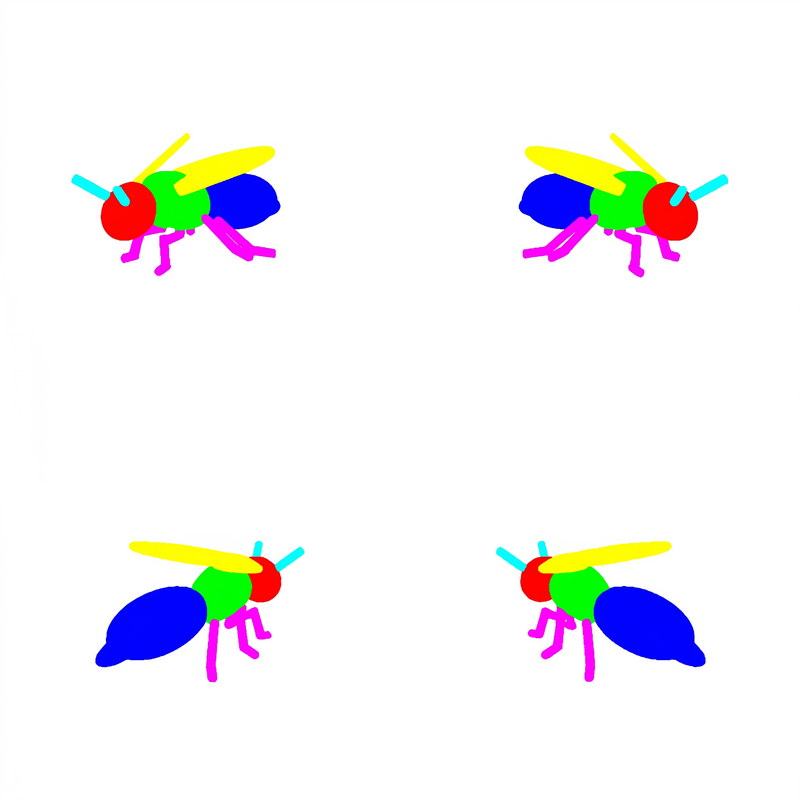}%
  \hfill
  \includegraphics[width=\segw]{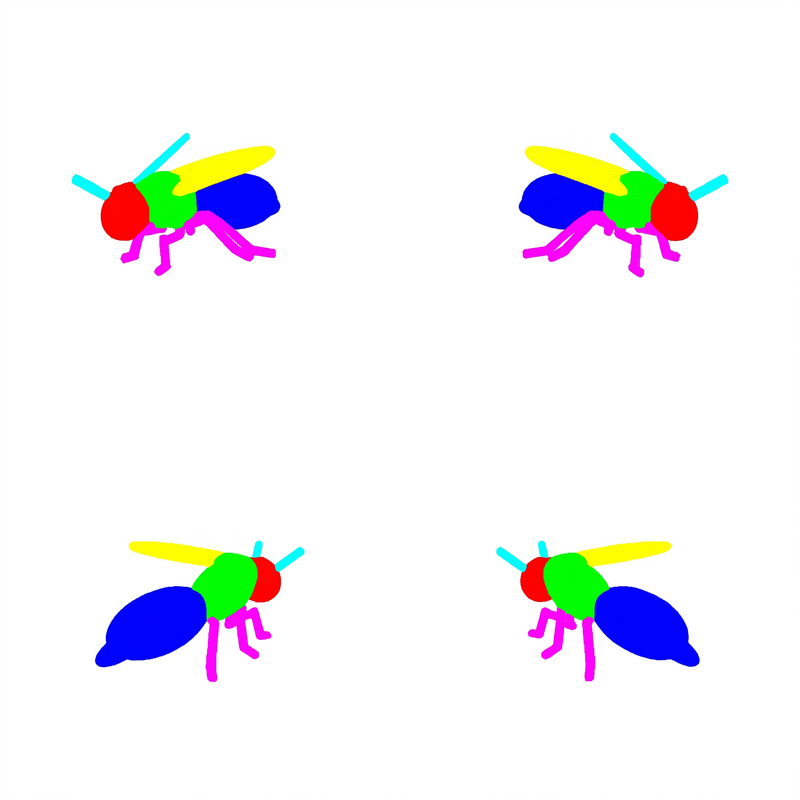}%

  \caption{\textbf{Run-to-run examples.}
    Part masks for a chair (top) and an insect (bottom) across five independent runs. Granularity and boundaries vary.}
  \label{fig:determinism}
\end{figure}

Three failure modes are most prominent (\Cref{fig:failure_cases}), and the supplementary material discusses further cases.
On the clustering side, clusters below a minimum size are discarded as outliers, which can inadvertently remove small but semantically important structures such as a thin chair leg.
We also do not control the granularity of the decomposition explicitly.
The number and coarseness of parts depend on the model's interpretation of the input, and repeated generation on the same input changes part granularity and boundary placement (\Cref{fig:determinism}).
Reliably steering it through prompting is untested.
On the generation side, large backgrounds can cause the model to hallucinate additional phantom views not present in the input.
Their votes fall outside the object silhouette and are discarded, so the abstraction is unaffected.
A practical limitation is the runtime of 40--50\,s per object, two thirds of it in the two model calls, far slower than feed-forward models.

\section{Conclusion}
\label{sec:conclusion}
Pretrained models carry visual part knowledge that can be reused for compact 3D primitive abstraction without task-specific training.
Generated part masks connect this knowledge to classical superquadric fitting: with five to eight primitives, our abstractions match the strongest learned baseline in Chamfer distance on HumanPrim, improve on it on Toys4K, and keep chair legs, rungs and wheels as separate parts.
On Toys4K, whose objects mostly lie outside the learned methods' training classes, these methods often lose appendages such as tails and limbs (\Cref{fig:qual_toys4k}), which our abstractions retain.

The controlled studies show that part knowledge alone is not sufficient.
The decomposition must match the granularity of the primitive representation, and unobserved surface limits how well the fitter recovers part thickness.
Because part prediction and fitting are separate, a better image generator can replace the current one without retraining any 3D model.

\bibliographystyle{eg-alpha}
\bibliography{references}

\setcounter{section}{0}\setcounter{figure}{0}\setcounter{table}{0}\setcounter{equation}{0}\setcounter{algorithm}{0}
\renewcommand{\thesection}{S\arabic{section}}
\renewcommand{\thefigure}{S\arabic{figure}}
\renewcommand{\thetable}{S\arabic{table}}
\renewcommand{\theequation}{S\arabic{equation}}
\renewcommand{\thealgorithm}{S\arabic{algorithm}}
\renewcommand{\theHsection}{S\arabic{section}}
\renewcommand{\theHfigure}{S\arabic{figure}}
\renewcommand{\theHtable}{S\arabic{table}}
\renewcommand{\theHequation}{S\arabic{equation}}
\renewcommand{\theHalgorithm}{S\arabic{algorithm}}
\raggedbottom
\vspace{2em}
\begin{center}{\Large\bfseries Supplementary Material}\end{center}
\noindent\textit{This supplement gives the superquadric formalism, hyperparameters and clustering algorithm, the fitter and selection ablations, the full view-layout study, the prompts and model comparison, the instance-color experiment, the ShapeNet results with the split by training classes, run-to-run variability, extended failure cases and future work. Sections, figures and tables carry the prefix S.}

\section{Superquadric Formalism}
\label{S-sec:sq_formalism}

A superquadric surface~\cite{barr1981superquadrics} is defined by the implicit inside-outside function:
\begin{equation}
  F(x, y, z) = \left( \left| \frac{x}{a_1} \right|^{\frac{2}{\epsilon_2}} + \left| \frac{y}{a_2} \right|^{\frac{2}{\epsilon_2}} \right)^{\frac{\epsilon_2}{\epsilon_1}} + \left| \frac{z}{a_3} \right|^{\frac{2}{\epsilon_1}} = 1
  \label{S-eq:sq_implicit}
\end{equation}
where $a_1, a_2, a_3$ control the size along each axis and the shape parameters $\epsilon_1, \epsilon_2 \in (0, 2]$ continuously interpolate between cuboids ($\epsilon \to 0$), ellipsoids ($\epsilon = 1$), cylinders, and octahedra ($\epsilon = 2$).
The corresponding parametric form is:
\begin{equation}
  \mathbf{s}(\eta, \omega) = \begin{pmatrix} a_1 \cos^{\epsilon_1}\!\eta \; \cos^{\epsilon_2}\!\omega \\ a_2 \cos^{\epsilon_1}\!\eta \; \sin^{\epsilon_2}\!\omega \\ a_3 \sin^{\epsilon_1}\!\eta \end{pmatrix}
  \label{S-eq:sq_parametric}
\end{equation}
with $\eta \in [-\frac{\pi}{2}, \frac{\pi}{2}]$ and $\omega \in [-\pi, \pi)$, where $\cos^{\epsilon}\!\theta = \text{sign}(\cos\theta)|\cos\theta|^{\epsilon}$ denotes the signed power function.

\begin{figure}[H]
  \centering
  \includegraphics[width=\columnwidth]{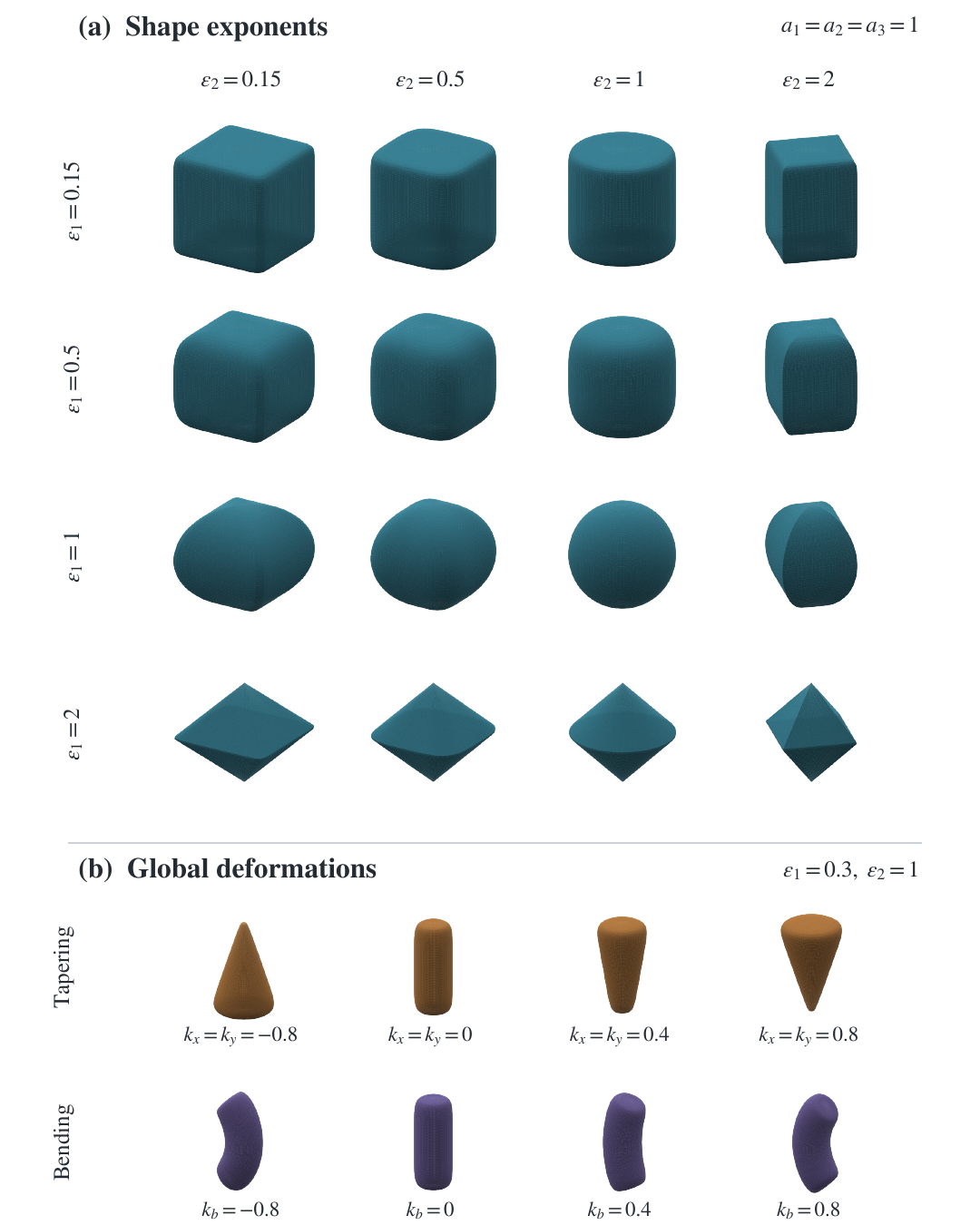}
  \caption{\textbf{Superquadric parameters visualization.}
  \textbf{(a)} Varying $\epsilon_1$ (rows) and $\epsilon_2$ (columns).
  \textbf{(b)} Tapering and bending of an elongated primitive with $(a_1,a_2,a_3)=(0.4,0.4,1)$; $\alpha_b=0$. Deformations are applied separately; zero is the undeformed reference.}
  \label{S-fig:sq_parameters}
\end{figure}

We additionally use the two global deformations of Solina and Bajcsy~\cite{solina1990recovery}.
Tapering scales the cross-section linearly along the $z$-axis via parameters $k_x, k_y$:
\begin{equation}
  \begin{pmatrix} x' \\ y' \\ z' \end{pmatrix} = \begin{pmatrix} x \cdot (k_x \frac{z}{a_3} + 1) \\ y \cdot (k_y \frac{z}{a_3} + 1) \\ z \end{pmatrix}
  \label{S-eq:taper}
\end{equation}
Bending curves the $z$-axis along an arc of radius $R = a_3 / k_b$ in the direction given by the angle $\alpha_b$ and models curved structures such as handles or bent limbs.
\Cref{S-fig:sq_parameters} illustrates the independent effects of the two shape exponents and of tapering and bending, with all remaining parameters fixed within each panel.

\section{Hyperparameters}
\label{S-sec:hyperparams}

We use the hyperparameters listed in \Cref{S-tab:hyperparams}. Selected parameters were explored in a sequential search on 40 shapes (20 HumanPrim and 20 Toys4K, one per category within each dataset). The remaining settings were fixed manually. The search column identifies which parameters were varied.

\begin{table*}[t]
  \centering
  \caption{Hyperparameters of the pipeline as used for all reported results. $\checkmark$: varied during parameter search. For exponents and deformations, the search varied whether optimization was enabled. Objects are normalized to the cube $[-1,1]^3$, and all lengths are given in these units.}
  \label{S-tab:hyperparams}
  \small
  \setlength{\tabcolsep}{4pt}
  \begin{tabular}{@{}p{0.12\textwidth}p{0.23\textwidth}p{0.49\textwidth}c@{}}
    \toprule
    Stage & Parameter & Value & Search \\
    \midrule
    \multirow{8}{*}{Rendering}
      & camera model & perspective, FOV $35^\circ$, distance $5$ & \\
      & viewpoints (4 views) & azimuth $45^\circ, 315^\circ, 225^\circ, 135^\circ$, elevation $15^\circ$ & \\
      & viewpoints (6 views) & the four ring views plus a top view (azimuth $90^\circ$, elevation $60^\circ$) and a bottom view ($270^\circ$, $-60^\circ$) & \\
      & resolution per view & $1024 \times 1024$ px & \\
      & composite & $2 \times 2$ grid ($2048 \times 2048$ px) or $3 \times 2$ grid ($3072 \times 2048$ px); white background & \\
      & point clouds & oriented splats (if normals available), else discs & \\
      & meshes & flat shading & \\
      & appearance & untextured gray, three-point lighting & \\
    \midrule
    \multirow{4}{*}{Segmentation}
      & analysis model & Gemini 3 Flash (text + image input) & \\
      & mask model & Nano Banana 2 (Gemini 3.1 Flash Image) & \\
      & output & \texttt{imageSize} 2K, aspect $1{:}1$ (four views); 4K, aspect $3{:}2$ (six views) & \\
      & prompts & see \emph{Segmentation Prompts} in this appendix & \\
    \midrule
    \multirow{4}{*}{Reprojection}
      & mask filtering & $3 \times 3$ median filter per view & \\
      & background & all channels $> 200$ (white) or $< 20$ (black) & \\
      & label assignment & per-point majority vote over the palette labels of the mask pixels it projects to & \\
      & palette merging & mean-shift in CIELAB, bandwidth $\Delta E = 40$ & \\
    \midrule
    \multirow{3}{*}{Clustering}
      & color quantization & 6 levels per RGB channel & \\
      & flood-fill radius $r$ & $0.05$ (KD-tree neighbor query) & $\checkmark$ \\
      & minimum cluster size & $1\,\%$ of the object's points & $\checkmark$ \\
    \midrule
    \multirow{12}{*}{Fitting}
      & input points per segment & $2048$ (random subsample) & \\
      & initializations & ApproxMVBB $\times$ \{cuboid, ellipsoid, cylinder\} $\times$ 3 axes & \\
      & surface samples $|\mathcal{S}|$ & $2500$ random points per iteration, area-weighted & \\
      & loss weight $\lambda$ & $0.1$ & $\checkmark$ \\
      & optimizer & L-BFGS, learning rate $1.0$ & \\
      & optimization iterations & $80$ & $\checkmark$ \\
      & pose and size & optimized & \\
      & shape exponents & optimized, $\epsilon_{1,2} \in [0.1, 1.9]$ & $\checkmark$ \\
      & deformations & taper and bend optimized & $\checkmark$ \\
      & latent mapping & $\tanh$ (taper, bend), clamping (size, $\epsilon$) & \\
      & minimum axis size & $2\,\%$ of the segment's largest OBB half-extent & \\
      & candidates & 9 poses $\times$ 5 deformation starts $= 45$ & \\
      & selection & largest axis product $a_1a_2a_3$ among the 10 lowest-cost candidates & $\checkmark$ \\
    \midrule
    \multirow{3}{*}{Metrics}
      & Chamfer distance & $2048$ surface samples per shape (FPS from 4096) & \\
      & volumetric IoU & $128^3$ grid, winding numbers & \\
      & HumanPrim ground truth & Poisson reconstruction (depth 7) of the 10k points & \\
    \bottomrule
  \end{tabular}
\end{table*}

\section{Clustering Algorithm}
\label{S-sec:clustering_alg}

\begin{algorithm}[htbp]
\caption{Color-restricted spatial clustering}
\label{S-alg:clustering}
\begin{algorithmic}[1]
\Require colored points $\{(\mathbf{x}_i, \mathbf{c}_i)\}_{i=1}^{N}$, radius $r$, minimum size $m$
\Ensure list of part point clouds
\State $\mathcal{G} \gets$ group indices by quantized color $\lfloor 6\,\mathbf{c}_i \rfloor$, drop background (unassigned gray)
\For{each color group $G \in \mathcal{G}$}
  \State build KD-tree on $\{\mathbf{x}_i\}_{i \in G}$; mark all $i \in G$ unvisited
  \For{each unvisited seed $s \in G$}
    \State $C \gets \{s\}$, queue $Q \gets [s]$, mark $s$ visited
    \While{$Q$ not empty}
      \State pop $i$ from $Q$; for every unvisited $j \in G$ with $\|\mathbf{x}_j - \mathbf{x}_i\| \le r$: mark visited, add $j$ to $C$ and $Q$
    \EndWhile
    \If{$|C| \ge m$} emit $C$ as one part \EndIf
  \EndFor
\EndFor
\end{algorithmic}
\end{algorithm}

\Cref{S-alg:clustering} lists the procedure used in \Cref{sec:clustering}.
It is single-linkage connected-component extraction on the $r$-neighborhood graph, run separately per quantized color.
With $r = 0.05$ and $m = 1\,\%$ of the points it takes well under a second per object.

\section{Minimum-Axis Floor}
\label{S-sec:fitter_floor}

The Chamfer objective of \Cref{sec:fitting} is a surface distance in both directions, so a segment that is observed from one side only (a table top, a mattress, a screen) admits a degenerate optimum.
A zero-thickness sheet on the observed surface has zero Chamfer distance but no volume.
We therefore floor every axis at a fixed fraction of the segment's largest oriented-bounding-box half-extent.
\Cref{S-tab:floor} varies this fraction on the 50 objects of the view-layout study.
A floor of $2\,\%$ leaves CD nearly unchanged ($0.074$ to $0.073$) and raises IoU by $2.1$ points with four views, where one-sided segments are frequent.
With six views, where the top and bottom views expose both sides of most parts, its aggregate effect is negligible ($69.2$ to $69.1\,\%$).
The median per-object change is zero in both layouts and 8 of the 50 objects lose slightly, i.e.\ the floor mainly binds in the degenerate cases and is neutral for most segments.
A larger floor of $5\,\%$ starts to hurt genuinely thin parts (screens, rotor blades, table tops seen from both sides).
CD rises and IoU drops by $1.5$ points with four views and by $1.0$ with six.

\begin{table}[h]
  \centering
  \small
  \begin{tabular}{@{}lccccccc@{}}
    \toprule
    & \multicolumn{3}{c}{4 ring views} & & \multicolumn{3}{c}{6 views (ring, top, bottom)} \\
    \cmidrule{2-4} \cmidrule{6-8}
    Floor & CD $\downarrow$ & IoU $\uparrow$ & \#P & & CD $\downarrow$ & IoU $\uparrow$ & \#P \\
    \midrule
    none        & 0.074 & 66.3\,\% & 7.5 & & \textbf{0.073} & \textbf{69.2}\,\% & 7.6 \\
    \rowcolor{gray!15}
    2\,\%       & \textbf{0.073} & \textbf{68.4}\,\% & 7.5 & & \textbf{0.073} & 69.1\,\% & 7.6 \\
    5\,\%        & 0.075 & 66.9\,\% & 7.5 & & 0.074 & 68.1\,\% & 7.6 \\
    \bottomrule
  \end{tabular}
  \caption{\textbf{Minimum-axis floor} as a fraction of the segment's largest bounding-box half-extent, 50 HumanPrim objects, same segmentations and fitter otherwise. The gray row is our configuration, bold marks the best value per column and layout.}
  \label{S-tab:floor}
\end{table}

\section{Fitter Ablations}
\label{S-sec:fitter_ablations}

Both ablations use the 50 objects of the view-layout study with the stored segmentations, so only the fitter changes.

\paragraph*{Multi-start initialization.}
\Cref{S-tab:multistart} compares the ApproxMVBB box~\cite{barequet2001mvbb} itself (no optimization), a single seed (cuboid, box orientation), the three cuboid orientations, all nine seeds (three shape families $\times$ three axes) without deformations, and the full configuration with 45 candidates.
Every stage helps.
Optimization from one seed closes more than half of the gap between the box and the full model, and the three orientations add $3.3$ and $2.2$ IoU points.
The ellipsoid and cylinder families add another $1$--$1.5$ points in IoU, and they are selected for $72$--$75\,\%$ of the parts.
Tapering and bending, with the 45 candidates of \Cref{sec:fitting}, improve both metrics in both layouts, CD from $0.079$ and $0.080$ to $0.073$ and IoU by $2.5$ and $3.3$ points.

\begin{table}[h]
  \centering
  \footnotesize
  \setlength{\tabcolsep}{3pt}
  \resizebox{\columnwidth}{!}{%
  \begin{tabular}{@{}lcccccc@{}}
    \toprule
    & & \multicolumn{2}{c}{4 ring views} & & \multicolumn{2}{c}{6 views} \\
    \cmidrule{3-4} \cmidrule{6-7}
    Initialization & seeds & CD $\downarrow$ & IoU $\uparrow$ & & CD $\downarrow$ & IoU $\uparrow$ \\
    \midrule
    bounding box & 1 & 0.113 & 52.1\,\% &  & 0.116 & 48.4\,\% \\
    single cuboid & 1 & 0.082 & 61.5\,\% &  & 0.082 & 62.1\,\% \\
    three cuboid orientations & 3 & 0.080 & 64.8\,\% &  & 0.081 & 64.3\,\% \\
    three families, rigid & 9 & 0.079 & 65.9\,\% &  & 0.080 & 65.8\,\% \\
    \rowcolor{gray!15}
    three families + taper/bend & 45 & \textbf{0.073} & \textbf{68.4}\,\% &  & \textbf{0.073} & \textbf{69.1}\,\% \\
    \bottomrule
  \end{tabular}%
  }
  \caption{\textbf{Multi-start ablation.} Means over 50 HumanPrim objects. The gray row is our configuration, bold marks the best value per column and layout.}
  \label{S-tab:multistart}
\end{table}

\paragraph*{Deformations.}
\Cref{S-tab:deformation} switches tapering and bending off individually.
With both, CD and IoU are best in both layouts, $2.5$ and $3.3$ IoU points above the rigid fit.
With four views, tapering alone recovers most of the gain and bending alone less. With six views, both deformations alone are on par, and the two combine in both layouts.

\begin{table}[h]
  \centering
  \footnotesize
  \setlength{\tabcolsep}{3pt}
  \resizebox{\columnwidth}{!}{%
  \begin{tabular}{@{}ccccccc@{}}
    \toprule
    & & \multicolumn{2}{c}{4 ring views} & & \multicolumn{2}{c}{6 views} \\
    \cmidrule{3-4} \cmidrule{6-7}
    Tapering & Bending & CD $\downarrow$ & IoU $\uparrow$ & & CD $\downarrow$ & IoU $\uparrow$ \\
    \midrule
    $\times$ & $\times$ & 0.079 & 65.9\,\% &  & 0.080 & 65.8\,\% \\
    $\checkmark$ & $\times$ & 0.075 & 67.5\,\% &  & 0.075 & 67.8\,\% \\
    $\times$ & $\checkmark$ & 0.077 & 66.1\,\% &  & 0.077 & 67.9\,\% \\
    \rowcolor{gray!15}
    $\checkmark$ & $\checkmark$ & \textbf{0.073} & \textbf{68.4}\,\% &  & \textbf{0.073} & \textbf{69.1}\,\% \\
    \bottomrule
  \end{tabular}%
  }
  \caption{\textbf{Deformation ablation.} Means over 50 HumanPrim objects. $\checkmark$ enabled, $\times$ disabled. The gray row is our configuration, bold marks the best value per column and layout. Primitive count is constant and omitted.}
  \label{S-tab:deformation}
\end{table}

\section{Candidate Selection}
\label{S-sec:selection}

The multi-start fitter of \Cref{sec:fitting} optimizes 45 candidates per segment and keeps one.
\Cref{S-tab:selection} compares three rules for that choice on the same optimized candidates.
The Chamfer objective is a surface criterion in both directions.
For a part observed from one side only (a seat, a backrest, a table top) it does not determine the thickness: a thin and a thick primitive that both follow the observed surface differ by a fraction of a percent in cost and several-fold in volume.
Keeping the lowest-cost candidate then decides the thickness by whichever candidate happens to converge marginally lower.
Preferring the largest volume among the ten lowest-cost candidates raises IoU by $2.8$ points on the 50 HumanPrim objects and by $1.3$ points on all 315 Toys4K objects, which were not used to choose the rule, while CD stays within $0.001$.
The same contrast between a surface-only and a surface-plus-volume criterion separates SuperDec from SuperFlex~\cite{fedele2025superdec,tavernini2026superflex}, where the volumetric term is trained rather than selected.
A softer variant, a cost threshold of $10\,\%$ instead of a fixed count of ten, recovers less, and a multiplicative volume reward on the cost recovers less still and becomes unstable for larger weights.

The oracle row shows how much the candidate set contains: choosing, per segment, the candidate that maximizes the object's IoU against the reference reaches $74.4\,\%$ on the 50 HumanPrim objects where the lowest-cost rule gives $65.6\,\%$.
The volume rule recovers about one third of that gap.
The remaining gap shows further headroom in candidate selection. Part of it is the thickness ambiguity of one-sided segments, which the six-view layout of \Cref{S-sec:view_layout_full} addresses at the input.

\begin{table}[h]
  \centering
  \small
  \setlength{\tabcolsep}{5pt}
  \begin{tabular}{@{}lcccccc@{}}
    \toprule
    & \multicolumn{2}{c}{HumanPrim (50)} & & \multicolumn{2}{c}{Toys4K (315)} & \\
    \cmidrule{2-3} \cmidrule{5-6}
    Rule & CD $\downarrow$ & IoU $\uparrow$ & & CD $\downarrow$ & IoU $\uparrow$ & $\Delta$seg \\
    \midrule
    \textsc{Cost}          & \textbf{0.073} & 65.6\,\% & & \textbf{0.083} & 60.1\,\% & -- \\
    \textsc{Vol-10\%}      & \textbf{0.073} & 67.3\,\% & & \textbf{0.083} & 60.9\,\% & 6.4 / 3.7 \\
    \rowcolor{gray!15}
    \textsc{Vol-Top10}     & \textbf{0.073} & \textbf{68.4}\,\% & & \textbf{0.083} & \textbf{61.4}\,\% & 7.2 / 4.3 \\
    \midrule
    \textsc{Oracle}        & 0.071 & 74.4\,\% & & -- & -- & 7.0 \\
    \bottomrule
  \end{tabular}
  \caption{\textbf{Candidate selection.} All rows share the same 45 optimized candidates per segment and differ only in which one is kept. \textsc{Cost}: lowest Chamfer cost. \textsc{Vol-10\%}: largest axis product $a_1a_2a_3$ among the candidates within $10\,\%$ of the lowest cost. \textsc{Vol-Top10}: largest axis product among the ten lowest-cost candidates (ours). \textsc{Oracle}: the candidate that maximizes the object's IoU against the reference, an upper bound computed on HumanPrim only. $\Delta$seg: mean number of segments per object whose winner differs from \textsc{Cost} (HumanPrim / Toys4K). Toys4K was not used to choose the rule. CD is computed against the stored fitting points and is therefore not comparable to \Cref{tab:results}.}
  \label{S-tab:selection}
\end{table}

\section{View Layout and Coverage}
\label{S-sec:view_layout_full}

\begin{table}[t]
\centering
\setlength{\tabcolsep}{4pt}
\resizebox{\columnwidth}{!}{%
\begin{tabular}{l c c c c c c c}
\toprule
Layout & $V$ & Mask output & VC $\uparrow$ & CD $\downarrow$ & IoU $\uparrow$ & \#P & s/obj \\
\midrule
single view            & 1  & 1K       & 46.5\,\% & 0.134 & 36.0\,\% & 5.7 & 33 \\
ring, 2 of 4 views     & 2  & --       & 65.3\,\% & 0.085 & 57.9\,\% & 7.0 & -- \\
ring, 3 of 4 views     & 3  & --       & 78.3\,\% & 0.076 & 66.3\,\% & 7.3 & -- \\
\rowcolor{gray!15}
ring                   & 4  & 2K       & 81.6\,\% & \textbf{0.073} & 68.4\,\% & 7.5 & 39 \\
\rowcolor{gray!15}
ring + top/bottom      & 6  & 4K (3:2) & 97.5\,\% & \textbf{0.073} & \textbf{69.1}\,\% & 7.6 & 52 \\
two rings + top        & 9  & 4K       & 99.1\,\% & 0.074 & 68.8\,\% & 7.3 & 56 \\
four rings             & 16 & 4K       & 99.8\,\% & 0.074 & 68.8\,\% & 7.2 & 55 \\
\bottomrule
\end{tabular}%
}
\caption{%
  \textbf{View layout.} Means over 50 HumanPrim objects, identical renderer, prompt and fitter.
  VC = view coverage, s/obj = time per object for the two model calls and the fitting.
  Gray rows are the layouts of \Cref{tab:results}, bold is best per column.
  Rows with 2 and 3 views use the first $k$ sequential masks of the ring layout, so mask output and runtime do not apply.
}
\label{S-tab:view_layout}
\end{table}

\paragraph*{Setup.}
We compare five camera layouts on the 50-object HumanPrim subset of \Cref{tab:view_coverage}:
(i) a single view, (ii) the four ring views of the main pipeline (azimuths $45^\circ, 135^\circ, 225^\circ, 315^\circ$ at $15^\circ$ elevation),
(iii) the ring plus a top ($+60^\circ$) and a bottom ($-60^\circ$) view, (iv) two rings of four at $+35^\circ$ and $-20^\circ$ plus a top view (9),
and (v) four rings of four at $\pm 60^\circ$ and $\pm 20^\circ$ (16).
Views are tiled into one composite ($2 \times 2$, $3 \times 2$, $3 \times 3$, $4 \times 4$) at $1024^2$ per tile.
The generative image model returns the composite at the closest supported resolution and aspect ratio (1K, 2K, 4K), so the mask resolution per view stays between $1024^2$ and about $1300^2$ for all layouts.
\emph{View coverage} is computed by ray casting the 10k surface points against a Poisson reconstruction of the object and counting a point as covered if it is inside the frustum and unoccluded in at least one view.
Prompt, reprojection, clustering and fitter are identical for all layouts.

\paragraph*{Results.}
\Cref{S-tab:view_layout} summarizes the outcome, \Cref{tab:view_coverage} in the main paper shows a subset of its columns.
Ring layouts saturate at about $80\,\%$ coverage regardless of how many views they contain, because no ring view sees undersides.
Only steep views add coverage.
Six views with top and bottom reach $97.5\,\%$ and improve IoU from $68.4$ to $69.1\,\%$.
Layouts with 9 and 16 views reach $99.1$ and $99.8\,\%$ and are indistinguishable from six ($\Delta$IoU $\le 0.3$ points, $\Delta$CD $\le 0.001$).
CD changes little because it is a surface metric and the visible surface is already well fitted.
Additional views help mainly where they expose new surface.

\paragraph*{Composite versus sequential masks.}
The pipeline generates one mask for the whole composite.
The alternative is one mask per view, generated sequentially with the previous views and masks in the model's context so that colors stay consistent.
On the same 50 objects the two variants are equivalent (ring layout, CD $0.073$ vs.\ $0.074$, IoU $68.4$ vs.\ $68.2\,\%$), while the composite needs one image-generation call instead of four.
Using only the first $k$ of the four sequential masks gives the rows with 2 and 3 views in \Cref{S-tab:view_layout}.
With $k = 1, 2, 3$ views the masks cover $46$, $65$ and $78\,\%$ of the surface.
The corresponding IoU values are $36$, $58$ and $66\,\%$, on the same curve as the composite layouts.
Per-view mask resolution is not the bottleneck either.
A 1K and a 2K composite give the same result.

\begin{figure}[t]
  \centering
  \includegraphics[width=\columnwidth]{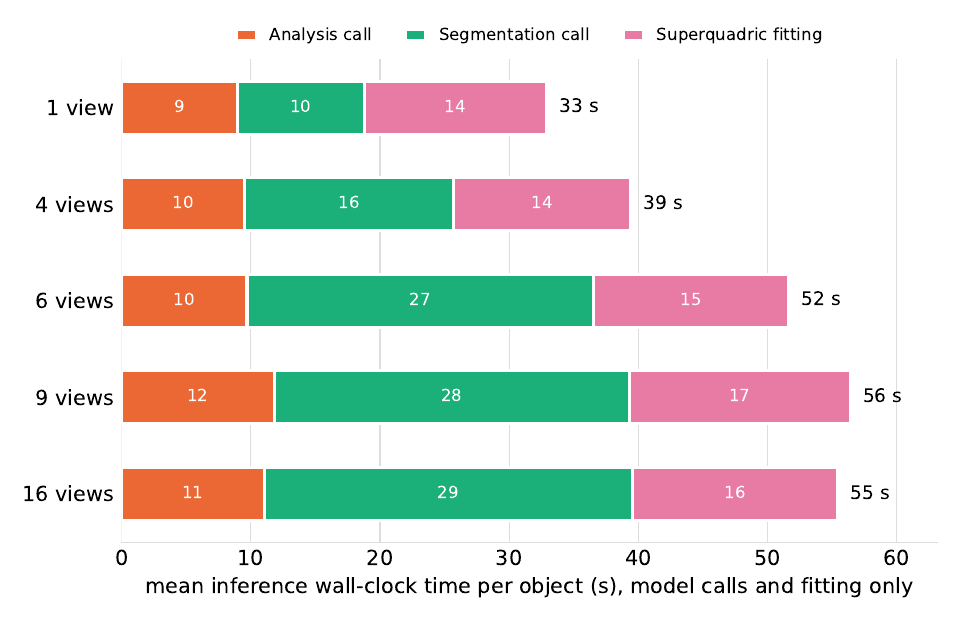}
  \caption{%
    \textbf{Inference time per object} by pipeline stage and view layout (means over 50 HumanPrim objects). Rendering, reprojection and metrics are not shown.
    The two model calls dominate.
  }
  \label{S-fig:runtime}
\end{figure}

\paragraph*{Cost and time.}
\Cref{S-fig:runtime} reports per-object times of $33$, $39$, $52$, $56$ and $55$\,s for 1, 4, 6, 9 and 16 views, counting the two model calls and the fitting ($14$--$17$\,s per object with 45 candidates per segment).
The model calls account for $57$--$71\,\%$ of it.
The mask call alone takes $10$--$29$\,s and grows with the output resolution class.
Rendering and reprojection are not counted.
Image-generation cost grows with the output resolution class (1K, 2K, 4K), so the six-view layout costs about $1.3\times$ the ring layout per object.

\paragraph*{Limitation.}
Steep bottom views are darker under the fixed lighting.
This has not caused segmentation errors in the cases we inspected, but a view-dependent light would remove the effect.
With full coverage, a second effect becomes visible.
Parts whose underside is hidden by other parts (a mattress on a bed frame) remain one-sided observations and are fitted as thin sheets unless the minimum-axis floor of \Cref{sec:fitting} binds (\Cref{S-sec:fitter_floor}).

\section{Segmentation Prompts and Model Comparison}
\label{S-sec:prompts}

The pipeline issues two sequential model calls per object, and both receive the same multi-view composite image as input.
The prompts are reproduced verbatim below.
The analysis prompt is layout-independent.
The segmentation prompt is a template in which the view count and grid size are substituted ($\{N\}=4$, $\{COLS\}\times\{ROWS\} = 2\times2$ for four views and $6$, $3\times2$ for six).

\begin{figure*}[!p]
  \centering
  \definecolor{pjred}{HTML}{FF0000}
  \definecolor{pjgrn}{HTML}{00FF00}
  \definecolor{pjblu}{HTML}{0000FF}
  \definecolor{pjylw}{HTML}{FFFF00}
  \definecolor{pjmag}{HTML}{FF00FF}
  \definecolor{pjcyn}{HTML}{00FFFF}
  \newcommand{\csqj}[1]{\tikz[baseline=0.25ex]\draw[fill=#1,draw=gray,line width=0.2pt](0,0)rectangle(1.5ex,1.5ex);}%
  \newcommand{\pipefboximg}[1]{\begin{minipage}[c]{0.28\linewidth}\centering\setlength{\fboxsep}{2pt}\setlength{\fboxrule}{0.5pt}\fcolorbox{gray!60}{white}{\includegraphics[width=\dimexpr\linewidth-5pt\relax,keepaspectratio]{#1}}\end{minipage}}%
  \newcommand{\pipeimg}[1]{\begin{minipage}[c]{0.28\linewidth}\centering\setlength{\fboxsep}{2pt}\setlength{\fboxrule}{0.5pt}\fcolorbox{white}{white}{\includegraphics[width=\dimexpr\linewidth-5pt\relax,keepaspectratio]{#1}}\end{minipage}}%
  \newcommand{\pipearrow}{\makebox[0.06\linewidth][c]{$\rightarrow$}}%

  \begin{minipage}[t]{0.48\textwidth}
  \centering
  {\small\bfseries Chair}\\[6pt]
  \begin{minipage}{\linewidth}
\begin{lstlisting}[style=jsonbox]
{
  "object": "chair",
  "parts": [
    {
      "name": "seat",
      "description": "A flat, square-shaped cuboid
        that forms the sitting surface of the chair.",
      "color": (*@\csqj{pjred}@*) "#FF0000"
    },
    {
      "name": "legs",
      "description": "Four vertical cuboids located at
        each corner of the seat that provide support
        and height.",
      "color": (*@\csqj{pjgrn}@*) "#00FF00"
    },
    {
      "name": "backrest_posts",
      "description": "Two vertical cuboids extending
        upwards from the rear of the seat to form the
        outer frame of the backrest.",
      "color": (*@\csqj{pjblu}@*) "#0000FF"
    },
    {
      "name": "backrest_rails",
      "description": "Horizontal cuboids that connect
        the backrest posts at the top and bottom to
        complete the backrest frame.",
      "color": (*@\csqj{pjylw}@*) "#FFFF00"
    },
    {
      "name": "backrest_slats",
      "description": "Multiple narrow vertical cuboids
        arranged within the backrest frame for back
        support.",
      "color": (*@\csqj{pjmag}@*) "#FF00FF"
    },
    {
      "name": "stretchers",
      "description": "Horizontal cuboids connecting the
        legs below the seat to provide additional
        structural stability.",
      "color": (*@\csqj{pjcyn}@*) "#00FFFF"
    }
  ]
}
\end{lstlisting}
  \end{minipage}
  \end{minipage}%
  \hfill
  \begin{minipage}[t]{0.48\textwidth}
  \centering
  {\small\bfseries Insect}\\[6pt]
  \begin{minipage}{\linewidth}
\begin{lstlisting}[style=jsonbox]
{
  "object": "bee",
  "parts": [
    {
      "name": "head",
      "description": "The spherical front section of
        the bee containing the eyes and antennae.",
      "color": (*@\csqj{pjred}@*) "#FF0000"
    },
    {
      "name": "thorax",
      "description": "The spherical middle section of
        the body where the legs and wings are
        attached.",
      "color": (*@\csqj{pjgrn}@*) "#00FF00"
    },
    {
      "name": "abdomen",
      "description": "The elongated spherical or
        ellipsoidal rear section of the bee's body.",
      "color": (*@\csqj{pjblu}@*) "#0000FF"
    },
    {
      "name": "wings",
      "description": "Thin, flat cuboid-shaped
        structures attached to the thorax for flight.",
      "color": (*@\csqj{pjylw}@*) "#FFFF00"
    },
    {
      "name": "legs",
      "description": "Multiple cylindrical segments
        extending from the thorax used for
        locomotion.",
      "color": (*@\csqj{pjmag}@*) "#FF00FF"
    },
    {
      "name": "antennae",
      "description": "Thin cylindrical sensory
        structures extending forward from the head.",
      "color": (*@\csqj{pjcyn}@*) "#00FFFF"
    }
  ]
}
\end{lstlisting}
  \end{minipage}
  \end{minipage}

  \vspace{10pt}

  {\setlength{\fboxsep}{2pt}\setlength{\fboxrule}{0.5pt}%
  \begin{tabular}{@{}%
    >{\centering\arraybackslash}m{0.14\textwidth}@{}%
    >{\centering\arraybackslash}m{0.03\textwidth}@{}%
    >{\centering\arraybackslash}m{0.14\textwidth}@{}%
    >{\centering\arraybackslash}m{0.03\textwidth}@{}%
    >{\centering\arraybackslash}m{0.14\textwidth}@{\hspace{0.04\textwidth}}%
    >{\centering\arraybackslash}m{0.14\textwidth}@{}%
    >{\centering\arraybackslash}m{0.03\textwidth}@{}%
    >{\centering\arraybackslash}m{0.14\textwidth}@{}%
    >{\centering\arraybackslash}m{0.03\textwidth}@{}%
    >{\centering\arraybackslash}m{0.14\textwidth}@{}}
  {\scriptsize\bfseries Input} & & {\scriptsize\bfseries Segmentation} & & {\scriptsize\bfseries Abstraction} &
  {\scriptsize\bfseries Input} & & {\scriptsize\bfseries Segmentation} & & {\scriptsize\bfseries Abstraction}\\[2pt]
  \fcolorbox{gray!60}{white}{\includegraphics[width=\dimexpr0.14\textwidth-5pt\relax,keepaspectratio]{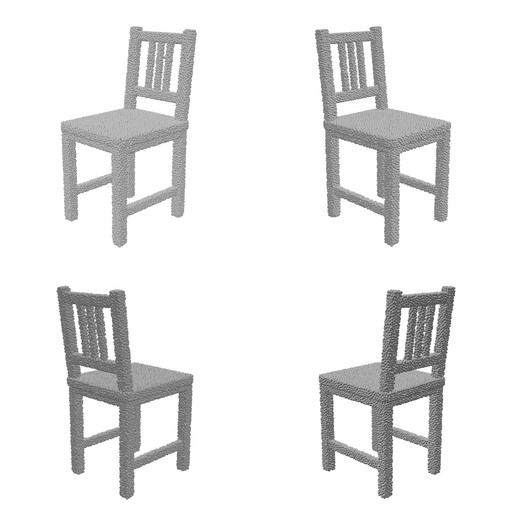}} &
  $\rightarrow$ &
  \fcolorbox{gray!60}{white}{\includegraphics[width=\dimexpr0.14\textwidth-5pt\relax,keepaspectratio]{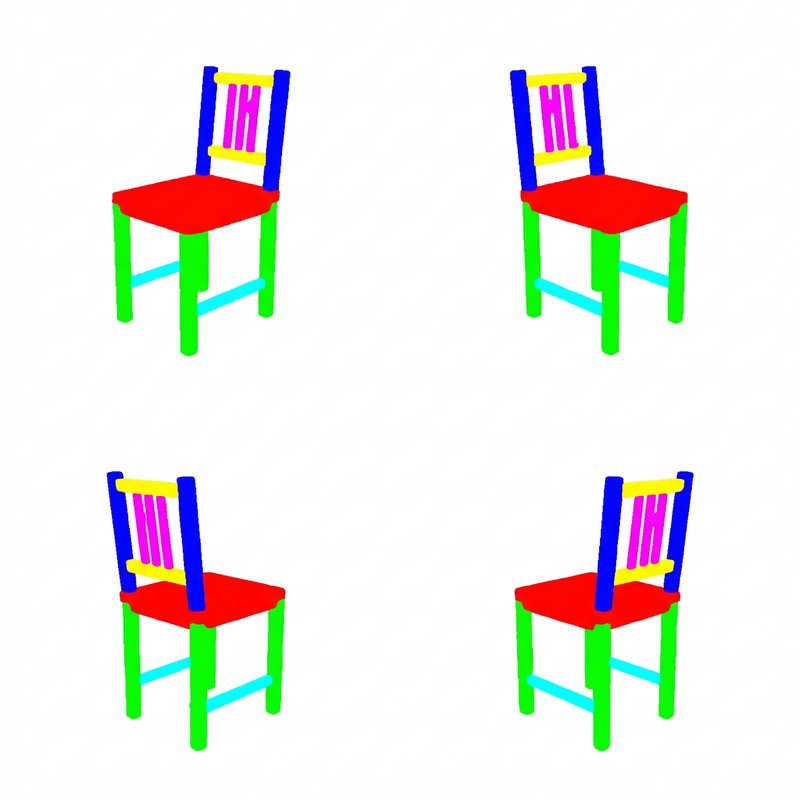}} &
  $\rightarrow$ &
  \includegraphics[width=0.14\textwidth,keepaspectratio]{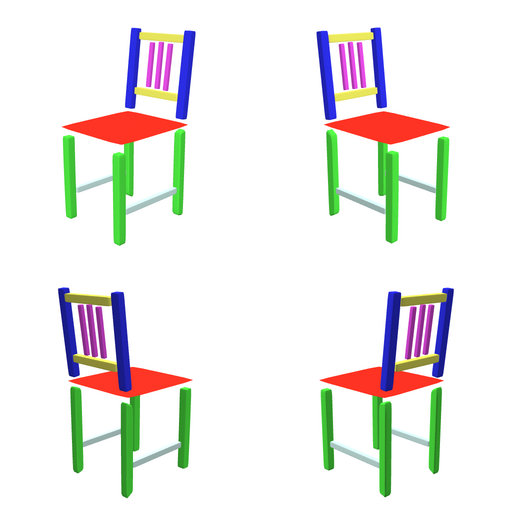} &
  \fcolorbox{gray!60}{white}{\includegraphics[width=\dimexpr0.14\textwidth-5pt\relax,keepaspectratio]{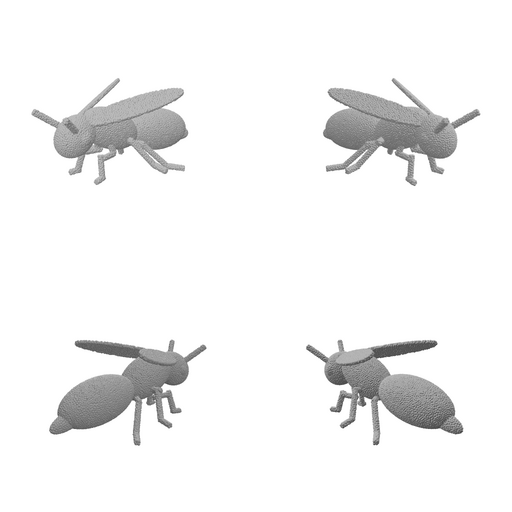}} &
  $\rightarrow$ &
  \fcolorbox{gray!60}{white}{\includegraphics[width=\dimexpr0.14\textwidth-5pt\relax,keepaspectratio]{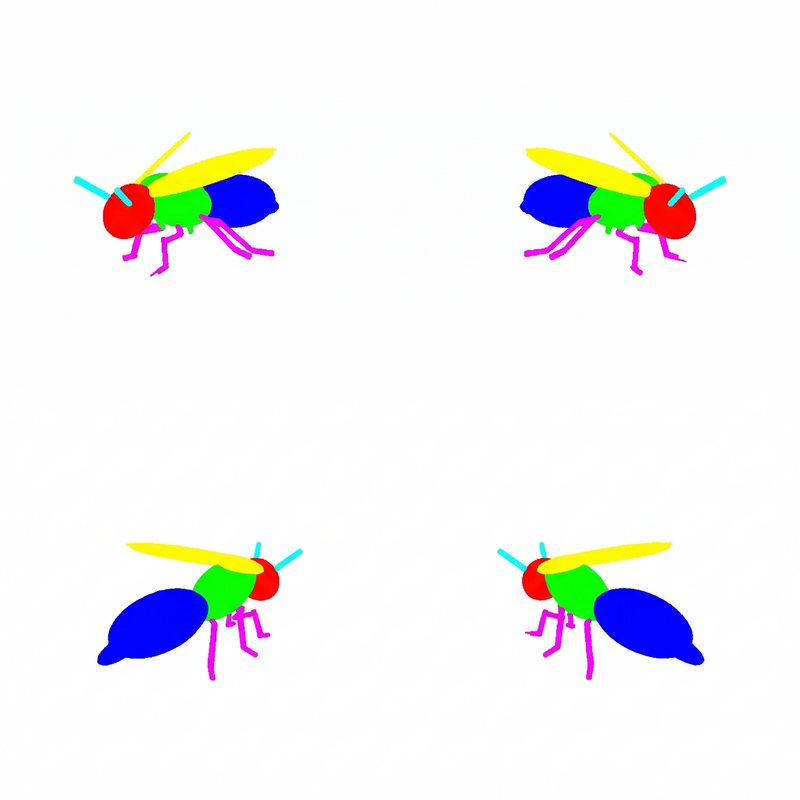}} &
  $\rightarrow$ &
  \includegraphics[width=0.14\textwidth,keepaspectratio]{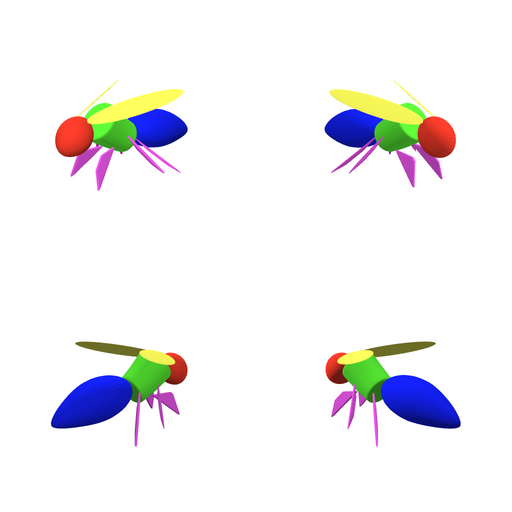}\\
  \end{tabular}}

  \caption{%
    \textbf{Vision-language model analysis output and pipeline results for two representative objects.}
    For each object, the box shows the structured JSON returned by Call~1 (Analysis)
    with the semantic parts, their exact descriptions and the assigned colors.
    The row below shows the four-view input, the part mask from the generative image model (Call~2) and the final superquadric abstraction.
  }
  \label{S-fig:humanprim_appendix}
\end{figure*}

\textbf{Call~1: Vision-Language Model Analysis.}
We ask a vision-language model to identify the depicted object and to propose a decomposition into semantically meaningful parts (\Cref{S-fig:humanprim_appendix}).
Each part receives a unique vibrant color other than white, which is reserved for the background.
The model returns a structured JSON with part names, short descriptions and hex color codes.
The prompt explicitly instructs the model not to distinguish left from right instances of symmetric parts, so all chair legs, for example, share one color.
This lets the subsequent reprojection step produce a consistent 3D labeling across opposing views.

\begin{lstlisting}[style=jsonbox]
# TASK

Analyze the following multi-view images of an object:
- What object is depicted?

Segment it into distinct parts.
- In which parts can it be subdivided, e.g. legs, seat, wings, arms, ...
- Subdivide the object using simple geometric shapes (cuboids, cylinders, spheres, ...)
- Each part should be represented by a single geometric shape
- If the object is composed of multiple instances of the same part, assign the same color to all instances.
- Do NOT differentiate between left and right instances of the same part


Assign a single color to each unique instance.
- Use vibrant colors like #FF0000 or #FF00FF
- Do NOT use #FFFFFF as this is the background color

# OUTPUT

Answer in a structured JSON format.
- Answer only with the JSON string, nothing else.

{
    "object": "",
    "parts": [
        {
            "name": "",
            "description": "",
            "color": ""
        }
    ]
}
\end{lstlisting}

\textbf{Call~2: Generative Image Model Mask Generation.}
We pass the same multi-view image to a generative image model together with the JSON output of Call~1 as a structured prefix.
The prompt instructs the model to paint a color-coded part mask over the input with exactly the colors specified in the JSON and to keep the colors consistent across all views.
It must not alter the object geometry, must use a white background, and must preserve the original aspect ratio.

\begin{lstlisting}[style=jsonbox]
{JSON Analysis Output}

# TASK

Generate a multi-view segmentation mask of the object:
- color the distinct parts exactly as defined above.
- maintain color consistency across all {N} perspective views (tracking the correct part as the camera moves around, above and below the object)
- the image is a grid of {COLS} columns x {ROWS} rows; each of the {N} cells shows exactly one view of the same object
- keep every view in its own cell at the same position, do NOT add, remove, merge or move views
- do NOT change the object in the image
- use a white background #FFFFFF
- keep the original aspect ratio
\end{lstlisting}

\begin{figure*}[p]
  \centering
  \providecommand{\vc}[1]{}\renewcommand{\vc}[1]{%
    \begin{minipage}[c]{0.185\textwidth}\centering
      \includegraphics[width=\linewidth,keepaspectratio]{#1}%
    \end{minipage}}
  \providecommand{\vh}[1]{}\renewcommand{\vh}[1]{%
    \makebox[0.185\textwidth][c]{\scriptsize #1}}
  \providecommand{\vho}[1]{}\renewcommand{\vho}[1]{%
    \makebox[0.185\textwidth][c]{\scriptsize\bfseries #1}}
  \providecommand{\vrulsep}{}\renewcommand{\vrulsep}{%
    \hspace{3pt}{\color{gray!60}\rule[-0.0925\textwidth]{0.5pt}{0.185\textwidth}}\hspace{3pt}}
  \providecommand{\vhsep}{}\renewcommand{\vhsep}{\hspace{6.5pt}}
  \providecommand{\priceval}[1]{}\renewcommand{\priceval}[1]{%
    \makebox[0.185\textwidth][c]{\scriptsize #1}}
  \providecommand{\pricena}{}\renewcommand{\pricena}{%
    \makebox[0.185\textwidth][c]{\scriptsize + VLM-output}}
  \makebox[0.185\textwidth][c]{\small\textit{Input}}\vhsep
  \makebox[0.802\textwidth][c]{\small\textit{Open-source}}\\[-3pt]
  {\color{gray!60}\rule{0.183\textwidth}{0.4pt}}\vhsep
  {\color{gray!60}\rule{0.800\textwidth}{0.4pt}}\\[2pt]
  \vh{Multi-View}\vhsep
  \vh{FLUX.2 Klein}\hfill
  \vh{HiDream-O1}\hfill
  \vh{Qwen Image Edit+}\hfill
  \vh{HunyuanImage 3.0 Instruct}\\[2pt]
  \pricena\vhsep
  \priceval{\$0.017}\hfill
  \priceval{\$0.010}\hfill
  \priceval{\$0.060}\hfill
  \priceval{\$0.090}\\[3pt]
  \vc{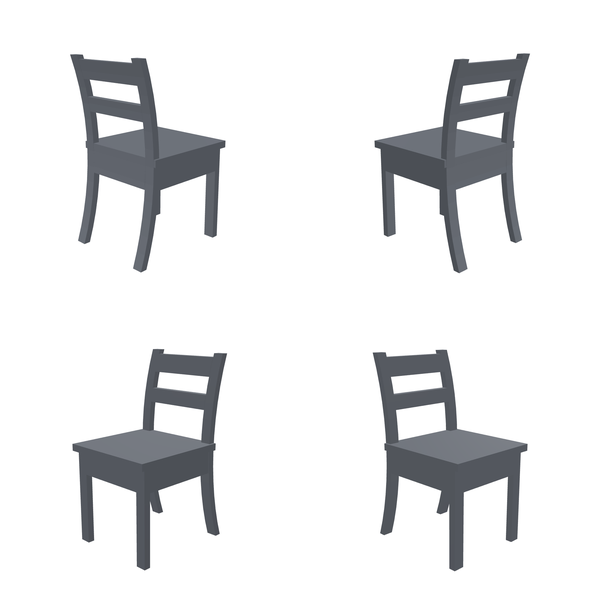}\vrulsep
  \vc{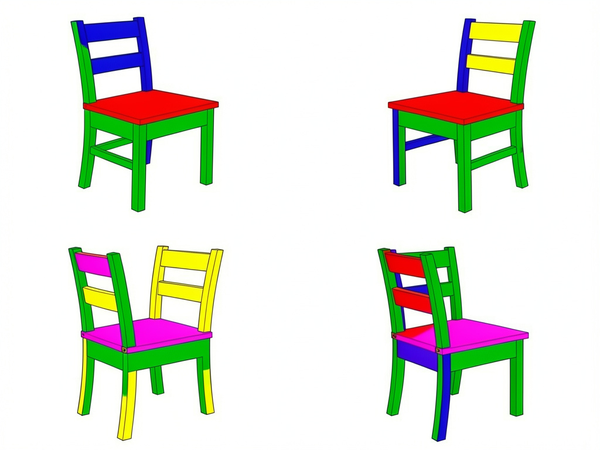}\hfill
  \vc{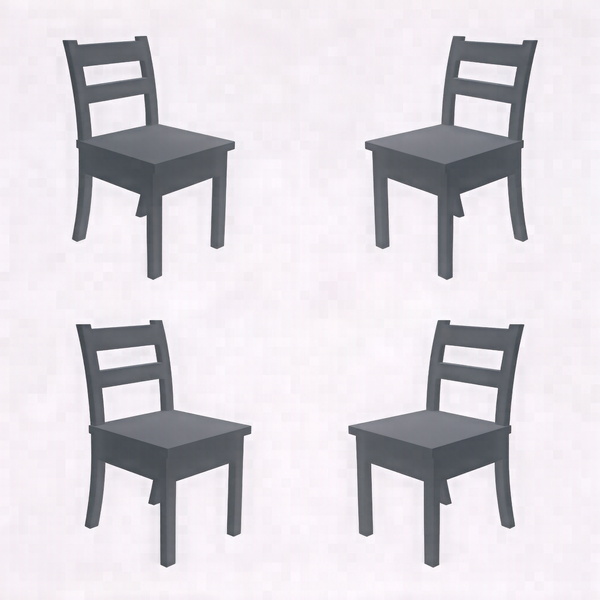}\hfill
  \vc{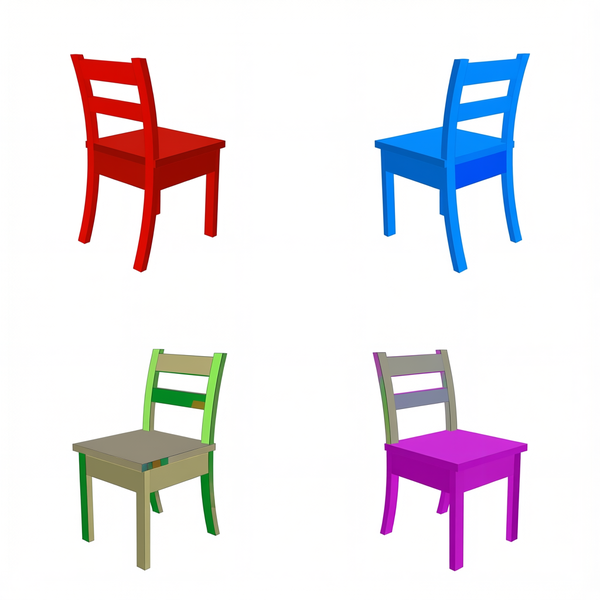}\hfill
  \vc{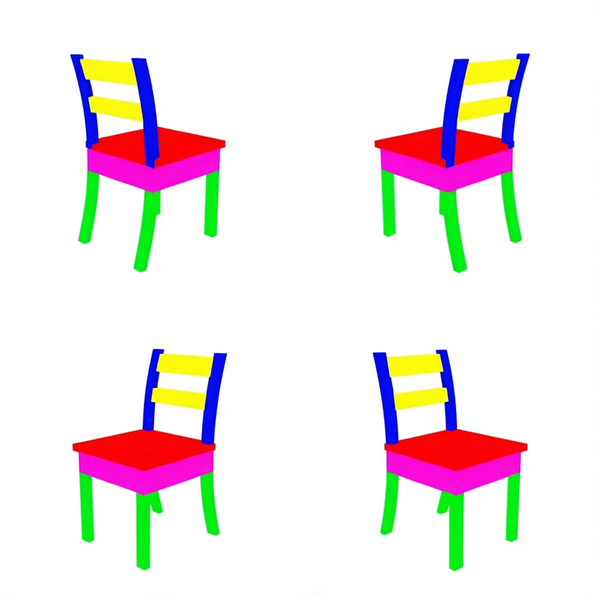}\\[4pt]
  \vc{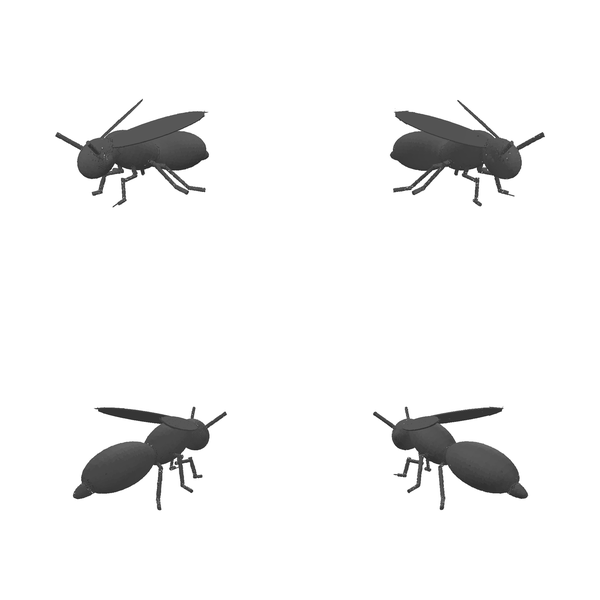}\vrulsep
  \vc{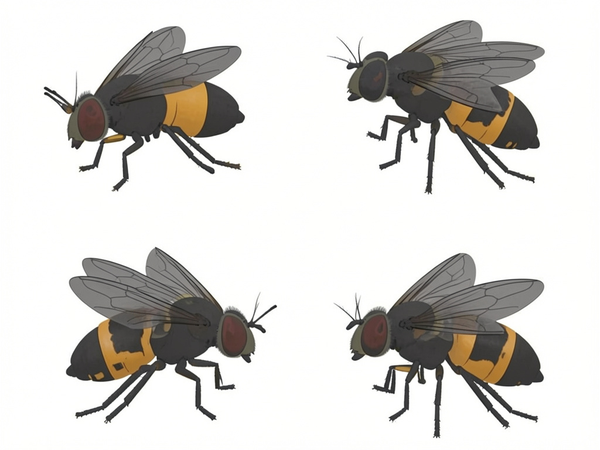}\hfill
  \vc{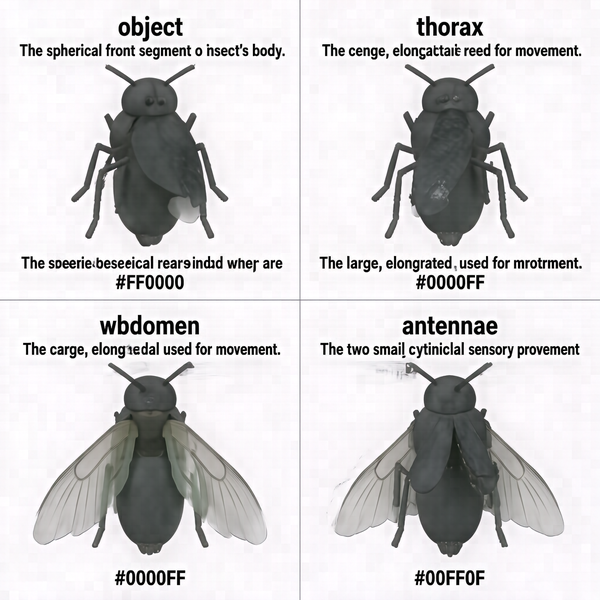}\hfill
  \vc{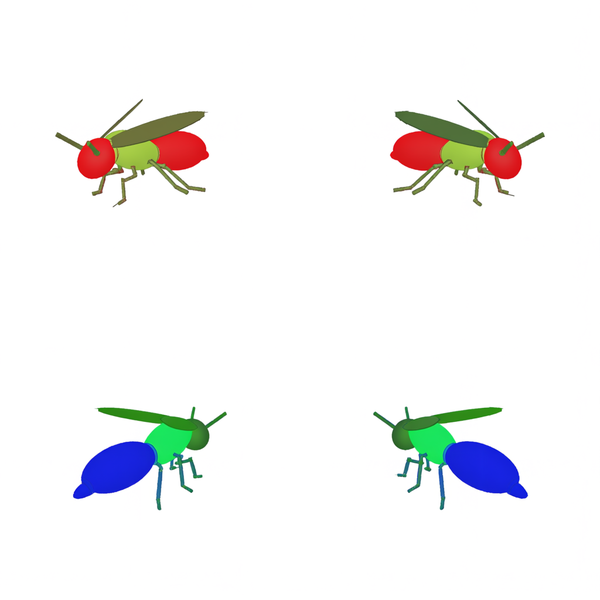}\hfill
  \vc{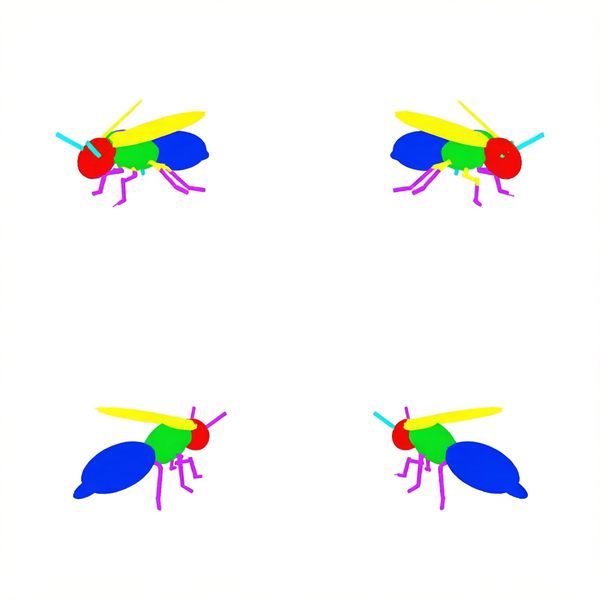}\\[10pt]
  \makebox[0.985\textwidth][c]{\small\textit{Proprietary}}\\[-3pt]
  {\color{gray!60}\rule{0.985\textwidth}{0.4pt}}\\[2pt]
  \vh{FLUX.2 Pro}\hfill
  \vh{GPT Image 1.5}\hfill
  \vh{GPT Image 2.0}\hfill
  \vh{Nano Banana Pro}\hfill
  \vho{Nano Banana 2}\\[2pt]
  \priceval{\$0.045}\hfill
  \priceval{\$0.133}\hfill
  \priceval{\$0.211}\hfill
  \priceval{\$0.134}\hfill
  \priceval{\$0.067}\\[3pt]
  \vc{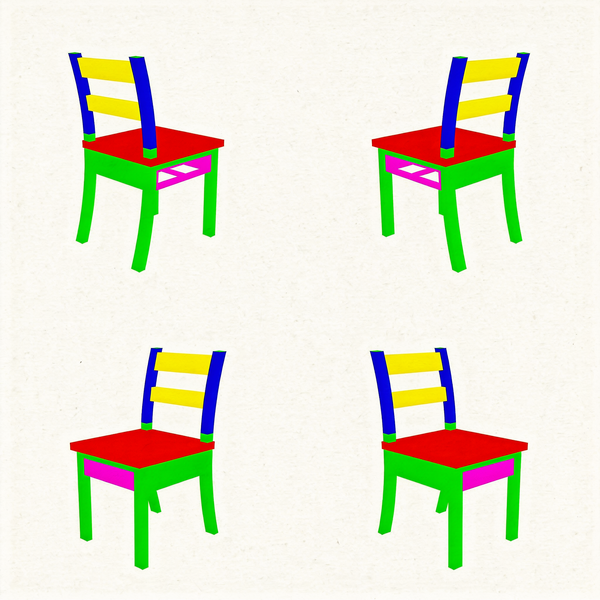}\hfill
  \vc{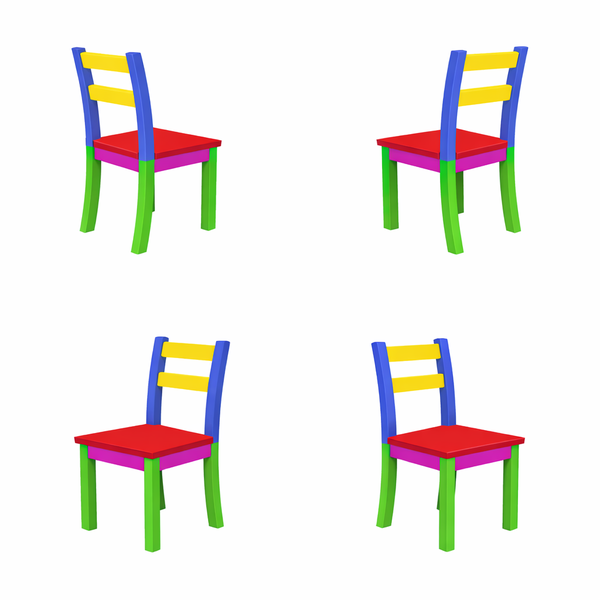}\hfill
  \vc{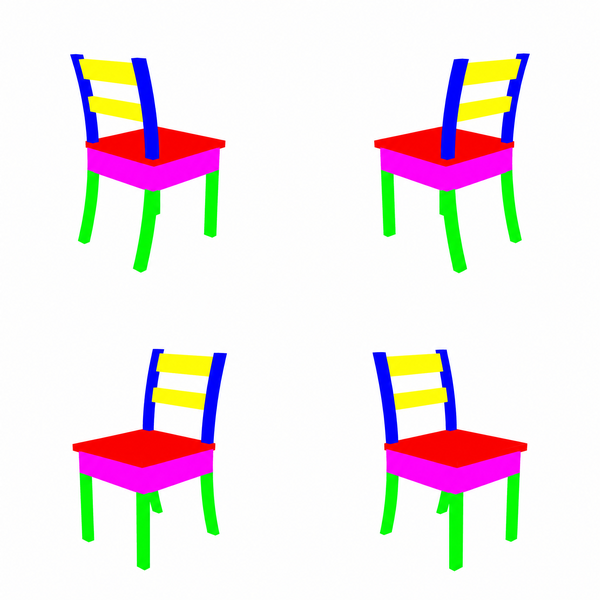}\hfill
  \vc{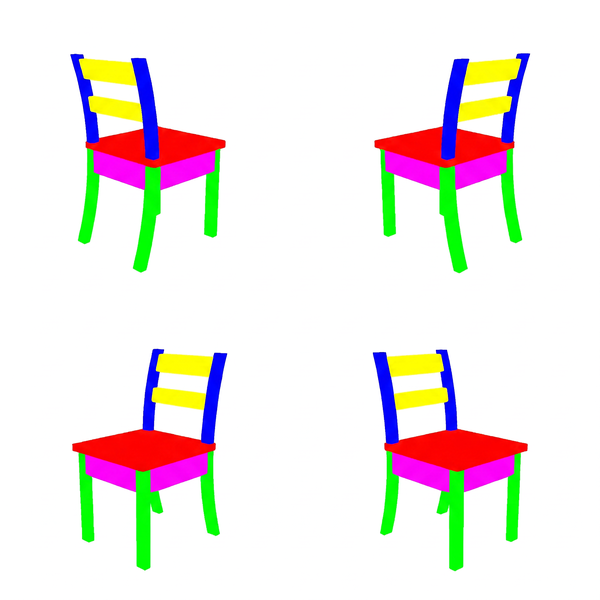}\hfill
  \vc{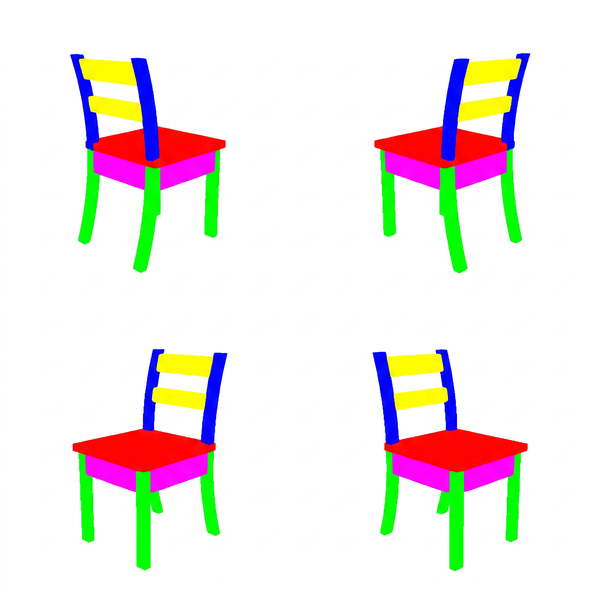}\\[4pt]
  \vc{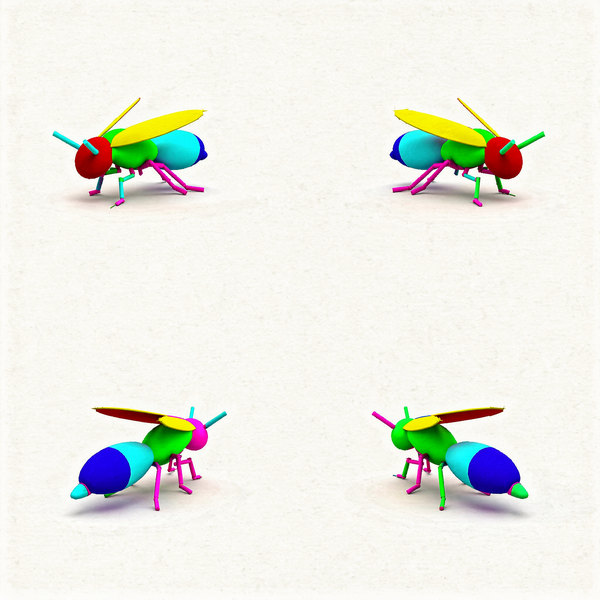}\hfill
  \vc{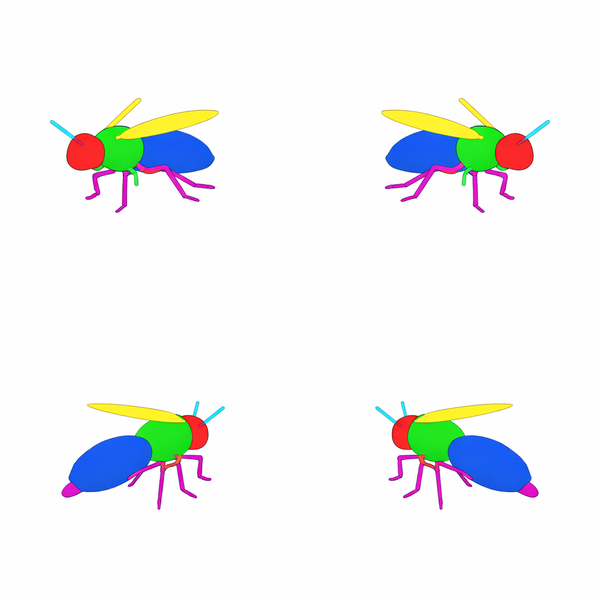}\hfill
  \vc{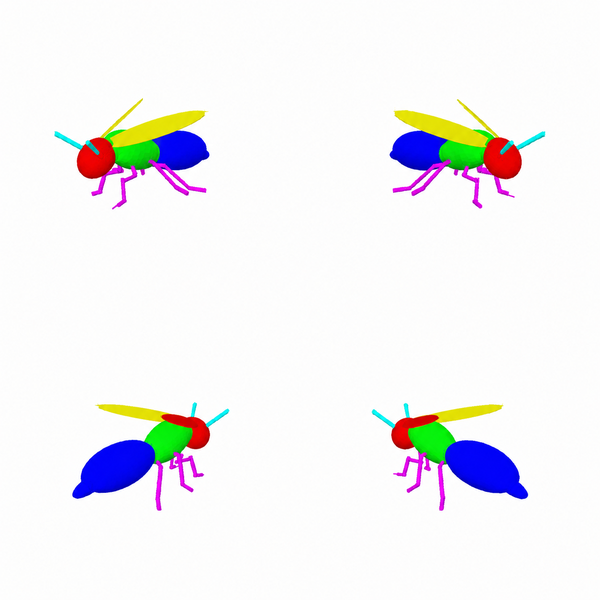}\hfill
  \vc{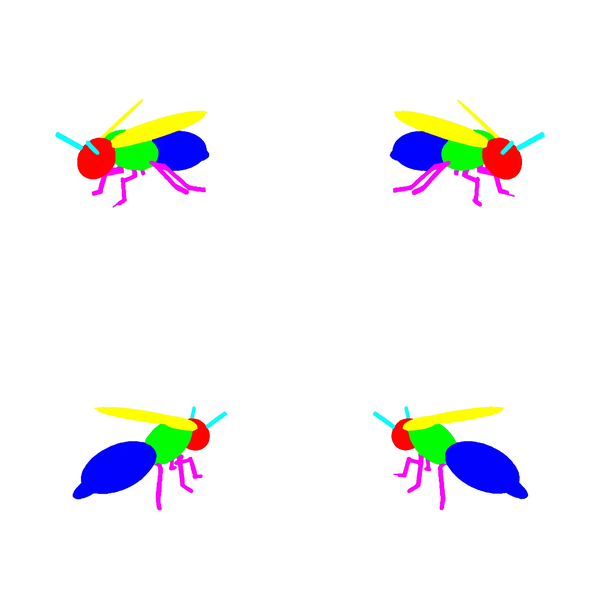}\hfill
  \vc{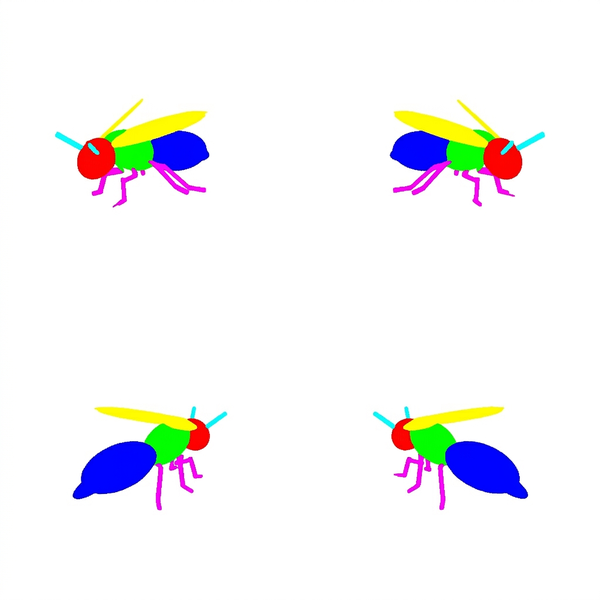}%
  \caption{%
    \textbf{Qualitative comparison of generative image models on part-mask generation.}
    Color-coded part masks for two hand-picked objects (rows within each block), grouped into open-source and proprietary models.
    All models receive the identical input and the same prompt (\Cref{S-sec:prompts}).
    The price above each column is the cost per edited image (USD), taken from the Artificial Analysis website~\cite{artificialanalysis}.
    Among the open-source models only HunyuanImage 3.0 Instruct produces usable masks. Among the proprietary models GPT Image and Nano Banana do, and Nano Banana 2 gives the cleanest masks at the lowest price among the successful models on these two objects. We use it for all experiments.
  }
  \label{S-fig:vlm_comparison}
\end{figure*}

Mask generation is the harder of the two calls, because the model must follow the given palette exactly and keep it consistent across views.
We evaluated nine candidate models on this task with the identical two-call prompt (\Cref{S-fig:vlm_comparison}).
On the two objects shown, five models from three families succeed.
GPT Image 1.5 and 2.0, Nano Banana Pro and 2, and HunyuanImage 3.0 preserve a single color per part across all four views and respect part boundaries.
The remaining models fail in one or more ways, namely inconsistent colors across opposing views, hallucinated phantom views, color bleed across boundaries, or geometry alteration.
Among these, Nano Banana 2~\cite{nanobana2} gave the cleanest masks at the lowest price per image, and we use it for all results in this paper.

\section{Instance Segmentation Ablation}
\label{S-sec:instance_seg}

Our pipeline assigns one color per semantic part \emph{type}, so all instances of the same part share a color.
As an ablation, we test whether asking the model to assign a unique color per instance improves part separation.
We replace the corresponding Call~1 prompt instructions as follows:

\begin{lstlisting}[style=jsonbox]
...
(*@{\scriptsize\ttfamily\sout{- If the object is composed of multiple instances of the same}}@*)
(*@{\scriptsize\ttfamily\sout{  part, assign the same color to all instances.}}@*)
(*@{\scriptsize\ttfamily\sout{- Do NOT differentiate between left and right instances.}}@*)
(*@{\scriptsize\ttfamily\textcolor{green!50!black}{- Differentiate multiple instances of the same category,}}@*)
(*@{\scriptsize\ttfamily\textcolor{green!50!black}{  like left and right or front and back.}}@*)
...
\end{lstlisting}

\begin{figure}[!t]
\centering
\definecolor{isRed}{HTML}{FF0000}
\definecolor{isGrn}{HTML}{00FF00}
\definecolor{isBlu}{HTML}{0000FF}
\definecolor{isYlw}{HTML}{FFFF00}
\definecolor{isMag}{HTML}{FF00FF}
\definecolor{isCyn}{HTML}{00FFFF}
\definecolor{isOrg}{HTML}{FFA500}
\definecolor{isPrp}{HTML}{800080}
\definecolor{isDkg}{HTML}{008000}
\definecolor{isPnk}{HTML}{FFC0CB}
\definecolor{isBrn}{HTML}{A52A2A}
\definecolor{isAqu}{HTML}{7FFFD4}
\definecolor{isInk}{HTML}{4B0082}
\definecolor{isLim}{HTML}{7FFF00}
\definecolor{isChc}{HTML}{D2691E}
\definecolor{isRdO}{HTML}{FF4500}
\newcommand{\csi}[1]{\tikz[baseline=0.25ex]\draw[fill=#1,draw=gray,line width=0.2pt](0,0)rectangle(1.5ex,1.5ex);}%

{\small\bfseries Instance Segmentation Output of Call~1 for a Chair}\\[4pt]
\begin{lstlisting}[style=jsonbox]
{ "object": "chair", "parts": [
  { "name": "seat",                 "color": (*@\csi{isRed}@*) "#FF0000" },
  { "name": "front left leg",       "color": (*@\csi{isGrn}@*) "#00FF00" },
  { "name": "front right leg",      "color": (*@\csi{isBlu}@*) "#0000FF" },
  { "name": "back left leg",        "color": (*@\csi{isYlw}@*) "#FFFF00" },
  { "name": "back right leg",       "color": (*@\csi{isMag}@*) "#FF00FF" },
  { "name": "left backrest post",   "color": (*@\csi{isCyn}@*) "#00FFFF" },
  { "name": "right backrest post",  "color": (*@\csi{isOrg}@*) "#FFA500" },
  { "name": "top backrest rail",    "color": (*@\csi{isPrp}@*) "#800080" },
  { "name": "bottom backrest rail", "color": (*@\csi{isDkg}@*) "#008000" },
  { "name": "backrest slat 1",      "color": (*@\csi{isPnk}@*) "#FFC0CB" },
  { "name": "backrest slat 2",      "color": (*@\csi{isBrn}@*) "#A52A2A" },
  { "name": "backrest slat 3",      "color": (*@\csi{isAqu}@*) "#7FFFD4" },
  { "name": "front leg stretcher",  "color": (*@\csi{isInk}@*) "#4B0082" },
  { "name": "back leg stretcher",   "color": (*@\csi{isLim}@*) "#7FFF00" },
  { "name": "left side stretcher",  "color": (*@\csi{isChc}@*) "#D2691E" },
  { "name": "right side stretcher", "color": (*@\csi{isRdO}@*) "#FF4500" } ] }
\end{lstlisting}

\vspace{4pt}

{\small\bfseries Generative Image Model Segmentation (Call~2)}\\[4pt]
\setlength{\tabcolsep}{1pt}
\renewcommand{\arraystretch}{0.5}
\begin{tabular}{ccc}
  {\setlength{\fboxsep}{1pt}\setlength{\fboxrule}{0.5pt}\fcolorbox{gray!60}{white}{\includegraphics[width=0.31\columnwidth]{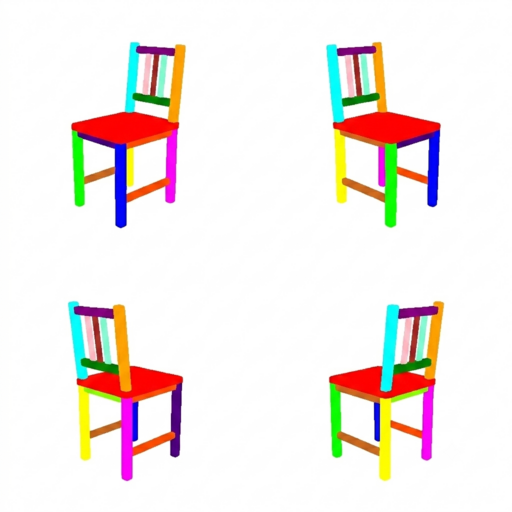}}} &
  \includegraphics[width=0.31\columnwidth]{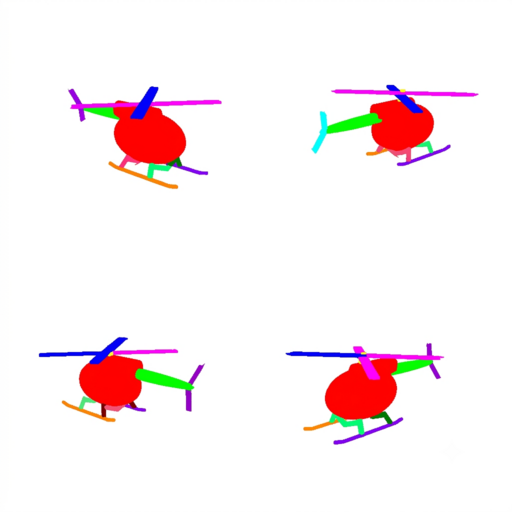} &
  \includegraphics[width=0.31\columnwidth]{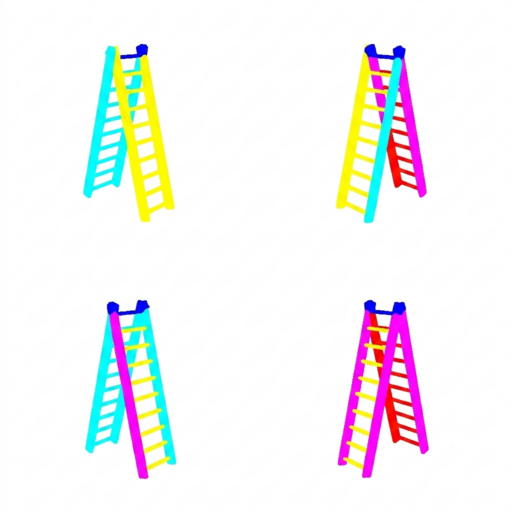} \\
  \includegraphics[width=0.31\columnwidth]{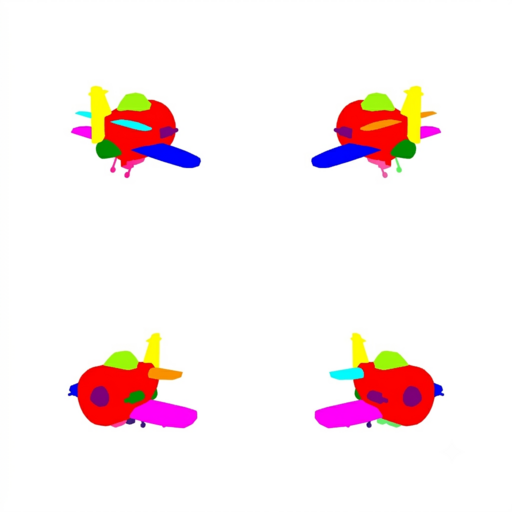} &
  \includegraphics[width=0.31\columnwidth]{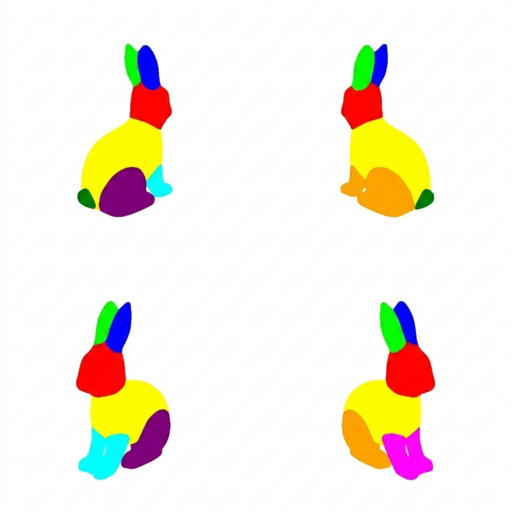} &
  \includegraphics[width=0.31\columnwidth]{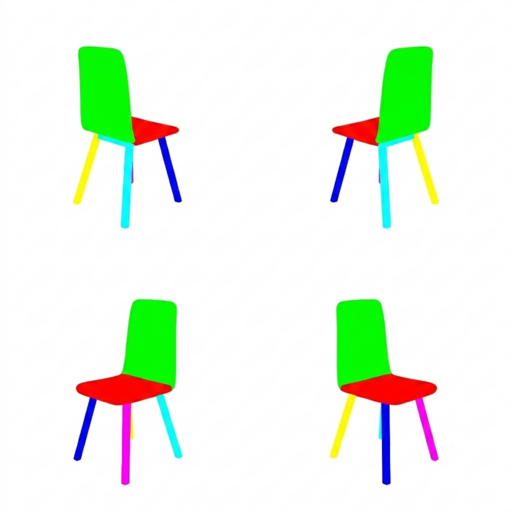} \\
  \includegraphics[width=0.31\columnwidth]{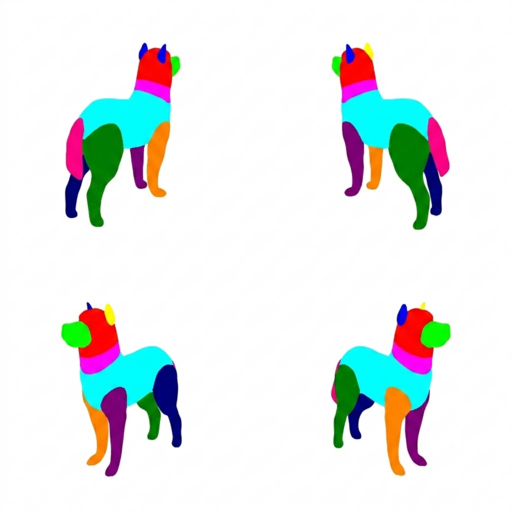} &
  \includegraphics[width=0.31\columnwidth]{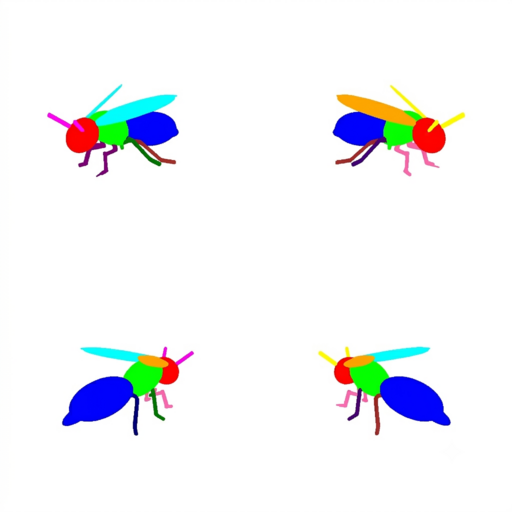} &
  \includegraphics[width=0.31\columnwidth]{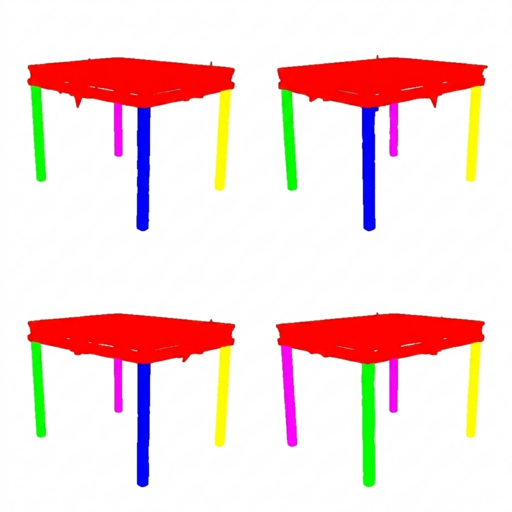} \\
\end{tabular}

\caption{%
  \textbf{Instance segmentation ablation.}
  An explicit instance-segmentation prompt assigns a distinct color to each part
  instance (16 colors for the chair shown above).
  The resulting masks (top left) show that the generative image model cannot
  reliably track instance identities across views. Left/right and front/back
  relationships are especially likely to be swapped or lost.
  We therefore assign a single color per semantic part \emph{type} instead.%
}
\label{S-fig:instance_seg}
\end{figure}

Call~1 reliably adopts the new instruction.
The chair example in \Cref{S-fig:instance_seg} receives 16 distinct colors, and each leg, stretcher and backrest slat gets its own color.
The generative image model (Call~2), however, cannot maintain these instance identities across viewpoints.
Left/right and front/back relationships are especially prone to being swapped or lost between views.
The resulting masks are inconsistent across views and cannot be reprojected into a coherent 3D labeling, which is why our pipeline retains the original same-color-per-type prompt.
The limitation lies in the tested image models.
A model that keeps instance identities consistent across views could use the instance prompt without any other change to the method.

\section{ShapeNet Results and the Split by Training Classes}
\label{S-sec:shapenet_results}
\begin{figure*}[p]
\centering
\setlength{\qw}{0.097\textwidth}
\setlength{\tabcolsep}{0.5pt}
\renewcommand{\arraystretch}{0.3}
\begin{tabular}{c@{\hspace{2pt}} ccccccccc}
 & \multicolumn{3}{c}{Airplane} & \multicolumn{3}{c}{Chair} & \multicolumn{3}{c}{Table} \\[2pt]
\qlab{EMS} &
\qIairn{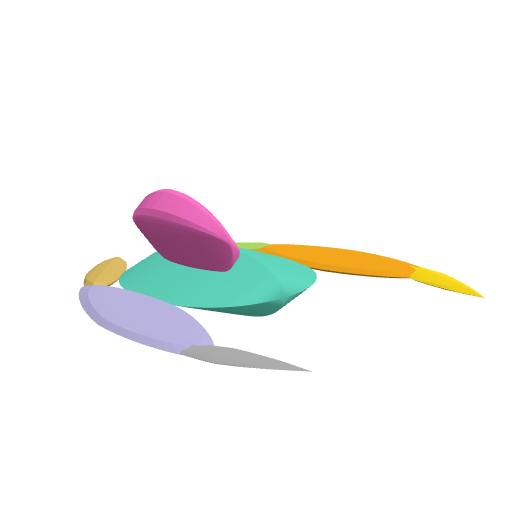}{8} &
\qIairn{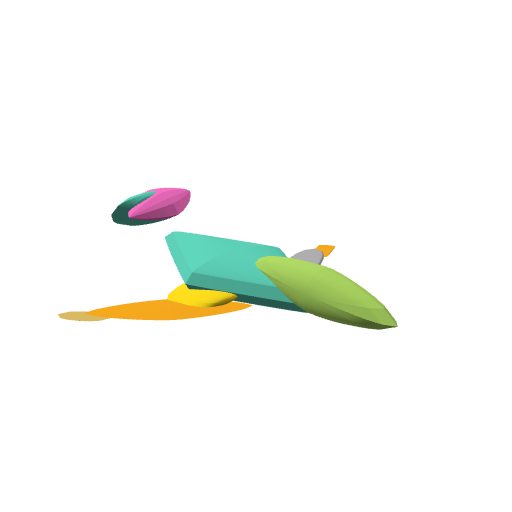}{8} &
\qIairn{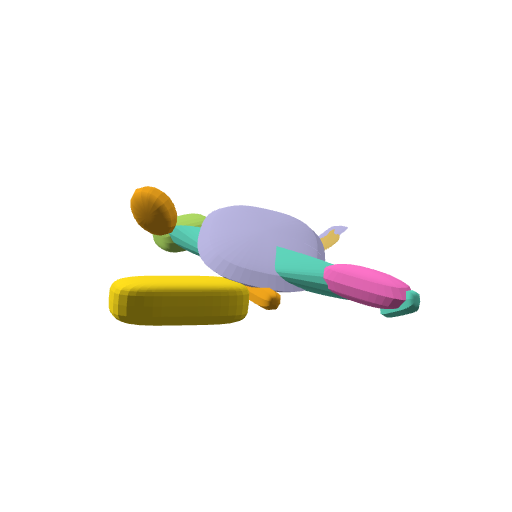}{8} &
\qIfailed &
\qIsn{0.85}{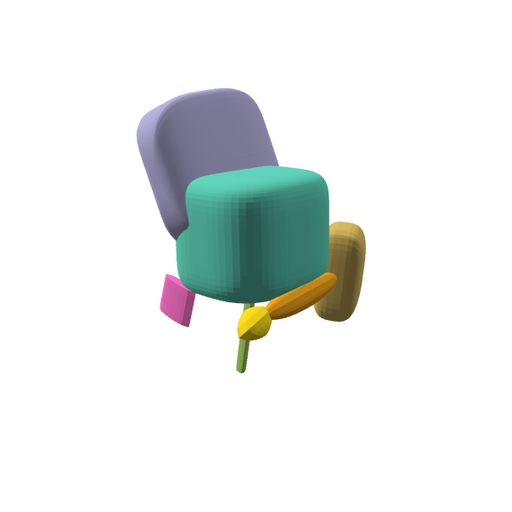}{8} &
\qIsn{0.85}{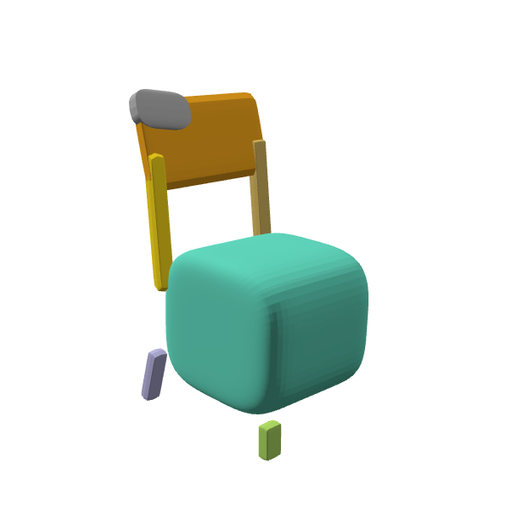}{8} &
\qIfailed &
\qIn{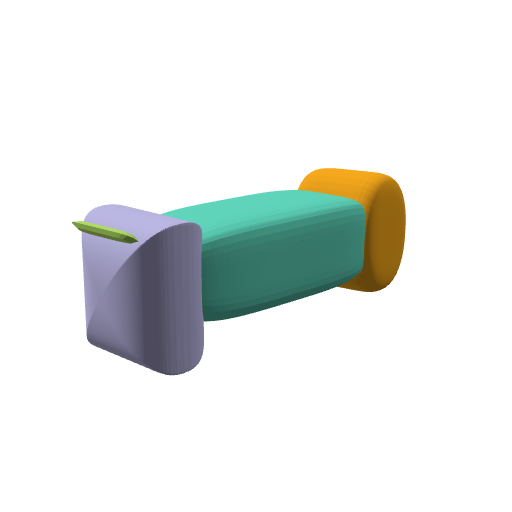}{5} &
\qIn{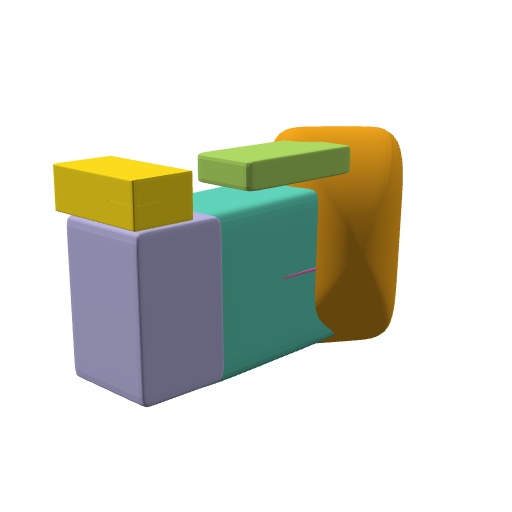}{6} \\
\qlab{Light-SQ} &
\qIairn{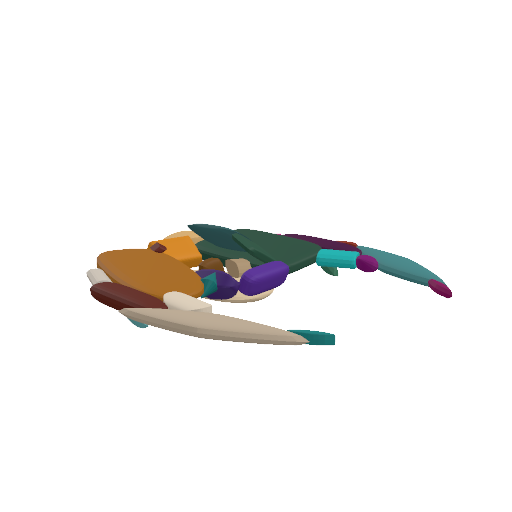}{28} &
\qIairn{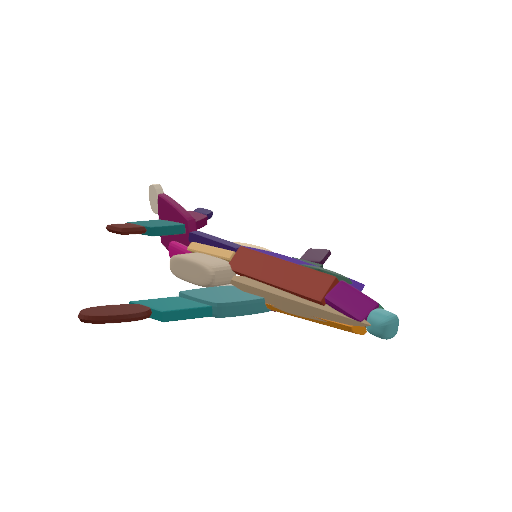}{30} &
\qIairn{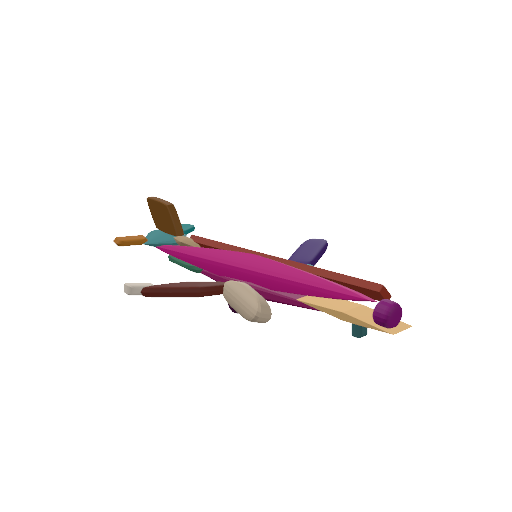}{24} &
\qIn{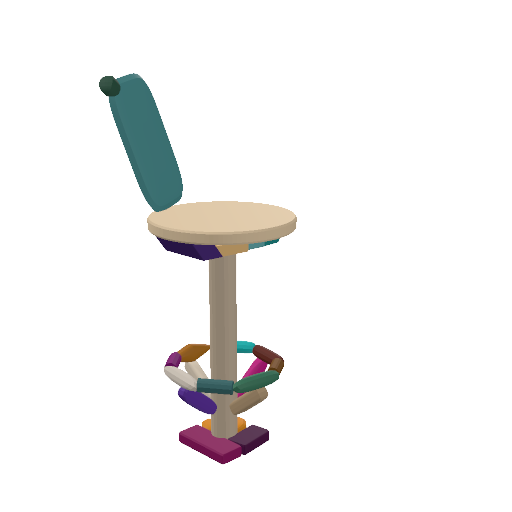}{24} &
\qIsn{0.85}{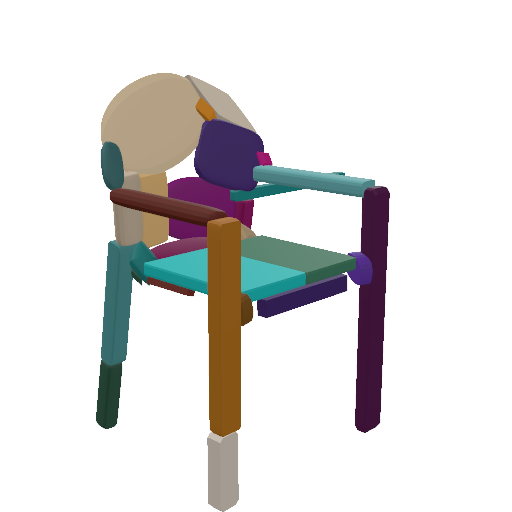}{30} &
\qIsn{0.85}{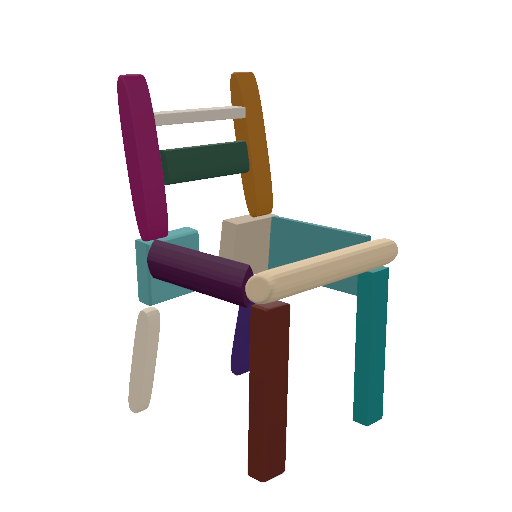}{13} &
\qIsn{0.75}{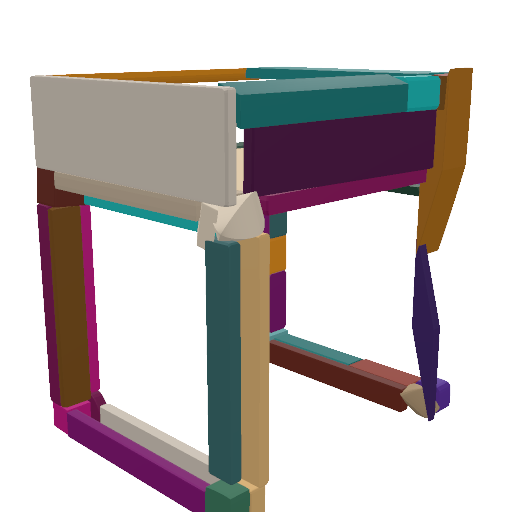}{46} &
\qIn{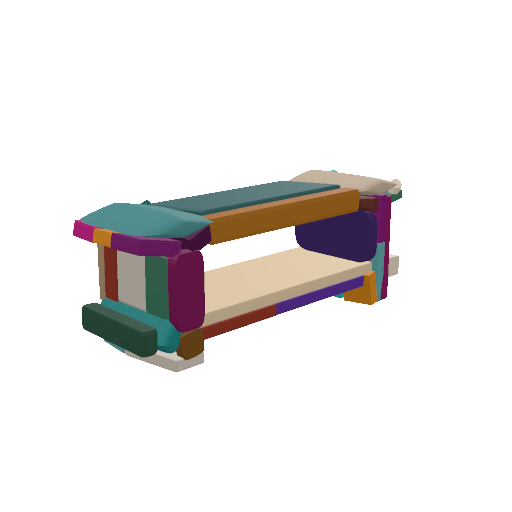}{41} &
\qIn{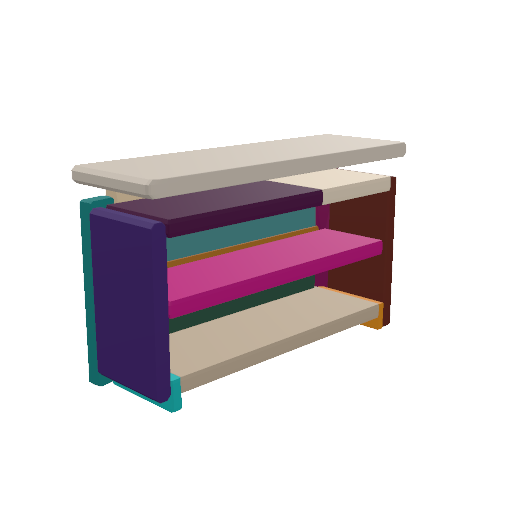}{18} \\
\midrule
\qlab{PrimAny} &
\qIairn{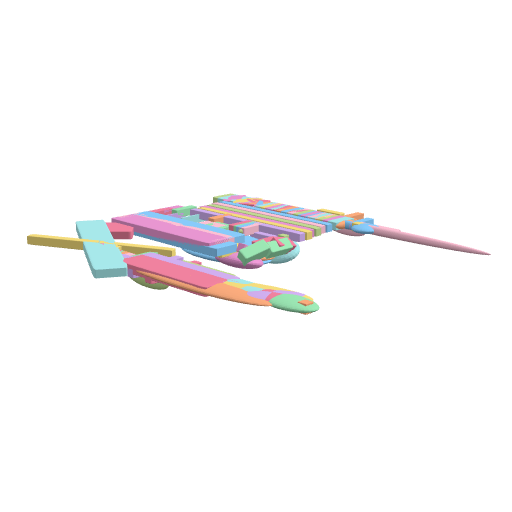}{144} &
\qIairn{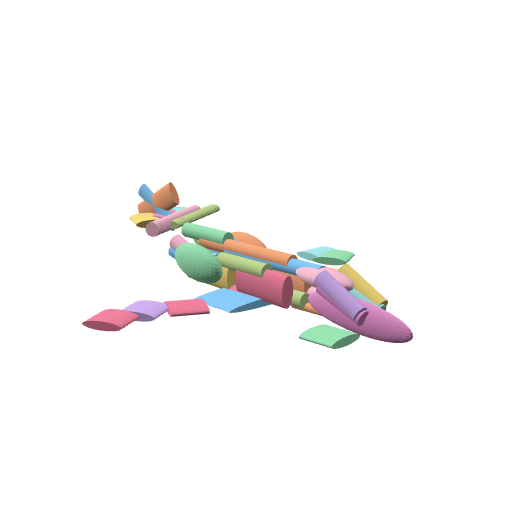}{47} &
\qIairn{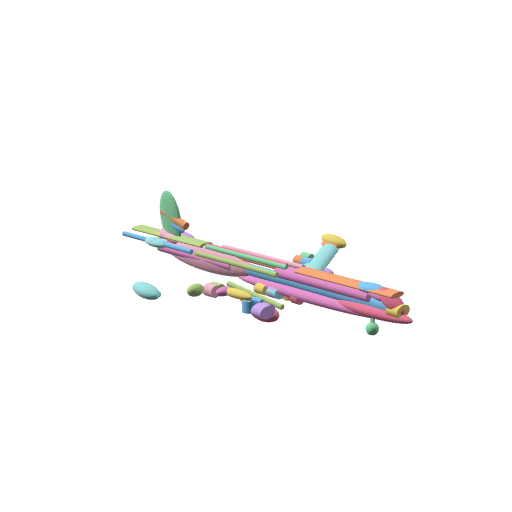}{66} &
\qIn{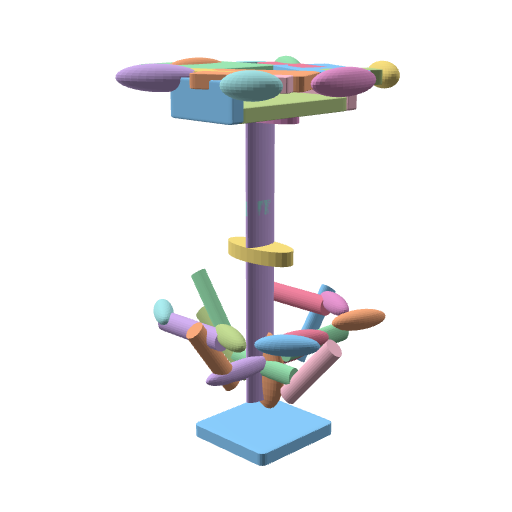}{43} &
\qIsn{0.85}{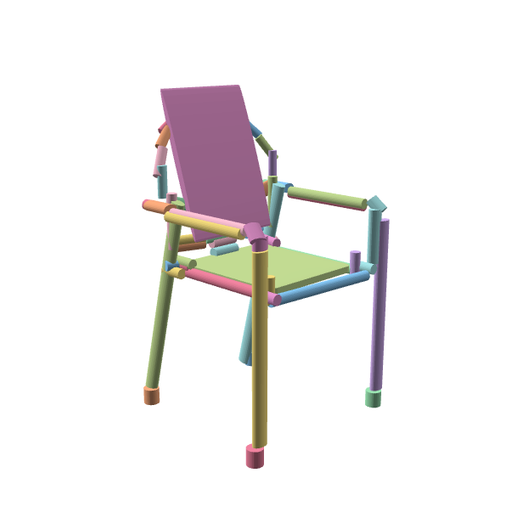}{61} &
\qIsn{0.85}{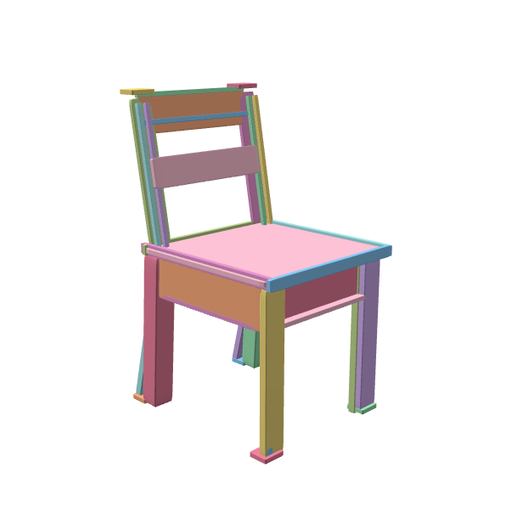}{41} &
\qIsn{0.75}{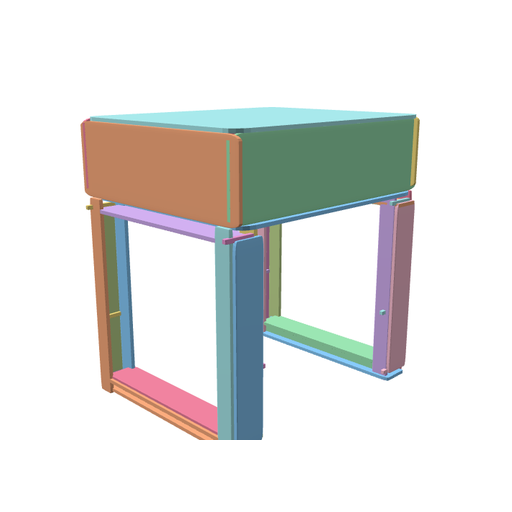}{42} &
\qIn{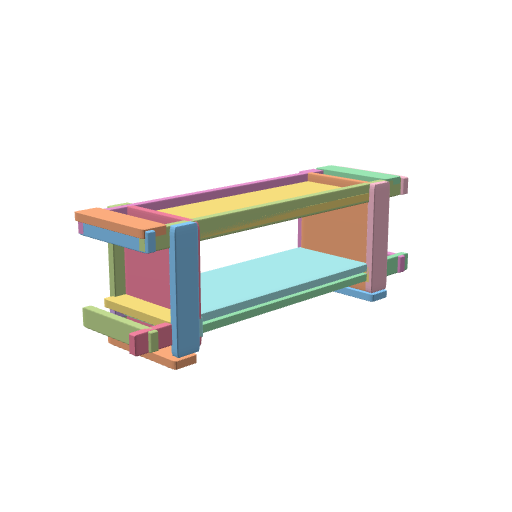}{30} &
\qIn{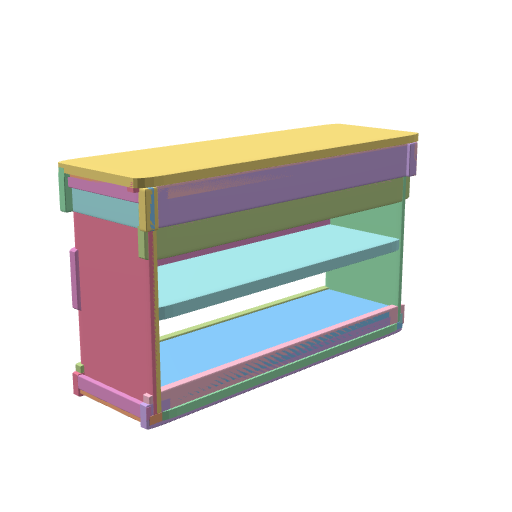}{41} \\
\qlab{F2C} &
\qIairn{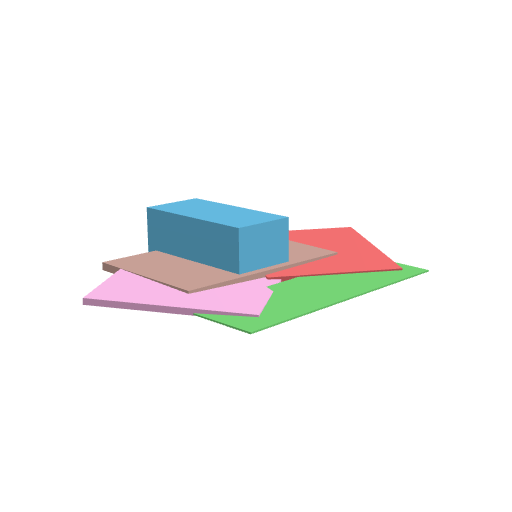}{7} &
\qIairn{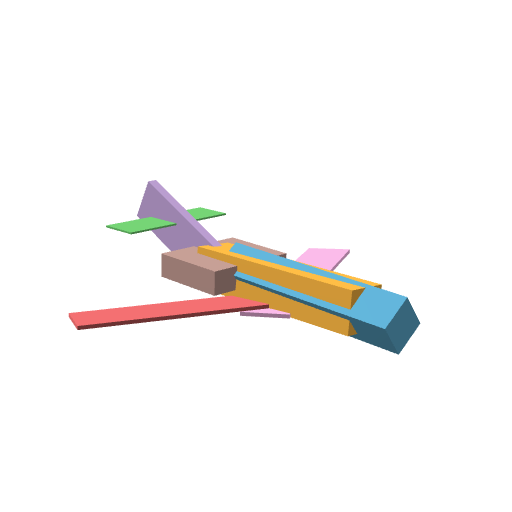}{7} &
\qIairn{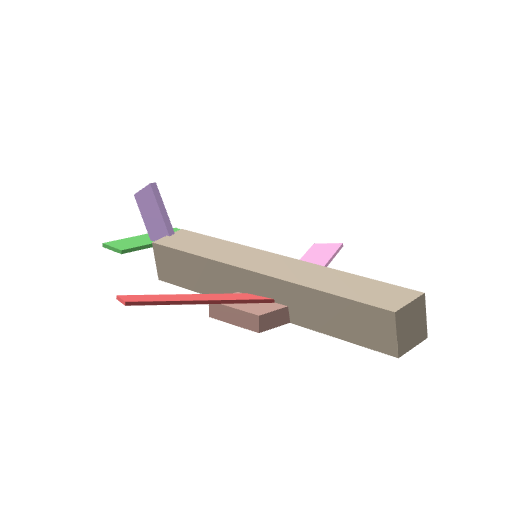}{7} &
\qIn{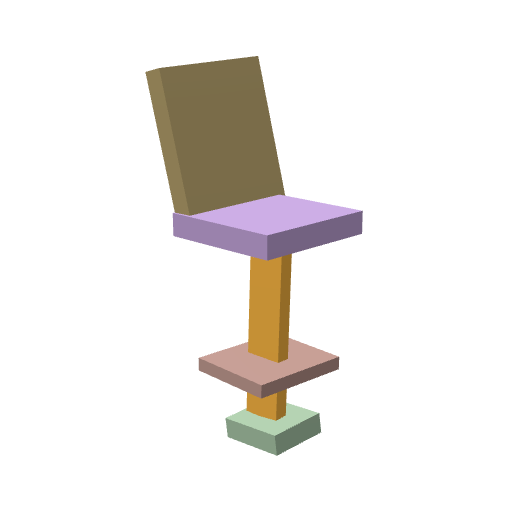}{10} &
\qIsn{0.85}{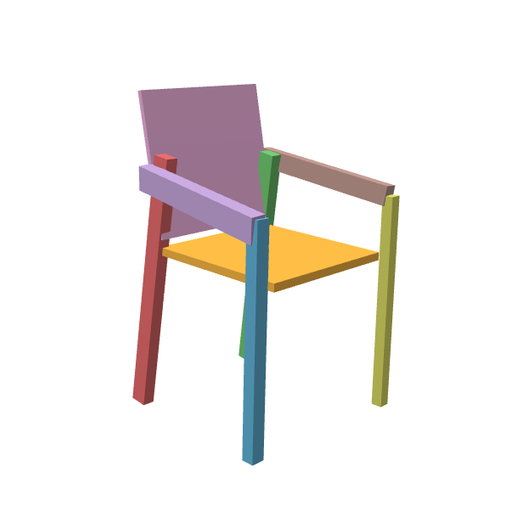}{10} &
\qIsn{0.85}{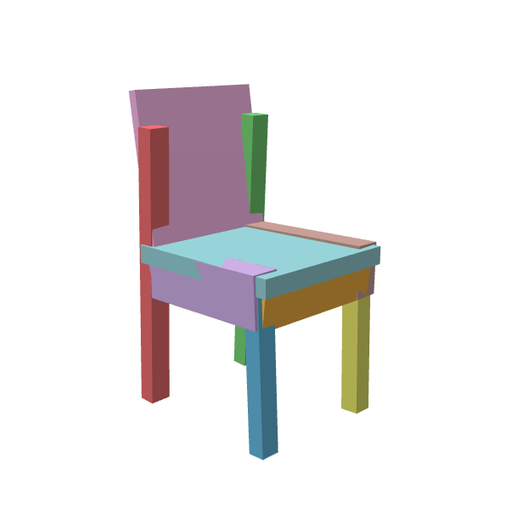}{10} &
\qIsn{0.75}{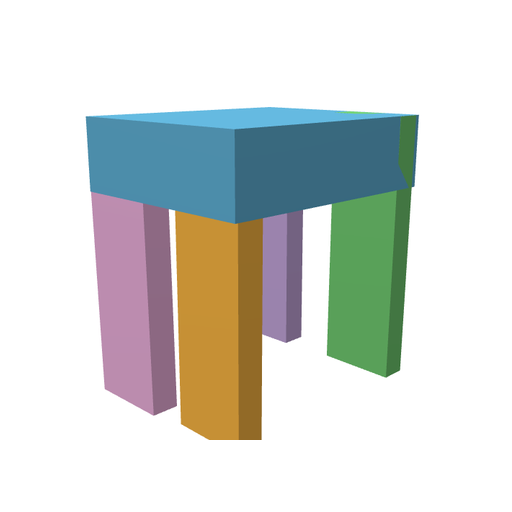}{7} &
\qIn{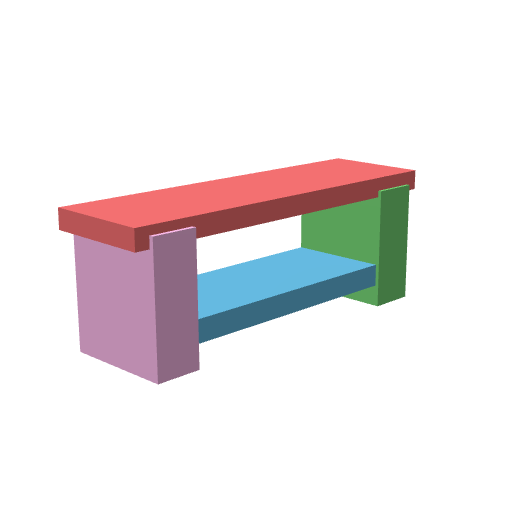}{7} &
\qIn{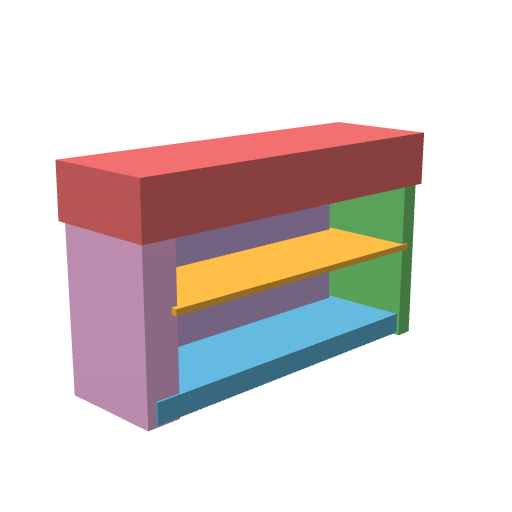}{7} \\
\qlab{SuperDec} &
\qIairn{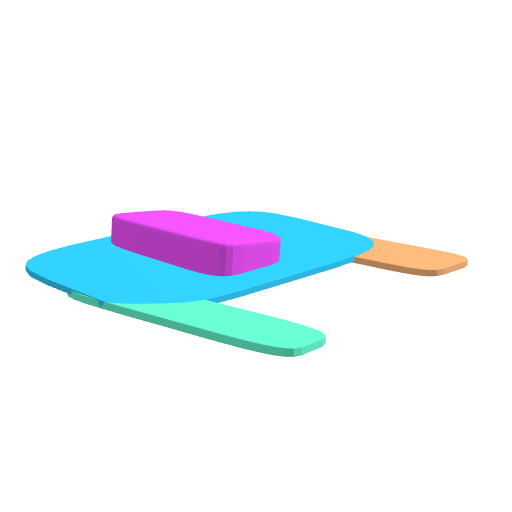}{4} &
\qIairn{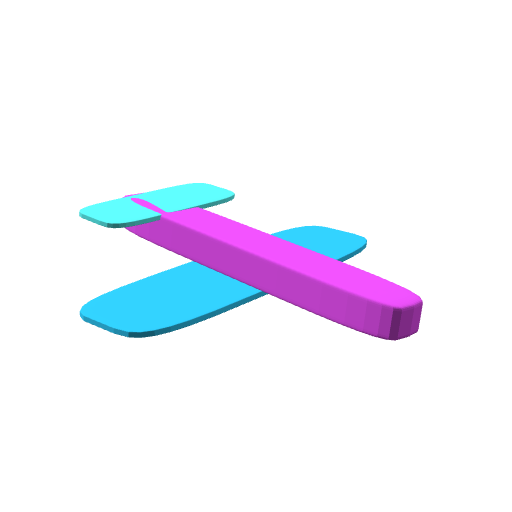}{3} &
\qIairn{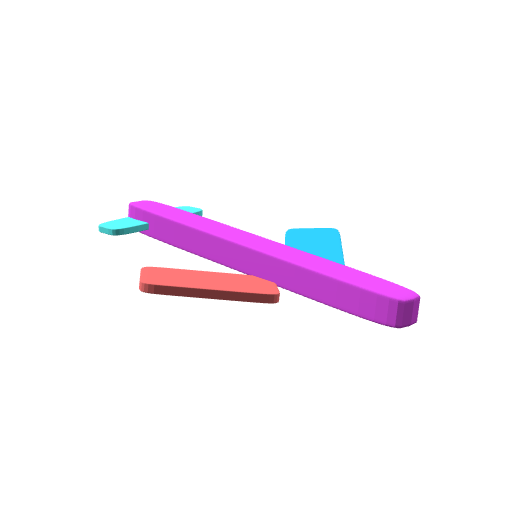}{4} &
\qIn{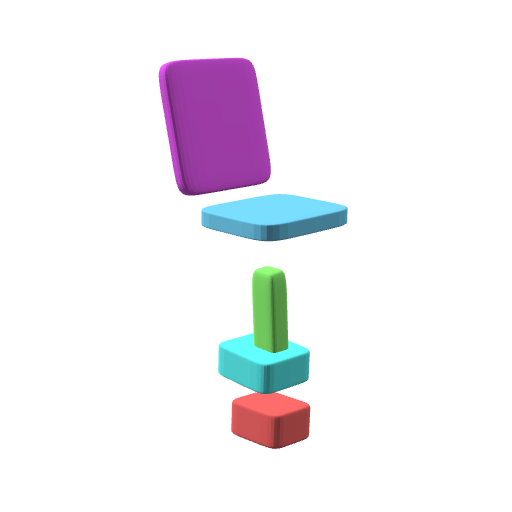}{5} &
\qIsn{0.85}{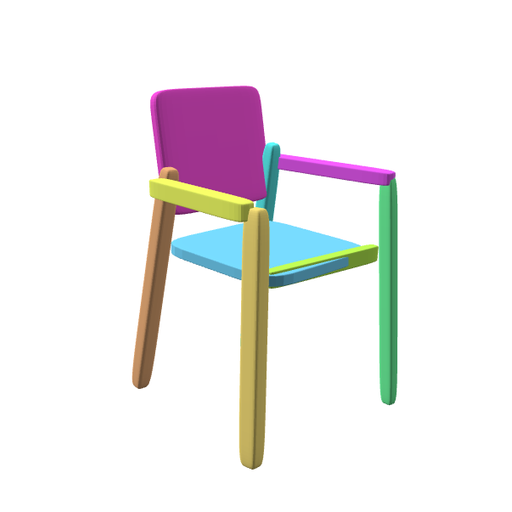}{9} &
\qIsn{0.85}{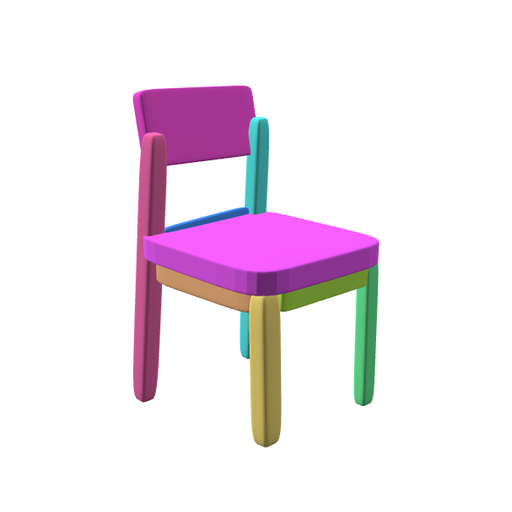}{10} &
\qIsn{0.75}{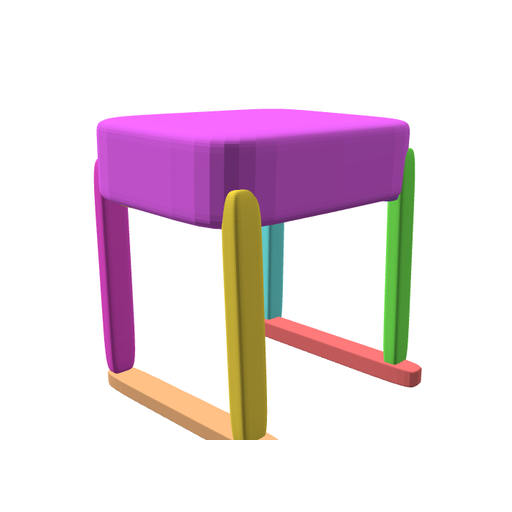}{9} &
\qIn{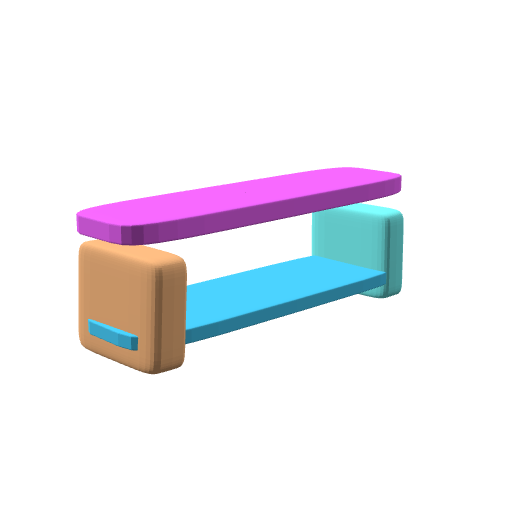}{4} &
\qIn{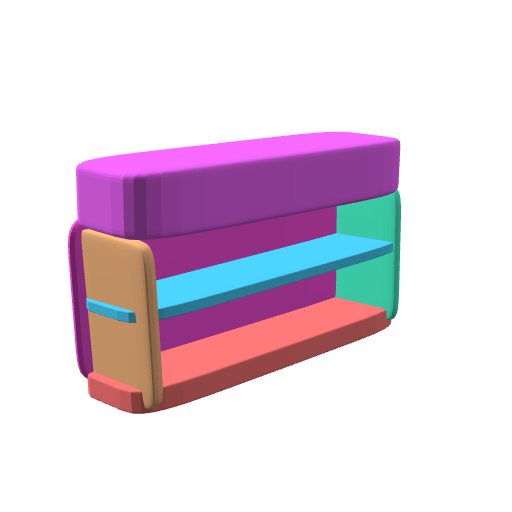}{6} \\
\qlab{SuperFlex} &
\qIairn{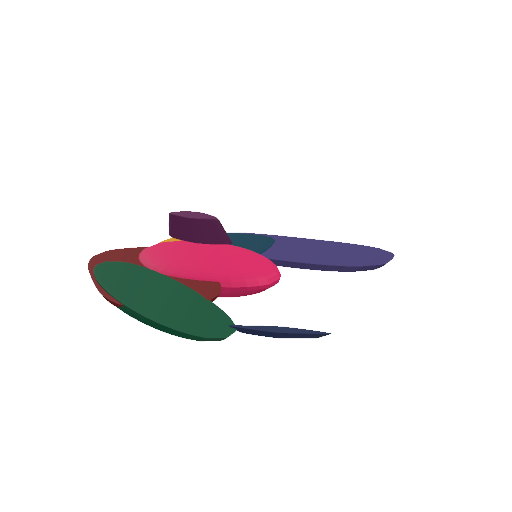}{8} &
\qIairn{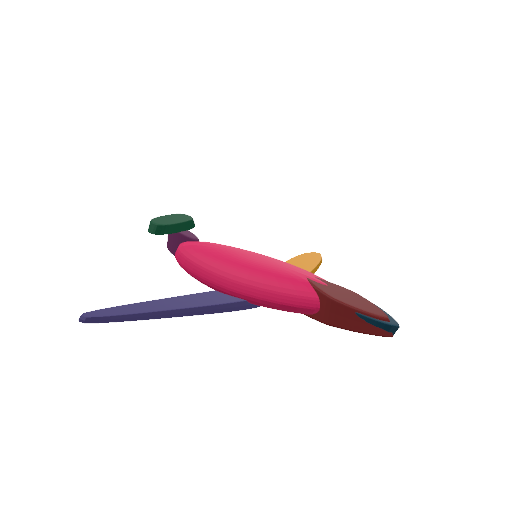}{7} &
\qIairn{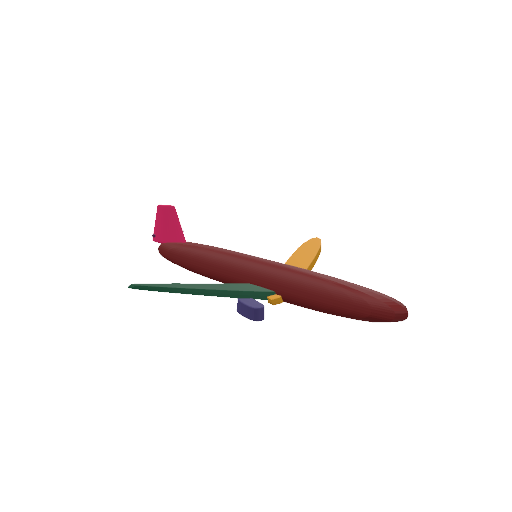}{7} &
\qIn{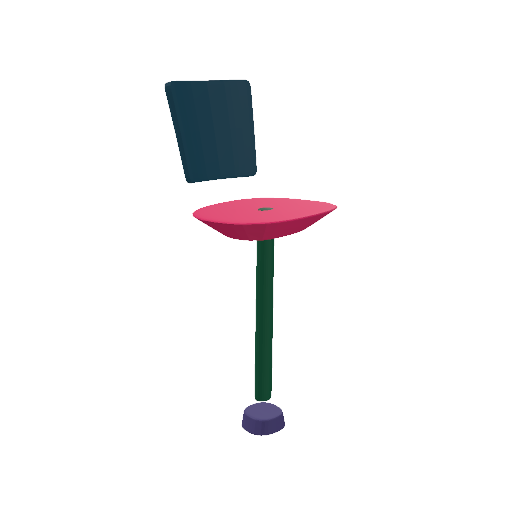}{4} &
\qIsn{0.85}{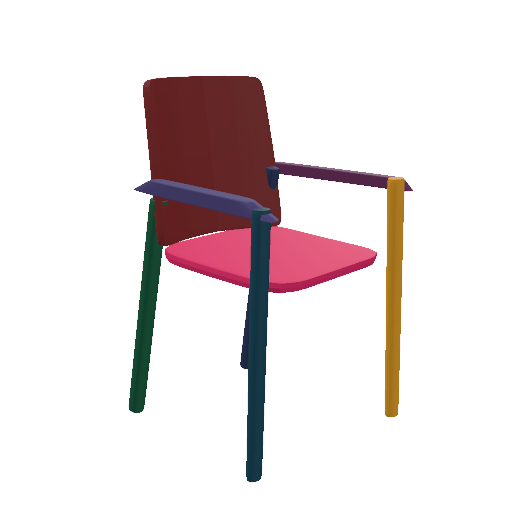}{8} &
\qIsn{0.85}{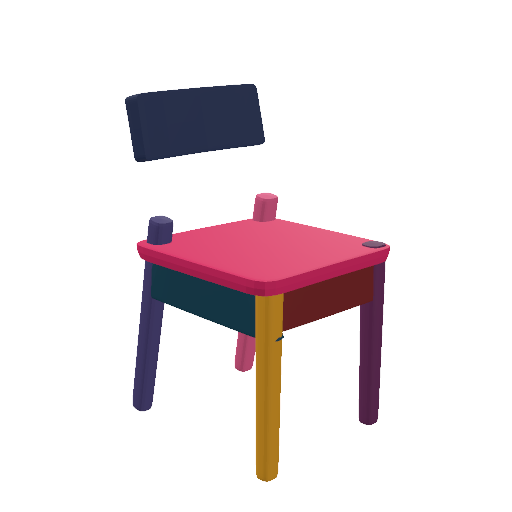}{10} &
\qIsn{0.75}{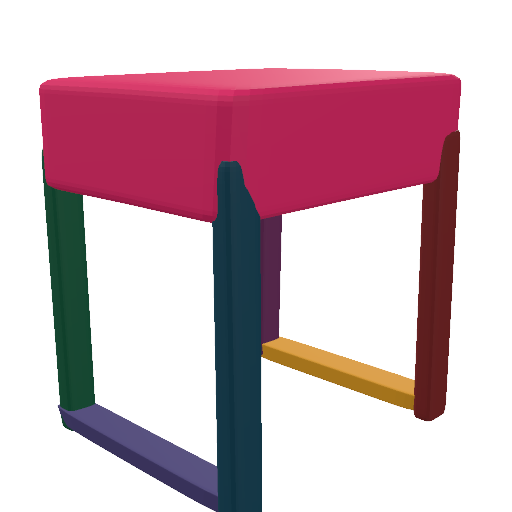}{7} &
\qIn{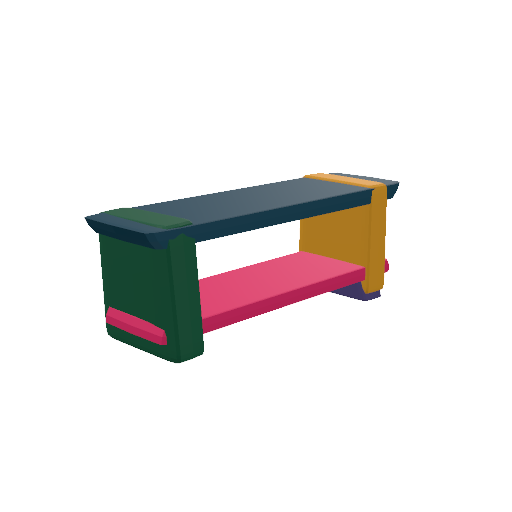}{5} &
\qIn{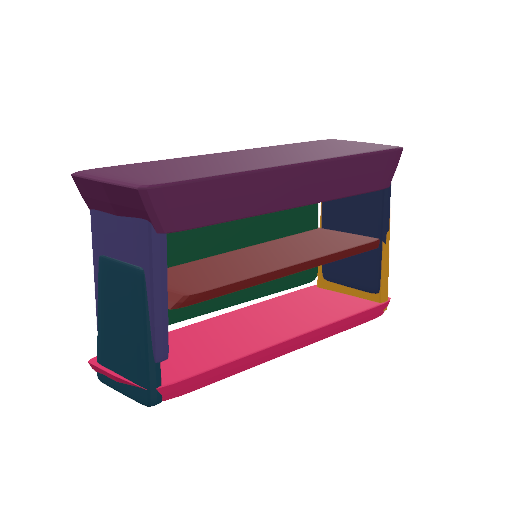}{8} \\
\midrule
\qlab{P3-SAM$^\ast$} &
\qIairn{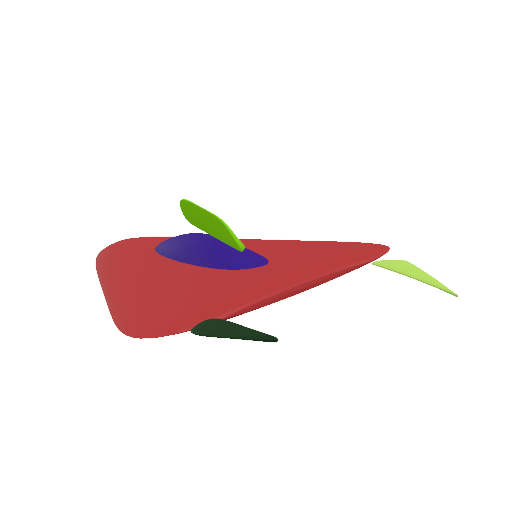}{5} &
\qIairn{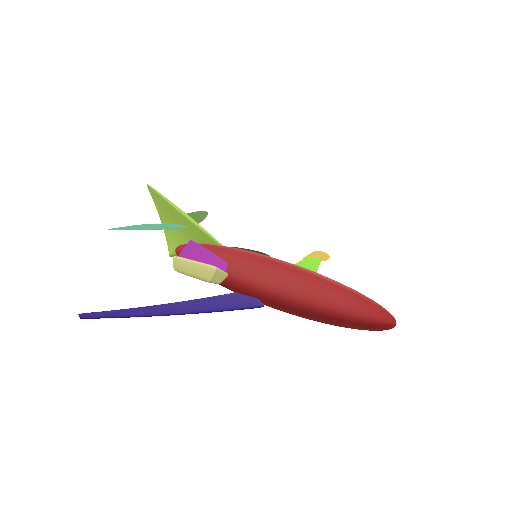}{10} &
\qIairn{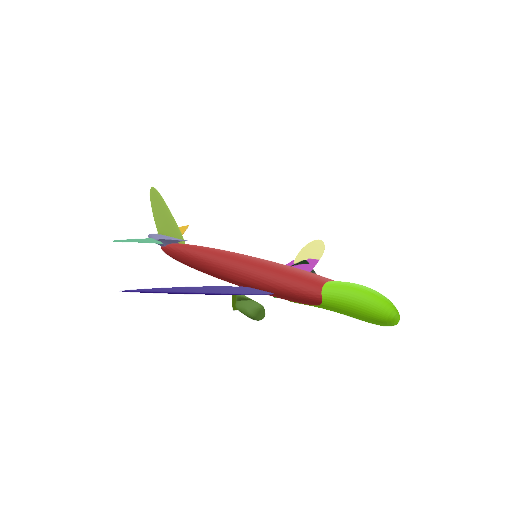}{13} &
\qIn{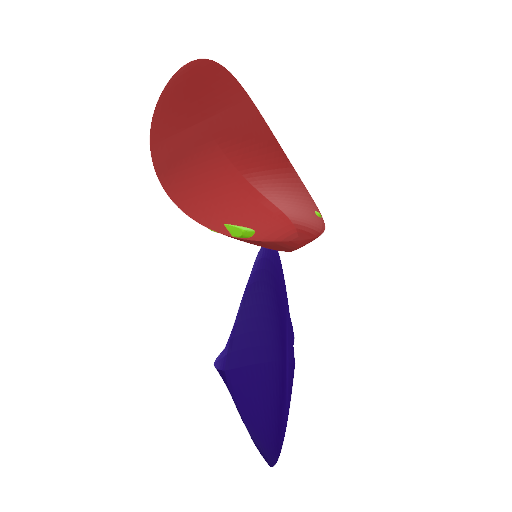}{3} &
\qIsn{0.75}{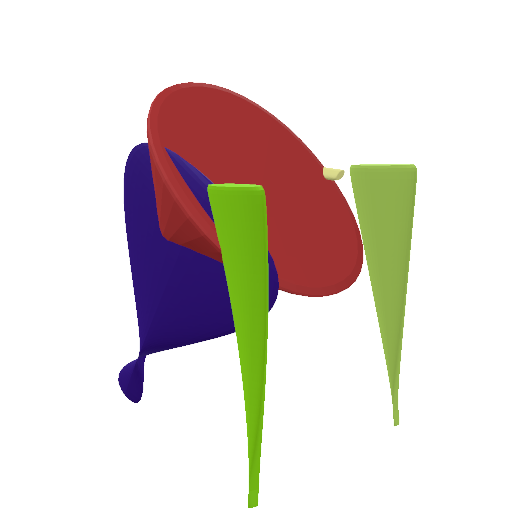}{5} &
\qIsn{0.75}{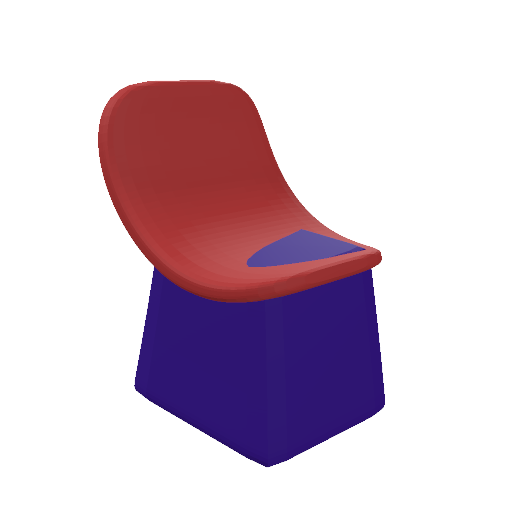}{3} &
\qIsn{0.70}{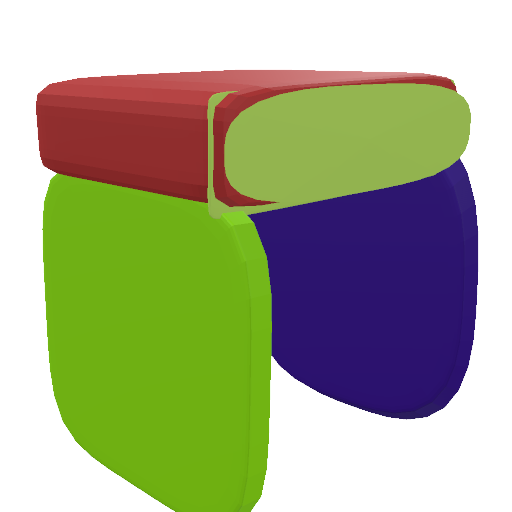}{5} &
\qIn{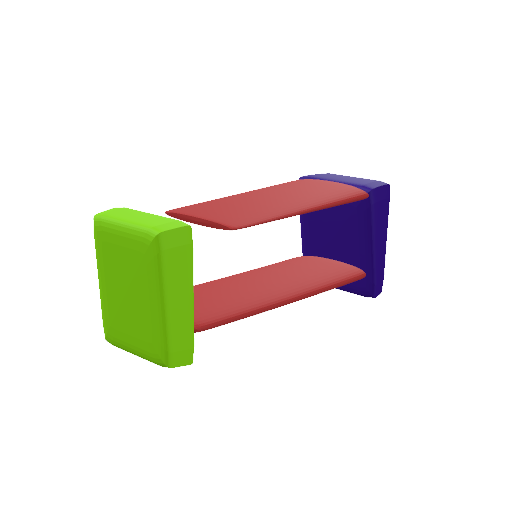}{4} &
\qIn{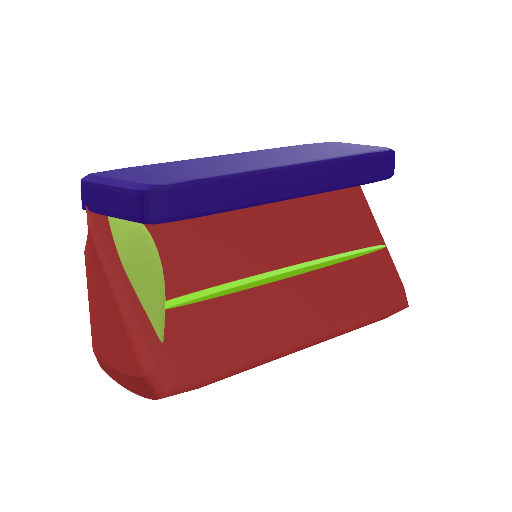}{4} \\
\qlab{PartField$^\ast$} &
\qIairn{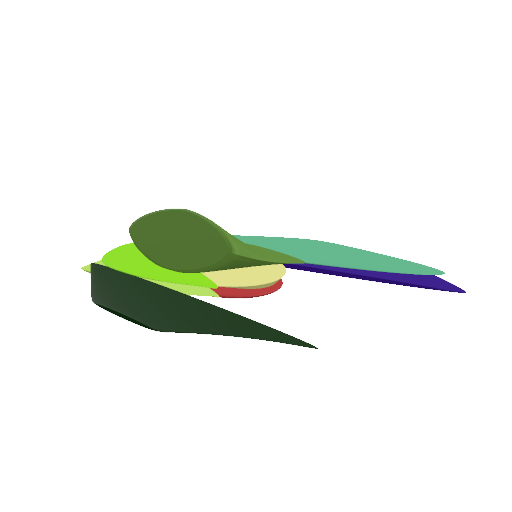}{8} &
\qIairn{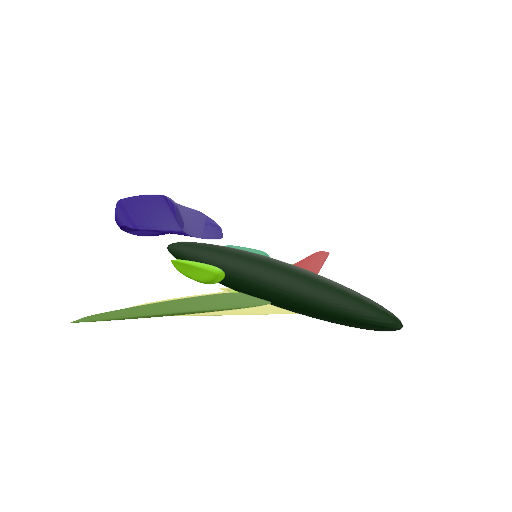}{8} &
\qIairn{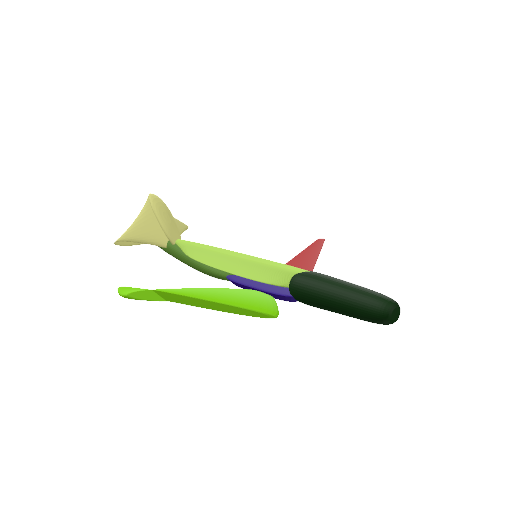}{8} &
\qIn{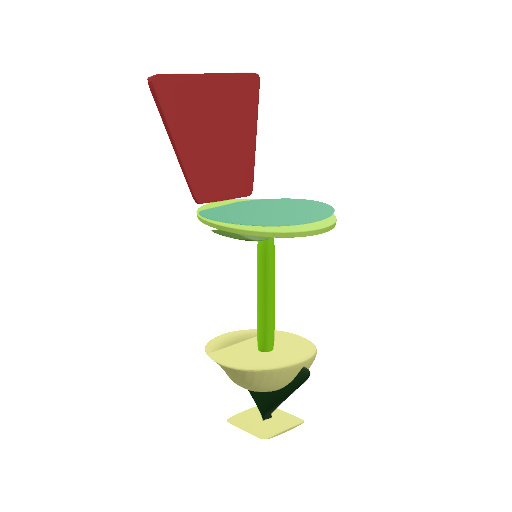}{9} &
\qIsn{0.75}{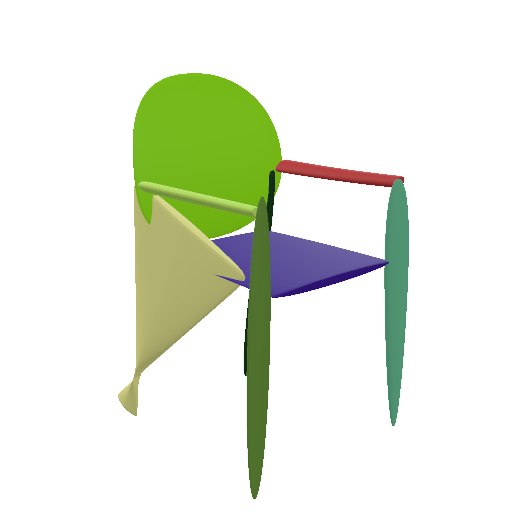}{8} &
\qIsn{0.75}{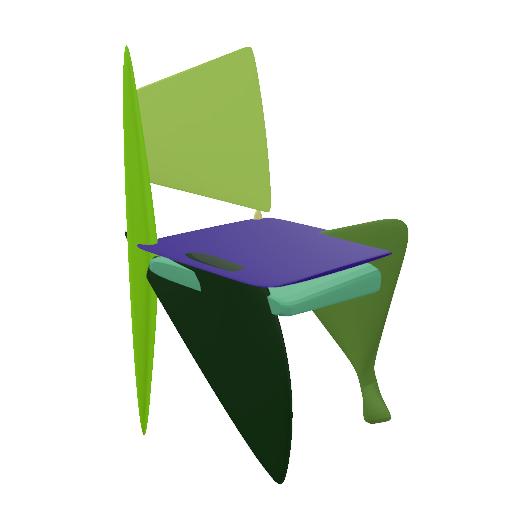}{8} &
\qIsn{0.70}{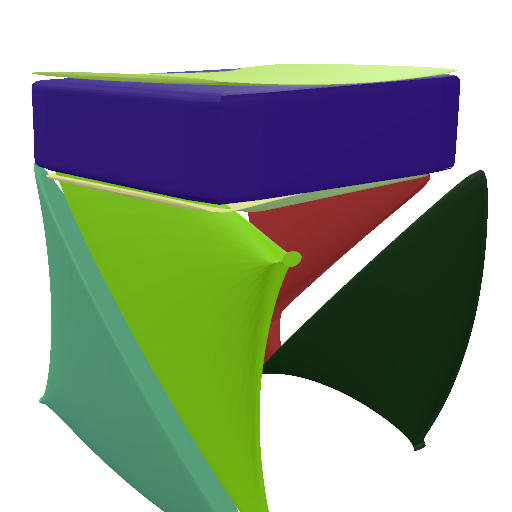}{8} &
\qIn{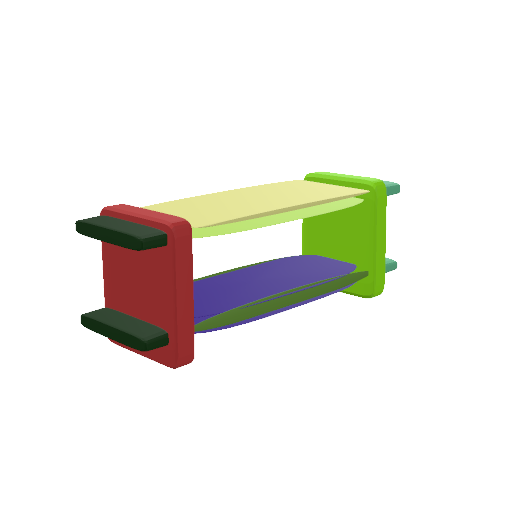}{12} &
\qIn{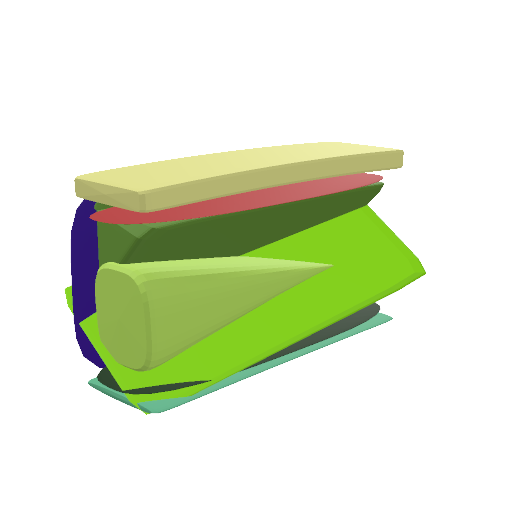}{8} \\
\qlab{PartSAM$^\ast$} &
\qIairn{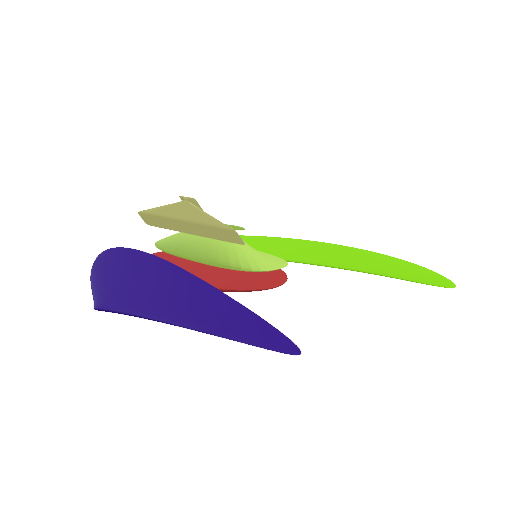}{5} &
\qIairn{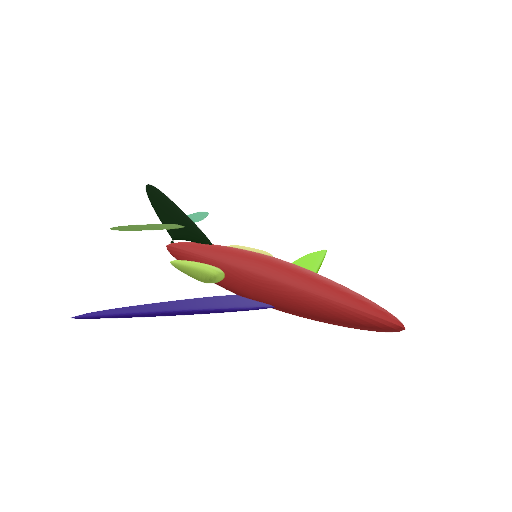}{8} &
\qIairn{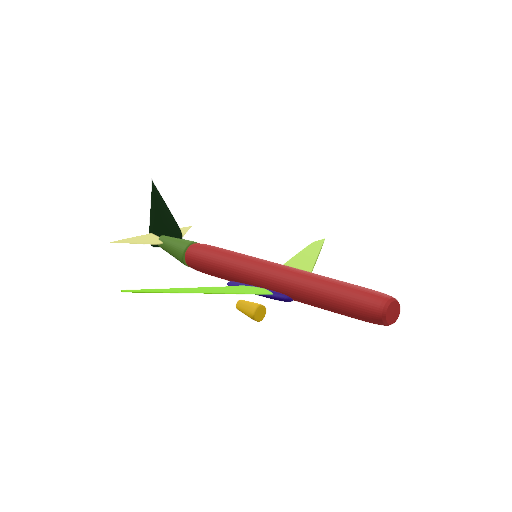}{9} &
\qIn{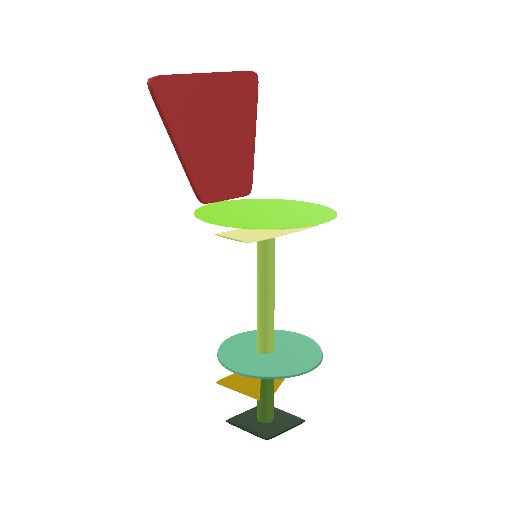}{8} &
\qIsn{0.75}{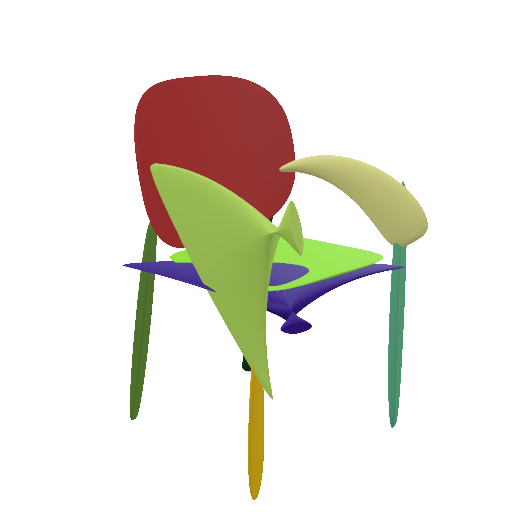}{9} &
\qIsn{0.75}{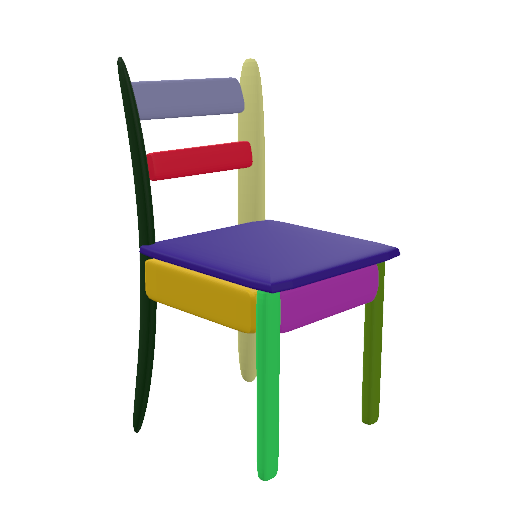}{11} &
\qIsn{0.70}{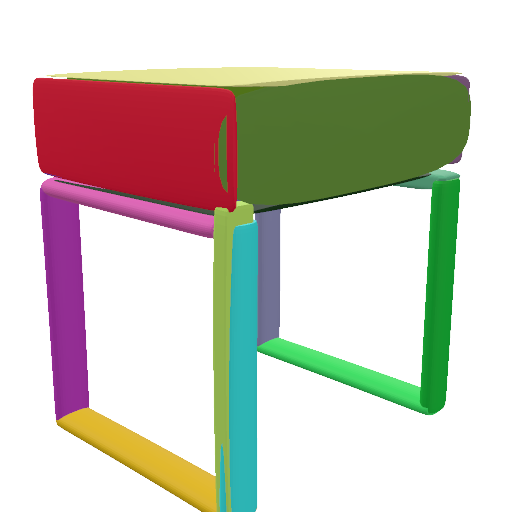}{17} &
\qIn{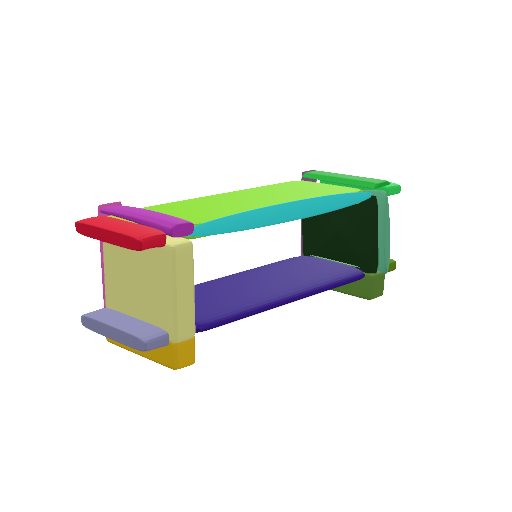}{19} &
\qIn{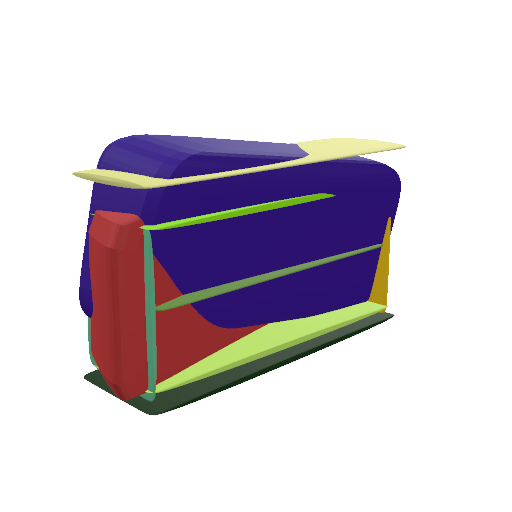}{9} \\
\midrule
\qlab{Ours} &
\qIairn{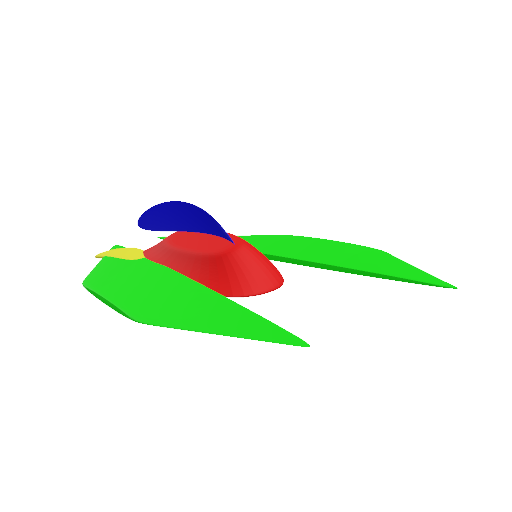}{6} &
\qIairn{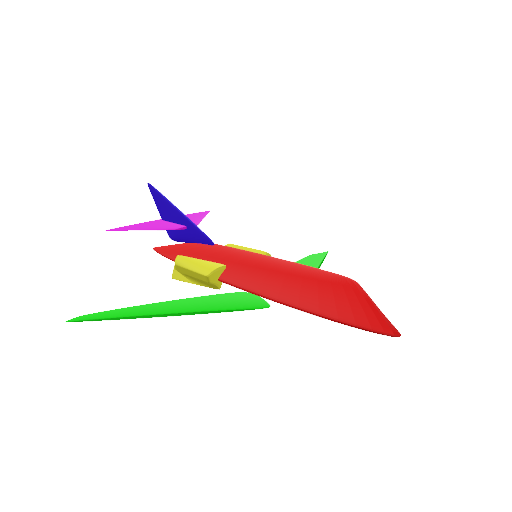}{7} &
\qIairn{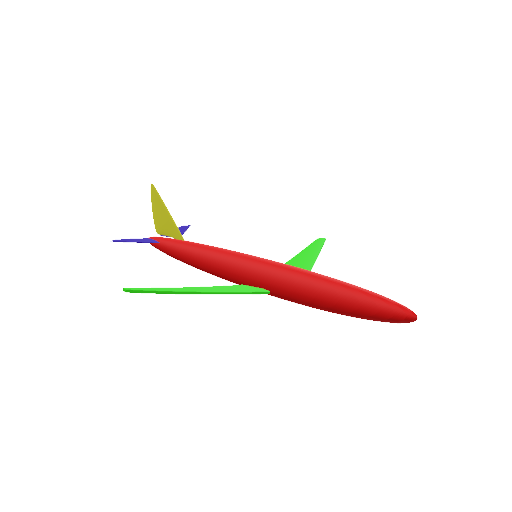}{7} &
\qIn{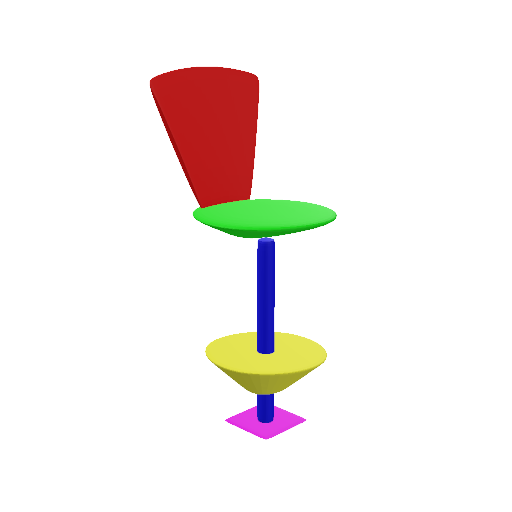}{5} &
\qIsn{0.75}{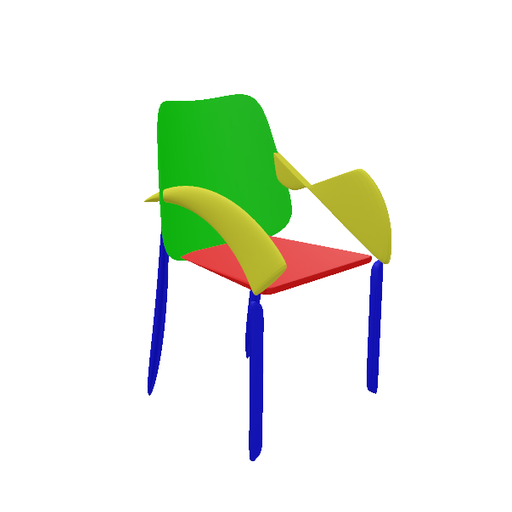}{8} &
\qIsn{0.75}{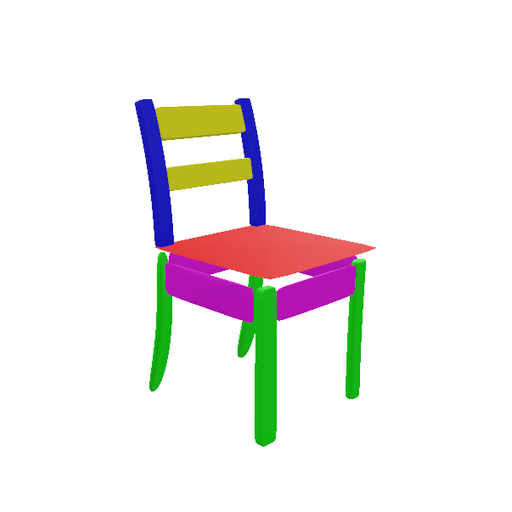}{13} &
\qIsn{0.70}{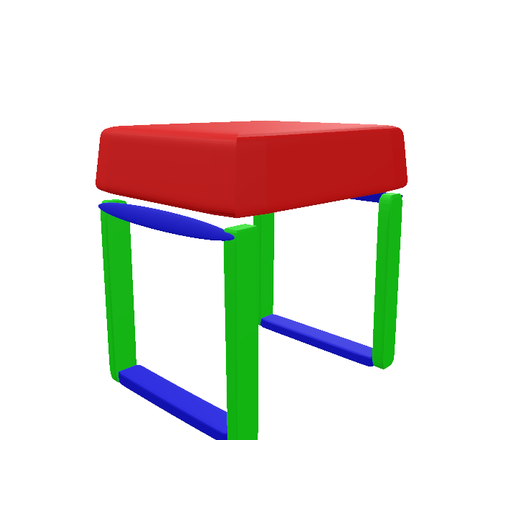}{9} &
\qIn{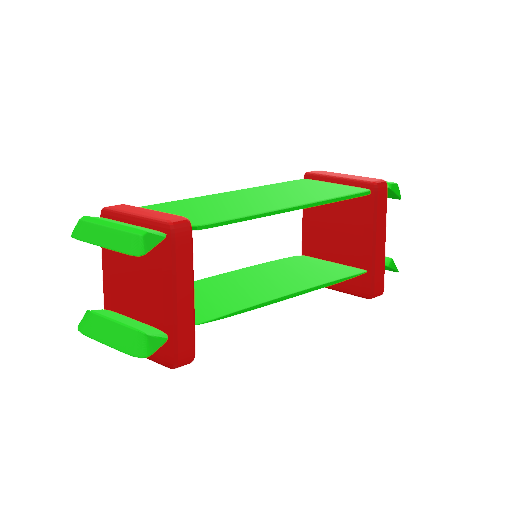}{8} &
\qIn{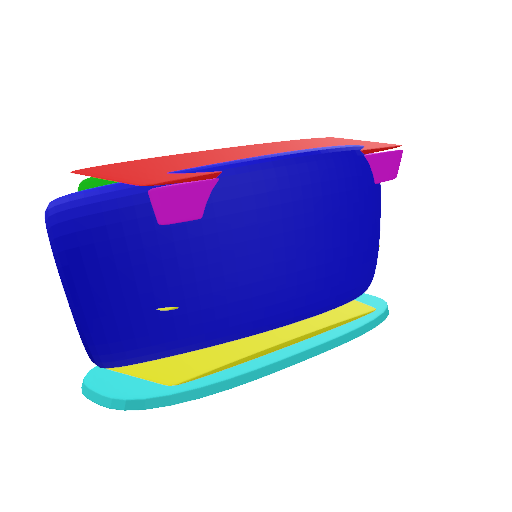}{7} \\
\qlab{GT} &
\qIair{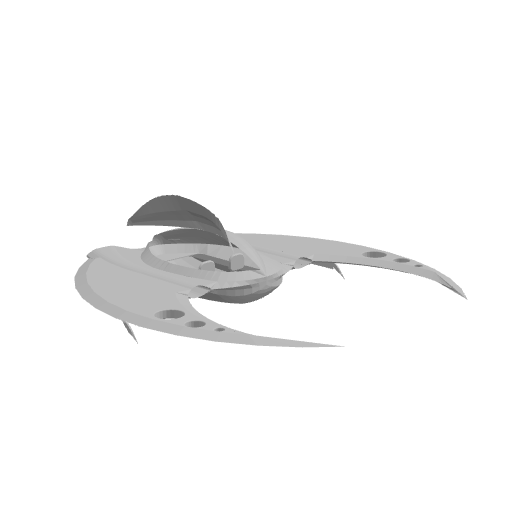} &
\qIair{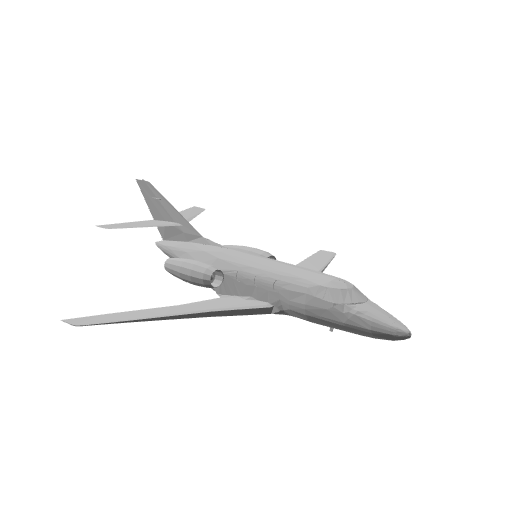} &
\qIair{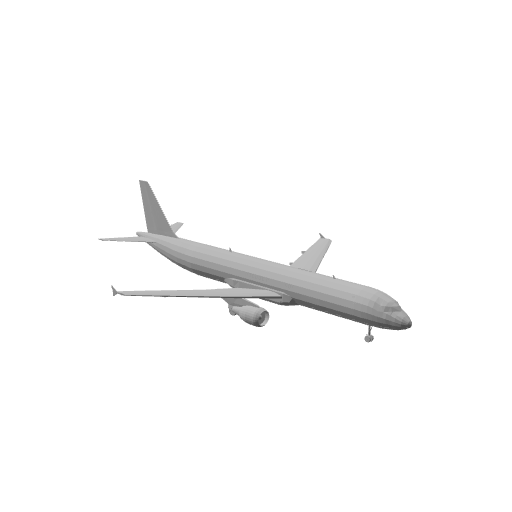} &
\qI{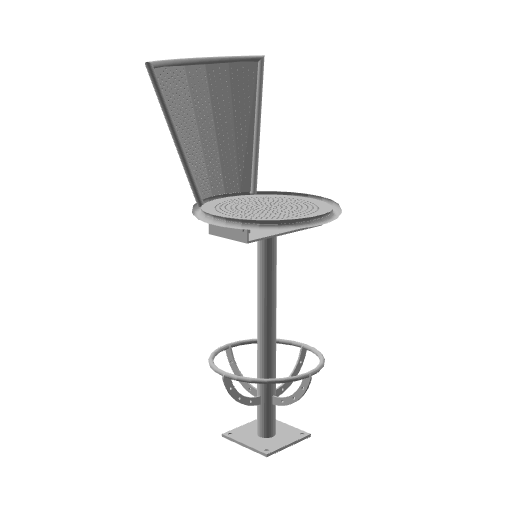} &
\qIs{0.85}{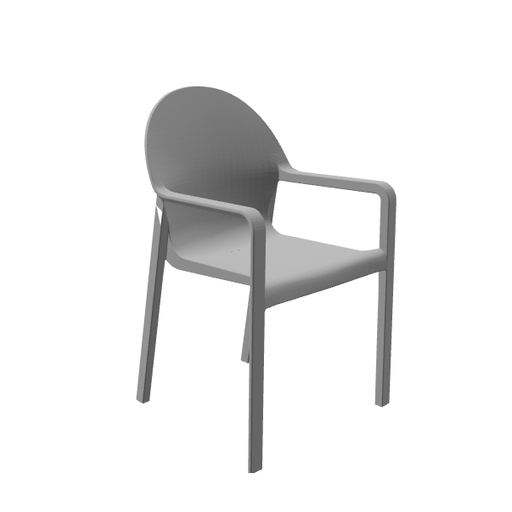} &
\qIs{0.85}{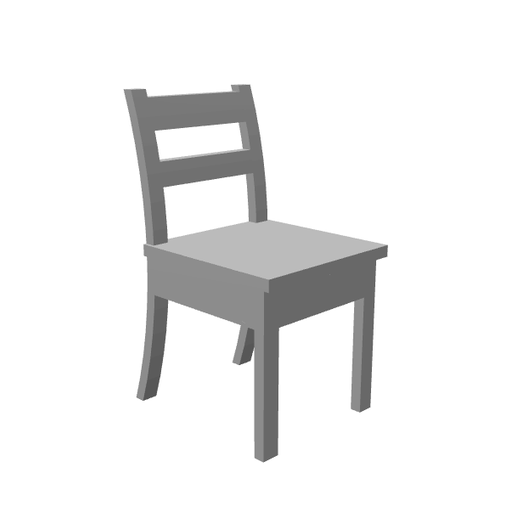} &
\qIs{0.75}{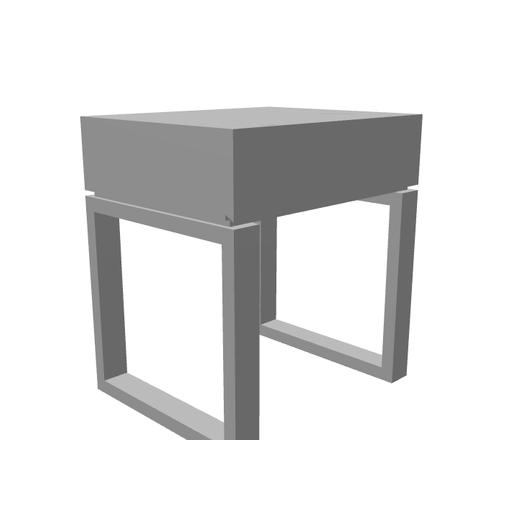} &
\qI{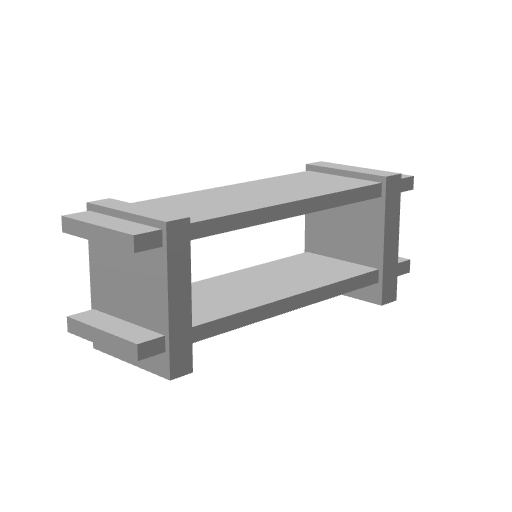} &
\qI{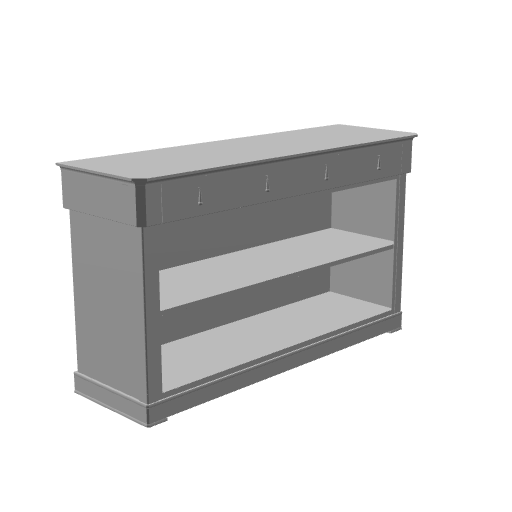} \\
\end{tabular}
\caption{\textbf{Qualitative comparison on ShapeNet.} F2C, SuperDec and SuperFlex were trained on these classes. F2C and SuperDec are comparable to ours there, while SuperFlex is more accurate and fills the volume of seats and table tops more completely. Light-SQ uses 13--46 superquadrics per object. EMS fails on some instances or produces geometrically arbitrary splits. Primitive Anything reaches finer surface coverage with about $7\times$ more primitives. $^\ast$Part labels of that segmenter fitted with our clustering and fitter (\Cref{sec:seg_source}).}
\label{S-fig:qualitative}
\end{figure*}

\Cref{S-tab:results_shapenet} and \Cref{S-fig:qualitative} give the ShapeNet results that the main paper refers to, for EMS~\cite{liu2022ems}, Light-SQ~\cite{wang2025lightsq}, Primitive Anything~\cite{ye2025primitiveanything}, F2C~\cite{kobsik2025f2c}, SuperDec~\cite{fedele2025superdec}, SuperFlex~\cite{tavernini2026superflex} and our fitter on labels of P3-SAM~\cite{li2025p3sam}, PartField~\cite{qian2025partfield} and PartSAM~\cite{zhu2026partsam}.
The ShapeNet~\cite{chang2015shapenet} set contains 100 airplanes, 100 chairs and 100 tables, evaluated with the protocol of \Cref{sec:setup}.
These are the classes F2C was trained on and three of the 13 training classes of SuperDec and SuperFlex.
On the classes the learned methods were trained on, SuperFlex leads the abstraction methods with a CD of 0.081 and an IoU of 63.3\,\%, and F2C is second in IoU (51.9\,\%).
Our method reaches a CD of 0.095 and an IoU of 48.6\,\% with six views.
Among the abstraction methods it is second in CD, ahead of F2C, SuperDec, EMS and Light-SQ, uses 5.9--6.2 primitives, tied with SuperDec for the fewest, and has the lowest overlap rate.
Per class, our CD is lower than SuperFlex's on airplanes (0.054 vs.\ 0.061).
It is higher on chairs (0.122 vs.\ 0.094) and tables (0.108 vs.\ 0.087), the classes with thin legs, stretchers and lattice backs, where a superquadric fitted to a one-sided segmented surface comes out too thin (\Cref{sec:view_layout}).

ShapeNet is therefore the best case for the learned methods, and the split of \Cref{S-tab:ood_split} asks how much of their lead survives on HumanPrim and Toys4K objects outside these classes.
A category counts as in-distribution there if it maps to one of the 13 ShapeNet classes of the 3D-R2N2 split~\cite{choy20163d} (\Cref{S-tab:ood_category_map}).
By this mapping, 62\,\% of HumanPrim~\cite{ye2025primitiveanything} (195 of 314 objects in 13 of 30 categories) but only 10\,\% of Toys4K~\cite{stojanov21cvpr} (33 of 315 objects in 11 of 105 categories) lie inside the training classes, so Toys4K is the largely out-of-distribution benchmark.
The split is exact for SuperDec and SuperFlex.
F2C was trained on three of these classes and Primitive Anything on HumanPrim itself, so for these two methods the split is only approximate and their numbers are reported for completeness.

\begin{table}[t]
\centering
\small
\setlength{\tabcolsep}{9pt}
\begin{tabular}{@{}p{2.3cm} cccc@{}}
\toprule
Method & CD $\downarrow$ & IoU $\uparrow$ & OR $\downarrow$ & \#P \\
\midrule
\multicolumn{5}{c}{\emph{ShapeNet} (chair, airplane, table; 300 objects)} \\
EMS$^\dagger$ & 0.173 & 34.4\% & 1.159 & 7.5 \\
\textcolor{black!55}{Light-SQ} & \textcolor{black!55}{0.098} & \textcolor{black!55}{50.5\%} & \textcolor{black!55}{1.011} & \textcolor{black!55}{\underline{25.5}} \\
\midrule
\textcolor{black!55}{PrimAny} & \textcolor{black!55}{0.099} & \textcolor{black!55}{41.3\%} & \textcolor{black!55}{1.413} & \textcolor{black!55}{\underline{64.7}} \\
F2C & 0.110 & 51.9\% & 1.190 & 6.3 \\
SuperDec & 0.117 & 39.9\% & 1.073 & 5.9 \\
SuperFlex & 0.081 & \textbf{63.3\%} & 1.110 & 6.6 \\
\midrule
P3-SAM$^\ast$ & 0.111 & 42.3\% & 1.060 & \textbf{5.5} \\
PartField$^\ast$ & 0.080 & 44.6\% & 1.067 & 8.7 \\
PartSAM$^\ast$ & \textbf{0.075} & 50.7\% & 1.103 & 10.3 \\
\midrule
\rowcolor{gray!15}
Ours (4 views) & 0.093 & 44.1\% & \textbf{1.039} & 5.9 \\
\rowcolor{gray!15}
Ours (6 views) & 0.095 & 48.6\% & 1.058 & 6.2 \\
\bottomrule
\end{tabular}
\caption{\textbf{Quantitative results on ShapeNet} (100 airplanes, 100 chairs, 100 tables), the classes F2C was trained on and three of the 13 training classes of SuperDec and SuperFlex. Bold marks the best value per column among the compact methods, including the shared-fitter rows. Primitive Anything and Light-SQ (gray) use 4--11$\times$ more primitives (underlined) and are not ranked. $^\dagger$EMS failed on 22 objects, which are excluded from its means. $^\ast$Part labels of that segmenter fitted with our clustering and fitter (\Cref{sec:seg_source}).}
\label{S-tab:results_shapenet}
\end{table}

\begin{table}[t]
\centering\small
\setlength{\tabcolsep}{3.2pt}
\begin{tabular}{@{}l cc cc cc cc@{}}
\toprule
 & \multicolumn{4}{c}{\emph{HumanPrim}} & \multicolumn{4}{c}{\emph{Toys4K}} \\
\cmidrule(lr){2-5}\cmidrule(lr){6-9}
 & \multicolumn{2}{c}{CD $\downarrow$} & \multicolumn{2}{c}{IoU $\uparrow$} & \multicolumn{2}{c}{CD $\downarrow$} & \multicolumn{2}{c}{IoU $\uparrow$} \\
\cmidrule(lr){2-3}\cmidrule(lr){4-5}\cmidrule(lr){6-7}\cmidrule(lr){8-9}
Method & in & out & in & out & in & out & in & out \\
\midrule
EMS & 0.134 & 0.113 & 37.0 & 47.8 & 0.107 & 0.114 & 60.8 & 58.8 \\
\textcolor{black!55}{Light-SQ} & \textcolor{black!55}{0.076} & \textcolor{black!55}{0.079} & \textcolor{black!55}{66.4} & \textcolor{black!55}{69.4} & \textcolor{black!55}{0.123} & \textcolor{black!55}{0.125} & \textcolor{black!55}{66.5} & \textcolor{black!55}{66.0} \\
\midrule
\textcolor{black!55}{PrimAny} & \textcolor{black!55}{0.082} & \textcolor{black!55}{0.094} & \textcolor{black!55}{71.6} & \textcolor{black!55}{67.7} & \textcolor{black!55}{0.130} & \textcolor{black!55}{0.147} & \textcolor{black!55}{42.6} & \textcolor{black!55}{46.3} \\
F2C & 0.123 & 0.132 & 38.3 & 38.7 & 0.136 & 0.143 & 46.2 & 42.5 \\
SuperDec & 0.086 & 0.098 & 60.0 & 55.7 & 0.098 & 0.104 & 53.8 & 52.0 \\
SuperFlex & \textbf{0.078} & 0.089 & \textbf{68.5} & \textbf{64.1} & 0.094 & 0.097 & \textbf{67.8} & \textbf{67.6} \\
\midrule
\rowcolor{gray!15}
Ours (4 views) & 0.081 & \textbf{0.084} & 63.2 & 63.1 & 0.091 & 0.089 & 58.4 & 61.7 \\
\rowcolor{gray!15}
Ours (6 views) & 0.079 & 0.085 & 66.6 & 63.1 & \textbf{0.087} & \textbf{0.086} & 61.2 & 64.0 \\
\bottomrule
\end{tabular}
\caption{\textbf{Inside versus outside the training classes.} Mean CD and IoU (\%) on the objects whose category is one of the 13 ShapeNet classes the learned methods were trained on (in, 195 HumanPrim and 33 Toys4K objects) and on the remaining objects (out, 119 and 282). The category mapping is given in \Cref{S-tab:ood_category_map}. Bold marks the best compact method per column. Primitive Anything and Light-SQ (gray) are not ranked. On HumanPrim the out-of-class objects are harder for the learned methods and for ours, but the CD of SuperDec and SuperFlex rises by 0.011--0.012 against 0.003--0.006 for ours. On Toys4K our CD is unchanged outside the training classes while every learned method worsens. SuperFlex keeps its IoU lead on both benchmarks.}
\label{S-tab:ood_split}
\end{table}

\begin{table}[t]
\centering\small
\begin{tabular}{@{}lll@{}}
\toprule
Benchmark & Category (object-id prefix) & ShapeNetCore class \\
\midrule
HumanPrim & biplane & airplane \\
 & modern fighter & airplane \\
 & old fighter & airplane \\
 & a police car & car \\
 & a sports car & car \\
 & the car & car \\
 & a chair & chair \\
 & tv & display \\
 & lights & lamp \\
 & sofa & sofa \\
 & the table & table \\
 & a ship & watercraft \\
 & a submarine & watercraft \\
\midrule
Toys4K & airplane & airplane \\
 & closet & cabinet \\
 & car & car \\
 & chair & chair \\
 & monitor & display \\
 & tv & display \\
 & light bulb & lamp \\
 & sofa & sofa \\
 & phone & telephone \\
 & boat & watercraft \\
 & submarine & watercraft \\
\bottomrule
\end{tabular}
\caption{\textbf{Categories counted as in-distribution.} Object categories of HumanPrim and Toys4K mapped to one of the 13 ShapeNet classes the learned methods were trained on. All other categories count as out of distribution in \Cref{S-tab:ood_split}. In distribution are 13 of 30 HumanPrim categories (195 of 314 objects) and 11 of 105 Toys4K categories (33 of 315 objects).}
\label{S-tab:ood_category_map}
\end{table}

\section{Segmentation Source with a Shared Fitter}
\label{S-sec:seg_source_full}
\Cref{tab:seg_source} of the main paper compares part segmentations under a shared clustering and fitter on HumanPrim and Toys4K.
The asterisked rows of \Cref{S-tab:results_shapenet} give the ShapeNet numbers, and \Cref{S-fig:qualitative} shows the resulting primitives next to the baselines, as \Cref{fig:qual_humanprim,fig:qual_toys4k} do for HumanPrim and Toys4K.
PartField labels are $k$-means clusters ($k=8$) of the learned point features and split objects into pieces of similar size regardless of structure, which explains its low IoU on the articulated HumanPrim and Toys4K objects and its competitive numbers on the more uniform ShapeNet classes.
P3-SAM returns the fewest parts and misses thin structures, most visibly on chairs.
PartSAM returns the most parts and covers thin structures with additional primitives, which lowers CD and raises IoU at the cost of the highest overlap rate.
Our generated masks sit between P3-SAM and PartSAM in granularity and give the lowest overlap rate.
For three objects (two Toys4K, one airplane) PartSAM returned no mask at its default score threshold of 0.65, so we lowered it to 0.5 for these.

\section{Run-to-Run Variability}
\label{S-sec:determinism_full}

Repeated model calls on the same input may produce different part lists and masks, which propagate into different abstractions.
To quantify this variability, we run the full pipeline five times on a subset of 50 HumanPrim objects with identical inputs and report the mean, the per-object standard deviation across runs and the standard deviation of the 50-object means (\Cref{S-tab:determinism}). The fitter is fixed, so the spread measures the generative stages alone.
For a single object the variation is substantial ($0.006$ CD, $4.6$ IoU points, about one primitive).
This is of the same order as the gaps between the top methods in \Cref{tab:results}, so single-object differences of that size should not be over-interpreted.
The standard deviation of the run means is $0.001$ CD and $0.6$ IoU points on this subset, so aggregate scores are far more repeatable than individual abstractions.
The segmentation used for the studies of \Cref{S-sec:fitter_floor,S-sec:fitter_ablations,S-sec:selection,S-sec:view_layout_full} was generated separately with the final prompt. With CD $0.073$ and IoU $68.4\,\%$ it lies at the favorable end of this spread, comparable to the best of the five runs.
The variability is also headroom.
Selecting, per object, the run with the lowest Chamfer distance out of the five four-view runs lowers the mean CD from $0.079$ to $0.073$ and raises IoU from $64.0$ to $68.2\,\%$, and the best of two runs reaches $0.076$ / $66.2\,\%$.
This selection uses Chamfer distance to the input geometry, which is already available at inference time, and requires no part annotations.
Generating two to five candidates multiplies the model calls accordingly and requires evaluating each candidate against the input, so the selection trades computation for quality.

\begin{table}[htbp]
  \centering
  \begin{tabular}{lcccc}
    \toprule
    & CD $\downarrow$ & IoU $\uparrow$ & OR $\downarrow$ & \#P \\
    \midrule
    Mean                    & $0.079$       & $64.0\%$       & $1.021$        & $6.5$ \\
    Std.\ dev.\ per object   & $\pm 0.006$   & $\pm 4.6\%$    & $\pm 0.019$    & $\pm 1.1$ \\
    Std.\ dev.\ of run means & $\pm 0.001$   & $\pm 0.6\%$    & $\pm 0.006$    & $\pm 0.4$ \\
    \midrule
    Best of 5 runs           & $0.073$       & $68.2\%$       & $1.016$        & $6.5$ \\
    Best of 2 runs           & $0.076$       & $66.2\%$       & $1.018$        & $6.6$ \\
    \rowcolor{gray!15}
    Single run               & $0.079$       & $64.0\%$       & $1.021$        & $6.5$ \\
    \bottomrule
  \end{tabular}
  \caption{\textbf{Run-to-run variability} over 5 independent runs on the 50 HumanPrim objects of \Cref{tab:view_coverage}: mean, the per-object standard deviation across runs (averaged over objects), and the standard deviation of the 50-object means across runs. The lower rows select, per object, the run with the lowest Chamfer distance to the input among 5, 2 or 1 runs (averaged over all run subsets).}
  \label{S-tab:determinism}
\end{table}

\Cref{fig:determinism} of the main paper shows the five segmentations of two objects.
The runs differ in part granularity (the chair back is one part or frame plus slats) and in boundary placement, and the fitted abstractions inherit these differences.
Color assignments differ as well, since the vision-language model selects an independent palette each time.
The insect example (bottom row) also shows typical failure modes that vary across runs.
Wings are occasionally represented as blocky, volumetric structures rather than flat surfaces, and may be misclassified or merged with the body in some views.

\section{Failure Cases and Limitations}
\label{S-sec:failure_full}

\Cref{sec:failure_cases} in the main paper summarizes the three main failure modes.
Instance identity across views is discussed separately in \Cref{S-sec:instance_seg}.
Here we describe each in more detail, together with structural limitations not discussed there.

\textbf{Phantom view hallucination.}
When the multi-view composite image contains large areas of white background, the generative image model occasionally fills that space with additional repetitive views that were not present in the input.
The resulting part mask is locally invalid, but this does not propagate to the 3D abstraction, because votes from hallucinated pixels land outside the object silhouette and are silently discarded during reprojection.
Cropping the composite more tightly around each object reduces the frequency of this failure.

\textbf{Uncontrollable part granularity.}
The model decides how finely to segment.
The same object may receive different part counts in different runs, with a per-object standard deviation of about one primitive (\Cref{S-sec:determinism_full}).
The default granularity is usually semantically reasonable.
Controlling the number of parts through prompting is untested.

\textbf{Instance identity across views.}
The generative image model cannot keep left/right identities consistent across opposing views, so instance-colored masks do not reproject into a coherent 3D labeling (\Cref{S-sec:instance_seg}).
Our prompt therefore assigns one color per part \emph{type}.
Instance-level distinctions such as individual legs collapse into one semantic cluster, and the flood fill recovers them only where the instances are spatially separated.

\textbf{Small-cluster loss and color bleed.}
The flood-fill step discards clusters below a minimum size, tuned once on the development subset (\Cref{S-sec:hyperparams}), which can remove small but semantically important structures.
An object-adaptive threshold could reduce such losses.
Adjacent parts with similar or bleeding colors merge into a single cluster fitted by one primitive, which a joint segmentation-and-fitting algorithm such as EMS~\cite{liu2022ems} could split again.

\textbf{Surface-only fitting.}
The optimizer minimizes surface Chamfer distance, and only the visible surface can be rendered and segmented, so the thickness of a part seen from one side is unobservable.
The Chamfer-optimal superquadric is a sheet of zero thickness, which leaves the surface metrics untouched but removes the part's volume.
The minimum-axis floor (\Cref{sec:fitting}) prevents the degenerate solution, but it is a lower bound, not a reconstruction of the true thickness.
This contributes to the lower IoU values relative to methods that supervise volumetric overlap directly.

\textbf{Fixed viewpoints.}
The four ring viewpoints leave undersides of wings, seats and table tops unobserved, which costs volume rather than surface accuracy (\Cref{S-sec:view_layout_full}).
A top and a bottom view expose most of this surface, and beyond about $98\,\%$ coverage further views do not help.

\section{Future Work and Improvements}
\label{S-sec:future_work_full}

\textbf{Ensemble segmentation.}
Running the segmentation $k$ times on the same object, aligning the part labels across runs and aggregating them per point by majority vote before clustering would average out the run-to-run variation in part boundaries (\Cref{S-sec:determinism_full}), at the cost of $k-1$ additional pairs of model calls.

\textbf{Adaptive view selection.}
In \Cref{S-sec:view_layout_full}, IoU follows surface coverage more closely than the number of views.
A per-object strategy that adds views only where surface remains unobserved would reach the coverage of the six-view layout at lower cost and handle flat or strongly concave objects that a fixed layout leaves partly unobserved.

\textbf{Volume-aware fitting.}
The fitter sees surfaces only, so one-sided parts receive the minimum thickness rather than their true one.
A weak volumetric prior, for example a penalty on empty space between a part's visible surface and the neighboring parts, could recover thickness where the geometry constrains it.

\end{document}